\documentclass[letterpaper,12pt,3p]{elsarticle} 
\usepackage{style}

\begin{document}
\vspace*{0.5em}
\title{{\color{uiucbluedark}\small{\href{https://doi.org/10.1016/j.arcontrol.2021.10.001}{\textbf{TUTORIAL ARTICLE}}}} \vspace{1.5mm}\\ \textbf{Contraction Theory for Nonlinear Stability Analysis and Learning-based Control: A Tutorial Overview}}
\journal{Annual Reviews in Control}

\author[1]{\textcolor{uiucbluedark}{\href{https://hirotsukamoto.com/}{Hiroyasu Tsukamoto}}}
\ead{hiroyasu@alumni.caltech.edu; hiroyasu@illinois.edu}
\author[1,3]{\textcolor{uiucbluedark}{\href{https://aerospacerobotics.caltech.edu/}{Soon-Jo Chung}}}
\ead{sjchung@caltech.edu}

\author[2]{\textcolor{uiucbluedark}{\href{https://web.mit.edu/nsl/www/}{Jean-Jacques Slotine}}}
\ead{jjs@mit.edu}
\address[1]{Division of Engineering and Applied Science, California Institute of Technology, Pasadena, CA, USA}
\address[3]{Jet Propulsion Laboratory, California Institute of Technology, Pasadena, CA, USA}
\address[2]{Nonlinear Systems Laboratory, Massachusetts Institute of Technology, Cambridge MA, USA}


\tnotetext[1]{This work is in part funded by NASA's Jet Propulsion Laboratory, California Institute of Technology. Lecture notes: {\color{uiucbluedark}{\href{https://hirotsukamoto.com/contraction/}{https://hirotsukamoto.com/contraction/}}}.}
\maketitle
\thispagestyle{titlepagefancy}
\section*{\textbf{\normalsize{Abstract}}}
\noindent Contraction theory is an analytical tool to study differential dynamics of a non-autonomous (i.e., time-varying) nonlinear system under a contraction metric defined with a uniformly positive definite matrix, the existence of which results in a necessary and sufficient characterization of incremental exponential stability of multiple solution trajectories with respect to each other. By using a squared differential length as a Lyapunov-like function, its nonlinear stability analysis boils down to finding a suitable contraction metric that satisfies a stability condition expressed as a linear matrix inequality, indicating that many parallels can be drawn between well-known linear systems theory and contraction theory for nonlinear systems. Furthermore, contraction theory takes advantage of a superior robustness property of exponential stability used in conjunction with the comparison lemma. This yields much-needed safety and stability guarantees for neural network-based control and estimation schemes, without resorting to a more involved method of using uniform asymptotic stability for input-to-state stability. Such distinctive features permit the systematic construction of a contraction metric via convex optimization, thereby obtaining an explicit exponential bound on the distance between a time-varying target trajectory and solution trajectories perturbed externally due to disturbances and learning errors. The objective of this paper is, therefore, to present a tutorial overview of contraction theory and its advantages in nonlinear stability analysis of deterministic and stochastic systems, with an emphasis on deriving formal robustness and stability guarantees for various learning-based and data-driven automatic control methods. In particular, we provide a detailed review of techniques for finding contraction metrics and associated control and estimation laws using deep neural networks.


\pagestyle{myfancy}
\newpage
\renewcommand{\contentsname}{Table of Contents}
\tableofcontents
\newpage
\section{Introduction}
Lyapunov theory is one of the most widely-used approaches to stability analysis of a nonlinear system~\cite{Isidori:1995:NCS:545735,marino1995nonlinear,Khalil:1173048,Vidyasagar,Ref_Slotine,Nijmeijer}, which provides a condition for stability with respect to an equilibrium point, a target trajectory, or an invariant set. Contraction theory~\cite{Ref:contraction1,Ref:contraction4,Ref:contraction5,Ref:contraction2,Ref:contraction3} rewrites suitable Lyapunov stability conditions using a quadratic Lyapunov function of the differential states, defined by a Riemannian contraction metric and its uniformly positive definite matrix, thereby characterizing a necessary and sufficient condition for incremental exponential convergence of the multiple nonlinear system trajectories to one single trajectory. It can be regarded as a generalization of Krasovskii's theorem~\cite[p. 83]{Ref_Slotine} applied to nonlinear incremental stability analysis~\cite{Ref:contraction1,989067}, where the differential formulation permits a pure differential coordinate change with a non-constant metric for simplifying its stability proofs~\cite{Ref:contraction1}. 

The differential nature of contraction theory implies we can exploit the Linear Time-Varying (LTV) systems-type techniques for nonlinear stability analysis and control/estimation synthesis~\cite{mypaperTAC,ncm,nscm,Ref:Stochastic,mypaper} (see Table~\ref{tab:contraction_summary}). We emphasize that some of these methodological simplifications in contraction theory are accomplished by its extensive use of exponential stability along with the comparison lemma~\cite[pp. 102-103, pp. 350-353]{Khalil:1173048}, in lieu of Input-to-State Stability (ISS) or uniform asymptotic stability which often renders nonlinear stability analysis more involved~\cite{Isidori:1995:NCS:545735,marino1995nonlinear,Khalil:1173048,Vidyasagar,Ref_Slotine,Nijmeijer}. Several studies related to the notion of contraction, although not based on direct differential analysis, can be traced back to~\cite{Hart,Demi,Lew}.

The objective of this tutorial paper is to elucidate how contraction theory may be utilized as a method of providing provable incremental exponential robustness and stability guarantees of learning-based and data-driven automatic control techniques. In pursuit of this goal, we also provide an overview of the advantages of contraction theory and present a systematic convex optimization formulation to explicitly construct an optimal contraction metric and a differential Lyapunov function for general nonlinear deterministic and stochastic systems.
\begin{table}[tb]
\caption{Differences between Contraction Theory and Lyapunov Theory. \label{tab:contraction_summary}}
\footnotesize
\begin{center}
\renewcommand{\arraystretch}{1.2}
\rowcolors{1}{uiucbluedark!5}{uiucbluedark!10}
\begin{tabular}{ l  m{5cm}  m{5cm} } 
 \hline\hline
  & Contraction theory (with positive definite matrix $M(x,t)$ that defines contraction metric) & Lyapunov direct method (with Lyapunov function $V(x,t)$) \\
 \hline
 1. Lyapunov function & Always quadratic function of differential state $\delta x$ ($V=\delta x^{\top} M(x,t)\delta x$) & Any function of $x$, including $V(x,t)=x^{\top}M(x,t)x$\\
 \arrayrulecolor{mygray}\hline
 2. Stability condition & Exponential stability of trajectories including points/invariant sets$^*$ & Asymptotic or exponential stability of points and invariant sets \\ 
 \hline
 3. Incremental stability & Incremental stability of trajectories using differential displacements ($\lim_{t\to\infty}\delta x(t)=0$) & Incremental stability via stability of points ($\lim_{t\to\infty}(x(t)-x_d(t))=0$ for given $x_d(t)$) \\ 
 \hline
 4. Non-autonomous system & Same theory as for autonomous systems & Additional conditions required for non-autonomous stability analysis \\
 \hline
 5. Robustness analysis \ & Intuitive both for ISS and finite-gain $\mathcal{L}_p$ stability due to extensive use of exponential stability & 
 Same as contraction theory if exponentially stable; more involved if uniformly asymptotically stable \\
 \hline
 6. Analogy to linear system & LTV-like differential dynamics for global convergence & Indirect methods use linearization for local stability (direct methods use motion integrals) \\
 \hline
 7. $\mathcal{L}_2$ stability condition & Reduces to LMI conditions in terms of contraction metric defined by positive definite matrix $M$ & Hamilton-Jacobi inequality (PDE) in terms of Lyapunov function $V$~\cite[p. 211]{Khalil:1173048}\\
 \hline
 8. Modular stability & Differential analysis handles hierarchical, feedback, and parallel combinations~\cite{Ref:contraction2} & Passivity is not intuitive for hierarchical combinations \\
 \arrayrulecolor{black}\hline\hline
\end{tabular}
\end{center}
\footnotesize{$^*$A semi-contracting system with a negative semi-definite generalized Jacobian matrix can be used to analyze asymptotic stability (see Sec.~\ref{Sec:adaptive}).}
\end{table}
\renewcommand{\arraystretch}{1.0}
\subsection{Paper Organization}
\label{Sec:organization}
This tutorial paper is organized into the following two groups of sections.
\subsubsection*{\hyperref[part1NSA]{Part~I: Nonlinear Stability Analysis (Sec.~\ref{sec:ContractionCH2}--\ref{sec:convex})}}
In Sec.~\ref{sec:ContractionCH2}, we present fundamental results of contraction theory for nonlinear robustness and stability analysis. Sections~\ref{sec:HinfKYP} and~\ref{sec:convex} consider nonlinear optimal feedback control and estimation problems from the perspective of contraction theory, deriving and delineating a convex optimization-based method for constructing contraction metrics. Section~\ref{sec:HinfKYP} also presents some new results on relating contraction theory to the bounded real lemma~\cite{lmi} and Kalman–Yakubovich–Popov (KYP) lemma~\cite[p. 218]{kypbook}.
\subsubsection*{\hyperref[part2LBC]{Part~II: Learning-based Control (Sec.~\ref{Sec:learning_stability}--\ref{Sec:datadriven})}}
In Sec.~\ref{Sec:learning_stability}, we derive several theorems which form the basis of learning-based control using contraction theory. Sections~\ref{Sec:ncm} and~\ref{Sec:lagros} present frameworks for learning-based control, estimation, and motion planning via contraction theory using deep neural networks for designing contraction metrics, and Section~\ref{Sec:adaptive} extends these results to parametric uncertain nonlinear systems with adaptive control techniques. In Sec.~\ref{Sec:datadriven}, we propose model-free versions of contraction theory for learning-based and data-driven control.
\subsection{Related Work}
In the remainder, we give an overview of each section (Sec.~\ref{sec:ContractionCH2}--\ref{Sec:datadriven}) as well as a survey of related work.
\subsubsection*{Contraction Theory (Sec.~\ref{sec:ContractionCH2})}
\label{sec:contraction_lyapunov}
According to contraction theory, all the solution trajectories of a given nonlinear system converge to one single trajectory incrementally and exponentially, regardless of the initial conditions, if the system has a contraction metric and its associated quadratic Lyapunov function of the differential state~\cite{Ref:contraction1} (see $1$--$3$ of Table~\ref{tab:contraction_summary}). This paper primarily considers this generalized notion of stability, called incremental exponential stability, which enables systematic learning-based and data-driven control synthesis with formal robustness and stability guarantees. The purpose of this section is not for proposing that other notions of stability, such as traditional Lyapunov-based stability or incremental asymptotic stability for semi-contraction systems, should be replaced by incremental exponential stability, but for clarifying its advantages to help determine which of these concepts is the best fit when analyzing nonlinear robustness and stability (see Sec.~\ref{sec:ContractionCH2} for the illustrative examples).

In keeping with the use of the comparison lemma~\cite[pp. 102-103, pp. 350-353]{Khalil:1173048}, incremental exponential stability naturally holds for non-autonomous nonlinear systems without any additional conditions or modifications unlike Lyapunov techniques (see, \eg{}, the examples and theorems in~\cite{Rantzer}). Such aspects of contraction theory, including the extensive use of exponential stability, result in intuitive proofs on ISS and finite-gain $\mathcal{L}_p$ stability both for autonomous and non-autonomous nonlinear systems, without resorting to uniform asymptotic stability which makes stability analysis much more involved than necessary~\cite{Isidori:1995:NCS:545735,marino1995nonlinear,Khalil:1173048,Vidyasagar,Ref_Slotine,Nijmeijer}. In particular, perturbed systems with a time-varying target trajectory are non-autonomous, and thus contraction theory allows us to easily obtain an explicit exponential bound on its tracking error~\cite{Ref:contraction1,Ref:phasesync,mypaperTAC,Ref:Stochastic,mypaper}, leveraging incremental stability of the perturbed system trajectories with respect to the target trajectory (see $4$ of Table~\ref{tab:contraction_summary}). Having such analytical bounds on the tracking error is almost essential for the safe and robust implementation of automatic control schemes in real-world scenarios.

Contraction theory also simplifies input-output stability analysis, such as $\mathcal{L}_p$ gain analysis of nonlinear systems including the $\mathcal{H}_{\infty}$ nonlinear optimal control problem~\cite{256331,159566,29425,85062,151101,599969,doi:10.1137/S0363012996301336,1259458}. For example, in Lyapunov theory, the problem of finding a suitable Lyapunov function with the smallest $\mathcal{L}_2$ gain boils down to solving a Partial Differential Equation (PDE) called the Hamilton-Jacobi inequality~\cite[p. 211]{Khalil:1173048},~\cite{256331} in terms of its associated Lyapunov function. In essence, since contraction theory utilizes a quadratic Lyapunov function of the differential state for stability analysis, the problem could be solved with a Linear Matrix Inequality (LMI) constraint~\cite{lmi} analogous to the KYP lemma~\cite[p. 218]{kypbook} in LTV systems theory~\cite{mypaperTAC,Ref:Stochastic,mypaper,rccm}, as shall be shown in Sec.~\ref{sec:HinfKYP}--\ref{sec:convex} (see $5$--$7$ of Table~\ref{tab:contraction_summary}). There exist stochastic analogues of these stability results for nonlinear systems with stochastic perturbations~\cite{Pham2009,mypaperTAC,Ref:Stochastic,mypaper,han2021incremental}, as shall be outlined also in this paper.

Another notable feature of contraction theory is modularity, which preserves contraction through parallel, feedback, and hierarchical combinations~\cite{Ref:contraction2,Ref:contraction5}, specific time-delayed feedback communications~\cite{1618853}, synchronized coupled oscillations~\cite{Ref:contraction3,Ref:contraction_robot}, and synchronized networks~\cite{Ref:contraction2,Ref:phasesync,Ref:contraction_sync,Ref:contraction_robot,6327337,Ref:phasesync,Ref:ChungTRO}, expanding the results obtainable with the passivity formalism of Lyapunov theory~\cite{passivity},~\cite[p. 227]{Khalil:1173048},~\cite[p. 132]{Ref_Slotine} (see $8$ of Table~\ref{tab:contraction_summary}). Due to all these useful properties, extensions of contraction theory have been considered in many different settings. These include, but are not limited to, stochastic contraction (Gaussian white noise~\cite{Pham2009,mypaperTAC,Ref:Stochastic,mypaper}, Poisson shot noise and L\'{e}vy noise~\cite{han2021incremental}), contraction for discrete and hybrid nonlinear systems~\cite{Ref:contraction1,Ref:contraction_robot,Ref:contraction4,4795665,mypaperTAC,mypaper,wei2021discretetime}, partial contraction~\cite{Ref:contraction3}, transverse contraction~\cite{MANCHESTER201432}, incremental stability analysis of nonlinear estimation (the Extended Kalman Filter (EKF)~\cite{6849943}, nonlinear observers~\cite{zhao2005discrete,Ref:Stochastic}, Simultaneous Localization And Mapping (SLAM)~\cite{doi:10.1177/0278364917710541}), generalized gradient descent based on geodesical convexity~\cite{beyondconvexity}, contraction on Finsler and Riemannian manifolds~\cite{6632882,scrm,SIMPSONPORCO201474,finsler_ccm}, contraction on Banach and Hilbert spaces for PDEs~\cite{6580717,6161268,pdecontraction}, non-Euclidean contraction~\cite{sontag_contraction,davydov2021noneuclidean}, contracting learning with piecewise-linear basis functions~\cite{lohmiller2018notes}, incremental quadratic stability analysis~\cite{quad_stability}, contraction after small transients~\cite{MARGALIOT2016178}, immersion and invariance stabilizing controller design~\cite{7581040,1193738}, and Lipschitz-bounded neural networks for robustness and stability guarantees~\cite{neurallander,revay2020lipschitz,revay2021recurrent,cdc_systemid}.
\subsubsection*{Construction of Contraction Metrics (Sec.~\ref{sec:HinfKYP}--\ref{sec:convex})}
\label{sec:contraction_search}
The benefits of contraction theory reviewed so far naturally lead to a discussion on how to design a contraction metric and corresponding Lyapunov function. There are some cases in which we can analytically find them using special structures of systems in question~\cite{Isidori:1995:NCS:545735,Ref_Slotine,Khalil:1173048}. Among these are Lagrangian systems~\cite[p. 392]{Ref_Slotine}, where one easy choice of positive definite matrices that define a contraction metric is the inertia matrix, or feedback linearizable systems~\cite{532343,Primbs99nonlinearoptimal,1425952,AYLWARD20082163,5443730}, where we could solve the Riccati equation for a contraction metric as in LTV systems. This is also the case in the context of state estimation (\eg{}, the nonlinear SLAM problem can be reformulated as an LTV estimation problem using virtual synthetic measurements~\cite{doi:10.1177/0278364917710541,Ref:Stochastic}). Once we find a contraction metric and Lyapunov function of a nominal nonlinear system for the sake of stability, they could be used as a Control Lyapunov Function (CLF) to attain stabilizing feedback control~\cite{doi:10.1137/0321028,SONTAG1989117,Khalil:1173048} or could be augmented with an integral control law called adaptive backstepping to recursively design a Lyapunov function for strict- and output-feedback systems~\cite{backstepping,DENG1997143,Deng1999,5160111}. However, deriving an analytical form of contraction metrics for general nonlinear systems is challenging, and thus several search algorithms have been developed for finding them at least numerically using the LMI nature of the contraction condition.

The simplest of these techniques is the method of State-Dependent Riccati Equation (SDRE)~\cite{sddre,Banks2007,survey_SDRE}, which uses the State-Dependent Coefficient (SDC) parameterization (also known as extended linearization) of nonlinear systems for feedback control and state estimation synthesis. Motivating optimization-based approaches to design a contraction metric, it is proposed in~\cite{mypaperTAC,Ref:Stochastic,mypaper} that the Hamilton-Jacobi inequality for the finite-gain $\mathcal{L}_2$ stability condition can be expressed as an LMI when contraction theory is equipped with the extended linearity of the SDC formulation. Specifically, in~\cite{mypaperTAC,ncm,nscm,mypaper}, a convex optimization-based framework for robust feedback control and state estimation, named ConVex optimization-based Steady-state Tracking Error Minimization (CV-STEM), is derived to find a contraction metric that minimizes an upper bound of the steady-state distance between perturbed and unperturbed system trajectories. In this context, we could utilize Control Contraction Metrics (CCMs)~\cite{ccm,7989693,47710,WANG201944,vccm,rccm} for extending contraction theory to the systematic design of differential feedback control $\delta u = k(x,\delta x,u,t)$ via convex optimization, achieving greater generality at the expense of computational efficiency in obtaining $u$. Applications of the CCM to estimation, adaptive control, and motion planning are discussed in~\cite{estimation_ccm},~\cite{9109296,lopez2021universal,regretboffi}, and~\cite{7989693,8814758,9303957,L1contraction2,sun2021uncertaintyaware}, respectively, using geodesic distances between trajectories~\cite{scrm}. It is also worth noting that the objective function of CV-STEM has the condition number of a positive definite matrix that defines a contraction metric as one of its arguments, rendering it applicable and effective even to machine learning-based automatic control frameworks as shall be seen in Sec.~\ref{Sec:learning_stability}--\ref{Sec:datadriven}.
\subsubsection*{Contraction Theory for Learning-based Control (Sec.~\ref{Sec:learning_stability})}
\label{sec:contraction_machine_learning}
One drawback of these numerical schemes is that they require solving optimization problems or nonlinear systems of equations at each time instant online, which is not necessarily realistic in practice. In Lyapunov theory, approximating functions in a given hypothesis space has therefore been a standard technique~\cite{664157,JOHANSEN20001617,LyapunovRBF,1184414,spencer18lyapunovnn,NIPS2019_8587,8263816,9115021,9146356,9147615,gaby2021lyapunovnet}, where examples of its function classes include piecewise quadratic functions~\cite{664157}, linearly parameterized non-quadratic functions~\cite{JOHANSEN20001617}, a linear combination of radial basis functions~\cite{LyapunovRBF}, Sum-Of-Squares (SOS) functions~\cite{1184414}, and neural networks~\cite{spencer18lyapunovnn,NIPS2019_8587,gaby2021lyapunovnet}. In~\cite{AYLWARD20082163,ccm}, the SOS approximation is investigated for the case of contraction theory, showing that the contraction condition can be relaxed to SOS constraints for dynamics with polynomial or rational vector fields. Although computationally tractable, it still has some limitations in that the problem size grows exponentially with the number of variables and basis functions~\cite{sos_dissertation}. Learning-based and data-driven control using contraction theory~\cite{ncm,nscm,ancm,lagros} has been developed to refine these ideas, using the high representational power of DNNs~\cite{neural1,neural2,neural3} and their scalable training realized by stochastic gradient descent~\cite{sgd,beyondconvexity}.

The major advantage of using contraction theory for learning-based and data-driven control is that, by regarding its internal learning error as an external disturbance, we can ensure the distance between the target and learned trajectories to be bounded exponentially with time as in the CV-STEM results~\cite{ncm,nscm,ancm,lagros}, with its steady-state upper bound proportional to the learning error. Such robustness and incremental stability guarantees are useful for formally evaluating the performance of machine learning techniques such as reinforcement learning~\cite{sutton,ndp,8593871,NIPS2017_766ebcd5}, imitation learning~\cite{9001182,glas,NIPS2016_cc7e2b87,8578338,7995721}, or neural networks~\cite{neural1,neural2,neural3}. This implies contraction theory could be utilized as a central tool in realizing safe and robust operations of learning-based and data-driven control, estimation, and motion planning schemes in real-world scenarios. We especially focus on the following areas of research.
\subsubsection*{Learning-based Robust Control and Estimation (Sec.~\ref{Sec:ncm})}
In order to achieve real-time computation of a contraction metric, mathematical models based on a Deep Neural Network (DNN) called a Neural Contraction Metric (NCM)~\cite{ncm} and Neural Stochastic Contraction Metric (NSCM)~\cite{nscm} are derived to compute optimal CV-STEM contraction metrics for nonlinear systems perturbed by deterministic and stochastic disturbances, respectively. It can be proven that the NCM and NSCM still yield robustness and optimality associated with the CV-STEM framework despite having non-zero modeling errors~\cite{cdc_ncm}. These metrics could also be synthesized and learned simultaneously with their feedback control laws directly by DNNs~\cite{chuchu,sun2021uncertaintyaware,cdc_ncm} at the expense of the convex property in the CV-STEM formulation.
\subsubsection*{Learning-based Robust Motion Planning (Sec.~\ref{Sec:lagros})}
In~\cite{7989693,zhao2021tubecertified}, contraction theory is leveraged to develop a tracking
feedback controller with an optimized control invariant tube, solving the problem of robust motion planning under bounded external disturbances. This problem is also considered for systems with changing operating conditions~\cite{8814758} and parametric uncertainty~\cite{9303957,L1contraction2,sun2021uncertaintyaware} for its broader use in practice. As these methods still require online computation of a target trajectory, Learning-based Autonomous Guidance with RObustness and Stability (LAG-ROS)~\cite{lagros} is developed to model such robust control laws including the CV-STEM by a DNN, without explicitly requiring the target or desired trajectory as its input. While this considers motion planning algorithms only implicitly to avoid solving them in real-time, it is shown that contraction theory still allows us to assure a property of robustness against deterministic and stochastic disturbances following the same argument as in the NCM and NSCM work~\cite{lagros}. Note that LAG-ROS using contraction theory is not intended to derive new learning-based motion planning, but rather to augment any existing motion planner with a real-time method of guaranteeing formal incremental robustness and stability, and thus still applicable to other methods such as tube-based robust Model Predictive Control (MPC)~\cite{tube_nmpc,tube_mpc,7989693,10.1007/BFb0109870,zhao2021tubecertified,learningmpc}, its dynamic and adaptive counterparts~\cite{8814758,9303957,L1contraction2,sun2021uncertaintyaware}, and CCM-based learning certified control~\cite{chuchu}.
\subsubsection*{Learning-based Adaptive Control (Sec.~\ref{Sec:adaptive})}
Adaptive control using contraction theory is studied in~\cite{5160111} for parametric strict-feedback nonlinear systems, and recently generalized to deal with systems with unmatched parametric uncertainty by means of parameter-dependent CCM feedback control~\cite{9109296,lopez2021universal}. This method is further explored to develop an adaptive Neural Contraction Metric (aNCM)~\cite{ancm}, a parameter-dependent DNN model of the adaptive CV-STEM contraction metric. As the name suggests, the aNCM control makes adaptive control of~\cite{9109296,lopez2021universal} implementable in real-time for asymptotic stabilization, while maintaining the learning-based robustness and CV-STEM-type optimality of the NCM. Although it is designed to avoid the computation for evaluating integrals involving geodesics unlike~\cite{9109296,lopez2021universal}, these differential state feedback schemes could still be considered, trading off added computational cost for generality. It is demonstrated in~\cite{ancm} that the aNCM is applicable to many types of systems such as robotics systems~\cite[p. 392]{Ref_Slotine}, spacecraft high-fidelity dynamics~\cite{battin,doi:10.2514/1.55705}, and systems modeled by basis function approximation and DNNs~\cite{Nelles2001,SannerSlotine1992}. Discrete changes could be incorporated in this framework using~\cite{regretboffi,1469901}.
\subsubsection*{Contraction Theory for Learned Models (Sec.~\ref{Sec:datadriven})}
Recent applications of machine learning often consider challenging scenarios in the field of systems and control theory, where we only have access to system trajectory data generated by unknown underlying dynamics, and the assumptions in the aforementioned adaptive control techniques are no longer valid. For situations where the data is used for system identification of full/residual dynamics~\cite{neurallander,47710,cdc_systemid,7992901} by a spectrally-normalized DNN~\cite{miyato2018spectral}, we can show by contraction theory that the model-based approaches (\eg{} CV-STEM and NCM) are still utilizable to guarantee robustness against dynamics modeling errors and external disturbances. It is also proposed in~\cite{boffi2020learning} that we could directly learn certificate functions such as contraction metrics and their associated Lyapunov functions using trajectory data. Note that some of the theoretical results on gradient descent algorithms, essential in the field of data-driven machine learning, can be replaced by more general ones based on contraction and geodesical convexity~\cite{beyondconvexity}.
\subsection{Notation}
\label{notation}
For a square matrix $A\in\mathbb{R}^{n\times n}$, we use the notation $A \succ 0$, $A \succeq 0$, $A \prec 0$, and $A \preceq 0$ for the positive definite, positive semi-definite, negative definite, and negative semi-definite matrices, respectively. The $\mathcal{L}_p$ norm in the extended space $\mathcal{L}_{pe}$~\cite[pp. 196-197]{Khalil:1173048}, $p \in [1,\infty]$, is defined as
$\|(y)_{\tau}\|_{\mathcal{L}_p} = \left(\int_0^\tau \|y(t)\|^p\right)^{{1}/{p}} < \infty$ for $p\in[1,\infty)$ and $\|(y)_{\tau}\|_{\mathcal{L}_{\infty}} = \sup_{t\geq 0}\|(y(t))_{\tau}\| < \infty$ for $p =\infty$,
where $(y(t))_{\tau}$ is a truncation of $y(t)$, \ie, $(y(t))_{\tau} = 0$ for $t > \tau$ and $(y(t))_{\tau} = y(t)$ for $0 \leq t \leq \tau$ with $\tau \in \mathbb{R}_{\geq 0}$. Furthermore, we use $f_{x} = \partial f/\partial x$, $M_{x_i} = \partial M/\partial x_i$, and $M_{x_ix_j} = \partial^2 M/(\partial x_i\partial x_j)$, where $x_i$ and $x_j$ ate the $i$th and $j$th elements of $x \in \mathbb{R}^n$, for describing partial derivatives in a limited space. The other notations are given in Table~\ref{tab:notations_in_this_paper}.
\begin{table}[tb]
\caption{Notations used in this paper. \label{tab:notations_in_this_paper}}
\footnotesize
\begin{center}
\renewcommand{\arraystretch}{1.2}
\rowcolors{1}{uiucbluedark!5}{uiucbluedark!10}
\begin{tabular}{ l l} 
\hline \hline
$\|x\|$ & Euclidean norm of $x \in \mathbb{R}^n$ \\ \arrayrulecolor{mygray}\hline
$\delta x$ & Differential displacement of $x \in \mathbb{R}^n$ \\ \hline
$\|A\|$ & Induced $2$-norm of $A \in \mathbb{R}^{n\times m}$ \\ \hline
$\|A\|_F$ & Frobenius norm of $A \in \mathbb{R}^{n\times m}$ \\ \hline
$\sym(A)$ & Symmetric part of $A\in\mathbb{R}^{n\times n}$, \ie{}, $(A+A^{\top})/2$ \\ \hline
$\lambda_{\min}(A)$  & Minimum eigenvalue of $A\in\mathbb{R}^{n\times n}$ \\ \hline
$\lambda_{\max}(A)$  & Maximum eigenvalue of $A\in\mathbb{R}^{n\times n}$ \\ \hline $\mathrm{I}$ & Identity matrix of appropriate dimensions \\ \hline
$\mathop{\mathbb{E}}$ & Expected value operator \\ \hline
$\mathop{\mathbb{P}}$ & Probability measure \\ \hline
$\mathbb{R}_{>0}$ & Set of positive reals, \ie{}, $\{a\in\mathbb{R}|a\in(0,\infty)\}$ \\ \hline
$\mathbb{R}_{\geq 0}$ & Set of non-negative reals, \ie{}, $\{a\in\mathbb{R}|a\in[0,\infty)\}$ \\ \arrayrulecolor{black}\hline\hline
\end{tabular}
\end{center}
\end{table}
\newpage
\vspace*{20em}
\part*{\huge \normalfont{Part I: Nonlinear Stability Analysis}}
\label{part1NSA}
\vspace{2em}
\newpage
\section{Contraction Theory}
\label{sec:ContractionCH2}
We present a brief review of the results from~\cite{Ref:contraction1,Ref:contraction2,Ref:contraction3,Pham2009,Ref:phasesync,mypaperTAC,Ref:Stochastic,mypaper}. They will be extensively used to provide formal robustness and stability guarantees for a variety of systems in the subsequent sections, simplifying and generalizing Lyapunov theory.
\subsection{Fundamentals}
Consider the following smooth non-autonomous (\ie{}, time-varying) nonlinear system:
\begin{equation}
\label{eq:xfx}
{{\dot
x}}(t)={f}({x}(t),t)
\end{equation}
where $t \in \mathbb{R}_{\geq 0}$ is time, ${x}:\mathbb{R}_{\geq 0}\rightarrow\mathbb{R}^n$ the system state, and ${f}:
\mathbb{R}^n\times\mathbb{R}_{\geq 0}\rightarrow\mathbb{R}^n$ is a smooth function. Note that the smoothness of $f(x,t)$ guarantees existence and uniqueness of the solution to \eqref{eq:xfx} for a given $x(0)=x_0$ at least locally~\cite[pp. 88-95]{Khalil:1173048}.
\begin{definition}{Differential Dynamics}{}
A differential displacement, $\delta{x}$, is defined as an
infinitesimal displacement at a fixed time as used in the calculus of variation~\cite[p. 107]{citeulike:802300}, and \eqref{eq:xfx} yields the following differential dynamics:
\begin{align}
\label{eq:original_differential}
\delta\dot{x}(t)=\frac{\partial f}{\partial x}({x}(t),t)\delta x(t)
\end{align}
where $f({x}(t),t)$ is given in \eqref{eq:xfx}.
\end{definition}

Let us first present a special case of the comparison lemma~\cite[pp. 102-103, pp. 350-353]{Khalil:1173048} to be used extensively throughout this paper.
\begin{lemma}
\label{Lemma:comparison}
Suppose that a continuously differentiable function $v\in\mathbb{R}_{\geq0}\rightarrow\mathbb{R}$ satisfies the following differential inequality:
\begin{align}
\label{cl_ivp}
\dot{v}(t) \leq -\gamma v(t)+c,~v(0)=v_0,~\forall t\in\mathbb{R}_{\geq 0}
\end{align}
where $\gamma \in \mathbb{R}_{>0}$, $c\in\mathbb{R}$, and $v_0\in\mathbb{R}$. Then we have
\begin{align}
v(t) \leq v_0e^{-\gamma t}+\frac{c}{\gamma}(1-e^{-\gamma t}),~\forall t\in\mathbb{R}_{\geq 0}.
\end{align}
\end{lemma}
\begin{proof}
See~\cite[pp. 659-660]{Khalil:1173048}. \qed
\end{proof}
\subsubsection{Contraction Theory and Contraction Metric}
\begin{figure}[tb]
  \centering
  \includegraphics[width=133.5mm]{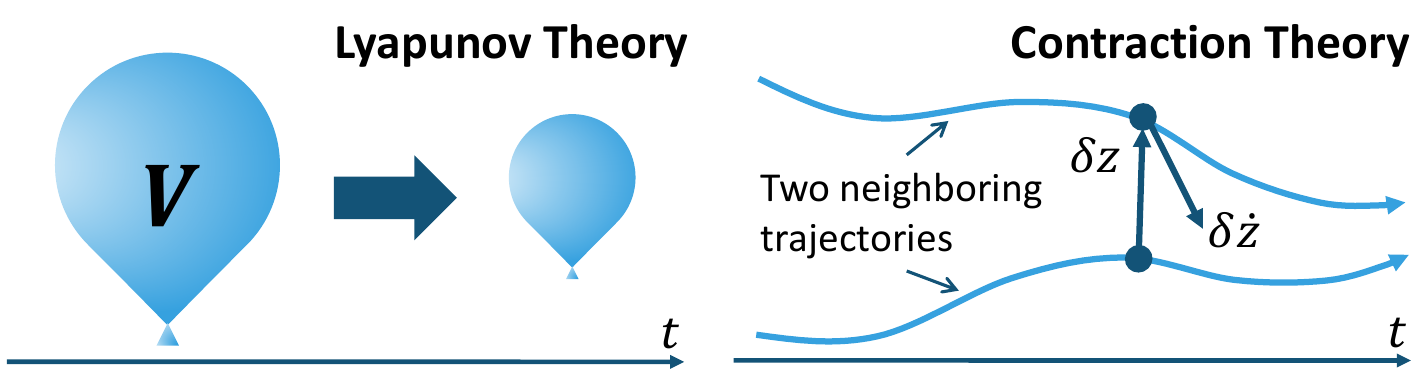}
  \caption{Lyapunov theory and contraction theory, where $V$ is a Lyapunov function and $\delta z = \Theta (x,t)\delta x$ for $M(x,t)=\Theta(x,t)\Theta(x,t)^{\top}\succ 0$ that defines a contraction metric (see Theorem~\ref{THM:Thm:contraction}).}
  \label{fig:contraction}
\end{figure}

In Lyapunov theory, nonlinear stability of \eqref{eq:xfx} is studied by constructing a Lyapunov function $V(x,t)$, one example of which is $V=x^\top P(x,t)x$. However, finding $V(x,t)$ for general nonlinear systems is challenging as $V(x,t)$ can be any scalar function of $x$ (\eg{}, a candidate $V(x,t)$ can be obtained by solving a PDE~\cite[p. 211]{Khalil:1173048}). In contrast, as summarized in Table~\ref{tab:contraction_summary}, contraction theory uses a differential Lyapunov function that is always a quadratic function of $\delta x$, \ie{}, $V(x,\delta x,t)=\delta x^{\top} M(x,t)\delta x$, thereby characterizing a necessary and sufficient condition for incremental exponential convergence of the multiple nonlinear system trajectories to one single trajectory. Thus, the problem of finding $V$ for stability analysis boils down to finding a finite-dimensional positive-definite matrix $M$, as illustrated in Figure~\ref{fig:contraction}~\cite{Ref:contraction1}. These properties to be derived in Theorem~\ref{THM:Thm:contraction}, which hold both for autonomous (\ie{} time-invariant) and non-autonomous systems, epitomize significant methodological simplifications of stability analysis in contraction theory.
\begin{theorem}{Deterministic Contraction}{Thm:contraction}
If there exists a uniformly positive definite matrix ${M}({x},t)={{\Theta}}({x},t)^{\top}{{\Theta}}({x},t) \succ 0,~\forall x,t$, where ${\Theta(x,t)}$ defines a smooth coordinate transformation of $\delta x$, \ie{}, $\delta{z}={\Theta}(x,t)\delta{x}$, \st{} either of the following equivalent conditions holds for $\exists \alpha \in \mathbb{R}_{>0}$, $\forall x,t$:
\begin{align}
&\lambda_{\max}(\sym(F(x,t))) =\lambda_{\max}\left(\sym\left(\left({\dot{\Theta}}+{{\Theta}}\frac{\partial
{f}}{\partial
{x}}\right){{\Theta}}^{-1}\right)\right) \leq - \alpha
\label{eq_MdotContracting_z} \\
&{\dot{M}}+M\frac{\partial
{f}}{\partial {x}}+\frac{\partial {f}}{\partial
{x}}^{\top}M \preceq -2\alpha M
\label{eq_MdotContracting}
\end{align}
where the arguments $(x,t)$ of $M(x,t)$ and $\Theta(x,t)$ are omitted for notational simplicity, then all the solution trajectories of \eqref{eq:xfx} converge to a single trajectory exponentially fast regardless of their initial conditions (\ie{}, contracting, see Definition~\ref{DEF:Def:contraction}), with an exponential convergence rate $\alpha$. The converse also holds.
\end{theorem}
\begin{proof}
The proof of this theorem can be found in~\cite{Ref:contraction1}, but here we emphasize the use of the comparison lemma given in Lemma~\ref{Lemma:comparison}. Taking the time-derivative of a differential Lyapunov function of $\delta x$ (or $\delta z$), $V=\delta {z}^{\top}\delta {z}=\delta{x}^{\top}M(x,t)\delta{x}$, using the differential dynamics \eqref{eq:original_differential}, we have
\begin{align}
\dot{V}(x,\delta x, t)&=2\delta
{z}^{\top}{F}\delta{z}=\delta{x}^{\top}\Bigl({\dot{M}}+\frac{\partial
{f}}{\partial
{x}}^{\top}{{M}}+{{M}}\frac{\partial
{f}}{\partial {x}}\Bigr)\delta{x}\nonumber \leq -2\alpha \delta{z}^{\top}\delta{z} =- 2 \alpha \delta {x}^{\top}{M} \delta {x}
\end{align}
where the conditions \eqref{eq_MdotContracting_z} and \eqref{eq_MdotContracting} are used with the generalized Jacobian $F$ in \eqref{eq_MdotContracting_z} obtained from $\delta \dot{z}=\dot{\Theta}\delta x+ \Theta\delta\dot{x}=F\delta z$. We get $d\|\delta {z}\|/dt \leq -\alpha\|\delta {z}\|$ by $d\|\delta {z}\|^2/dt=2\|\delta {z}\|d\|\delta {z}\|/dt$, which then yields $\|\delta {z}(t)\|\leq \|\delta {z}_0\|e^{-\alpha t}$ by the comparison lemma of Lemma~\ref{Lemma:comparison}. Hence, any infinitesimal length $\|\delta {z}(t)\|$ and $\|\delta {x}(t)\|$, as well as $\delta z$ and $\delta x$, tend to zero exponentially fast. By path integration (see Definition~\ref{DEF:Def:incremental} and Theorem~\ref{THM:Thm:path_integral}), this immediately implies that the length of any finite path converges exponentially to zero from any initial conditions.

Conversely, consider an exponentially convergent system, which implies the following for $\exists\beta >0$ and $\exists k \geq 1$:
\begin{align}
\label{converse_exp}
\|\delta x(t)\|^2 \leq -k\|\delta x(0)\|^2e^{-2 \beta t}
\end{align}
and define a matrix-valued function $\Xi(x(t),t)\in\mathbb{R}^{n\times n}$ (not necessarily $\Xi\succ0$) as
\begin{align}
\label{converse_Meq}
\dot{\Xi} = -2\beta \Xi-\Xi\frac{\partial
{f}}{\partial {x}}-\frac{\partial {f}}{\partial
{x}}^{\top}\Xi,~\Xi(x(0),0) = k\mathrm{I}.
\end{align}
Note that, for $V=\delta x^{\top}\Xi\delta x$, \eqref{converse_Meq} gives $\dot{V}=-2\beta V$, resulting in $V=-k\|\delta x(0)\|^2e^{-2 \beta t}$. Substituting this into \eqref{converse_exp} yields
\begin{align}
\label{converse_posdef}
\|\delta x(t)\|^2 = \delta x(t)^{\top}\delta x(t)\leq V=\delta x(t)^{\top}\Xi(x(t),t)\delta x(t)
\end{align}
which indeed implies that $\Xi \succeq \mathrm{I} \succ 0$ as \eqref{converse_posdef} holds for any $\delta x$. Thus, $\Xi$ satisfies the contraction condition \eqref{eq_MdotContracting}, \ie{}, $\Xi$ defines a contraction metric (see Definition~\ref{DEF:Def:contraction})~\cite{Ref:contraction1}. \qed
\end{proof}
\begin{remark}
Since $M$ of Theorem~\ref{THM:Thm:contraction} is positive definite, \ie{}, $v^{\top}Mv=\|\Theta v\|^2\geq0,~\forall v\in\mathbb{R}^n$, and $\|\Theta v\|^2=0$ if and only if $v=0$, the equation $\Theta v=0$ only has a trivial solution $v=0$. This implies that $\Theta$ is always non-singular (\ie{}, $\Theta(x,t)^{-1}$ always exists). 
\end{remark}
\begin{definition}{Incremental Exponential Stability}{Def:incremental}
Let $\xi_0(t)$ and $\xi_1(t)$ denote some solution trajectories of \eqref{eq:xfx}. We say that \eqref{eq:xfx} is incrementally exponentially stable if $\exists C,\alpha>0$ \st{} the following holds~\cite{989067}:
\begin{align}
\label{eq:incremental}
\|\xi_1(t)-\xi_0(t)\| \leq Ce^{-\alpha t}\|\xi_0(0)-\xi_1(0)\|
\end{align}
for any $\xi_0(t)$ and $\xi_1(t)$. Note that, since we have $\|\xi_1(t)-\xi_0(t)\|=\|\int_{\xi_0}^{\xi_1}\delta x\|$ (see Theorem~\ref{THM:Thm:path_integral}), Theorem~\ref{THM:Thm:contraction} implies incremental stability of the system \eqref{eq:xfx}.
\end{definition}
\begin{definition}{Contracting Systems}{Def:contraction}
The system \eqref{eq:xfx} satisfying the conditions in Theorem~\ref{THM:Thm:contraction} is said to be contracting, and a uniformly positive definite matrix $M$ that satisfies \eqref{eq_MdotContracting} defines a contraction metric. As to be discussed in Theorem~\ref{THM:Thm:path_integral} of Sec.~\ref{sec:robustpath}, the metric induces the Riemannian distance, and a contracting system is incrementally exponentially stable in the sense of Definition~\ref{DEF:Def:incremental}. 
\end{definition}
\begin{example}{Lyapunov Theory vs. Contraction Theory I}{ex:cont_ex1}
One of the distinct features of contraction theory in Theorem~\ref{THM:Thm:contraction} is incremental stability with exponential convergence. Consider the example given in~\cite{Ref:contraction6}:
\begin{align}
\label{ex_eq_incremental}
\dfrac{d}{dt}\begin{bmatrix}x_1\\x_2\end{bmatrix} = \begin{bmatrix}-1 & x_1\\-x_1 & -1\end{bmatrix}\begin{bmatrix}x_1\\x_2\end{bmatrix}.
\end{align}
A Lyapunov function $V = \|x\|^2/2$ for \eqref{ex_eq_incremental}, where $x=[x_1,x_2]^{\top}$, yields $\dot{V} \leq -2V$. Thus, \eqref{ex_eq_incremental} is exponentially stable with respect to $x=0$. The differential dynamics of \eqref{ex_eq_incremental} is given as
\begin{align}
\label{ex_eq_incremental_diff}
\dfrac{d}{dt}\begin{bmatrix}\delta x_1\\\delta x_2\end{bmatrix} = \begin{bmatrix}-1+x_2 & x_1\\-2x_1 & -1\end{bmatrix}\begin{bmatrix}\delta x_1\\\delta x_2\end{bmatrix}
\end{align}
and the contraction condition \eqref{eq_MdotContracting} for \eqref{ex_eq_incremental_diff} can no longer be proven by $V=\|\delta x\|^2/2$, due to the lack of the skew-symmetric property of \eqref{ex_eq_incremental} in \eqref{ex_eq_incremental_diff}. This difficulty illustrates the difference between Lyapunov theory and contraction theory, where the former considers stability of \eqref{ex_eq_incremental} with respect to the equilibrium point, while the latter analyzes exponential convergence of any couple of trajectories in \eqref{ex_eq_incremental} with respect to each other (\ie{}, incremental stability in Definition~\ref{DEF:Def:incremental})~\cite{Ref:contraction1,Ref:contraction6}.
\end{example}
\begin{example}{Lyapunov Theory vs. Contraction Theory II}{}
Contraction defined by Theorem~\ref{THM:Thm:contraction} guarantees incremental stability of their solution trajectories but does not require the existence of stable fixed points. Let us consider the following system for $x:\mathbb{R}_{\geq0}\rightarrow\mathbb{R}$:
\begin{align}
\label{easy_example}
\dot{x} = -x + e^t.
\end{align}
Using the condition \eqref{eq_MdotContracting} of Theorem~\ref{THM:Thm:contraction}, we can easily verify that \eqref{easy_example} is contracting as $M=\mathrm{I}$ defines its contraction metric with the contraction rate $\alpha=1$. However, since \eqref{easy_example} has $x(t) = e^t/2+(x(0)-1/2)e^{-t}$ as its unique solution, it is not stable with respect to any fixed point.
\end{example}
\begin{example}{Contraction in Linear Time-Invariant Systems}{ex:lyapunov_contraction}
Consider a Linear Time-Invariant (LTI) system, $\dot{x} = f(x) = Ax$. Lyapunov theory states that the origin is globally exponentially stable if and only if there exists a constant positive-definite matrix $P\in\mathbb{R}^{n\times n}$ \st{}~\cite[pp. 67-68]{lssbook}
\begin{align}
\label{lin_lyapunov}
\exists \epsilon >0\text{ \st{} }PA+A^{\top}P \preceq -\epsilon \mathrm{I}.
\end{align}
Now, let $\overline{p} = \|P\|$. Since $-\mathrm{I} \preceq -P/\overline{p}$, \eqref{lin_lyapunov} implies that $PA+A^{\top}P \leq -(\epsilon/\overline{p})P$, which shows that $M=P$ with $\alpha = \epsilon/(2\overline{p})$ satisfies \eqref{eq_MdotContracting} due to the relation $\partial f/\partial x = A$.

The contraction condition \eqref{eq_MdotContracting} can thus be viewed as a generalization of the Lyapunov stability condition \eqref{lin_lyapunov} (see the generalized Krasovskii's theorem~\cite[pp. 83-86]{Ref_Slotine}) in a nonlinear non-autonomous system setting, expressed in the differential formulation that permits a non-constant metric and pure differential coordinate change~\cite{Ref:contraction1}. Furthermore, if $f(x,t)=A(t)x$ and $M=\mathrm{I}$, \eqref{eq_MdotContracting} results in $A(t)+A(t)^{\top} \preceq -2\alpha \mathrm{I}$ (\ie{}, all the eigenvalues of the symmetric matrix $A(t)+A(t)^{\top}$ remain strictly in the left-half complex plane), which is a known sufficient condition for stability of Linear Time-Varying (LTV) systems~\cite[pp. 114-115]{Ref_Slotine}.
\end{example}
\begin{example}{Contraction in Gradient Descent}{}
Most of the learning-based techniques involving neural networks are based on optimizing their hyperparameters by gradient descent~\cite{sgd}. Contraction theory provides a generalized view on the analysis of such continuous-time gradient-based optimization algorithms~\cite{beyondconvexity}.

Let us consider a twice differentiable scalar output function $f:\mathbb{R}^n\times\mathbb{R}\rightarrow\mathbb{R}$, a matrix-valued function $M:\mathbb{R}^n\rightarrow\mathbb{R}^{n\times n}$ with $M(x)\succ0,~\forall x\in\mathbb{R}^n$, and the following natural gradient system~\cite{doi:10.1162/089976698300017746}:
\begin{align}
\label{eq_natural_grad}
    \dot{x} = h(x,t) = -M(x)^{-1}\nabla_x f(x,t).
\end{align}
Then, $f$ is geodesically $\alpha$-strongly convex for each $t$ in the metric defined by $M(x)$ (\ie{}, $H(x)\succeq \alpha M(x)$ with $H(x)$ being the Riemannian Hessian matrix of $f$ with respect to $M$~\cite{gconvex}), if and only if \eqref{eq_natural_grad} is contracting with rate $\alpha$ in the metric defined by $M$ as in \eqref{eq_MdotContracting} of Theorem~\ref{THM:Thm:contraction}, where $A = {\partial h}/{\partial x}$. More specifically, the Riemannian Hessian verifies $H(x) = -(\dot{M}+MA+A^{\top}M)/2$. See~\cite{beyondconvexity} for details.
\end{example}
\begin{remark}
Theorem~\ref{THM:Thm:contraction} can be applied to other vector norms of $\|\delta {z}\|_p$ with, \eg{}, $p=1$ or $p=\infty$~\cite{Ref:contraction1}. It can also be shown that for a contracting autonomous system of the form $\dot{x}={f}({x})$, all trajectories
converge to an equilibrium point exponentially fast. 
\end{remark}
\subsubsection{Partial Contraction}
Although satisfying the condition~\eqref{eq_MdotContracting} of Theorem~\ref{THM:Thm:contraction} guarantees exponential convergence of any couple of trajectories in \eqref{eq:xfx}, proving their incremental stability with respect to a subset of these trajectories possessing a specific property could be sufficient for some cases~\cite{mypaperTAC,ncm,nscm,mypaper}, leading to the concept of partial contraction~\cite{Ref:contraction3}.
\begin{theorem}{Partial Contraction}{Thm:partial_contraction}
Consider the following nonlinear system with the state $x\in\mathbb{R}_{\geq 0}\rightarrow\mathbb{R}^n$ and the auxiliary or virtual system with the state $q\in\mathbb{R}_{\geq 0}\rightarrow\mathbb{R}^n$:
\begin{align}
\label{xx_system}
\dot{x}(t) &= \textsl{g}(x(t),x(t),t) \\
\label{q_system}
\dot{q}(t) &= \textsl{g}(q(t),x(t),t)
\end{align}
where $\textsl{g}:
\mathbb{R}^n\times\mathbb{R}^n\times\mathbb{R}_{\geq 0}\rightarrow\mathbb{R}^n$ is a smooth function. Suppose that \eqref{q_system} is contracting with respect to $q$. If a particular solution of \eqref{q_system} verifies a smooth specific property, then all trajectories of \eqref{xx_system} verify this property exponentially.
\end{theorem}
\begin{proof}
This statement follows from Theorem~\ref{THM:Thm:contraction} and the fact that $q=x$ and a trajectory with the specific property are particular solutions of \eqref{q_system} (see~\cite{Ref:contraction3} for details). \qed
\end{proof}

The importance of this theorem lies in the fact that we can analyze contraction of some specific parts of the system~\eqref{xx_system} while treating the rest as a function of the time-varying parameter $x(t)$. Strictly speaking, the system \eqref{xx_system} is said to be partially contracting, but we will not distinguish partial contraction from contraction of Definition~\ref{DEF:Def:contraction} in this paper for simplicity. Instead, we will use the variable $q$ when referring to partial contraction of Theorem~\ref{THM:Thm:partial_contraction}.

Examples of a trajectory with a specific property include the target trajectory $x_d$ of feedback control in \eqref{sdc_dynamicsd}, or the trajectory of actual dynamics $x$ for state estimation in \eqref{sdc_dynamics_est} and \eqref{sdc_dynamics_est_sto} to be discussed in the subsequent sections. Note that contraction can be regarded as a particular case of partial contraction.
\begin{example}{Partial Contraction I}{ex:partial_cont}
Let us illustrate the role of partial contraction using the following nonlinear system~\cite{Ref:contraction6}:
\begin{align}
\dot{x} &= -D(x)x+u \\
\dot{x}_d &= -D(x_d)x_d
\end{align}
where $x$ is the system state, $x_d$ is the target state, $u$ is the control input designed as $u = -K(x)(x-x_d)+(D(x)-D(x_d))x_d$, and $D(x)+K(x) \succ 0$. We could define a virtual system, which has $q=x$ and $q=x_d$ as its particular solutions, as follows:
\begin{align}
\label{virtual_simple}
\dot{q} = -(D(x)+K(x))(q-x_d)-D(x_d)x_d.
\end{align}
Since we have $\delta\dot{q} = -(D(x)+K(x))\delta q$ and $D(x)+K(x)\succ0$, \eqref{virtual_simple} is contracting with $M=\mathrm{I}$ in \eqref{eq_MdotContracting}. However, if we consider the following virtual system:
\begin{align}
\label{virtual_complex}
\dot{q} = -(D(q)+K(q))(q-x_d)-D(x_d)x_d
\end{align}
which also has $q=x$ and $q=x_d$ as its particular solutions, proving contraction is no longer straightforward because of the terms $\partial D/\partial q_i$ and $\partial K/\partial q_i$ in the differential dynamics of \eqref{virtual_complex}. This is due to the fact that, in contrast to \eqref{virtual_simple}, \eqref{virtual_complex} has particular solutions nonlinear in $q$ in addition to $q=x$ and $q=x_d$, and the condition \eqref{eq_MdotContracting} becomes more involved for \eqref{virtual_complex} as it is for guaranteeing exponential convergence of any couple of these particular solution trajectories~\cite{Ref:contraction6}.
\end{example}
\begin{example}{Partial Contraction II}{ex:lag_metric}
As one of the key applications of partial contraction given in Theorem~\ref{THM:Thm:partial_contraction}, let us consider the following closed-loop Lagrangian system~\cite[p. 392]{Ref_Slotine}:
\begin{align}
\label{lagrange_partial}
&\mathcal{H}(\mathtt{q})\ddot{\mathtt{q}}+\mathcal{C}(\mathtt{q},\dot{\mathtt{q}})\dot{\mathtt{q}}+\mathcal{G}(\mathtt{q})=u(\mathtt{q},\dot{\mathtt{q}},t) \\
\label{lagrange_u}
&u(\mathtt{q},\dot{\mathtt{q}},t) = \mathcal{H}(\mathtt{q})\ddot{\mathtt{q}}_r+\mathcal{C}(\mathtt{q},\dot{\mathtt{q}})\dot{\mathtt{q}}_r+\mathcal{G}(\mathtt{q})-\mathcal{K}(t)(\dot{\mathtt{q}}-\dot{\mathtt{q}}_r)
\end{align}
where $\mathtt{q},\dot{\mathtt{q}}\in\mathbb{R}^n$, $\dot{\mathtt{q}}_r=\dot{\mathtt{q}}_d(t)-\Lambda(t)(\mathtt{q}-\mathtt{q}_d(t))$, $\mathcal{H}: \mathbb{R}^n\rightarrow\mathbb{R}^{n\times n}$, $\mathcal{C}: \mathbb{R}^n\times \mathbb{R}^n\rightarrow\mathbb{R}^{n\times n}$, $\mathcal{G}: \mathbb{R}^n\rightarrow\mathbb{R}^n$, $\mathcal{K}: \mathbb{R}_{\geq 0}\rightarrow\mathbb{R}^{n\times n}$, $\Lambda: \mathbb{R}_{\geq 0}\rightarrow\mathbb{R}^{n\times n}$, and $(\mathtt{q}_d,\dot{\mathtt{q}}_d)$ is the target trajectory of the state $(\mathtt{q},\dot{\mathtt{q}})$. Note that $\mathcal{K},\Lambda \succ 0$ are control gain matrices (design parameters), and $\dot{\mathcal{H}}-2\mathcal{C}$ is skew-symmetric with $\mathcal{H} \succ 0$ by construction.

By comparing with \eqref{lagrange_partial} and \eqref{lagrange_u}, we define the following virtual observer-like system of $q$ (not $\mathtt{q}$) that has $q = \dot{\mathtt{q}}$ and $q = \dot{\mathtt{q}}_r$ as its particular solutions:
\begin{align}
\label{lagrange_cl}
\mathcal{H}(\mathtt{q})\dot{q}+\mathcal{C}(\mathtt{q},\dot{\mathtt{q}}){q} +\mathcal{K}(t)({q}-\dot{\mathtt{q}})+\mathcal{G}(\mathtt{q}) =u(\mathtt{q},\dot{\mathtt{q}},t)
\end{align}
which gives $\mathcal{H}(\mathtt{q})\delta\dot{q}+(\mathcal{C}(\mathtt{q},\dot{\mathtt{q}})+\mathcal{K}(t))\delta q=0$. We thus have that
\begin{align}
\label{cont_lagrangian}
\dfrac{d}{dt} (\delta q^{\top} \mathcal{H}(\mathtt{q})\delta q) = \delta q^{\top} (\dot{\mathcal{H}}-2\mathcal{C}-2\mathcal{K})\delta q = -2\delta q^{\top}\mathcal{K}\delta q
\end{align}
where the skew-symmetry of $\dot{\mathcal{H}}-2\mathcal{C}$ is used to obtain the second equality. Since $\mathcal{K}\succ0$, \eqref{cont_lagrangian} indicates that the virtual system \eqref{lagrange_cl} is partially contracting in $q$ with $\mathcal{H}$ defining its contraction metric. Contraction of the full state $(\mathtt{q},\dot{\mathtt{q}})$ will be discussed in Example~\ref{EX:ex:hiera}. 

Note that if we treat the arguments $(\mathtt{q},\dot{\mathtt{q}})$ of $\mathcal{H}$ and $\mathcal{C}$ also as the virtual state $q$, we end up having additional terms such as $\partial\mathcal{H}/\partial{q_i}$, which makes proving contraction analytically more involved as in Example~\ref{EX:ex:partial_cont}.
\end{example}

As can be seen from Examples~\ref{EX:ex:partial_cont} and~\ref{EX:ex:lag_metric}, the role of partial contraction in theorem~\ref{THM:Thm:partial_contraction} is to provide some insight on stability even for cases where it is difficult to prove contraction for all solution trajectories as in Theorem~\ref{THM:Thm:contraction}. Although finding a contraction metric analytically for general nonlinear systems is challenging, we will see in Sec.~\ref{sec:HinfKYP} and Sec.~\ref{sec:convex} that the convex nature of the contraction condition \eqref{eq_MdotContracting} helps us find it numerically.
\subsection{Path-Length Integral and Robust Incremental Stability Analysis}\label{sec:robustpath}
Theorem~\ref{THM:Thm:contraction} can also be proven by using the transformed squared length integrated over two arbitrary solutions of \eqref{eq:xfx}~\cite{Ref:contraction1,Ref:phasesync,mypaperTAC,Ref:Stochastic,mypaper}, which enables formalizing its connection to incremental stability discussed in Definition~\ref{DEF:Def:contraction}. Note that the integral forms \eqref{eq_VSL} and \eqref{eq_VL} to be given in Theorem~\ref{THM:Thm:path_integral} are useful for handling perturbed systems with external disturbances as shall be seen in Theorems~\ref{THM:Thm:Robust_contraction_original}--\ref{THM:Thm:discrete_contraction}.
\begin{theorem}{Path Integral}{Thm:path_integral}
Let $\xi_0$ and $\xi_1$ be the two arbitrary solutions of \eqref{eq:xfx}, and define the transformed squared length with $M(x,t)$ of Theorem~\ref{THM:Thm:contraction} as follows:
\begin{equation}\label{eq_VSL}
V_{s\ell}\left(x,\delta x,t\right)=\int^{\xi_1}_{\xi_0} \|\delta z\|^2 =\int_{0}^{1}\frac{\partial x}{\partial\mu}^{\top}M\left(x,t\right)\frac{\partial x}{\partial\mu}d\mu
\end{equation}
where $x$ is a smooth path parameterized as $x(\mu=0,t)=\xi_0(t)$ and $x(\mu=1,t)=\xi_1(t)$ by $\mu \in [0,1]$. Also, define the path integral with the transformation $\Theta(x,t)$ for $M(x,t) = \Theta(x,t)^{\top}\Theta(x,t)$ as follows:
\begin{equation}
V_\ell(x,\delta x,t)=\int^{\xi_1}_{\xi_0} \|\delta {z}\|=\int^{\xi_1}_{\xi_0} \|\Theta(x,t)\delta {x}\|. \label{eq_VL}
\end{equation}
Then \eqref{eq_VSL} and \eqref{eq_VL} are related as
\begin{align}
\|\xi_1 -\xi_0\|=\left\|\int^{\xi_1}_{\xi_0} \delta {x}\right\|\leq \frac{V_{\ell}}{\sqrt{\underline{m}}} \leq \sqrt{\frac{V_{s\ell}}{\underline{m}}} \label{eq_VL_VSL}
\end{align}
where $M(x,t)\succeq\underline{m}\mathrm{I},~\forall x,t$ for $\exists\underline{m}\in\mathbb{R}_{>0}$, and Theorem~\ref{THM:Thm:contraction} can also be proven by using \eqref{eq_VSL} and \eqref{eq_VL} as a Lyapunov-like function, resulting in incremental exponential stability of the system \eqref{eq:xfx} (see Definition~\ref{DEF:Def:incremental}).
Note that the shortest path integral $V_\ell$ of \eqref{eq_VL} with a parameterized state $x$ (\ie{}, $\inf V_\ell=\sqrt{\inf V_{s\ell}}$) defines the Riemannian distance and the path integral of a minimizing geodesic~\cite{ccm}.
\end{theorem}
\begin{proof}
Using $M(x,t)\succeq\underline{m}\mathrm{I}$ which gives $\sqrt{\underline{m}}\|\xi_1 -\xi_0\|\leq V_{\ell}$, we have $\|\xi_1 -\xi_0\|=\|\int^{\xi_1}_{\xi_0} \delta {x}\|\leq {V_{\ell}}/{\sqrt{\underline{m}}}$ of \eqref{eq_VL_VSL}. The inequality $V_{\ell} \leq \sqrt{V_{s\ell}}$ of \eqref{eq_VL_VSL} can be proven by applying the Cauchy–Schwarz inequality~\cite[p. 316]{10.5555/2422911} to the functions $\psi_1(\mu)=\|\Theta(x,t)(\partial x/\partial \mu)\|$ and $\psi_2(\mu)=1$.

We can also see that computing $\dot{V}_{s\ell}$ of \eqref{eq_VSL} using the differential dynamics of \eqref{eq:original_differential} yields
\begin{align}
\dot{V}_{s\ell}=\int_{0}^{1}\frac{\partial x}{\partial\mu}^{\top}\left(\dot{M}+M\frac{\partial f}{\partial x}+\frac{\partial f}{\partial x}^{\top}M\right)\frac{\partial x}{\partial\mu}d\mu
\end{align}
to have $\dot{V}_{s\ell}\leq-2\alpha V_{s\ell}$ and $\dot{V}_{\ell}\leq -\alpha V_{\ell}$ by the contraction conditions \eqref{eq_MdotContracting_z} and \eqref{eq_MdotContracting}. Since these hold for any $\xi_0$ and $\xi_1$, the incremental exponential stability results in Theorem~\ref{THM:Thm:contraction} follow from the comparison lemma of Lemma~\ref{Lemma:comparison} (see also~\cite{mypaperTAC,Ref:Stochastic,mypaper}, and~\cite{ccm,7989693} for the discussion on the geodesic). \qed
\end{proof}
\subsubsection{Deterministic Perturbation}
Let $\xi_0(t)$ be a solution of the system \eqref{eq:xfx}. It is now perturbed as
\begin{equation}\label{Eq:fxt2-1}
\dot{{x}} = {f}({x},t) + {d}({x},t)
\end{equation}
and let $\xi_1(t)$ denote a trajectory of \eqref{Eq:fxt2-1}. Then a virtual system of a smooth path $q(\mu,t)$ parameterized by $\mu \in [0,1]$, which has $q(\mu=0,t)=\xi_0$ and $q(\mu=1,t)=\xi_1$ as its particular solutions is given as follows:
\begin{align}
\label{virtual_d}
\dot{q}(\mu,t) = f(q(\mu,t),t)+d_{\mu}(\mu,\xi_1,t)
\end{align}
where $d_{\mu}(\mu,\xi_1,t) = \mu d(\xi_1,t)$. 
Since contraction means exponential convergence, a contracting system exhibits a superior property of robustness~\cite{Ref:contraction1,Ref:phasesync}.
\begin{theorem}{Robust Deterministic Contraction}{Thm:Robust_contraction_original}
If the system~\eqref{eq:xfx} satisfies \eqref{eq_MdotContracting_z} and \eqref{eq_MdotContracting} of Theorem~\ref{THM:Thm:contraction} (\ie{}, the system \eqref{eq:xfx} is contracting), then the path integral $V_\ell(q,\delta q,t)=\int^{\xi_1}_{\xi_0}\|\Theta(q,t)\delta q\|$ of \eqref{eq_VL}, where $\xi_0$ is a solution of the contracting system \eqref{eq:xfx}, $\xi_1$ is a solution of the perturbed system \eqref{Eq:fxt2-1}, and $q$ is the virtual state of \eqref{virtual_d}, exponentially converges to a bounded error ball as long as ${\Theta}{d}\in \mathcal{L}_\infty$ (\ie{}, $\sup_{x,t}\|\Theta d\| < \infty$). Specifically, if $\exists \underline{m},\overline{m} \in \mathbb{R}_{>0}$ and $\exists \bar{d}\in\mathbb{R}_{\geq 0}$ \st{} $\bar{d}=\sup_{{x},t}\|
{d}(x,t)\|$ and
\begin{align}
\label{Mcon}
    \underline{m}\mathrm{I} \preceq M(x,t) \preceq \overline{m}\mathrm{I},~\forall x,t
\end{align}
then we have the following relation:
\begin{equation}\label{Eq:Robust_contraction21}
\|\xi_1(t)-\xi_0(t)\|\ \leq \frac{V_\ell(0)}{\sqrt{\underline{m}}}e^{-\alpha t}+\frac{\bar{d}}{\alpha}\sqrt{\frac{\overline{m}}{\underline{m}}}(1-e^{-\alpha t})
\end{equation}
where $V_\ell(t) = V_\ell(q(t),\delta q(t),t)$ for notational simplicity.
\end{theorem}
\begin{proof}
Using the contraction condition \eqref{eq_MdotContracting}, we have for $M=\Theta^{\top}\Theta$ given in Theorem~\ref{THM:Thm:contraction} that
\begin{align}
\dfrac{d}{dt}\|\Theta(q,t)\partial_{\mu} q\| &= (2\|\Theta(q,t)\partial_{\mu} q\|)^{-1}\dfrac{d}{dt}\partial_{\mu} q^{\top}M(q,t)\partial_{\mu} q \leq-\alpha\|\Theta(q,t)\partial_{\mu} q\|+\|\Theta(q,t)\partial_{\mu}d_{\mu}\|
\end{align}
where $\partial_{\mu} q = {\partial q}/{\partial \mu}$ and $\partial_{\mu} d_{\mu} = {\partial d_{\mu}}/{\partial \mu}=d(\xi_1,t)$. Taking the integral with respect to $\mu$ gives
\begin{align}
\dfrac{d}{dt}\int_{0}^1\|\Theta\partial_{\mu} q\|d\mu \leq \int_{0}^1-\alpha\|\Theta\partial_{\mu} q\|+\|\Theta d(\xi_1,t)\|d\mu
\end{align}
which implies $\dot{V}_{\ell}(t) \leq -\alpha V_{\ell}(t)+\sup_{q,\xi_1,t}{\|{\Theta}({q},t)d(\xi_1,t)\|}$ for $V_{\ell}$ in \eqref{eq_VL} of Theorem~\ref{THM:Thm:path_integral}. Thus, applying the comparison lemma (see Lemma~\ref{Lemma:comparison}) results in
\begin{align}
\label{Eq:Robust_contraction3prev}
    V_\ell(t) \leq e^{-\alpha t}V_\ell(0)+\sup_{q,\xi_1,t}{\|{\Theta}({q},t)d(\xi_1,t)\|}\frac{1-e^{-\alpha t}}{\alpha }.
\end{align}
By using $\xi_1 -\xi_0=\int^{\xi_1}_{\xi_0} \delta {x}$ and $\|\xi_1 -\xi_0\|=\|\int^{\xi_1}_{\xi_0} \delta {x}\| \leq \int^{\xi_1}_{\xi_0} \|\delta {x}\| \leq \int^{\xi_1}_{\xi_0} \|\Theta^{-1}\|\|\delta z\| $, we obtain $\sqrt{\inf_t\lambda_\mathrm{min}({M})}\|\xi_1 -\xi_0\|\leq V_\ell= \int^{\xi_1}_{\xi_0} \|\delta {z}\|$, and thus
\begin{align}
\|\xi_1 -\xi_0\|\ \leq \frac{e^{-\alpha t}V_\ell(0)}{\sqrt{\inf_t\lambda_\mathrm{min}({M})}}+\frac{\sup_{q,\xi_1,t}{\|{\Theta}{d}\|}}{\sqrt{\inf_t\lambda_\mathrm{min}({M})}}\frac{1-e^{-\alpha t}}{\alpha}.\nonumber
\end{align}
This relation with the bounds on $M$ and $d$ gives \eqref{Eq:Robust_contraction21}. \qed
\end{proof}
\subsubsection{Stochastic Perturbation}
Next, consider the following dynamical system modeled by the It\^{o} stochastic differential equation:
\begin{align}
\label{stochastic_dynamics}
dx = f(x,t)dt + G(x,t)d\mathscr{W}(t)
\end{align}
where $G:\mathbb{R}^n\times\mathbb{R}_{\geq 0} \rightarrow \mathbb{R}^{n\times w}$ is a matrix-valued function and $\mathscr{W}:\mathbb{R}_{\geq 0} \rightarrow \mathbb{R}^{w}$ is a $w$-dimensional Wiener process~\cite[p. 100]{arnold_SDE} (see also~\cite[p. xii]{arnold_SDE} for the notations used). For the sake of the existence and uniqueness of the solution, we assume in \eqref{stochastic_dynamics} that
\begin{align}
&\begin{aligned}[c]
\exists L \in\mathbb{R}_{\geq 0} \text{ \st{} } &\|f(x,t)-f(x',t)\| + \|G(x,t)-G(x',t)\|_F \\
&\leq L \|x - x'\|,~\forall t \in \mathbb{R}_{\geq 0},~\forall x,x' \in \mathbb{R}^n
\end{aligned}
\label{stocon1} \\
&\exists \bar{L} \in\mathbb{R}_{\geq 0} \text{ \st{} }
\|f(x,t)\|^2 + \|G(x,t)\|_F^2 \leq \bar{L}(1+\|x\|^2),~\forall t \in \mathbb{R}_{\geq 0},~\forall x \in \mathbb{R}^n.
\label{stocon2}
\end{align}
In order to analyze the incremental stability property of \eqref{stochastic_dynamics} as in Theorem~\ref{THM:Thm:path_integral}, we consider two trajectories $\xi_0(t)$ and $\xi_1(t)$ of stochastic nonlinear systems with Gaussian white noise, driven by two independent Wiener processes $\mathscr{W}_0(t)$ and $\mathscr{W}_1(t)$:
\begin{align}
\label{eq4}
d\xi_i &= f(\xi_i,t) dt + G_i(\xi_i,t) d\mathscr{W}_i(t),~i=0,1.
\end{align}
One can show that \eqref{stochastic_dynamics} has a unique solution $x(t)$ which is continuous with probability one under the conditions \eqref{stocon1} and \eqref{stocon2} (see~\cite[p. 105]{arnold_SDE} and \cite{Pham2009,Ref:Stochastic}), leading to the following lemma as in the comparison lemma of Lemma~\ref{Lemma:comparison}.
\begin{lemma}
\label{Lemma:comparison_sto}
Suppose that $V_{s\ell}$ of \eqref{eq_VSL} satisfies the following inequality:
\begin{align}
\label{cl_ivp_sto}
\mathscr{L}V_{s\ell} \leq -\gamma V_{s\ell}+c
\end{align}
where $\gamma \in \mathbb{R}_{>0}$, $c \in \mathbb{R}_{\geq0}$, and $\mathscr{L}$ denotes the infinitesimal differential generator of the It\^{o} process given in~\cite[p. 15]{sto_stability_book}. Then we have the following bound~\cite{Pham2009}:
\begin{align}
\label{bound_comparison_sto}
\mathop{\mathbb{E}}\left[\|\xi_1(t)-\xi_0(t)\|^2\right] \leq \frac{1}{\underline{m}}\left( \mathop{\mathbb{E}}[V_{s\ell}(0)]e^{-\gamma t}+\frac{c}{\gamma}\right)
\end{align}
where $V_{s\ell}(0) = V_{s\ell}(x(0),\delta x(0),0)$ for $V_{s\ell}$ in \eqref{eq_VSL}, $\underline{m}$ is given in \eqref{Mcon}, $\xi_{0}$ and $\xi_{1}$ are given in \eqref{eq4}, and $\mathop{\mathbb{E}}$ denotes the expected value operator. Furthermore, the probability that $\|\xi_1-\xi_0\|$ is greater than or equal to $\varepsilon\in\mathbb{R}_{> 0}$ is given as
\begin{align}
\label{bound_comparison_sto_prob}
\mathop{\mathbb{P}}\left[\|\xi_1(t)-\xi_0(t)\| \geq \varepsilon\right] \leq \frac{1}{\varepsilon^2\underline{m}}\left( \mathop{\mathbb{E}}[V_{s\ell}(0)]e^{-\gamma t}+\frac{c}{\gamma}\right).
\end{align}
\end{lemma}
\begin{proof}
The bound \eqref{bound_comparison_sto} follows from Theorem~2 of~\cite{Pham2009} (see also~\cite{mypaperTAC,Ref:Stochastic,old_observer,doi:10.1137/0305037,mypaper} and~\cite[p. 10]{sto_stability_book} (Dynkin's formula)). The probability tracking error bound \eqref{bound_comparison_sto_prob} then follows from Markov's inequality~\cite[pp. 311-312]{probbook}. \qed
\end{proof}
\begin{remark}
Although Lemma~\ref{Lemma:comparison_sto} considers the second moment of $\|\xi_1(t)-\xi_0(t)\|$, \ie{}, $\mathbb{E}[\|\xi_1(t)-\xi_0(t)\|^2]$, it can be readily generalized to the $p$-th moment of $\|\xi_1(t)-\xi_0(t)\|$, \ie{}, $\mathbb{E}[\|\xi_1(t)-\xi_0(t)\|^p]$, applying the Lyapunov-based technique proposed in~\cite{9304491}.
\end{remark}

A virtual system of a smooth path $q(\mu,t)$ parameterized by $\mu \in [0,1]$, which has $q(\mu=0,t)=\xi_0$ and $q(\mu=1,t)=\xi_1$ of \eqref{eq4} as its particular solutions, is given as follows:
\begin{align}
\label{virtual_d_sto}
dq(\mu,t) = f(q(\mu,t),t)dt+G(\mu,\xi_0,\xi_1,t)d\mathscr{W}(t)
\end{align}
where $G(\mu,\xi_0,\xi_1,t) = [(1-\mu) G_0(\xi_0,t),\mu G_1(\xi_1,t)]$ and $\mathscr{W} = [\mathscr{W}_0^{\top},\mathscr{W}_1^{\top}]^{\top}$. As a consequence of Lemma~\ref{Lemma:comparison_sto}, showing stochastic incremental stability between $\xi_0$ and $\xi_1$ of \eqref{eq4} reduces to proving the relation \eqref{cl_ivp_sto}, similar to the deterministic case in Theorems~\ref{THM:Thm:contraction} and~\ref{THM:Thm:Robust_contraction_original}.
\begin{theorem}{Robust Stochastic Contraction}{Thm:robuststochastic}
Suppose that $\exists \bar{g}_{0}\in\mathbb{R}_{\geq 0}$ and $\exists \bar{g}_{1}\in\mathbb{R}_{\geq 0}$ \st{} $\sup_{x,t}\|G_1(x,t)\|_F = \bar{g}_0$ and $\sup_{x,t}\|G_1(x,t)\|_F = \bar{g}_1$ in \eqref{eq4}. Suppose also that there exists $M(x,t) \succ 0,~\forall x,t$, \st{} $M_{x_i}=\partial M/\partial x_i$ is Lipschitz with respect to $x$ for all $i=1,\cdots,n$, \ie{}, $\exists L_m \in\mathbb{R}_{\geq 0}$ \st{}
\begin{align}
\label{eq:Mlipschitz}
\|M_{x_i}(x,t)-M_{x_i}(x',t)\|\leq L_m\|x-x'\|,~\forall x,x',t,i.
\end{align}
Also, suppose that $M$ of \eqref{eq:Mlipschitz} satisfies \eqref{Mcon} and \eqref{eq_MdotContracting} with its right-hand side replaced by $-2\alpha M-\alpha_s \mathrm{I}$ for $\alpha_s = L_m(\bar{g}_0^2+\bar{g}_1^2)(\alpha_G+{1}/{2})$, \ie{},
\begin{align}
\label{eq_MdotContracting_sto}
{\dot{M}}+M\frac{\partial
{f}}{\partial {x}}+\frac{\partial {f}}{\partial
{x}}^{\top}M \preceq - 2 \alpha M-\alpha_s\mathrm{I}
\end{align}
where $\alpha_G\in\mathbb{R}_{>0}$ is an arbitrary constant (see \eqref{Vsl_comp_0} for its details). Then, the following error bound of incremental stability holds:
\begin{align}
\label{Eq:boundnewsto}
\mathop{\mathbb{E}}\left[\|\xi_1(t)-\xi_0(t)\|^2\right] \leq \frac{\mathop{\mathbb{E}}[V_{s\ell}(0)]}{\underline{m}}e^{-2\alpha t}+\frac{C}{2\alpha}\frac{\overline{m}}{\underline{m}}
\end{align}
where $\xi_0$ and $\xi_1$ are the trajectories given in \eqref{eq4}, $V_{s\ell}(t)=V_{s\ell}(q(t),\delta q(t),t)=\int^{\xi_1}_{\xi_0}\delta q^{\top}M(q,t)\delta q$ is given in \eqref{eq_VSL} with the virtual state $q$ of \eqref{virtual_d_sto}, $\underline{m}$ and $\overline{m}$ are given in \eqref{Mcon}, $C = (\bar{g}_0^2+\bar{g}_1^2)({2}{\alpha_G}^{-1}+1)$, and $\mathop{\mathbb{E}}$ denotes the expected value operator. Furthermore, the probability that $\|\xi_1-\xi_0\|$ is greater than or equal to $\varepsilon\in\mathbb{R}_{> 0}$ is given as
\begin{align}
\label{Eq:boundnewsto_prob}
\mathop{\mathbb{P}}\left[\|\xi_1(t)-\xi_0(t)\|\geq\varepsilon\right] \leq \frac{1}{\varepsilon^2}\left(\frac{\mathop{\mathbb{E}}[V_{s\ell}(0)]}{\underline{m}}e^{-2\alpha t}+\frac{C}{2\alpha}\frac{\overline{m}}{\underline{m}}\right).~~~~~~
\end{align}
\end{theorem}
\begin{proof}
By definition of the infinitesimal differential generator given in Lemma~\ref{Lemma:comparison_sto}~\cite[p. 15]{sto_stability_book}, we have~\cite{mypaperTAC,mypaper}
\begin{align}
\mathscr{L}V_{s\ell} &=\int_{0}^{1} V_t+\sum^n_{i=1}\left(V_{q_i}f_i+V_{\partial_{\mu} q_i}\left(\frac{\partial f}{\partial q}\partial_{\mu} q\right)_i\right)+\frac{1}{2}\sum^n_{i,j=1}\Bigl(V_{q_iq_j}(GG^{\top})_{ij}+2V_{q_i\partial_{\mu} q_j}(G\partial_{\mu} G^{\top})_{ij} \nonumber \\
& +V_{\partial_{\mu} q_i\partial_{\mu} q_j} (\partial_{\mu} G\partial_{\mu} G^{\top})_{ij} \Bigr) d\mu\label{LV0}
\end{align}
where $V = \partial_{\mu} q^{\top}M(q,t)\partial_{\mu} q$, $\partial_{\mu} q=\partial q/\partial \mu$, $\partial_{\mu} G=\partial G/\partial \mu$, $V_{p} = \partial V/\partial p$, and $V_{p_1p_2}=\partial^2V/(\partial p_1\partial p_2)$.

Since $M_{x_i}$ is Lipschitz as in \eqref{eq:Mlipschitz}, we have $\|M_{x_i x_j}\| \leq L_m$ and $\left\|M_{x_i}\right\| \leq \sqrt{{2L_m}{\overline{m}}}$ using \eqref{Mcon} as derived~\cite{nscm}. Computing $\mathscr{L}V_{s\ell}$ of \eqref{LV0} using these bounds, the bounds of $\|G_{0}\|_F$ and $\|G_{1}\|_F$, and $\partial_{\mu} G=\partial G/\partial \mu = [-G_0,G_1]$ as in \cite{mypaperTAC,mypaper} yields
\begin{align}
\mathscr{L}V_{s\ell} &\leq \int_0^1\partial_{\mu} q^{\top}(\dot{M}+2\sym(Mf_{x}))\partial_{\mu} qd\mu +(\bar{g}_0^2+\bar{g}_1^2)(L_m\|\partial_{\mu} q\|^2/2+2\sqrt{{2L_m}{\overline{m}}}\|\partial_{\mu} q\|+\overline{m}) \nonumber \\ 
&\leq \int_0^1\partial_{\mu}q^{\top}(\dot{M}+2\sym(Mf_{x})+\alpha_s\mathrm{I})\partial_{\mu} qd\mu+{C}{\overline{m}}~~~~~~~~
\label{Vsl_comp_0}
\end{align}
where $\alpha_s = L_m(\bar{g}_0^2+\bar{g}_1^2)(\alpha_G+{1}/{2})$, $C = (\bar{g}_0^2+\bar{g}_1^2)({2}{\alpha_G}^{-1}+1)$, and the relation $2ab \leq \alpha_G^{-1}a^2+\alpha_G b^2$, which holds for any $a,b\in\mathbb{R}$ and $\alpha_G \in \mathbb{R}_{>0}$, is used with $a = \sqrt{2\overline{m}}$ and $b = \sqrt{L_m}\|\partial_{\mu} x\|$ to get the second inequality. This reduces to $\mathscr{L}V_{s\ell} \leq -2\alpha V_{s\ell}+\underline{m}C$ under the condition \eqref{eq_MdotContracting_sto}, resulting in \eqref{Eq:boundnewsto} and \eqref{Eq:boundnewsto_prob} as a result of 
\eqref{bound_comparison_sto} and \eqref{bound_comparison_sto_prob} in Lemma~\ref{Lemma:comparison_sto}. \qed
\end{proof}
\begin{remark}
Although we consider the Gaussian white noise stochastic differential equation \eqref{eq4} when referring to stochastic systems in this paper, other types of stochastic noises, including compound Poisson shot noise and bounded-measure L\'{e}vy noise, could be considered as in Theorem~\ref{THM:Thm:robuststochastic} using contraction theory~\cite{han2021incremental}.
\end{remark}
\subsection{Finite-Gain \texorpdfstring{$\mathcal{L}_p$}{$\mathcal{L}_p$} Stability and Hierarchical Contraction}
Due to the simpler stability analysis of Theorem~\ref{THM:Thm:contraction} when compared with Lyapunov theory, Input-to-State Stability (ISS) and input-output stability in the sense of finite-gain $\mathcal{L}_p$ stability can be easily studied using contraction theory~\cite{Ref:phasesync}.
\begin{theorem}{$\mathcal{L}_p$ Robust Contraction}{Thm:Robust_contraction}
If \eqref{Eq:fxt2-1} is perturbed by ${d}({x},t)\in \mathcal{L}_{pe}$ (\ie{}, $\|(d)_{\tau}\|_{\mathcal{L}_p} < \infty$ for $\tau\in\mathbb{R}_{\geq0}$ and $p\in[1,\infty]$, see Sec.~\ref{notation}) and Theorem~\ref{THM:Thm:contraction} holds, then \eqref{Eq:fxt2-1} is finite-gain $\mathcal{L}_p$ stable with $p\in [1,\infty]$ for an output ${y}={h}({x},{d},t)$ with $\int^{Y_1}_{Y_0}\|\delta {y}\|\leq \eta_0 \int^{\xi_1}_{\xi_0}\|\delta {x}\|+\eta_1\|{d}\|$, $\exists\eta_0,\eta_1\geq0$, \ie{}, $\forall \tau\in\mathbb{R}_{\geq 0}$~\cite{Ref:phasesync}
\begin{equation}\label{Eq:Robust_contractionLp}
\left\|{\left(\int^{Y_1}_{Y_0} \|\delta {y}\|\right)_\tau}\right\|_{\mathcal{L}_p} \leq \left(\frac{\eta_0}{\alpha}+{\eta_1}\right)\frac{\|({\Theta}{d})_\tau \|_{\mathcal{L}_p}}{\sqrt{\underline{m}}}+\frac{\eta_0\zeta V_\ell(0)}{\sqrt{\underline{m}}}
\end{equation}
where $Y_0$ and $Y_1$ denote the output trajectories of the original contracting system \eqref{eq:xfx} and its perturbed system \eqref{Eq:fxt2-1}, respectively, $\underline{m}$ is defined as ${M}({x},t)\succeq \underline{m}{\mathrm{I}}$, $\forall x,t$, as in \eqref{Mcon}, and $V_\ell(0) = V_\ell(x(0),\delta x(0),0)$ for $V_{\ell}$ in \eqref{eq_VL}. Also, $\zeta=1$ if $p=\infty$ and $\zeta=1/(\alpha p)^{1/p}$ if $p\in[1,\infty)$. The perturbed system \eqref{Eq:fxt2-1} also exhibits ISS.
\end{theorem}
\begin{proof}
In keeping with Theorem 5.1 of~\cite{Khalil:1173048}, \eqref{Eq:Robust_contraction3prev} also implies the following relation for $M=\Theta^{\top}\Theta$:
\begin{align}\label{Eq:Robust_contractionLp2}
\|(V_\ell)_\tau\|_{\mathcal{L}_p}&\leq V_\ell(0)\|(e^{-\alpha t})_\tau\|_{\mathcal{L}_p}  +\|e^{-\alpha t}\|_{\mathcal{L}_1} \|({\Theta}{d})_\tau \|_{\mathcal{L}_p} \leq V_\ell(0)\zeta +\|({\Theta}{d})_\tau \|_{\mathcal{L}_p}/\alpha.
\end{align}
Since we have $\|{\Theta}^{-1}\|\leq 1/\sqrt{\underline{m}}$ for $M=\Theta^{\top}\Theta$, \eqref{Eq:Robust_contractionLp} can be obtained by using both \eqref{Eq:Robust_contractionLp2} and the known bound of $\|\delta {y}\|$, thereby yielding a finite $\mathcal{L}_p$ gain independently of $\tau$. ISS can be guaranteed by Lemma~4.6 of~\cite[p. 176]{Khalil:1173048}, which states that exponential stability of an unperturbed system results in ISS. \qed
\end{proof}

Theorems~\ref{THM:Thm:contraction}--\ref{THM:Thm:Robust_contraction} are also applicable to the hierarchically combined system of two contracting dynamics due to Theorem~\ref{THM:Thm:Robust_contraction_hierc}~\cite{Ref:contraction2,Ref:contraction5}.
\begin{theorem}{Robust Hierarchical Contraction}{Thm:Robust_contraction_hierc}
Consider the following hierarchically combined system of two contracting dynamics:
\begin{align}
\label{deltaz_hiera}
\dfrac{d}{dt}\begin{bmatrix}\delta z_0\\\delta z_1\end{bmatrix}=\begin{bmatrix}{F}_{00} & {0}\\ {F}_{10} & {F}_{11}\end{bmatrix}\begin{bmatrix}\delta z_0\\\delta z_1\end{bmatrix}
\end{align}
where $F=F_{00}$ and $F=F_{11}$ both satisfy the contraction condition \eqref{eq_MdotContracting_z}, and suppose that it is subject to perturbation $[{d}_0,{d}_1]^{\top}$. Then the path length integral $V_{\ell,i}(t)=\int^{\xi_1}_{\xi_0}\|\delta {z}_i\|$ with $i=0,1$ between the original and perturbed dynamics trajectories, $\xi_0$ and $\xi_1$, respectively, verifies~\cite{Ref:contraction1}
\begin{align}\label{Eq:hierclemma}
\dot{V}_{\ell,0}+\alpha_0V_{\ell,0}&\leq\|{\Theta}_0{d}_0\| \\
\dot{V}_{\ell,1}+\alpha_1V_{\ell,1}&\leq\|{\Theta}_1{d}_1\|+\int^{\xi_1}_{\xi_0}\|{F}_{10}\|\|\delta {z}_1\| \label{Eq:hierclemmab}
\end{align}
where $\alpha_i=\sup_{x,t}|\lambda_\mathrm{max}({F}_{ii})|,~i=0,1$. Hence, the error bounds of $V_{\ell,0}(t)$ and $V_{\ell,1}(t)$ can be obtained using Theorem~\ref{THM:Thm:Robust_contraction_original} if $\|{F}_{10}\|$ is bounded. In particular, if $\|{\Theta}_i d_i\|\leq \sqrt{\overline{m}_i}\bar{d}_i$, $i=0,1$, the relations \eqref{Eq:hierclemma} and \eqref{Eq:hierclemmab} yield $V_{\ell,0}(t) \leq V_{\ell,0}(0)e^{-\alpha_0t}+(\sqrt{\overline{m}_0}\bar{d}_0/{\alpha_0})(1-e^{-\alpha_0t})$ as in \eqref{Eq:Robust_contraction21}, and
\begin{align}
\label{robust_hiera_bound}
V_{\ell,1}(t) &\leq V_{\ell,1}(0)e^{-\alpha_1t}+\frac{\sqrt{\overline{m}_1}\bar{d}_1+\bar{f}_{10}\bar{V}_{0}}{\alpha_1}(1-e^{-\alpha_1t})~~~~
\end{align}
where $\bar{f}_{10}=\sup_{t\in\mathbb{R}_{\geq 0}}\|{F}_{10}\|$ and $\bar{V}_{0}=V_{\ell,0}(0)+\sqrt{\overline{m}_0}\bar{d}_0/{\alpha_0}$.

Also, similar to Theorem~\ref{THM:Thm:Robust_contraction}, we have $\|(V_{\ell,0})_\tau\|_{\mathcal{L}_p} \leq V_{\ell,0}(0)\zeta_0+{\|({\Theta}_0{d}_0)_\tau \|_{\mathcal{L}_p}}/{\alpha_0}$, and thus a hierarchical connection for finite-gain $\mathcal{L}_p$ stability can be established as follows~\cite{Ref:phasesync}:
\begin{align}
\|(V_{\ell,1})_\tau\|_{\mathcal{L}_p} &\leq V_{\ell,1}(0)\zeta_1 +\frac{\|({\Theta}_1{d}_1)_\tau \|_{\mathcal{L}_p}+\bar{f}_{10}\|(V_{\ell,0})_\tau\|_{\mathcal{L}_p}}{\alpha_1}~~~~\label{Eq:hierclemma2}
\end{align}
where $\zeta_i=1$ if $p=\infty$ and $\zeta_i=1/(\alpha_i p)^{1/p}$ if $p\in[1,\infty)$ for $i=0,1$. By recursion, this result can be extended to an arbitrary number of hierarchically combined groups.
\end{theorem}
\begin{proof}
Applying the comparison lemma of Lemma~\ref{Lemma:comparison} to \eqref{Eq:hierclemma} and \eqref{Eq:hierclemmab}, which follow from the differenital dynamics \eqref{deltaz_hiera} with the condition \eqref{eq_MdotContracting_z}, we get \eqref{robust_hiera_bound} as in Theorem~\ref{THM:Thm:Robust_contraction_original}. Similarly, \eqref{Eq:hierclemma2} follows from obtaining $\|(V_{\ell,0})_\tau\|_{\mathcal{L}_p}$ by \eqref{Eq:Robust_contractionLp2} using \eqref{Eq:hierclemma}, and then recursively obtaining $\|(V_{\ell,1})_\tau\|_{\mathcal{L}_p}$ using \eqref{Eq:hierclemmab} as in Theorem~\ref{THM:Thm:Robust_contraction}~\cite{Ref:phasesync}. \qed
\end{proof}
\begin{example}{Contraction in Lagrangian Systems}{ex:hiera}
As demonstrated in Example~\ref{EX:ex:lag_metric}, the Lagrangian virtual system \eqref{lagrange_cl} is contracting with respect to $q$, having $q=\dot{\mathtt{q}}$ and $q = \dot{\mathtt{q}}_r = \dot{\mathtt{q}}_d-\Lambda (\mathtt{q}-\mathtt{q}_d)$ as its particular solutions. Let $q_0=q$ for such $q$. The virtual system of $q_1$ which has $[q_0^{\top},q_1^{\top}]^{\top}=[\dot{\mathtt{q}}^{\top},\mathtt{q}^{\top}]^{\top}$ and $[\dot{\mathtt{q}}_r^{\top},\mathtt{q}_d^{\top}]^{\top}$ as its particular solutions is given as
\begin{align}
\dot{q}_1 = q_0-\Lambda(q_1-\mathtt{q})
\end{align}
resulting in $\delta \dot{q}_1=\mathrm{I}\delta q_0-\Lambda\delta q_1$. Since the virtual system of $q_0$ is contracting as in Example~\ref{EX:ex:lag_metric} and the virtual system $\delta \dot{q}_1=-\Lambda\delta q_1$ is contracting in $q_1$ due to $\Lambda \succ 0$, \eqref{robust_hiera_bound} of Theorem~\ref{THM:Thm:Robust_contraction_hierc} implies that the whole system \eqref{lagrange_partial} for $[q_0^{\top},q_1^{\top}]^{\top}$ is hierarchically contracting and robust against perturbation in the sense of Theorem~\ref{THM:Thm:Robust_contraction_hierc} ($F_{10}=\mathrm{I}$ in this case). Also, see~\cite{Ref:ChungTRO,Ref:phasesync} for the hierarchical multi-timescale separation of tracking and synchronization control for multiple Lagrangian systems.
\end{example}
\subsection{Contraction Theory for Discrete-time Systems}
Let us consider the following nonlinear system with bounded deterministic perturbation $d_k:\mathbb{R}^n\times\mathbb{N}\rightarrow\mathbb{R}^n$ with $\bar{d}\in\mathbb{R}_{\geq 0}$ \st{} $\bar{d}=\sup_{{x},k}\|
{d}_k(x,k)\|$:
\begin{align}
\label{discrete_system}
x(k+1) = f_k(x(k),k)+{d}_k(x(k),k)
\end{align}
where $k\in\mathbb{N}$, $x:\mathbb{N}\rightarrow\mathbb{R}^n$ is the discrete system state, and $f_k:\mathbb{R}^n\times\mathbb{N}\rightarrow\mathbb{R}^n$ is a smooth function. Although this tutorial paper focuses mainly on continuous-time nonlinear systems, let us briefly discuss contraction theory for \eqref{discrete_system} to imply that the techniques in the subsequent sections are applicable also to discrete-time nonlinear systems.

Let $\xi_{0}(k)$ and $\xi_{1}(k)$ be solution trajectories of \eqref{discrete_system} with $d_k=0$ and $d_k\neq 0$, respectively. Then a virtual system of $q(\mu,k)$ parameterized by $\mu\in[0,1]$, which has $q(\mu=0,k)=\xi_{0}(k)$ and $q_k(\mu=1,k)=\xi_{1}(k)$ as its particular solutions, can be expressed as follows:
\begin{align}
\label{virtual_system_discrete}
q(\mu,k+1) = f_k(q(\mu,k),k)+\mu d_{k}(\xi_1(k),k).
\end{align}
The discrete version of robust contraction in Theorem~\ref{THM:Thm:Robust_contraction_original} is given in the following theorem.
\begin{theorem}{Discrete-Time Contraction}{Thm:discrete_contraction}
Let ${x}(k)=x_k$ and $q(\mu,k)=q_k$ for any $k\in\mathbb{N}$ for simplicity. If there exists a uniformly positive definite matrix ${M}_k({x}_k,k)={{\Theta_k}}({x}_k,k)^{\top}{{\Theta_k}}({x}_k,k) \succ 0,~\forall x_k,k$, where ${\Theta_k}$ defines a smooth coordinate transformation of $\delta x_k$, \ie{}, $\delta{z}_k={\Theta_k}(x_k,k)\delta{x}_k$, \st{} either of the following equivalent conditions holds for $\exists \alpha \in (0,1)$, $\forall x_k,k$:
\begin{align}
&\left\|{\Theta_{k+1}}({x}_{k+1},k+1)\frac{\partial
{f}_k}{\partial
{x}_k}{{\Theta_k}({x}_k,k)^{-1}}\right\| \leq \alpha
\label{eq_MdotContracting_discrete_z} \\
&\frac{\partial
{f}_k}{\partial
{x}_k}^{\top}{{M_{k+1}}({x}_{k+1},k+1)}\frac{\partial
{f}_k}{\partial
{x}_k} \preceq \alpha^2 {M_k}({x}_{k},k)
\label{eq_MdotContracting_discrete}
\end{align}
then we have the following bound as long as we have $\underline{m}\mathrm{I}\preceq M_x(x_k,k)\preceq \overline{m}\mathrm{I},~\forall x_k,k$, as in \eqref{Mcon}:
\begin{align}
\label{discrete_bound_disturbance}
\|\xi_{1}(k)-\xi_{0}(k)\| \leq \frac{V_{\ell}(0)}{\sqrt{\underline{m}}}\alpha^k+\frac{\bar{d}(1-\alpha^k)}{1-\alpha}\sqrt{\frac{\overline{m}}{\underline{m}}}
\end{align}
where $V_{\ell}(k)=\int_{\xi_{0}}^{\xi_{1}}\|\Theta_k(q_k,k)\delta q_k\|$ as in \eqref{eq_VL} for the unperturbed trajectory $\xi_{0}$, perturbed trajectory $\xi_1$, and virtual state $q_k=q(k)$ given in \eqref{virtual_system_discrete}.
\end{theorem}
\begin{proof}
If \eqref{eq_MdotContracting_discrete_z} or \eqref{eq_MdotContracting_discrete} holds, we have that
\begin{align}
V_{\ell}(k+1) &\leq \int^1_0\|\Theta_{k+1}(\partial_{q_k} f_k(q_k,k)\partial_{\mu} q_k+d_k(x_k,k))\|d\mu \\
&\leq \alpha\int^1_0\|\Theta_{k}(q_k,k)\partial_{\mu} q_k\|d\mu+\bar{d}\sqrt{\overline{m}} = \alpha V_{\ell}(k)+\bar{d}\sqrt{\overline{m}} \nonumber
\end{align}
where $\Theta_{k+1}=\Theta_{k+1}(q_{k+1},k+1)$, $\partial_{q_k} f_k(q_k,k)=\partial f_k/\partial q_k$, and $\partial_{\mu} q_k=\partial q_k/\partial \mu$. Applying this inequality iteratively results in \eqref{discrete_bound_disturbance}.
\end{proof}

Theorem~\ref{THM:Thm:discrete_contraction} can be used with Theorem~\ref{THM:Thm:Robust_contraction_original} for stability analysis of hybrid nonlinear systems~\cite{Ref:contraction_robot,Ref:contraction4,4795665}, or with Theorem~\ref{THM:Thm:robuststochastic} for stability analysis of discrete-time stochastic nonlinear systems~\cite{mypaperTAC,mypaper,4795665}. For example, it is shown in~\cite{mypaperTAC} that if the time interval in discretizing \eqref{eq:xfx} as \eqref{discrete_system} is sufficiently small, contraction of discrete-time systems with stochastic perturbation reduces to that of continuous-time systems. 

We finally remark that the steady-state upper bounds of \eqref{Eq:Robust_contraction21} in Theorem~\ref{THM:Thm:Robust_contraction_original}, \eqref{Eq:boundnewsto} in Theorem~\ref{THM:Thm:robuststochastic}, and \eqref{discrete_bound_disturbance} in Theorem~\ref{THM:Thm:discrete_contraction} are all functions of $\overline{m}/\underline{m}$. This property is to be used extensively in Sec.~\ref{sec:convex} for designing a convex optimization-based control and estimation synthesis algorithm via contraction theory.
\newpage
\section{Robust Nonlinear Control and Estimation}
\label{sec:HinfKYP}
As shown in Theorem~\ref{THM:Thm:Robust_contraction_original} for deterministic disturbance and in Theorem~\ref{THM:Thm:robuststochastic} for stochastic disturbance, contraction theory provides explicit bounds on the distance of any couple of perturbed system trajectories. This property is useful in designing robust and optimal feedback controllers for a nonlinear system such as $\mathcal{H}_{\infty}$ control~\cite{Basar1995,256331,159566,29425,85062,151101,599969,1259458,doi:10.1137/S0363012996301336,doi:10.1137/S0363012903423727,rccm}, which attempts to minimize the system $\mathcal{L}_2$ gain for optimal disturbance attenuation.

Since most of such feedback control and estimation schemes are based on the assumption that we know a Lyapunov function candidate, this section delineates one approach to solve a nonlinear optimal feedback control problem via contraction theory~\cite{mypaperTAC,mypaper}, thereby proposing one explicit way to construct a Lyapunov function and contraction metric for general nonlinear systems for the sake of robustness. This approach is also utilizable for optimal state estimation problems as shall be seen in Sec.~\ref{sec:convex}.

We consider the following smooth nonlinear system, perturbed by bounded deterministic disturbances $d_c(x,t)$ with $\sup_{x,t}\|d_c(x,t)\|=\bar{d}_c\in\mathbb{R}_{\geq 0}$ or by Gaussian white noise $\mathscr{W}(t)$ with $\sup_{x,t}\|G_c(x,t)\|_F = \bar{g}_c\in\mathbb{R}_{\geq 0}$:
\begin{align}
\label{original_dynamics}
\dot{x} &= f(x,t) + B(x,t)u + d_c(x,t) \\
\label{sdc_dynamics}
dx &= (f(x,t)+B(x,t)u)dt+G_c(x,t)d\mathscr{W}(t) \\
\label{sdc_dynamicsd}
\dot{x}_d &= f(x_d,t)+B(x_d,t)u_d
\end{align}
where ${x}:\mathbb{R}_{\geq 0}\mapsto\mathbb{R}^n$ is the system state, $u \in \mathbb{R}^m$ is the system control input, ${f}:
\mathbb{R}^n\times\mathbb{R}_{\geq 0}\mapsto\mathbb{R}^n$ and $B:\mathbb{R}^n\times\mathbb{R}_{\geq 0}\mapsto\mathbb{R}^{n\times m}$ are known smooth functions, $d_c:\mathbb{R}^n\times\mathbb{R}_{\geq 0} \rightarrow\mathbb{R}^{n}$ and $G_c:\mathbb{R}^n\times\mathbb{R}_{\geq 0} \rightarrow\mathbb{R}^{n\times w}$ are unknown bounded functions for external disturbances, and $\mathscr{W}:\mathbb{R}_{\geq 0} \rightarrow\mathbb{R}^{w}$ is a $w$-dimensional Wiener process. Also, for \eqref{sdc_dynamicsd}, $x_d:\mathbb{R}_{\geq 0} \rightarrow\mathbb{R}^{n}$ and $u_d:\mathbb{R}_{\geq 0} \rightarrow\mathbb{R}^{m}$ denote the desired target state and control input trajectories, respectively. 
\begin{remark}
We consider control-affine nonlinear systems \eqref{original_dynamics}--\eqref{sdc_dynamicsd} in Sec.~\ref{sec:HinfKYP}, \ref{sec:convex}, and~\ref{Sec:ncm}--\ref{Sec:adaptive}. This is primarily because the controller design techniques for control-affine nonlinear systems are less complicated than those for control non-affine systems (which often result in $u$ given implicitly by $u = k(x,u,t)$~\cite{vccm,WANG201944}), but still utilizable even for the latter, \eg{}, by treating $\dot{u}$ as another control input (see Example~\ref{EX:ex:nonaffine1}), or by solving the implicit equation $u = k(x,u,t)$ iteratively with a discrete-time controller (see Example~\ref{EX:ex:nonaffine2} and Remark~\ref{remark_nonaffine}).
\end{remark}
\begin{example}{Control Non-Affine Systems I}{ex:nonaffine1}
By using $\dot{u}$ instead of $u$ in \eqref{original_dynamics} and \eqref{sdc_dynamics}, a control non-affine system, $\dot{x}=f(x,u,t)$, can be rewritten as
\begin{align}
\dfrac{d}{dt}\begin{bmatrix}x\\u\end{bmatrix}=\begin{bmatrix}f(x,u,t)\\0\end{bmatrix}+\begin{bmatrix}0\\\mathrm{I}\end{bmatrix}\dot{u}
\end{align}
which can be viewed as a control-affine nonlinear system with the state $[x^{\top},u^{\top}]^{\top}$ and control $\dot{u}$. 
\end{example}
\begin{example}{Control Non-Affine Systems II}{ex:nonaffine2}
One drawback of the technique in Example~\ref{EX:ex:nonaffine1} is that we have to control $\dot{u}$ instead of $u$, which could be difficult in practice. In this case, we can utilize the following control non-affine nonlinear system decomposed into control-affine and non-affine parts:
\begin{align}
\dot{x}=f(x,u,t)=f_a(x,t)+B_a(x,t)u+r(x,u,t)
\end{align}
where $r(x,u,t)=f(x,u,t)-f_a(x,t)-B_a(x,t)u$. The controller $u$ can now be designed implicitly as
\begin{align}
\label{ex_implicit_u}
B_a(x,t)u = B_a(x,t)u^*-r(x,u,t)
\end{align}
where $u^*$ is a stabilizing controller for the control-affine system $\dot{x}=f_a(x,t)+B_a(x,t)u^*$. Since solving such an implicit equation in \eqref{ex_implicit_u} in real-time could be unrealistic in practice, we will derive a learning-based approach to solve it iteratively for unknown $r(x,u,t)$, without deteriorating its stability performance (see Lemma~\ref{lemma:discrete_neurallander} and Theorem~\ref{THM:Thm:neurallander} of Sec.~\ref{Sec:datadriven}).
\end{example}
\subsection{Overview of Nonlinear Control and Estimation}
\label{Sec:overview}
We briefly summarize the advantages and disadvantages of existing nonlinear feedback control and state estimation schemes, so that one can identify which strategy is appropriate for their study and refer to the relevant parts of this tutorial paper.
\begin{table}[tb]
\caption{Comparison between SDC and CCM Formulation (note that $\gamma(\mu=0,t)=x_d$ and $\gamma(\mu=1,t)=x$). \label{tab:sdcccm_summary}}
\footnotesize
\begin{center}
\renewcommand{\arraystretch}{1.2}
\rowcolors{1}{uiucbluedark!5}{uiucbluedark!10}
\begin{tabular}{ l m{6.1cm} m{6.1cm} } 
 \hline\hline
  & SDC formulation (Theorem~\ref{THM:Thm:CV-STEM})~\cite{mypaperTAC,ncm,nscm,Ref:Stochastic,mypaper} & CCM formulation (Theorem~\ref{THM:Thm:ccm_cvstem})~\cite{ccm,7989693} \\
 \hline
 Control law & $u = u_d-K(x,x_d,u_d,t)(x-x_d)$ or $u_d-K(x,t)(x-x_d)$ & $u = u_d+\int_{0}^1k(\gamma(\mu,t),\partial_{\mu}\gamma(\mu,t),u,t)d\mu$  \\
 \arrayrulecolor{mygray}\hline
 Computation & Evaluates $K$ for given $(x,x_d,u_d,t)$ as in LTV systems & Computes geodesics $\gamma$ for given $(x,x_d,t)$ and integrates $k$ \\
 \hline
 Generality & Captures nonlinearity by (multiple) SDC matrices & Handles general differential dynamics \\
 \hline
 Contraction & Depends on $(x,x_d,u_d,t)$ or $(x,t)$ (partial contraction) & Depends on $(x,t)$ (contraction) \\
 \arrayrulecolor{black}\hline\hline
\end{tabular}
\end{center}
\end{table}
\renewcommand{\arraystretch}{1.0}
\subsubsection{Systems with Known Lyapunov Functions}
As discussed in Sec.~\ref{sec:contraction_search}, there are several nonlinear systems equipped with a known contraction metric/Lyapunov function, such as Lagrangian systems~\cite[p. 392]{Ref_Slotine}, whose inertia matrix $\mathcal{H}(\mathtt{q})$ defines its contraction metric (see Example~\ref{EX:ex:lag_metric}), or the nonlinear SLAM problem~\cite{doi:10.1177/0278364917710541,Ref:Stochastic} with virtual synthetic measurements, which can be reduced to an LTV estimation problem~\cite{doi:10.1177/0278364917710541}. Once we have a contraction metric/Lyapunov function, stabilizing control and estimation laws can be easily derived by using, \eg{},~\cite{doi:10.1137/0321028,SONTAG1989117,Khalil:1173048}. Thus, those dealing primarily with such nonlinear systems should skip this section and proceed to \hyperref[part2LBC]{Part~II} of this paper (Sec.~\ref{Sec:learning_stability}--\ref{Sec:datadriven}) on learning-based and data-driven control using contraction theory. Note that these known contraction metrics are not necessarily optimal, and the techniques to be derived in Sec.~\ref{sec:HinfKYP} and Sec.~\ref{sec:convex} are for obtaining contraction metrics with an optimal disturbance attenuation property~\cite{mypaperTAC,mypaper}.
\subsubsection{Linearization of Nonlinear Systems}
\label{Sec:overviewlinear}
If a contraction metric of a given nonlinear system is unknown, we could linearize it to apply methodologies inspired by LTV systems theory such as $\mathcal{H}_{\infty}$ control~\cite{29425,85062,151101,599969,doi:10.1137/S0363012996301336,1259458}, iterative Linear Quadratic Regulator (iLQR)~\cite{ilqr,silqr}, or Extended Kalman Filter (EKF). Their stability is typically analyzed by decomposing $f(x,t)$ as $f(x,t) = Ax+(f(x,t)-Ax)$ assuming that the nonlinear part $f(x,t)-Ax$ is bounded, or by finding a local contraction region for the sake of local
exponential stability as in~\cite{6849943,ncm}. Since the decomposition $f(x,t) = Ax+(f(x,t)-Ax)$ allows applying the result of Theorem~\ref{THM:Thm:Robust_contraction_original}, we could exploit the techniques in Sec.~\ref{sec:HinfKYP} and Sec.~\ref{sec:convex} for providing formal robustness and optimality guarantees for the LTV systems-type approaches. For systems whose nonlinear part $f(x,t)-Ax$ is not necessarily bounded, Sec.~\ref{Sec:neurallander} elucidates how contraction theory can be used to stabilize them with the learned residual dynamics for control synthesis. 
\subsubsection{State-Dependent Coefficient (SDC) Formulation}
\label{Sec:overviewsdc}
It is shown in~\cite{mypaperTAC,Ref:Stochastic,nscm,ncm,mypaper} that the SDC-based control and estimation~\cite{sddre,Banks2007,chang2009exponential,survey_SDRE}, which capture nonlinearity using a state-dependent matrix $A(x,t)$ \st{} $f(x,t)=A(x,t)x$ (\eg{}, we have $A(x,t)=\cos x$ for $f(x,t) = x\cos x$), result in exponential boundedness of system trajectories both for deterministic and stochastic systems due to Theorems~\ref{THM:Thm:Robust_contraction_original} and \ref{THM:Thm:robuststochastic}~\cite{ncm}. Because of the extended linear form of SDC (see Table~\ref{tab:sdcccm_summary}), the results to be presented in Sec.~\ref{sec:HinfKYP}--\ref{sec:convex} based on the SDC formulation are applicable to linearized dynamics that can be viewed as an LTV system with some modifications (see Remark~\ref{remark_other_forms}).

This idea is slightly generalized in~\cite{nscm} to explicitly consider incremental stability with respect to a target trajectory (\eg{}, $x_d$ for control and $x$ for estimation) instead of using $A(x,t)x=f(x,t)$. Let us derive the following lemma for this purpose~\cite{nscm,survey_SDRE,mypaperTAC,Ref:Stochastic,mypaper}.
\begin{lemma}
\label{sdclemma}
Let $f:\mathbb{R}^n\times\mathbb{R}_{\geq0}\mapsto\mathbb{R}^n$ and $B:\mathbb{R}^n\times\mathbb{R}_{\geq0}\mapsto\mathbb{R}^{n\times m}$ be piecewise continuously differentiable functions. Then there exists a matrix-valued function $A:\mathbb{R}^n\times\mathbb{R}^n\times\mathbb{R}^m\times\mathbb{R}_{\geq0}\mapsto\mathbb{R}^{n\times n}$ \st{}, $\forall {s}\in\mathbb{R}^n$, $\bar{s}\in\mathbb{R}^n$, $\bar{u}\in\mathbb{R}^m$, and $t\in\mathbb{R}_{\geq0}$,
\begin{align}
A({s},\bar{s},\bar{u},t)\mathtt{e}&=f({s},t)+B({s},t)\bar{u}-f(\bar{s},t)-B(\bar{s},t)\bar{u}
\end{align}
where $\mathtt{e} = {s}-\bar{s}$, and one such $A$ is given as follows:
\begin{align}
\label{sdcAc}
A({s},\bar{s},\bar{u},t) &= \int_0^1\frac{\partial \bar{f}}{\partial {s}}(c {s}+(1-c)\bar{s},\bar{u},t)dc
\end{align}
where $\bar{f}({s},\bar{u},t)=f({s},t)+B({s},t)\bar{u}$. We call $A$ an SDC matrix if it is constructed to satisfy the controllability (or observability for estimation) condition. Also, the choice of $A$ is not unique for $n \geq 2$, where $n$ is the number of states, and the convex combination of such non-unique SDC matrices also verifies extended linearization as follows:
\begin{align}
\begin{aligned}
&f({s},t)+B({s},t)\bar{u}X-f(\bar{s},t)-B(\bar{s},t)\bar{u} \\&=A(\varrho,{s},\bar{s},\bar{u},t)({s}-\bar{s})=\sum_{i=1}^{s_A}\varrho_iA_i({s},\bar{s},\bar{u},t)({s}-\bar{s})
\end{aligned}
\label{varrho_def}
\end{align}
where $\varrho=(\varrho_1,\cdots,\varrho_{s_A})$, $\sum_{i=1}^{s_A}\varrho_i=1,~\varrho_i\geq 0$, and each $A_i$ satisfies the relation $\bar{f}({s},\bar{u},t)-\bar{f}(\bar{s},\bar{u},t)=A_i({s},\bar{s},\bar{u},t)({s}-\bar{s})$.
\end{lemma}
\begin{proof}
The first statement on \eqref{sdcAc} follows from the integral relation given as
\begin{align}
\int_0^1\frac{d\bar{f}}{dc}(c {s}+(1-c)\bar{s},\bar{u},t)dc=\bar{f}({s},\bar{u},t)-\bar{f}(\bar{s},\bar{u},t) .
\end{align}
If there are multiple SDC matrices $A_i$, we clearly have $\varrho_iA_i({s},\bar{s},\bar{u},t)({s}-\bar{s}) = \varrho_i(\bar{f}({s},\bar{u},t)-\bar{f}(\bar{s},\bar{u},t)),~\forall i$, and therefore, the relation $\sum_{i=1}^{s_A}\varrho_i=1,~\varrho_i\geq 0$ gives \eqref{varrho_def}.
\qed
\end{proof}
\begin{example}{SDC Formulation}{}
Let us illustrate how Lemma~\ref{sdclemma} can be used in practice, taking the following nonlinear system as an example:
\begin{align}
\label{ex_dynamics}
\dot{x} = [x_2, -x_1x_2]^{\top}+[0,\cos x_1]^{\top}u
\end{align}
where $x = [x_1,x_2]^{\top}$. If we use $({s},\bar{s},\bar{u})=(x,x_d,u_d)$ in Lemma~\ref{sdclemma} for a given target trajectory $(x_d,u_d)$ that satisfies \eqref{ex_dynamics}, evaluating the integral of \eqref{sdcAc} gives
\begin{align}
\label{Aexample}
A_1(x,x_d,u_d,t) = -\begin{bmatrix}0 & 1\\\frac{x_2+x_{2d}}{2}-\frac{u_d(\cos x_1-\cos x_{d1})}{x_1-x_{d1}}&\frac{x_1+x_{1d}}{2}\end{bmatrix}~~~~~~~~
\end{align}
due to the relation $\partial \bar{f}/\partial {s} = \left[\begin{smallmatrix}0 & 1\\ -{s}_2 & -{s}_1\end{smallmatrix}\right]+\left[\begin{smallmatrix}0 & 0\\ -u_d\sin {s}_1 & 0\end{smallmatrix}\right]$ for $\bar{f}({s},u_d,t)=f({s},t)+B({s},t)u_d$, where $x_d = [x_{1d},x_{2d}]^{\top}$. Note that we have 
\begin{align}
\frac{(\cos x_1-\cos x_{d1})}{x_1-x_{d1}} = -\sin\left(\frac{x_1+x_{1d}}{2}\right)\sinc\left(\frac{x_1-x_{1d}}{2}\right)
\end{align}
and thus $A(x,x_d,u_d,t)$ is defined for all $x$, $x_d$, $u_d$, and $t$. The SDC matrix \eqref{Aexample} indeed verifies $A_1(x,x_d,u_d,t)(x-x_d)=\bar{f}(x,t)-\bar{f}(x_d,t)$.

We can see that the following is also an SDC matrix of the nonlinear system \eqref{ex_dynamics}:
\begin{align}
\label{Aexample2}
A_2(x,x_d,u_d,t) = -\begin{bmatrix}0 & 1\\x_2-\frac{u_d(\cos x_1-\cos x_{d1})}{x_1-x_{d1}}&x_{1d}\end{bmatrix}.
\end{align}
Therefore, the convex combination of $A_1$ in \eqref{Aexample} and $A_2$ in \eqref{Aexample2}, $A = \varrho_1A_1+\varrho_2A_2$ with $\varrho_1+\varrho_2=1$, $\varrho_1,\varrho_2 \geq 0$, is also an SDC matrix due to Lemma~\ref{sdclemma}. 
\end{example}

The major advantage of the formalism in Lemma~\ref{sdclemma} lies in its systematic connection to LTV systems based on uniform controllability and observability, adequately accounting for the nonlinear nature of underlying dynamics through $A(\varrho,x,x_d,u_d,t)$ for global stability, as shall be seen in Sec.~\ref{sec:HinfKYP} and Sec.~\ref{sec:convex}. Since $A$ depends also on $(x_d,u_d)$ in this case unlike the original SDC matrix, we could consider contraction metrics using a positive definite matrix $M(x,x_d,u_d,t)$ instead of $M(x,t)$ in Definition~\ref{DEF:Def:contraction}, to improve the representation power of $M$ at the expense of computational efficiency. Another interesting point is that the non-uniqueness of $A$ in Lemma~\ref{sdclemma} for $n \geq 2$ creates additional degrees of freedom for selecting the coefficients $\varrho$, which can also be treated as decision variables in constructing optimal contraction metrics as proposed in~\cite{mypaperTAC,Ref:Stochastic,mypaper}.

We focus mostly on the generalized SDC formulation in Sec.~\ref{sec:HinfKYP} and Sec.~\ref{sec:convex}, as it yields optimal control and estimation laws with global stability~\cite{nscm} while keeping the analysis simple enough to be understood as in LTV systems theory.
\begin{remark}
\label{remark_other_forms}
This does not mean that contraction theory works only for the SDC parameterized nonlinear systems but implies that it can be used with the other techniques discussed in Sec.~\ref{Sec:overview}. For example, due to the extended linear form given in Table~\ref{tab:sdcccm_summary}, the results to be presented in Sec.~\ref{sec:HinfKYP} and in Sec.~\ref{sec:convex} based on the SDC formulation are applicable to linearized dynamics that can be viewed as an LTV system with some modifications, regarding the dynamics modeling error term as an external disturbance as in Sec.~\ref{Sec:overviewlinear}. Also, the original SDC formulation with respect to a fixed point (\eg{}, $(s,\bar{s},\bar{u})=(x,0,0)$ in Lemma~\ref{sdclemma}) can still be used to obtain contraction conditions independent of a target trajectory $(x_d,u_d)$ (see Theorem~\ref{THM:Thm:fixed_sdc} for details).
\end{remark}
\subsubsection{Control Contraction Metric (CCM) Formulation}
We could also consider using the partial derivative of $f$ of the dynamical system directly for control synthesis through differential state feedback $\delta u = k(x,\delta x,u,t)$. This idea, formulated as the concept of a CCM~\cite{ccm,7989693,47710,WANG201944,vccm,rccm}, constructs contraction metrics with global stability guarantees independently of target trajectories, achieving greater generality while requiring added computation in evaluating integrals involving minimizing geodesics. Similar to the CCM, we could design a state estimator using a general formulation based on geodesics distances between trajectories~\cite{scrm,estimation_ccm}. These approaches are well compatible with the convex optimization-based schemes in Sec.~\ref{sec:convex}, and hence will be discussed in Sec.~\ref{Sec:ccm}.

The differences between the SDC and CCM formulation are summarized in Table~\ref{tab:sdcccm_summary}. Considering such trade-offs would help determine which form of the control law is the best fit when using contraction theory for nonlinear stabilization.
\begin{remark}
\label{remark_nonaffine}
For control non-affine nonlinear systems, we could find $f(x,u,t)-f(x_d,u_d,t)=A(x,x_d,u,u_d,t)(x-x_d)+B(x,x_d,u,u_d,t)(u-u_d)\allowbreak$ by Lemma~\ref{sdclemma} on the SDC formulation and use it in Theorem~\ref{THM:Thm:CV-STEM}, although \eqref{controller} has to be solved implicitly as $B$ depends on $u$ in this case. A similar approach for the CCM formulation can be found in~\cite{vccm,WANG201944}. As discussed in Example~\ref{EX:ex:nonaffine2}, designing such implicit control laws will be discussed in Lemma~\ref{lemma:discrete_neurallander} and Theorem~\ref{THM:Thm:neurallander} of Sec.~\ref{Sec:neurallander}.
\end{remark}
\subsection{LMI Conditions for Contraction Metrics}
We design a nonlinear feedback tracking control law parameterized by a matrix-valued function $M(x,x_d,u_d,t)$ (or $M(x,t)$, see Theorem~\ref{THM:Thm:fixed_sdc}) as follows:
\begin{align}
\label{controller}
\begin{aligned}
u &= u_d-K(x,x_d,u_d,t)(x-x_d) \\
&= u_d-R(x,x_d,u_d,t)^{-1}B(x,t)^{\top}M(x,x_d,u_d,t)(x-x_d) 
\end{aligned}
\end{align}
where $R(x,x_d,u_d,t) \succ 0$ is a weight matrix on the input $u$ and $M(x,x_d,u_d,t) \succ 0$ is a positive definite matrix (which satisfies the matrix inequality conditions for a contraction metric, to be given in Theorem~\ref{THM:Thm:CVSTEM:LMI}). As discussed in Sec.~\ref{Sec:overviewsdc}, the extended linear form of the tracking control \eqref{controller} enables LTV systems-type approaches to Lyapunov function construction, while being general enough to capture the nonlinearity of the underlying dynamics due to Lemma~\ref{u_equivalence_lemma}~\cite{lagros}. 
\begin{lemma}
\label{u_equivalence_lemma}
Consider a general feedback controller $u$ defined as $u = k(x,x_d,u_d,t)$ with $k(x_d,x_d,u_d,t)=u_d$, where $k:\mathbb{R}^n\times\mathbb{R}^n\times\mathbb{R}^m\times\mathbb{R}_{\geq 0}\mapsto\mathbb{R}^{m}$. If $k$ is piecewise continuously differentiable, then $\exists K:\mathbb{R}^n\times\mathbb{R}^n\times\mathbb{R}^m\times\mathbb{R}_{\geq 0}\mapsto\mathbb{R}^{m\times n}$ \st{} $u = k(x,x_d,u_d,t) = u_d-K(x,x_d,u_d,t)(x-x_d)$.
\end{lemma}
\begin{proof}
Using $k(x_d,x_d,u_d,t)=u_d$, $u$ can be decomposed as $u=u_d+(k(x,x_d,u_d,t)-k(x_d,x_d,u_d,t))$. Since we have $k(x,x_d,u_d,t)-k(x_d,x_d,u_d,t)=\int_0^1(dk(c x+(1-c)x_d,x_d,u_d,t)/dc)dc$, selecting $K$ as 
\begin{align}
K = -\int_0^1\frac{\partial k}{\partial x}(c x+(1-c)x_d,x_d,u_d,t)dc
\end{align}
gives the desired relation~\cite{lagros}. \qed
\end{proof}
\begin{remark}
\label{u_equivalence_remark}
Lemma~\ref{u_equivalence_lemma} implies that designing optimal $k$ of $u = k(x,x_d,u_d,t)$ reduces to designing the optimal gain $K(x,x_d,u_d,t)$ of $u=u_d-K(x,x_d,u_d,t)(x-x_d)$. We could also generalize this idea further using the CCM-based differential feedback controller $\delta u = k(x,\delta x,u,t)$~\cite{ccm,7989693,47710, WANG201944,vccm,rccm} (see Theorem~\ref{THM:Thm:ccm_cvstem}).
\end{remark}

Substituting \eqref{controller} into \eqref{original_dynamics} and \eqref{sdc_dynamics} yields the following virtual system of a smooth path $q(\mu,t)$, parameterized by $\mu \in [0,1]$ to have $q(0,t)=x_d$ and $q(1,t)=x$, for partial contraction in Theorem~\ref{THM:Thm:partial_contraction}:
\begin{align}
\label{closed_loop_e}
\dot{q}(\mu,t) &= \zeta(q(\mu,t),x,x_d,u_d,t)+d(\mu,x,t) \\
\label{closed_loop_e_sto}
dq(\mu,t) &= \zeta(q(\mu,t),x,x_d,u_d,t)dt+G(\mu,x,t)d\mathscr{W}(t)
\end{align}
where $d(\mu,x,t) = \mu d_c(x,t)$, $G(\mu,x,t) = \mu G_c(x,t)$, and $\zeta(q,x,x_d,u_d,t)$ is defined as
\begin{align}
    \label{Eq:virtual_sys_sto}
    \zeta &= (A(\varrho,x,x_d,u_d,t)-B(x,t)K(x,x_d,u_d,t))(q-x_d)+f(x_d,t)+B(x_d,t)u_d
\end{align}
where $A$ is the SDC matrix of Lemma~\ref{sdclemma} with $({s},\bar{s},\bar{u})=(x,x_d,u_d)$. Setting $\mu=1$ in \eqref{closed_loop_e} and \eqref{closed_loop_e_sto} results in \eqref{original_dynamics} and \eqref{sdc_dynamics}, respectively, and setting $\mu=0$ simply results in \eqref{sdc_dynamicsd}. Consequently, both $q=x$ and $q=x_d$ are particular solutions of \eqref{closed_loop_e} and \eqref{closed_loop_e_sto}. If there is no disturbance acting on the dynamics \eqref{original_dynamics} and \eqref{sdc_dynamics}, the differential dynamics of \eqref{closed_loop_e} and \eqref{closed_loop_e_sto} for $\partial_{\mu}q=\partial q/\partial \mu$ is given as
\begin{align}
\label{eq:differential_dyn_sto}
\partial_{\mu}\dot{q}= \left(A(\varrho,x,x_d,u_d,t)-B(x,t)K(x,x_d,u_d,t) \right)\partial_{\mu}q.
\end{align}
In~\cite{mypaperTAC,nscm,ncm,mypaper}, it is proposed that the contraction conditions of Theorems~\ref{THM:Thm:contraction} and~\ref{THM:Thm:robuststochastic} for the closed-loop dynamics \eqref{closed_loop_e} and \eqref{closed_loop_e_sto} can be expressed as convex conditions as summarized in Theorem~\ref{THM:Thm:CVSTEM:LMI}.
\begin{theorem}{Convexity of Contraction Conditions for Control}{Thm:CVSTEM:LMI}
Let $\beta$ be defined as $\beta = 0$ for deterministic systems \eqref{original_dynamics} and
\begin{align}
\beta = \alpha_s=L_m\bar{g}_c^2(\alpha_G+{1}/{2})
\end{align}
for stochastic systems \eqref{sdc_dynamics}, respectively, where $\bar{g}_c$ is given in \eqref{sdc_dynamics}, $L_m$ is the Lipschitz constant of $\partial M/\partial x_i$ for $M$ of \eqref{controller}, and $\alpha_G\in\mathbb{R}_{>0}$ is an arbitrary constant as in Theorem~\ref{THM:Thm:robuststochastic}. Also, let $W = M(x,x_d,u_d,t)^{-1}$ (or $W=M(x,t)^{-1}$, see Theorem~\ref{THM:Thm:fixed_sdc}), $\bar{W}=\nu W$, and $\nu=\overline{m}$. Then the following three matrix inequalities are equivalent:
\begin{align}
\label{riccatiFV}
\dot{M}+M\frac{\partial \zeta}{\partial q}+\frac{\partial \zeta}{\partial q}^{\top}M &\preceq -2 \alpha M-\beta \mathrm{I},~\forall \mu\in[0,1] \\
\label{riccati}
\dot{M}+2\sym(MA)-2MBR^{-1}B^{\top}M&\preceq -2 \alpha M-\beta \mathrm{I} \\
\label{riccatiW}
-\dot{\bar{W}}+2\sym(A\bar{W})-2\nu BR^{-1}B^{\top} &\preceq -2\alpha \bar{W}-\frac{\beta}{\nu}\bar{W}^2
\end{align}
where $\zeta$ is as defined in \eqref{Eq:virtual_sys_sto}. For stochastic systems with $\beta=\alpha_s>0$, these inequalities are also equivalent to
\begin{equation}
\begin{bmatrix}
-\dot{\bar{W}}+2\sym{}(A\bar{W})-2\nu BR^{-1}B^{\top}+2\alpha \bar{W}& \bar{W} \\
\bar{W} & -\frac{\nu}{\beta}\mathrm{I} \end{bmatrix} \preceq 0.\label{alpha_cond_convex_3}
\end{equation}
Note that $\nu$ and $\bar{W}$ are required for \eqref{riccatiW} and \eqref{alpha_cond_convex_3}, and the arguments $(x,x_d,u_d,t)$ for each matrix are suppressed for notational simplicity.

Furthermore, under these equivalent contraction conditions, Theorems~\ref{THM:Thm:Robust_contraction_original} and \ref{THM:Thm:robuststochastic} hold for the virtual systems \eqref{closed_loop_e} and \eqref{closed_loop_e_sto}, respectively. In particular, if $\underline{m}\mathrm{I} \preceq M\preceq \overline{m}\mathrm{I}$ of \eqref{Mcon} holds, or equivalently
\begin{align}
\label{lambda_con}
\mathrm{I} \preceq \bar{W} (x,x_d,u_d,t)\preceq \chi \mathrm{I}
\end{align}
holds for $\chi=\overline{m}/\underline{m}$, then we have the following bounds:
\begin{align}
\label{Eq:CVSTEM_control_error_bound}
\|x(t)-x_d(t)\| &\leq \frac{V_{\ell}(0)}{\sqrt{\underline{m}}}e^{-\alpha t}+\frac{\bar{d}_c}{\alpha}\sqrt{\chi}(1-e^{-\alpha t}) \\
\label{Eq:CVSTEM_control_error_bound_sto}
\mathop{\mathbb{E}}\left[\|x(t)-x_d(t)\|^2\right] &\leq \frac{\mathop{\mathbb{E}}[V_{s\ell}(0)]}{\underline{m}}e^{-2\alpha t}+\frac{C_C}{2\alpha}\chi
\end{align}
where $V_{s\ell}=\int^x_{x_d}\delta q^{\top}M\delta q$ and $V_{\ell}=\int^x_{x_d}\|\Theta\delta q\|$ are as given in Theorem~\ref{THM:Thm:path_integral} with $M=\Theta^{\top}\Theta$, the disturbance bounds $\bar{d}_c$ and $\bar{g}_c$ are given in \eqref{original_dynamics} and \eqref{sdc_dynamics}, respectively, and $C_C = \bar{g}_c^2({2}{\alpha_G}^{-1}+1)$. Note that for stochastic systems, the probability that $\|x-x_d\|$ is greater than or equal to $\varepsilon\in\mathbb{R}_{> 0}$ is given as
\begin{align}
\label{Eq:CVSTEM_control_error_bound_sto_prob}
\mathop{\mathbb{P}}\left[\|x(t)-x_d(t)\|\geq\varepsilon\right] \leq \frac{1}{\varepsilon^2}\left(\frac{\mathop{\mathbb{E}}[V_{s\ell}(0)]}{\underline{m}}e^{-2\alpha t}+\frac{C_C}{2\alpha}\chi\right).~~~~~~
\end{align}
\end{theorem}
\begin{proof}
Substituting \eqref{Eq:virtual_sys_sto} into \eqref{riccatiFV} gives \eqref{riccati}. Since $\nu > 0$ and $W \succ 0$, multiplying \eqref{riccati} by $\nu$ and then by $W$ from both sides preserves matrix definiteness. Also, the resultant inequalities are equivalent to the original ones~\cite[p. 114]{lmi}. These operations performed on \eqref{riccati} yield \eqref{riccatiW}. If $\beta=\alpha_s>0$ for stochastic systems, applying Schur's complement lemma~\cite[p. 7]{lmi} to \eqref{riccatiW} results in the Linear Matrix Inequality (LMI) condition \eqref{alpha_cond_convex_3} in terms of $\bar{W}$ and $\nu$. Therefore, \eqref{riccatiFV}--\eqref{alpha_cond_convex_3} are indeed equivalent.

Also, since we have $\|\partial_{\mu} d(\mu,x,t)\|\leq\bar{d}_c$ for $d$ in \eqref{closed_loop_e} and $\|\partial_{\mu} G(\mu,x,t)\|_F^2\leq \bar{g}_c^2$ for $G$ in \eqref{closed_loop_e_sto}, the virtual systems in \eqref{closed_loop_e} and \eqref{closed_loop_e_sto} clearly satisfy the conditions of Theorems~\ref{THM:Thm:Robust_contraction_original} and~\ref{THM:Thm:robuststochastic} if it is equipped with \eqref{riccatiFV}, which is equivalent to \eqref{riccati}--\eqref{alpha_cond_convex_3}. This implies the exponential bounds \eqref{Eq:CVSTEM_control_error_bound}--\eqref{Eq:CVSTEM_control_error_bound_sto_prob} rewritten using $\chi=\overline{m}/\underline{m}$, following the proofs of Theorems~\ref{THM:Thm:Robust_contraction_original} and~\ref{THM:Thm:robuststochastic}. \qed
\end{proof}

Because of the control and estimation duality in differential dynamics similar to that of the Kalman filter and Linear Quadratic Regulator (LQR) in LTV systems, we have an analogous robustness result for the contraction theory-based state estimator as to be derived in Sec.~\ref{Sec:cvstem_estimation}.

Although the conditions \eqref{riccatiFV}--\eqref{alpha_cond_convex_3} depend on $(x_d,u_d)$, we could also use the SDC formulation with respect to a fixed point~\cite{mypaperTAC,mypaper} in Lemma~\ref{sdclemma} to make them independent of the target trajectory as in the following theorem.
\begin{theorem}{Contraction Conditions with Fixed SDCs}{Thm:fixed_sdc}
Let $(\bar{x},\bar{u})$ be a fixed point selected arbitrarily in $\mathbb{R}^n\times\mathbb{R}^m$, \eg{}, $(\bar{x},\bar{u})=(0,0)$, and let $A(x,t)$ be an SDC matrix constructed with $({s},\bar{s},\bar{u})=(x,\bar{x},\bar{u})$ in Lemma~\ref{sdclemma}, \ie{}, 
\begin{align}
\label{fixed_sdc}
A(\varrho,x,t)(x-\bar{x})=f(x,t)+B(x,t)\bar{u}-f(\bar{x},t)-B(\bar{x},t)\bar{u}.~~~~
\end{align}
Suppose that the contraction metric of Theorem~\ref{THM:Thm:CVSTEM:LMI} is designed by $M(x,t)$ with $A$ of \eqref{fixed_sdc}, independently of the target trajectory $(x_d,u_d)$, and that the systems \eqref{original_dynamics} and \eqref{sdc_dynamics} are controlled by
\begin{align}
\label{fixed_sdc_control}
u=u_d-R(x,t)^{-1}B(x,t)^{\top}M(x,t)(x-x_d)
\end{align}
with such $M(x,t)$, where $R(x,t)\succ0$ is a weight matrix on $u$. If the function $\phi(x,x_d,u_d,t)=A(\varrho,x,t)(x_d-\bar{x})+B(x,t)(u_d-\bar{u})$ is Lipschitz in $x$ with its Lipschitz constant $\bar{L}$, then Theorem~\ref{THM:Thm:CVSTEM:LMI} still holds with $\alpha$ of the conditions \eqref{riccatiFV}--\eqref{alpha_cond_convex_3} replaced by $\alpha+\bar{L}\sqrt{\overline{m}/\underline{m}}$. The same argument holds for state estimation of Theorem~\ref{THM:Thm:CVSTEM-LMI-est} to be discussed in Sec.~\ref{Sec:cvstem_estimation}.
\end{theorem}
\begin{proof}
The unperturbed virtual system of \eqref{original_dynamics}, \eqref{sdc_dynamics}, and \eqref{sdc_dynamicsd} with $A$ of \eqref{fixed_sdc} and $u$ of \eqref{fixed_sdc_control} is given as follows:
\begin{align}
\label{virtual_fixed_sdc}
\begin{aligned}
\dot{q} &= (A(\varrho,x,t)-B(x,t)K(x,t))(q-x_d)\\
&+A(\varrho,q,t)(x_d-\bar{x})+B(q,t)(u_d-\bar{u})+f(\bar{x},t)+B(\bar{x},t)\bar{u}
\end{aligned}
\end{align}
where $K(x,t)=R(x,t)^{-1}B(x,t)^{\top}M(x,t)$. Following the proof of Theorem~\ref{THM:Thm:CVSTEM:LMI}, the computation of $\dot{V}$, where $V=\delta q^{\top} M(x,t)\delta q$, yields an extra term
\begin{align}
\label{fixed_sdc_extra_term}
2\delta q^{\top}M\frac{\partial\phi}{\partial q}(q,x_d,u_d,t)\delta q \leq 2\bar{L}\sqrt{\frac{\overline{m}}{\underline{m}}}\delta q^{\top} M\delta q
\end{align}
due to the Lipschitz condition on $\phi$, where $\phi(q,x_d,u_d,t)=A(q,t)(x_d-\bar{x})+B(q,t)(u_d-\bar{u})$. This indeed implies that the system \eqref{virtual_fixed_sdc} is contracting as long as the conditions \eqref{riccatiFV}--\eqref{alpha_cond_convex_3} hold with $\alpha$ replaced by $\alpha+\bar{L}\sqrt{\overline{m}/\underline{m}}$. The last statement on state estimation follows from the nonlinear control and estimation duality to be discussed in Sec.~\ref{Sec:cvstem_estimation}.
\end{proof}
\begin{remark}
\label{remark_fixed_sdc}
As demonstrated in~\cite{mypaperTAC}, we could directly use the extra term $2\delta q^{\top}M({\partial\phi}/{\partial q})\delta q$ of \eqref{fixed_sdc_extra_term} in \eqref{riccatiFV}--\eqref{alpha_cond_convex_3} without upper-bounding it, although now the conditions of Theorem~\ref{THM:Thm:fixed_sdc} depend on $(x,q,t)$ instead of $(x,t)$. Also, the following two inequalities given in~\cite{mypaperTAC} with $\bar{\gamma}=\nu \gamma$, $\gamma \in \mathbb{R}_{\geq 0}$:
\begin{align}
\label{riccati_lmi_con}
&-\dot{\bar{W}}+A\bar{W}+\bar{W}A^{\top}+\bar{\gamma} \mathrm{I}-\nu BR^{-1}B^{\top} \preceq 0 \\
&\begin{bmatrix}
\bar{\gamma} \mathrm{I}+\nu BR^{-1}B^{\top}-\bar{W}\phi^{\top}-\phi \bar{W}-2\alpha \bar{W}& \bar{W} \\
\bar{W} & \frac{\nu}{2\alpha_s}\mathrm{I} \end{bmatrix} \succeq 0\nonumber
\end{align}
are combined as one LMI \eqref{alpha_cond_convex_3} in Theorems~\ref{THM:Thm:CVSTEM:LMI} and~\ref{THM:Thm:fixed_sdc}.
\end{remark}
\begin{example}{Contraction in $\mathcal{H}_{\infty}$ Control}{ex:hinf_lmi}
The inequalities in Theorem~\ref{THM:Thm:CVSTEM:LMI} can be interpreted as in the Riccati inequality in $\mathcal{H}_{\infty}$ control. Consider the following system:
\begin{align}
\label{ex_hinf_dyn}
\begin{aligned}
\dot{x} &= Ax+B_uu+B_ww \\
z &= C_zx
\end{aligned}
\end{align}
where $A\in\mathbb{R}^{n\times n}$, $B_u\in\mathbb{R}^{n\times m}$, $B_w\in\mathbb{R}^{n\times w}$, and $C_z\in\mathbb{R}^{o\times n}$ are constant matrices, $w\in\mathbb{R}^w$ is an exogenous input, and $z\in\mathbb{R}^o$ is a system output. As shown in~\cite{256331} and~\cite[p. 109]{lmi}, there exists a state feedback gain $K=R^{-1}B_u^{\top}P$ such that the $\mathcal{L}_2$ gain of the closed-loop system \eqref{ex_hinf_dyn}, $\sup_{\|w\|\neq 0}{\|z\|}/{\|w\|}$, is less than or equal to $\gamma$ if
\begin{align}
\label{ex_riccati_robust}
2\sym{}(PA)-2PB_uR^{-1}B_u^{\top}P+\frac{PB_wB_w^{\top}P}{\gamma^2}+C_z^{\top}C_z \preceq 0~~~~~~
\end{align}
has a solution $P \succ 0$, where $R\succ 0$ is a constant weight matrix on the input $u$. If we select $B_w$ and $C_z$ to have $B_wB_w^{\top}\succeq (P^{-1})^2$ and $C_z^{\top}C_z \succeq 2\alpha P$ for some $\alpha >0$, the contraction condition \eqref{riccati} in Theorem~\ref{THM:Thm:CVSTEM:LMI} can be satisfied with $M=P$, $B = B_u$, and $\beta = 1/\gamma^2$ due to \eqref{ex_riccati_robust}.
\end{example}

In Sections~\ref{sec:brl_contraction} and~\ref{sec:kyp_contraction}, we will discuss the relationship to input-output stability theory as in Example~\ref{EX:ex:hinf_lmi}, using the results of Theorem~\ref{THM:Thm:CVSTEM:LMI}.
\subsection{Bounded Real Lemma in Contraction Theory}
\label{sec:brl_contraction}
The LMI \eqref{alpha_cond_convex_3} of Theorem~\ref{THM:Thm:CVSTEM:LMI} can be interpreted as the LMI condition for the bounded real lemma~\cite[p. 369]{brlbook}. Let us consider the following Hurwitz linear system with $\varsigma(t)\in \mathcal{L}_{2e}$, (\ie{}, $\|(\varsigma)_{\tau}\|_{\mathcal{L}_2}<\infty$ for $\tau \in\mathbb{R}_{\geq 0}$, see Sec.~\ref{notation}):
\begin{align}
    \begin{aligned}
    \dot{\kappa}&=\mathcal{A}\kappa+\mathcal{B}\varsigma(t) \\
    \upsilon&=\mathcal{C}\kappa+\mathcal{D}\varsigma(t).
    \end{aligned}\label{eq:linearsys}
\end{align}
Setting $\kappa = \delta q$ and viewing $\varsigma$ as external disturbance, this system can be interpreted as the differential closed-loop dynamics defined earlier in \eqref{eq:differential_dyn_sto}. The bounded real lemma states that this system is $\mathcal{L}_2$ gain stable with its $\mathcal{L}_2$ gain less than or equal to $\gamma$, \ie{}, $\|(\upsilon)_{\tau}\|_{\mathcal{L}_2}\leq \gamma\|(\varsigma)_{\tau}\|_{\mathcal{L}_2}+const.$, or equivalently, the $\mathcal{H}_{\infty}$ norm of the transfer function of \eqref{eq:linearsys} is less than or equal to $\gamma$, if the following LMI for $\mathcal{P}\succ 0$ holds (see~\cite[p. 369]{brlbook}):
\begin{equation}
    \begin{bmatrix}\dot{\mathcal{P}}+2\sym{}(\mathcal{P}\mathcal{A})+\mathcal{C}^\top\mathcal{C} & \mathcal{P}\mathcal{B}+\mathcal{C}^\top\mathcal{D} \\
    \mathcal{B}^\top\mathcal{P}+\mathcal{D}^\top\mathcal{C} & \mathcal{D}^\top\mathcal{D}-\gamma^2\mathrm{I} \end{bmatrix}\preceq 0.\label{eq:BoundedRealLemma}
\end{equation}
Theorem~\ref{THM:Thm:bounded_real_cvstem} introduces the bounded real lemma in the context of contraction theory.
\begin{theorem}{Bounded Real Lemma in Contraction Theory}{Thm:bounded_real_cvstem}
Let $\mathcal{A}= A-BR^{-1}B^{\top}M$, $\mathcal{B}=M^{-1}$, $C=\sqrt{2\alpha}\Theta$, and $\mathcal{D}=0$ in \eqref{eq:linearsys}, where $M=\Theta^{\top}\Theta$, and the other variables are as defined in Theorem~\ref{THM:Thm:CVSTEM:LMI}. Then \eqref{eq:BoundedRealLemma} with $\mathcal{P}=M$ and $\gamma=1/\sqrt{\beta}$ is equivalent to \eqref{alpha_cond_convex_3}, and thus
\eqref{alpha_cond_convex_3} implies $\mathcal{L}_2$ gain stability of \eqref{eq:linearsys} with its $\mathcal{L}_2$ gain less than or equal to $\gamma$.
\end{theorem}
\begin{proof}
Multiplying \eqref{alpha_cond_convex_3} by $\nu^{-1}$ and then by $\left[\begin{smallmatrix}M & 0\\ 0 & \mathrm{I}\end{smallmatrix}\right]$ from both sides gives the following matrix inequality:
\begin{align}
\nu\begin{bmatrix}
\dot{M}+2\sym{}(M(A-BR^{-1}B^{\top}M))+2\alpha M& \mathrm{I} \\
\mathrm{I} & -\frac{1}{\beta}\mathrm{I} \end{bmatrix} \preceq 0. \nonumber
\end{align}
This is indeed equivalent to \eqref{eq:BoundedRealLemma} if $\mathcal{A}= A-BR^{-1}B^{\top}M$, $\mathcal{B}=M^{-1}$, $C=\sqrt{2\alpha}\Theta$, $\mathcal{D}=0$, $\mathcal{P}=M$, and $\gamma=1/\sqrt{\beta}$. Now, multiplying \eqref{eq:BoundedRealLemma} by $[\delta q^{\top},\varsigma^{\top}]^{\top}=[\kappa^{\top},\varsigma^{\top}]^{\top}$ for such $(\mathcal{A},\mathcal{B},\mathcal{C},\mathcal{D})$ gives
\begin{align}
\label{eq:brl_cvstem_1}
\kappa^{\top}(\dot{\mathcal{P}}+2\mathcal{P}\mathcal{A}+\mathcal{C}^{\top}\mathcal{C})\kappa+2\kappa^{\top}\mathcal{P}\mathcal{B}\varsigma-\gamma^2\|\varsigma\|^2 \preceq 0
\end{align}
resulting in $\dot{V}+\|\upsilon\|^2-\gamma^2\|\varsigma\|^2 \preceq 0$ for $V = \kappa^{\top}\mathcal{P}\kappa = \delta q^{\top}M\delta q$. This implies $\mathcal{L}_2$ gain stability with its $\mathcal{L}_2$ gain less than or equal to $\gamma = 1/\sqrt{\beta}$~\cite[p. 209]{Khalil:1173048}. \qed
\end{proof}
\subsection{KYP Lemma in Contraction Theory}
\label{sec:kyp_contraction}
Analogously to Theorem~\ref{THM:Thm:bounded_real_cvstem}, the LMI \eqref{alpha_cond_convex_3} of Theorem~\ref{THM:Thm:CVSTEM:LMI} can be understood using the Kalman-Yakubovich-Popov (KYP) lemma~\cite[p. 218]{kypbook}. Consider the quadratic Lyapunov function $V=\kappa^\top \mathcal{Q} \kappa$ with $\mathcal{Q}\succ 0$, satisfying the following output strict passivity (dissipativity) condition:
\begin{equation}
    \dot{V}- 2 \varsigma^\top \upsilon+ \frac{2\upsilon^\top \upsilon}{\gamma}\leq 0\label{eq:VdotKYP}
\end{equation}
which can be expanded by completing the square to have 
\begin{align}
    \label{kyp_square}
    \dot{V}&\leq -\left\|\gamma\left(\varsigma-\frac{\upsilon}{\gamma}\right)\right\|^2+\gamma \|\varsigma\|^2-\frac{1}{\gamma} \|\upsilon\|^2 \leq \gamma \|\varsigma\|^2-\frac{1}{\gamma} \|\upsilon\|^2.
\end{align}
This implies that we have $\| (\upsilon)_\tau\|_{\mathcal{L}_2} \leq \gamma \|(\varsigma)_\tau\|_{\mathcal{L}_2}+\sqrt{\gamma V(0)}$ by the comparison lemma~\cite[pp. 102-103, pp. 350-353]{Khalil:1173048}, leading to $\mathcal{L}_2$ gain stability with its $\mathcal{L}_2$ gain less than or equal to $\gamma$~\cite[p. 218]{kypbook}. The condition \eqref{eq:VdotKYP} can be expressed equivalently as an LMI form as follows:
\begin{equation}
\begin{bmatrix}
\dot{\mathcal{Q}}+2\mathrm{sym}(\mathcal{Q}\mathcal{A})+\frac{2\mathcal{C}^\top\mathcal{C}}{\gamma} & \mathcal{Q}\mathcal{B}+\mathcal{C}^\top\left(\frac{2\mathcal{D}}{\gamma}-\mathrm{I}\right) \\
    \mathcal{B}^\top\mathcal{Q}+\left(\frac{2\mathcal{D}^\top}{\gamma}-\mathrm{I}\right)\mathcal{C} & -(\mathcal{D}^\top+D)+\frac{2\mathcal{D}^\top\mathcal{D}}{\gamma} 
\end{bmatrix}\preceq 0\label{eq:KYPLMI}
\end{equation}
where $\mathcal{A}$, $\mathcal{B}$, $\mathcal{C}$, and $\mathcal{D}$ are as defined in \eqref{eq:linearsys}.
\begin{theorem}{KYP Lemma in Contraction Theory}{Thm:kyp_cvstem}
If \eqref{eq:KYPLMI} holds, the LMI \eqref{eq:BoundedRealLemma} for the bounded real lemma holds with $\mathcal{P}=\gamma \mathcal{Q}$, \ie{}, the system \eqref{eq:linearsys} is $\mathcal{L}_2$ gain stable with its $\mathcal{L}_2$ gain less than or equal to $\gamma$. Thus, for systems with $\mathcal{A}$, $\mathcal{B}$, $\mathcal{C}$, and $\mathcal{D}$ defined in Theorem~\ref{THM:Thm:bounded_real_cvstem}, the condition \eqref{eq:KYPLMI} guarantees the contraction condition \eqref{alpha_cond_convex_3} of Theorem~\ref{THM:Thm:CVSTEM:LMI}.
\end{theorem}
\begin{proof}
Writing the inequality in \eqref{kyp_square} in a matrix form, we have that
\begin{align*}
&\begin{bmatrix}
\dot{\mathcal{Q}}+2\mathrm{sym}(\mathcal{Q}\mathcal{A})+\frac{2\mathcal{C}^\top\mathcal{C}}{\gamma} & \mathcal{Q}\mathcal{B}+\mathcal{C}^\top\left(\frac{2\mathcal{D}}{\gamma}-\mathrm{I}\right) \\
    \mathcal{B}^\top\mathcal{Q}+\left(\frac{2\mathcal{D}^\top}{\gamma}-\mathrm{I}\right)\mathcal{C} & -(\mathcal{D}^\top+D)+\frac{2\mathcal{D}^\top\mathcal{D}}{\gamma} 
\end{bmatrix}\\
&\succeq\begin{bmatrix}\dot{\mathcal{Q}}+2\mathrm{sym}(\mathcal{Q}\mathcal{A}) & \mathcal{Q}\mathcal{B} \\
    \mathcal{B}^\top\mathcal{Q} & 0\end{bmatrix}
    +\begin{bmatrix}\frac{\mathcal{C}^\top\mathcal{C}}{\gamma} & \frac{\mathcal{C}^\top\mathcal{D}}{\gamma} \\
    \frac{\mathcal{D}^\top\mathcal{C}}{\gamma} & \frac{\mathcal{D}^\top\mathcal{D}}{\gamma} \end{bmatrix}
       -\begin{bmatrix}0 & 0\\
0 & \gamma \mathrm{I}\end{bmatrix} \\
& =\begin{bmatrix}
\dot{\mathcal{Q}}+2\mathrm{sym}(\mathcal{Q}\mathcal{A})+\frac{\mathcal{C}^\top\mathcal{C}}{\gamma} & \mathcal{Q}\mathcal{B}+\frac{\mathcal{C}^\top\mathcal{D}}{\gamma} \\
    \mathcal{B}^\top\mathcal{Q}+\frac{\mathcal{D}^\top\mathcal{C}}{\gamma} & \frac{\mathcal{D}^\top\mathcal{D}}{\gamma}-\gamma \mathrm{I}
\end{bmatrix}.
\end{align*}
Therefore, a necessary condition of \eqref{eq:KYPLMI} reduces to \eqref{eq:BoundedRealLemma} if $\mathcal{Q}=\mathcal{P}/\gamma$. The rest follows from Theorem~\ref{THM:Thm:bounded_real_cvstem}. \qed
\end{proof}
\begin{example}{Contraction and Nonlinear Stability}{}
Theorems~\ref{THM:Thm:bounded_real_cvstem} and~\ref{THM:Thm:kyp_cvstem} imply the following statements.
\setlength{\leftmargini}{17pt}     
\begin{itemize}
	\setlength{\itemsep}{1pt}      
	\setlength{\parskip}{0pt}      
    \item If \eqref{eq:BoundedRealLemma} holds, the system is finite-gain $\mathcal{L}_2$ stable with $\gamma$ as its $\mathcal{L}_2$ gain.
    \item If \eqref{eq:KYPLMI} holds, the system is finite-gain $\mathcal{L}_2$ stable with $\gamma$ as its $\mathcal{L}_2$ gain.
    \item \eqref{eq:VdotKYP} is equivalent to \eqref{eq:KYPLMI}, and to the output strict passivity condition with dissipation $1/\gamma$~\cite[p. 231]{Khalil:1173048}.
    \item If \eqref{eq:KYPLMI} holds (KYP lemma), \eqref{eq:BoundedRealLemma} holds (bounded real lemma). The converse is not necessarily true.
\end{itemize}
\end{example}
\newpage
\section{Convex Optimality in Contraction Theory}
\label{sec:convex}
Theorem~\ref{THM:Thm:CVSTEM:LMI} indicates that the problem of finding contraction metrics for general nonlinear systems could be formulated as a convex feasibility problem. This section thus delineates one approach, called the method of ConVex optimization-based Steady-state Tracking Error Minimization (CV-STEM)~\cite{mypaperTAC,ncm,nscm,mypaper}, to optimally design $M$ of Theorem~\ref{THM:Thm:CVSTEM:LMI} that defines a contraction metric and minimizes an upper bound of the steady-state tracking error in Theorem~\ref{THM:Thm:Robust_contraction_original} or in Theorem~\ref{THM:Thm:robuststochastic} via convex optimization. In particular, we present an overview of the SDC- and CCM-based CV-STEM frameworks for provably stable and optimal feedback control and state estimation of nonlinear systems, perturbed by deterministic and stochastic disturbances. It is worth noting that the steady-state bound is expressed as a function of the condition number $\chi=\overline{m}/\underline{m}$ as to be seen in \eqref{stochasticmse_infty} and \eqref{stochasticmse_infty_est}, which renders the CV-STEM applicable and effective even to learning-based and data-driven control frameworks as shall be seen in Sec.~\ref{Sec:learning_stability}.
\subsection{CV-STEM Control}
\label{Sec:cvstem_control}
As a result of Theorems~\ref{THM:Thm:Robust_contraction_original},~\ref{THM:Thm:robuststochastic}, and~\ref{THM:Thm:CVSTEM:LMI}, the control law $u=u_d-K(x-x_d)$ of \eqref{controller} gives a convex steady-state upper bound of the Euclidean distance between the system trajectories, which can be used as an objective function for the CV-STEM control framework in Theorem~\ref{THM:Thm:CV-STEM}~\cite{mypaperTAC,ncm,nscm,mypaper}.
\begin{theorem}{Convexity of Steady-State Tracking Error Bounds}{Thm:CVSTEMobjective}
If one of the matrix inequalities of Theorem~\ref{THM:Thm:CVSTEM:LMI} holds, we have the following bound for  $\chi={\overline{m}}/{\underline{m}}$:
\begin{align}
\label{stochasticmse_infty}
&\lim_{t \to \infty}\sqrt{\mathop{\mathbb{E}}\left[\|x-x_d\|^2\right]} \leq c_0(\alpha,\alpha_G)\sqrt{\chi} \leq c_0(\alpha,\alpha_G)\chi
\end{align}
where $c_0(\alpha,\alpha_G)=\bar{d}_c
/\alpha$ for deterministic systems and $c_0(\alpha,\alpha_G)=\bar{g}_c\sqrt{(2\alpha_G^{-1}+1)/(2\alpha)}$ for stochastic systems, with the variables given in Theorem~\ref{THM:Thm:CVSTEM:LMI}.
\end{theorem}
\begin{proof}
Taking $\lim_{t\to\infty}$ in the exponential tracking error bounds \eqref{Eq:CVSTEM_control_error_bound} and \eqref{Eq:CVSTEM_control_error_bound_sto} of Theorem~\ref{THM:Thm:CVSTEM:LMI} gives the first inequality of \eqref{stochasticmse_infty}. The second inequality follows from the relation $1\leq\sqrt{\overline{m}/\underline{m}}\leq \overline{m}/\underline{m}=\chi$ due to $\underline{m}\leq\overline{m}$. \qed
\end{proof}

\begin{figure}
    \centering
    \includegraphics[width=127.5mm]{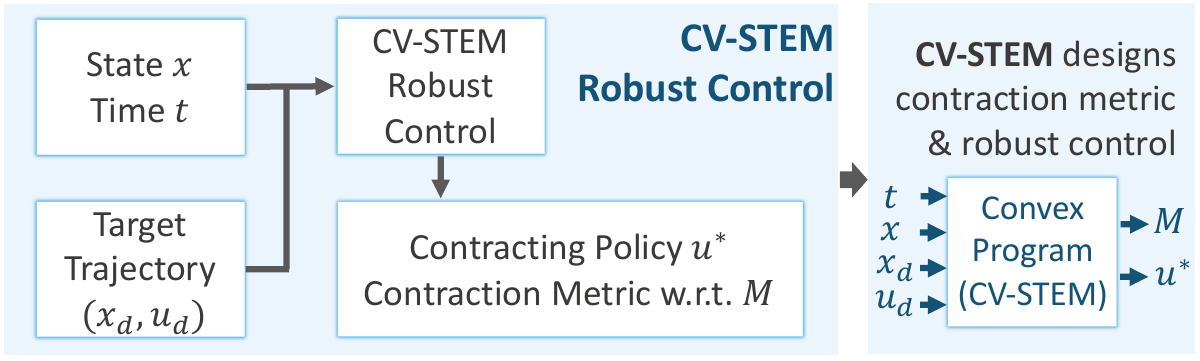}
    \caption{Block diagram of CV-STEM.}
    \label{slide2_cvstem}
\end{figure}

The convex optimization problem of minimizing \eqref{stochasticmse_infty}, subject to the contraction constraint of Theorem~\ref{THM:Thm:CVSTEM:LMI}, is given in Theorem~\ref{THM:Thm:CV-STEM}, thereby introducing the CV-STEM control~\cite{mypaperTAC,ncm,nscm,mypaper} for designing an optimal contraction metric and contracting control policy as depicted in Fig.~\ref{slide2_cvstem}.
\begin{theorem}{Robust Nonlinear Control via Convex Optimization}{Thm:CV-STEM}
Suppose that $\alpha$, $\alpha_G$, $\bar{d}_c$, and $\bar{g}_c$ in \eqref{stochasticmse_infty} and the Lipschitz constant $L_m$ of $\partial M/\partial x_i$ in Theorem~\ref{THM:Thm:CVSTEM:LMI} are given. If the pair $(A,B)$ is uniformly controllable, the non-convex optimization problem of minimizing the upper bound \eqref{stochasticmse_infty} is equivalent to the following convex optimization problem, with the convex contraction constraint \eqref{riccatiW} or \eqref{alpha_cond_convex_3} of Theorem~\ref{THM:Thm:CVSTEM:LMI} and $\mathrm{I}\preceq \bar{W}\preceq\chi\mathrm{I}$ of \eqref{lambda_con}:
\begin{align}
\label{cvstem_eq}
\begin{aligned}
{J}_{CV}^* &= \min_{\nu\in\mathbb{R}_{>0},\chi \in \mathbb{R},\bar{W} \succ 0} c_0\chi+c_1\nu+c_2 P(\chi,\nu,\bar{W}) \\
&\text{\rm\st{}~\eqref{riccatiW} and \eqref{lambda_con} for deterministic systems} \\
&\text{\rm{\color{white}\st{}}~\eqref{alpha_cond_convex_3} and \eqref{lambda_con} for stochastic systems} 
\end{aligned}
\end{align}
where $W = M(x,x_d,u_d,t)^{-1}$ for Theorem~\ref{THM:Thm:CVSTEM:LMI} or $W=M(x,t)^{-1}$ for Theorem~\ref{THM:Thm:fixed_sdc}, $\bar{W}=\nu W$, $\nu=\overline{m}$, $\chi=\overline{m}/\underline{m}$, $c_0$ is as defined in \eqref{stochasticmse_infty} of Theorem~\ref{THM:Thm:CVSTEMobjective}, $c_1,c_2\in\mathbb{R}_{\geq 0}$, and $P$ is a performance-based convex cost function.

The weight $c_1$ for $\nu$ can be viewed as a penalty on the $2$-norm of the feedback control gain $K$ of $u=u_d-K(x-x_d)$ in \eqref{controller}. Using non-zero $c_1$ and $c_2$ thus enables finding contraction metrics optimal in a different sense. Furthermore, the coefficients of the SDC parameterizations $\varrho$ in Lemma~\ref{sdclemma}, \ie{}, $A = \sum\varrho_iA_i$ in \eqref{varrho_def}, can also be treated as decision variables by convex relaxation, thereby adding design flexibility to mitigate the effects of external disturbances while verifying the system controllability.
\end{theorem}
\begin{proof}
As proven in Theorems~\ref{THM:Thm:CVSTEM:LMI} and~\ref{THM:Thm:CVSTEMobjective}, $\underline{m}\mathrm{I}\preceq M\preceq\overline{m}\mathrm{I}$ of \eqref{Mcon}, the contraction constraint \eqref{riccatiFV}, and the objective \eqref{stochasticmse_infty} reduce to \eqref{cvstem_eq} with $c_1=0$ and $c_2=0$. Since the resultant constraints are convex and the objective is affine in terms of the decision variables $\nu$, $\chi$, and $\bar{W}$, the problem \eqref{cvstem_eq} is indeed convex. Since we have $\sup_{x,t}\|M\|\leq \overline{m}=\nu$, $c_1$ can be used as a penalty to optimally adjust the induced $2$-norm of the control gain (see Example~\ref{EX:ex:lqrexample}). The SDC coefficients $\varrho$ can also be utilized as decision variables for controllability due to Proposition~1 of~\cite{mypaperTAC}. \qed
\end{proof}
\begin{example}{LQR Analogy}{ex:lqrexample}
The weights $c_0$ and $c_1$ of the CV-STEM control of Theorem~\ref{THM:Thm:CV-STEM} establish an analogous trade-off to the case of the LQR with the cost weight matrices of $Q$ for state and $R$ for control, since the upper bound of the steady-state tracking error \eqref{stochasticmse_infty} is proportional to $\chi$, and an upper bound of $\|K\|$ for the control gain $K$ in \eqref{controller}, $u=u_d-K(x-x_d)$, is proportional to $\nu$. In particular~\cite{ncm,nscm},
\setlength{\leftmargini}{17pt}     
\begin{itemize}
	\setlength{\itemsep}{1pt}      
	\setlength{\parskip}{0pt}      
    \item if $c_1$ is much greater than $c_0$ in \eqref{cvstem_eq} of Theorem~\ref{THM:Thm:CV-STEM}, we get smaller control effort but with a large steady-state tracking error, and
    \item if $c_1$ is much smaller than $c_0$ in \eqref{cvstem_eq} of Theorem~\ref{THM:Thm:CV-STEM}, we get a smaller steady-state state tracking error but with larger control effort.
\end{itemize}
\end{example}
This is also because the solution $P\succ0$ of the LQR Riccati equation~\cite[p. 89]{lssbook}, $-\dot{P}=PA+A^{\top}P-PBR^{-1}B^{\top}P+Q$, can be viewed as a positive definite matrix $M$ that defines a contraction metric as discussed in Example~\ref{EX:ex:lyapunov_contraction}.
\subsection{CV-STEM Estimation}
\label{Sec:cvstem_estimation}
We could also design an optimal state estimator analogously to the CV-STEM control of Theorem~\ref{THM:Thm:CV-STEM}, due to the differential nature of contraction theory that enables LTV systems-type approaches to stability analysis. In particular, we exploit the estimation and control duality in differential dynamics similar to that of the Kalman filter and LQR in LTV systems. 

Let us consider the following smooth nonlinear systems with a measurement $y(t)$, perturbed by deterministic disturbances $d_{e0}(x,t)$ and $d_{e1}(x,t)$ with $\sup_{x,t}\|d_{e0}(x,t)\|=\bar{d}_{e0} \in \mathbb{R}_{\geq 0}$ and $\sup_{x,t}\|d_{e1}(x,t)\|=\bar{d}_{e1} \in \mathbb{R}_{\geq 0}$, or by Gaussian white noise $\mathscr{W}_{0}(t)$ and $\mathscr{W}_{1}(t)$ with $\sup_{x,t}\|G_{e0}(x,t)\|_F = \bar{g}_{e0} \in \mathbb{R}_{\geq 0}$ and $\sup_{x,t}\|G_{e1}(x,t)\|_F = \bar{g}_{e1} \in \mathbb{R}_{\geq 0}$:
\begin{align}
\label{sdc_dynamics_est}
\dot{x} =& f(x,t)+d_{e0}(x,t),~y = h(x,t)+d_{e1}(x,t) \\
\label{sdc_dynamics_est_sto}
dx =& f(x,t)dt+G_{e0}d\mathscr{W}_0,~ydt = h(x,t)dt+G_{e1}d\mathscr{W}_1
\end{align}
where $t \in \mathbb{R}_{\geq 0}$ is time, ${x}:\mathbb{R}_{\geq 0}\mapsto\mathbb{R}^n$ is the system state, $y:\mathbb{R}_{\geq 0}\mapsto\mathbb{R}^m$ is the system measurement, ${f}:
\mathbb{R}^n\times\mathbb{R}_{\geq 0}\mapsto\mathbb{R}^n$ and $h:\mathbb{R}^n\times\mathbb{R}_{\geq 0} \rightarrow\mathbb{R}^{m}$ are known smooth functions, $d_{e0}:\mathbb{R}^n\times\mathbb{R}_{\geq 0} \rightarrow\mathbb{R}^{n}$, $d_{e1}:\mathbb{R}^n\times\mathbb{R}_{\geq 0} \rightarrow\mathbb{R}^{0}$, $G_{e0}:\mathbb{R}^n\times\mathbb{R}_{\geq 0} \rightarrow\mathbb{R}^{n\times w_{0}}$, and $G_{e1}:\mathbb{R}^n\times\mathbb{R}_{\geq 0} \rightarrow\mathbb{R}^{n\times w_{1}}$ are unknown bounded functions for external disturbances, $\mathscr{W}_{0}:\mathbb{R}_{\geq 0} \rightarrow\mathbb{R}^{w_{0}}$ and $\mathscr{W}_{1}:\mathbb{R}_{\geq 0} \rightarrow\mathbb{R}^{w_{1}}$ are two independent Wiener processes, and the arguments of $G_{e0}(x,t)$ and $G_{e1}(x,t)$ are suppressed for notational convenience. Let $A(\varrho_a,x,\hat{x},t)$ and $C(\varrho_c,x,\hat{x},t)$ be the SDC matrices given by Lemma~\ref{sdclemma} with $(f,{s},\bar{s},\bar{u})$ replaced by $(f,\hat{x},x,0)$ and $(h,\hat{x},x,0)$, respectively, \ie{}
\begin{align}
\label{est_A}
A(\varrho_a,x,\hat{x},t)(\hat{x}-x) &= f(\hat{x},t)-f(x,t) \\
\label{est_C}
C(\varrho_c,x,\hat{x},t)(\hat{x}-x) &= h(\hat{x},t)-h(x,t).
\end{align}
We design a nonlinear state estimation law parameterized by a matrix-valued function $M(\hat{x},t)$ as follows:
\begin{align}
\label{estimator}
\begin{aligned}
    \dot{\hat{x}} &= f(\hat{x},t)+L(\hat{x},t)(y-h(\hat{x},t)) \\
    &=f(\hat{x},t)+M(\hat{x},t)\bar{C}(\varrho_c,\hat{x},t)^{\top}R(\hat{x},t)^{-1}(y-h(\hat{x},t))
\end{aligned}
\end{align}
where $\bar{C}(\varrho_c,\hat{x},t)=C(\varrho_c,\hat{x},\bar{x},t)$ for a fixed trajectory $\bar{x}$ (\eg{}, $\bar{x}=0$, see Theorem~\ref{THM:Thm:fixed_sdc}), $R(\hat{x},t) \succ 0$ is a weight matrix on the measurement $y$, and $M(\hat{x},t) \succ 0$ is a positive definite matrix (which satisfies the matrix inequality constraint for a contraction metric, to be given in \eqref{convex_constraint_estimator1} of Theorem~\ref{THM:Thm:CV-STEM-estimation}). Note that we could use other forms of estimation laws such as the EKF~\cite{6849943,ncm,Brown:680442}, analytical SLAM~\cite{doi:10.1177/0278364917710541}, or SDC with respect to a fixed point~\cite{Ref:Stochastic,mypaperTAC,mypaper}, depending on the application of interest, which result in a similar stability analysis as in Theorem~\ref{THM:Thm:fixed_sdc}.
\subsubsection{Nonlinear Stability Analysis of SDC-based State Estimation}
Substituting \eqref{estimator} into \eqref{sdc_dynamics_est} and \eqref{sdc_dynamics_est_sto} yields the following virtual system of a smooth path $q(\mu,t)$, parameterized by $\mu \in [0,1]$ to have $q(\mu=0,t)=x$ and $q(\mu=1,t)=\hat{x}$:
\begin{align}
\label{closed_loop_e_est}
\dot{q}(\mu,t) &= \zeta(q(\mu,t),x,\hat{x},t)+d(\mu,x,\hat{x},t) \\
\label{closed_loop_e_est_sto}
dq(\mu,t) &= \zeta(q(\mu,t),x,\hat{x},t)dt+G(\mu,x,\hat{x},t)d\mathscr{W}(t)
\end{align}
where $d(\mu,x,\hat{x},t)=(1-\mu) d_{e0}(x,t)+\mu L(\hat{x},t)d_{e1}(x,t)$, $G(\mu,x,\hat{x},t)=[(1-\mu) G_{e0}(x,t),\mu L(\hat{x},t)G_{e1}(x,t)]$, $\mathscr{W} = [\mathscr{W}_0^{\top},\mathscr{W}_1^{\top}]^{\top}$, and $\zeta(q,x,x_d,u_d,t)$ is defined as
\begin{align}
\zeta(q,x,\hat{x},t) &= (A(\varrho_a,x,\hat{x},t)-L(\hat{x},t)C(\varrho_c,x,\hat{x},t))(q-x)+f(x,t).
\label{Eq:virtual_sys_est_sto}
\end{align}
Note that \eqref{Eq:virtual_sys_est_sto} is constructed to contain $q=\hat{x}$ and $q=x$ as its particular solutions of \eqref{closed_loop_e_est} and \eqref{closed_loop_e_est_sto}. If $d=0$ and $\mathscr{W}=0$, the differential dynamics of \eqref{closed_loop_e_est} and \eqref{closed_loop_e_est_sto} for $\partial_{\mu}q=\partial q/\partial \mu$ is given as
\begin{align}
\label{eq:differential_dyn_est_sto}
\partial_{\mu}\dot{q} = \left(A(\varrho_a,x,\hat{x},t)-L(\hat{x},t)C(\varrho_c,x,\hat{x},t) \right)\partial_{\mu}q.
\end{align}
The similarity between \eqref{eq:differential_dyn_sto} ($\partial_{\mu}\dot{q}= (A-BK)\partial_{\mu}q$) and \eqref{eq:differential_dyn_est_sto} leads to the following theorem~\cite{mypaperTAC,ncm,nscm,mypaper}. Again, note that we could also use the SDC formulation with respect to a fixed point as delineated in Theorem~\ref{THM:Thm:fixed_sdc} and as demonstrated in~\cite{mypaperTAC,Ref:Stochastic,mypaper}.
\begin{theorem}{Convexity of Contraction Conditions for Estimation}{Thm:CVSTEM-LMI-est}
Suppose $\exists \bar{\rho},\bar{c}\in\mathbb{R}_{\geq0}$ \st{} $\|R^{-1}(\hat{x},t)\|\leq\bar{\rho}$, $\|C(\varrho_c,x,\hat{x},t)\|\leq\bar{c},~\forall x,\hat{x},t$. Suppose also that $\underline{m}\mathrm{I}\preceq M \preceq \overline{m} I$ of \eqref{Mcon} holds, or equivalently, $\mathrm{I}\preceq \bar{W} \preceq \chi I$ of \eqref{lambda_con} holds with $W=M(\hat{x},t)^{-1}$, $\bar{W}=\nu W$, $\nu=\overline{m}$, and $\chi=\overline{m}/\underline{m}$. As in Theorem~\ref{THM:Thm:CVSTEM:LMI}, let $\beta$ be defined as $\beta = 0$ for deterministic systems \eqref{sdc_dynamics_est} and 
\begin{align}
\beta &= \alpha_s = \alpha_{e0}+\nu^2\alpha_{e1} = \frac{L_m\bar{g}_{e0}^2}{2}\left(\alpha_G+\frac{1}{2}\right)+\frac{\nu^2L_m\bar{\rho}^2\bar{c}^2\bar{g}_{e1}^2}{2}\left(\alpha_G+\frac{1}{2}\right)
\end{align}
for stochastic systems \eqref{sdc_dynamics_est_sto}, where $2\alpha_{e0} = L_m\bar{g}_{e0}^2(\alpha_G+{1}/{2})$, $2\alpha_{e1} = L_m\bar{\rho}^2\bar{c}^2\bar{g}_{e1}^2(\alpha_G+{1}/{2})$, $L_m$ is the Lipschitz constant of $\partial W/\partial x_i$, $\bar{g}_{e0}$ and $\bar{g}_{e1}$ are given in \eqref{sdc_dynamics_est_sto}, and $\exists\alpha_G\in\mathbb{R}_{>0}$ is an arbitrary constant as in Theorem~\ref{THM:Thm:robuststochastic}. 

If $M(\hat{x},t)$ in \eqref{estimator} is constructed to satisfy the following convex constraint for $\exists \alpha \in \mathbb{R}_{>0}$:
\begin{equation}
\label{convex_constraint_estimator1}
\dot{\bar{W}}+2\sym{}(\bar{W}A-\nu \bar{C}^{\top}R^{-1}C)\preceq -2\alpha \bar{W}-\nu\beta \mathrm{I}
\end{equation}
then Theorems~\ref{THM:Thm:Robust_contraction_original} and \ref{THM:Thm:robuststochastic} hold for the virtual systems \eqref{closed_loop_e_est} and \eqref{closed_loop_e_est_sto}, respectively, \ie{}, we have the following bounds for $\mathtt{e}=\hat{x}-x$ with $\nu=\overline{m}$ and $\chi=\overline{m}/\underline{m}$:
\begin{align}
\label{Eq:CVSTEM_estimation_error_bound}
\|\mathtt{e}(t)\| &\leq \sqrt{\overline{m}}{V_{\ell}(0)}e^{-\alpha t}+\frac{\bar{d}_{e0}\sqrt{\chi}+\bar{\rho}\bar{c}\bar{d}_{e1}\nu}{\alpha}(1-e^{-\alpha t}) \\
\label{Eq:CVSTEM_estimation_error_bound_sto}
\mathop{\mathbb{E}}\left[\|\mathtt{e}(t)\|^2\right] &\leq \overline{m}{\mathop{\mathbb{E}}[V_{s\ell}(0)]}e^{-2\alpha t}+\frac{C_{e0}\chi+C_{e1}\chi\nu^2}{2\alpha}
\end{align}
where $V_{s\ell}=\int^{\hat{x}}_{x}\delta q^{\top}W\delta q$ and $V_{\ell}=\int^{\hat{x}}_{x}\|\Theta\delta q\|$ are given in Theorem~\ref{THM:Thm:path_integral} with $W=M^{-1}=\Theta^{\top}\Theta$ defining a contraction metric, the disturbance bounds $\bar{d}_{e0}$, $\bar{d}_{e1}$, $\bar{g}_{e0}$, and $\bar{g}_{e1}$ are given in \eqref{sdc_dynamics_est} and \eqref{sdc_dynamics_est_sto}, respectively, $C_{e0} = \bar{g}_{e0}^2({2}{\alpha_G}^{-1}+1)$, and $C_{e1} = \bar{\rho}^2\bar{c}^2\bar{g}_{e1}^2({2}{\alpha_G}^{-1}+1)$. Note that for stochastic systems, the probability that $\|\mathtt{e}\|$ is greater than or equal to $\varepsilon\in\mathbb{R}_{> 0}$ is given as
\begin{align}
\label{Eq:CVSTEM_estimation_error_bound_sto_prob}
\mathop{\mathbb{P}}\left[\|\mathtt{e}(t)\|\geq\varepsilon\right] \leq \frac{1}{\varepsilon^2}\left(\overline{m}{\mathop{\mathbb{E}}[V_{s\ell}(0)]}e^{-2\alpha t}+\frac{C_E}{2\alpha}\right)
\end{align}
where $C_E=C_{e0}\chi+C_{e1}\chi\nu^2$.
\end{theorem}
\begin{proof}
Theorem~\ref{THM:Thm:CVSTEM:LMI} indicates that \eqref{convex_constraint_estimator1} is equivalent to
\begin{equation}
\label{constraint_estimator1}
\dot{W}+2\sym{}(WA-\bar{C}^{\top}R^{-1}C)\preceq -2\alpha W-\beta \mathrm{I}.
\end{equation}
Computing the time derivative of a Lyapunov function $V=\partial_{\mu} q^{\top}W\partial_{\mu} q$ with $\partial_{\mu} q=\partial q/\partial \mu$ for the unperturbed virtual dynamics \eqref{eq:differential_dyn_est_sto}, we have using \eqref{constraint_estimator1} that 
\begin{align}
\dot{V}&=\partial_{\mu} q^{\top}W\partial_{\mu} q = \partial_{\mu} q^{\top}(\dot{W}+2WA-2\bar{C}^{\top}R^{-1}C)\partial_{\mu} q \leq -2\alpha V-\beta\|\partial_{\mu} q\|^2
\end{align}
which implies that $W=M^{-1}$ defines a contraction metric. Since we have $\overline{m}^{-1}\mathrm{I}\preceq W \preceq \underline{m}^{-1}\mathrm{I}$, $V\geq\overline{m}^{-1}\|\partial_{\mu} q\|^2$, and
\begin{align}
\|\Theta(\hat{x},t)\partial_{\mu}d\|&\leq \frac{\bar{d}_{e0}}{\sqrt{\underline{m}}}+\bar{d}_{e1}\bar{\rho}\bar{c}\sqrt{\overline{m}} \nonumber \\
\|\partial_{\mu}G\|_F^2&\leq \bar{g}_{e0}^2+\bar{\rho}^2\bar{c}^2\bar{g}_{e1}^2\overline{m}^2 \nonumber
\end{align}
for $d$ in \eqref{closed_loop_e_est} and $G$ in \eqref{closed_loop_e_est_sto}, the bounds \eqref{Eq:CVSTEM_estimation_error_bound}--\eqref{Eq:CVSTEM_estimation_error_bound_sto_prob} follow from the proofs of Theorems~\ref{THM:Thm:Robust_contraction_original} and~\ref{THM:Thm:robuststochastic}~\cite{ncm,nscm}. \qed
\end{proof}
\begin{remark}
\label{convex_est_remark}
Although \eqref{convex_constraint_estimator1} is not an LMI due to the nonlinear term $-\nu\beta \mathrm{I}$ on its right-hand side for stochastic systems~\eqref{sdc_dynamics_est_sto}, it is a convex constraint as $-\nu \beta = -\nu\alpha_s = -\nu\alpha_{e0}-\nu^3\alpha_{e1}$ is a concave function for $\nu\in\mathbb{R}_{>0}$~\cite{citeulike:163662,nscm}.
\end{remark}
\subsubsection{CV-STEM Formulation for State Estimation}
The estimator \eqref{estimator} gives a convex steady-state upper bound of the Euclidean distance between $x$ and $\hat{x}$ as in Theorem~\ref{THM:Thm:CVSTEMobjective}~\cite{mypaperTAC,ncm,nscm,mypaper}.
\begin{theorem}{Convexity of Steady-State Estimation Error Bounds}{Thm:CV-STEMobjective-est}
If \eqref{convex_constraint_estimator1} of Theorem~\ref{THM:Thm:CVSTEM-LMI-est} holds, then we have the following bound:
\begin{align}
\label{stochasticmse_infty_est}
&\lim_{t \to \infty}\sqrt{\mathop{\mathbb{E}}\left[\|\hat{x}-x\|^2\right]} \leq c_0(\alpha,\alpha_G)\chi+c_1(\alpha,\alpha_G)\nu^s
\end{align}
where $c_0=\bar{d}_{e0}/\alpha$, $c_1=\bar{\rho}\bar{c}\bar{d}_{e1}/\alpha$, $s=1$ for deterministic systems \eqref{closed_loop_e_est}, and $c_0=\sqrt{C_{e0}/(2\alpha)}$, $c_1=C_{e1}/(2\sqrt{2\alpha C_{e0}})$, and $s=2$ for stochastic systems \eqref{closed_loop_e_est_sto}, with $C_{e0}$ and $C_{e0}$ given as $C_{e0} = \bar{g}_{e0}^2(2\alpha_G^{-1}+1)$ and $C_{e1} = \bar{\rho}^2\bar{c}^2\bar{g}_{e1}^2(2\alpha_G^{-1}+1)$.
\end{theorem}
\begin{proof}
The upper bound \eqref{stochasticmse_infty_est} for deterministic systems \eqref{closed_loop_e_est} follows from \eqref{Eq:CVSTEM_estimation_error_bound} with the relation $1\leq\sqrt{\chi}\leq \chi$ due to $\underline{m}\leq\overline{m}$. For stochastic systems, we have using \eqref{Eq:CVSTEM_estimation_error_bound_sto} that
\begin{align}
C_{e0}\chi+C_{e1}\nu^2\chi \leq C_{e0}\left(\chi+\frac{C_{e1}}{2C_{e0}}\nu^2\right)^2
\end{align}
due to $1\leq\chi\leq \chi^2$ and $\nu \in \mathbb{R}_{>0}$. This gives \eqref{stochasticmse_infty_est} for stochastic systems \eqref{closed_loop_e_est_sto}. \qed
\end{proof}

Finally, the CV-STEM estimation framework is summarized in Theorem~\ref{THM:Thm:CV-STEM-estimation}~\cite{mypaperTAC,ncm,nscm,mypaper}.
\begin{theorem}{Robust Nonlinear Estimation via Convex Optimization}{Thm:CV-STEM-estimation}
Suppose that $\alpha$, $\alpha_G$, $\bar{d}_{e0}$, $\bar{d}_{e1}$, $\bar{g}_{e0}$, $\bar{g}_{e1}$, and $L_m$ in \eqref{convex_constraint_estimator1} and \eqref{stochasticmse_infty_est} are given. If the pair $(A,C)$ is uniformly observable, the non-convex optimization problem of minimizing the upper bound \eqref{stochasticmse_infty_est} is equivalent to the following convex optimization problem with the contraction constraint \eqref{convex_constraint_estimator1} and $\mathrm{I}\preceq \bar{W}\preceq\chi\mathrm{I}$ of \eqref{lambda_con}:
\begin{align}
\label{cvstem_eq_est}
\begin{aligned}
{J}_{CV}^* &= \min_{\nu\in\mathbb{R}_{>0},\chi \in \mathbb{R},\bar{W} \succ 0} c_0\chi+c_1\nu^s+c_2 P(\chi,\nu,\bar{W}) \\
&\text{\rm\st{}~\eqref{convex_constraint_estimator1} and \eqref{lambda_con}}
\end{aligned}
\end{align}
where $c_0$, $c_1$, and $s$ are as defined in \eqref{stochasticmse_infty_est} of Theorem~\ref{THM:Thm:CV-STEMobjective-est}, $c_2\in\mathbb{R}_{\geq 0}$, and $P$ is some performance-based cost function as in Theorem~\ref{THM:Thm:CV-STEM}.

The weight $c_1$ for $\nu^s$ indicates how much we trust the measurement $y(t)$. Using non-zero $c_2$ enables finding contraction metrics optimal in a different sense in terms of $P$. Furthermore, the coefficients of the SDC parameterizations $\varrho_a$ and $\varrho_c$ in Lemma~\ref{sdclemma}, \ie{}, $A = \sum\varrho_{a,i}A_i$ and $C = \sum\varrho_{c,i}C_i$ in \eqref{est_A} and \eqref{est_C}, can also be treated as decision variables by convex relaxation~\cite{Ref:Stochastic}, thereby adding a design flexibility to mitigate the effects of external disturbances while verifying the system observability.
\end{theorem}
\begin{proof}
The proposed optimization problem is convex as its objective and constraints are convex in terms of decision variables $\chi$, $\nu$, and $\bar{W}$ (see Remark~\ref{convex_est_remark}). Also, larger $\bar{d}_{e1}$ and $\bar{g}_{e1}$ in \eqref{sdc_dynamics_est} and \eqref{sdc_dynamics_est_sto} imply larger measurement uncertainty. Thus by definition of $c_1$ in Theorem~\ref{THM:Thm:CV-STEMobjective-est}, the larger the weight of $\nu$, the less confident we are in $y(t)$ (see Example~\ref{EX:ex:kalmanexample}). The last statement on the SDC coefficients for guaranteeing observability follows from Proposition~1 of \cite{Ref:Stochastic} and Proposition~1 of~\cite{mypaperTAC}. \qed
\end{proof}
\begin{example}{Kalman Filter Analogy}{ex:kalmanexample}
The weights $c_0$ and $c_1$ of the CV-STEM estimation of Theorem~\ref{THM:Thm:CV-STEM-estimation} have an analogous trade-off to the case of the Kalman filter with the process and sensor noise covariance matrices, $Q$ and $P$, respectively, since the term $c_0\chi$ in upper bound of the steady-state tracking error in \eqref{stochasticmse_infty_est} becomes dominant if measurement noise is much smaller than process noise ($\bar{d}_{e0}\gg \bar{d}_{e1}$ or $\bar{g}_{e0}\gg \bar{g}_{e1}$), and the term $c_1\nu^s$ becomes dominant if measurement noise is much greater than process noise ($\bar{d}_{e0}\ll \bar{d}_{e1}$ or $\bar{g}_{e0}\ll \bar{g}_{e1}$). In particular~\cite{ncm,nscm},
\setlength{\leftmargini}{17pt}     
\begin{itemize}
	\setlength{\itemsep}{1pt}      
	\setlength{\parskip}{0pt}      
    \item if $c_1$ is much greater than $c_0$, large measurement noise leads to state estimation that responds slowly to unexpected changes in the measurement $y$, \ie{}, small estimation gain due to $\nu=\overline{m}\geq\|M\|$, and
    \item if $c_1$ is much smaller than $c_0$, large process noise leads to state estimation that responds fast to changes in the measurement, \ie{}, large $\nu=\overline{m}\geq\|M\|$.
\end{itemize}
This is also because the solution $Q = P^{-1}\succ0$ of the Kalman filter Riccati equation~\cite[p. 375]{Brown:680442}, $\dot{P}=AP+PA^{\top}-PC^{\top}R^{-1}CP+Q$, can be viewed as a positive definite matrix that defines a contraction metric as discussed in Example~\ref{EX:ex:lyapunov_contraction}.
\end{example}
\subsection{Control Contraction Metrics (CCMs)}
\label{Sec:ccm}
As briefly discussed in Remark~\ref{u_equivalence_remark}, the concept of a CCM~\cite{ccm,7989693,47710,WANG201944,vccm,rccm} is introduced to extend contraction theory to design differential feedback control laws for control-affine deterministic nonlinear systems \eqref{original_dynamics}, $\dot{x}=f(x,t)+B(x,t)u$. Let us show that the CCM formulation could also be considered in the CV-STEM control of Theorem~\ref{THM:Thm:CV-STEM}, where similar ideas have been investigated in~\cite{cdc_ncm,7989693,zhao2021tubecertified}.
\begin{theorem}{Control Contraction Metrics}{Thm:ccm_cvstem}
Suppose the CV-STEM control of Theorem~\ref{THM:Thm:CV-STEM} is designed with its contraction condition replaced by the following set of convex constraints along with $\mathrm{I}\preceq \bar{W}\preceq\chi\mathrm{I}$ of \eqref{lambda_con}:
\begin{align}
&B_{\bot}^{\top} \left( -\frac{\partial \bar{W}}{\partial t}-\partial_f \bar{W} + 2\sym{}\left({\frac{\partial f}{\partial x} \bar{W}}\right) + 2 \alpha \bar{W} \right)B_{\bot} \prec 0\label{ccm_con1}~~~~ \\
&B_{\bot}^{\top} \left( \partial_{b_i} \bar{W} - 2\sym{}\left({\frac{\partial b_i}{\partial x} \bar{W}}\right) \right) B_{\bot} = 0,\forall x,t,i
\label{ccm_con2}
\end{align}
where $B_{\bot}(x,t)$ is a matrix whose columns span the cokernel of $B(x,t)$ defined as $\mathrm{coker}(B)= \{a\in\mathbb{R}^n|B^{\top}a=0\}$ satisfying $B^{\top}B_{\bot} = 0$, $b_i(x,t)$ is the $i$th column of $B(x,t)$, $\bar{W}(x,t)=\nu W(x,t)$, $W(x,t) = M(x,t)^{-1}$ for $M(x,t)$ that defines a contraction metric, $\nu=\overline{m}$, $\chi=\overline{m}/\underline{m}$, and $\partial_{p} F = \sum_{k=1}^n(\partial F/\partial x_k)p_k$ for $p(x,t)\in\mathbb{R}^n$ and $F(x,t) \in \mathbb{R}^{n\times n}$. 
Then the controlled system \eqref{original_dynamics} with $d_c=0$, \ie{}, $\dot{x}=f(x,t)+B(x,t)u$, is
\setlength{\leftmargini}{17pt}     
\begin{enumerate}[label=\arabic*)]
	\setlength{\itemsep}{1pt}      
	\setlength{\parskip}{0pt}     
    \item universally exponentially open-loop controllable
    \item universally exponentially stabilizable via sampled-data feedback with arbitrary sample times
    \item universally exponentially stabilizable using continuous feedback defined almost everywhere, and everywhere in a neighborhood of the target trajectory
\end{enumerate}
all with rate $\alpha$ and overshoot $R = \sqrt{\overline{m}/\underline{m}}$ for $\underline{m}$ and $\overline{m}$ given in $\underline{m}\mathrm{I}\preceq M\preceq\overline{m}\mathrm{I}$ of \eqref{Mcon}. Such a positive definite matrix $M(x,t)$ defines a CCM.

Given a CCM, there exists a differential feedback controller $\delta u = k(x,\delta x,u,t)$ that stabilizes the following differential dynamics of \eqref{original_dynamics} with $d_c=0$ along all solutions (\ie{}, the closed-loop dynamics is contracting as in Definition~\ref{DEF:Def:contraction}):
\begin{align}
\label{ccm_differential_eq}
\delta\dot{x} = A(x,u,t)\delta x+B(x,t)\delta u
\end{align}
where $A=\partial f/\partial x+\sum_{i=1}^m(\partial b_i/\partial x)u_i$ and $\delta u$ is a tangent vector to a smooth path of controls at $u$. Furthermore, it can be computed as follows~\cite{ccm,7989693,rccm}:
\begin{align}
\label{ccm_controller}
u(x(t),t) = u_d+\int_{0}^1k(\gamma(\mu,t),\partial_{\mu}\gamma(\mu,t),u(\gamma(\mu,t),t),t)d\mu~~~~
\end{align}
where $\partial_{\mu}\gamma=\partial \gamma/\partial \mu$, $\gamma$ is the minimizing geodesic with $\gamma(0,t) = x_d$ and $\gamma(1,t) = x$ for $\mu\in[0,1]$, $(x_d,u_d)$ is a given target trajectory in $\dot{x}_d=f(x_d,t)+B(x_d,t)u_d$ of \eqref{sdc_dynamicsd}, and the computation of $k$ is given in~\cite{ccm,7989693} (see Remark~\ref{remark_ccm_kgamma}). Furthermore, Theorem~\ref{THM:Thm:Robust_contraction_original} for deterministic disturbance rejection still holds and the CCM controller \eqref{ccm_controller} thus possesses the same sense of oplimality as for the CV-STEM control in Theorem~\ref{THM:Thm:CV-STEM}.
\end{theorem}
\begin{proof}
Since the columns of $B_{\bot}(x,t)$ span the cokernel of $B(x,t)$, the constraints \eqref{ccm_con1} and \eqref{ccm_con2} can be equivalently written as $B_{\bot}(x,t)$ replaced by $a$ in $\mathrm{coker}(B)= \{a\in\mathbb{R}^n|B^{\top}a=0\}$. Let $a=M\delta x=W^{-1}\delta x$. Then multiplying \eqref{ccm_con1} and \eqref{ccm_con2} by $\nu^{-1} >0$ and rewriting them using $a$ yields
\begin{align}
\label{ccm_condition}
\delta x^{\top}MB=0~\Rightarrow~\delta x^{\top}(\dot{M}+2\sym(MA)+2\alpha M)\delta x < 0~~~~~~
\end{align}
where $A=\partial f/\partial x+\sum_{i=1}^m(\partial b_i/\partial x)u_i$ for the differential dynamics \eqref{ccm_differential_eq}. The relation \eqref{ccm_condition} states that $\delta x$ orthogonal to the span of actuated directions $b_i$ is naturally contracting, thereby implying the stabilizability of the system \eqref{original_dynamics} with $d_c=0$, \ie{}, $\dot{x}=f(x,t)+B(x,t)u$. See~\cite{ccm,7989693,rccm} (and~\cite{cdc_ncm} on the CV-STEM formulation) for the rest of the proof. \qed
\end{proof}
\begin{example}{Control Contraction Metrics I}{}
Let us consider a case where $k$ of the differential feedback controller does not depend on $u$, and is defined explicitly by $M(x,t) \succ 0$ as follows~\cite{cdc_ncm}:
\begin{align}
\label{differential_u}
&u = u_d-\int_{x_d}^xR(q,t)^{-1}B(q,t)^{\top}M(q,t)\delta q
\end{align}
where $R(x,t) \succ 0$ is a given weight matrix and $q$ is a smooth path that connects $x$ to $x_d$. Since \eqref{differential_u} yields $\delta u=-R(x,t)^{-1}B(x,t)^{\top}M(x,t)\delta x$, the contraction conditions \eqref{ccm_con1} and \eqref{ccm_con2} could be simplified as
\begin{align}
\label{ccm_con1_simple}
&-\frac{\partial \bar{W}}{\partial t}-\partial_f \bar{W}+2\sym{}\left(\frac{\partial f}{\partial x}\bar{W}\right)-\nu BR^{-1}B^{\top} \preceq -2\alpha \bar{W} \\
\label{ccm_con2_simple}
&-\partial_{b_i}\bar{W}+2\sym{}\left(\frac{\partial b_i}{\partial x}\bar{W}\right) = 0
\end{align}
yielding the convex optimization-based control synthesis algorithm of Theorem~\ref{THM:Thm:ccm_cvstem} independently of $(x_d,u_d)$ similar to that of Theorem~\ref{THM:Thm:CV-STEM} (see~\cite{cdc_ncm} for details).
\end{example}
\begin{example}{Control Contraction Metrics II}{}
If $B$ is of the form $[0,\mathrm{I}]^{\top}$ for the zero matrix $0\in\mathbb{R}^{n_1\times m}$ and identity matrix $\mathrm{I}\in\mathbb{R}^{n_2\times m}$ with $n=n_1+n_2$, the condition \eqref{ccm_con2_simple} says that $M$ should not depend on the last $n_2$ state variables~\cite{ccm}.
\end{example}
\begin{remark}
\label{remark_ccm_kgamma}
We could consider stochastic perturbation in Theorem~\ref{THM:Thm:ccm_cvstem} using Theorem~\ref{THM:Thm:robuststochastic}, even with the differential control law of the form \eqref{ccm_controller} or \eqref{differential_u} as demonstrated in~\cite{cdc_ncm}. Also, although the relation \eqref{ccm_con2} or \eqref{ccm_con2_simple} is not included as a constraint in Theorem~\ref{THM:Thm:CV-STEM} for simplicity of discussion, the dependence of $\dot{\bar{W}}$ on $u$ in Theorem~\ref{THM:Thm:CV-STEM} can be removed by using it in a similar way to~\cite{cdc_ncm}. 

As stated in \eqref{ccm_controller}, the computation of the differential feedback gain $k(x,\delta x,u,t)$ and minimizing geodesics $\gamma$ is elaborated in~\cite{ccm,7989693}. For example, if $M$ is state-independent, then geodesics are just straight lines.
\end{remark}

Let us again emphasize that, as delineated in Sec.~\ref{Sec:overview}, the differences between the SDC- and CCM-based CV-STEM frameworks in Theorems~\ref{THM:Thm:CV-STEM} and~\ref{THM:Thm:ccm_cvstem} arise only from their different form of controllers in $u=u_d-K(x-x_d)$ of \eqref{controller} and $u=u_d+\int^1_0kd\mu$ of \eqref{ccm_controller}, leading to the trade-offs outlined in Table~\ref{tab:sdcccm_summary}.
\subsection{Remarks in CV-STEM Implementation}
We propose some useful techniques for the practical application of the CV-STEM in Theorems~\ref{THM:Thm:CV-STEM},~\ref{THM:Thm:CV-STEM-estimation}, and~\ref{THM:Thm:ccm_cvstem}. Note that the CV-STEM requires solving a convex optimization problem at each time instant, but its solution can be approximated with formal stability guarantees to enable faster computation using machine learning techniques as shall be seen in Sec.~\ref{Sec:ncm}.
\subsubsection{Performance-based Cost Function}\label{sec:ChoiceObJ}
Selecting $c_2=0$ in Theorems~\ref{THM:Thm:CV-STEM}, \ref{THM:Thm:CV-STEM-estimation}, and~\ref{THM:Thm:ccm_cvstem} yields a convex objective function that allows for a systematic interpretation of its weights as seen earlier in Examples~\ref{EX:ex:lqrexample} and~\ref{EX:ex:kalmanexample}. One could also select $c_2>0$ to augment the CV-STEM with other control and estimation performances of interest, as long as $P(\nu,\chi,\bar{W})$ in \eqref{cvstem_eq} and \eqref{cvstem_eq_est} is convex. For example, we could consider a steady-state tracking error bound of parametric uncertain systems as $P$, using the adaptive control technique to be discussed in Sec.~\ref{Sec:adaptive}~\cite{ancm}. Such a modification results in a contraction metric optimal in a different sense.
\subsubsection{Selecting and Computing CV-STEM Parameters}
\label{given_params}
The CV-STEM optimization problems derived in Theorems~\ref{THM:Thm:CV-STEM},~\ref{THM:Thm:CV-STEM-estimation}, and~\ref{THM:Thm:ccm_cvstem} are convex if we assume that $\alpha$, $\alpha_G$, and $L_m$ are given. However, these parameters would also affect the optimality of resultant contraction metrics. In~\cite{ncm,nscm,7989693}, a line search algorithm is performed to find optimal $\alpha$ and $\alpha_G$, while the Lipschitz constraint given with $L_m$ is guaranteed by spectrally-normalization~\cite{miyato2018spectral,nscm} as shall be seen in detail in Sec.~\ref{Sec:SN} (see~\cite{revay2020lipschitz,revay2021recurrent,cdc_systemid} for contraction theory-based techniques for obtaining Lipschitz bounds). Also, the CV-STEM can be formulated as a finite-dimensional problem by using backward difference approximation on $\dot{\bar{W}}$, where we can then use $-\bar{W} \preceq -\mathrm{I}$ to get a sufficient condition of its constraints, or we could alternatively solve it along pre-computed trajectories $\{x(t_i)\}_{i=0}^M$ as in~\cite{ncm}. In Sec.~\ref{Sec:ncm}, we use a parameterized function such as neural networks~\cite{neural1,neural2,neural3} for approximating $M$ to explicitly compute $\dot{M}$.

\newpage
\vspace*{20em}
\part*{\huge \normalfont{Part II: Learning-based Control}}
\label{part2LBC}
\vspace{2em}
\newpage
\section{Contraction Theory for Learning-based Control}
\label{Sec:learning_stability}
Machine learning techniques, \eg{}, reinforcement learning~\cite{sutton,ndp,8593871,NIPS2017_766ebcd5}, imitation learning~\cite{9001182,glas,NIPS2016_cc7e2b87,8578338,7995721}, and neural networks~\cite{neural1,neural2,neural3}, have gained popularity due to their ability to achieve a large variety of innovative engineering and scientific tasks which have been impossible heretofore. Starting from this section, we will see how contraction theory enhances learning-based and data-driven automatic control frameworks providing them with formal optimality, stability, and robustness guarantees. In particular, we present Theorems~\ref{THM:Thm:contraction_learning} and~\ref{THM:Thm:contraction_learning_sto} for obtaining robust exponential bounds on trajectory tracking errors of nonlinear systems in the presence of learning errors, whose steady-state terms are again written as a function of the condition number $\chi=\overline{m}/\underline{m}$ of a positive definite matrix $M$ that defines a contraction metric, consistently with the CV-STEM frameworks of Sec.~\ref{sec:convex}. 
\subsection{Problem Formulation}
Let us consider the following virtual nonlinear system as in Theorem~\ref{THM:Thm:partial_contraction}:
\begin{align}
    \label{learning_x}
    \dot{q}(\mu,t) &= \textsl{g}(q(\mu,t),\varpi,t)+\mu\Delta_L(\varpi,t)+d(\mu,\varpi,t)
\end{align}
where $\mu\in[0,1]$, $q:[0,1]\times\mathbb{R}_{\geq0}\mapsto\mathbb{R}^n$ is a smooth path of the system states, $\varpi:\mathbb{R}_{\geq0}\mapsto\mathbb{R}^{p}$ is a time-varying parameter, $\textsl{g}:\mathbb{R}^n\times\mathbb{R}^p\times\mathbb{R}_{\geq 0}\mapsto\mathbb{R}^n$ is a known smooth function which renders $\dot{q} = \textsl{g}(q,\varpi,t)$ contracting with respect to $q$, $d:[0,1]\times\mathbb{R}^p\times\mathbb{R}_{\geq 0}\mapsto\mathbb{R}^n$ with $\sup_{\mu,\varpi,t}\|\partial d/\partial \mu\|=\bar{d}\in\mathbb{R}_{\geq 0}$ is unknown external disturbance parameterized by $\mu$ as in Theorem~\ref{THM:Thm:Robust_contraction_original}, and $\Delta_L:\mathbb{R}^p\times\mathbb{R}_{\geq 0}\mapsto\mathbb{R}^n$ is the part to be learned with machine learning-based methodologies such as neural networks~\cite{neural1,neural2,neural3}. We can also formulate this in a stochastic setting as follows:
\begin{align}
\label{learning_x_sto}
d{q}(\mu,t) &= (\textsl{g}(q(\mu,t),\varpi,t)+\mu\Delta_L(\varpi,t))dt+G(\mu,\varpi,t)d\mathscr{W}
\end{align}
where $\mathscr{W}:\mathbb{R}_{\geq0}\mapsto\mathbb{R}^{w}$ is a Wiener process, and $G:[0,1]\times\mathbb{R}^p\times\mathbb{R}_{\geq 0} \rightarrow\mathbb{R}^{n\times w}$ is a matrix-valued function parameterized by $\mu$ with $\sup_{\mu,\varpi,t}\|\partial G/\partial \mu\|_F = \bar{g}\in\mathbb{R}_{\geq 0}$ as in Theorem~\ref{THM:Thm:robuststochastic}. The objective of learning is to find $\Delta_L$ which satisfies the following, assuming that $\varpi\in\mathcal{S}_{\varpi}\subseteq\mathbb{R}^p$ and $t\in\mathcal{S}_t\subseteq\mathbb{R}_{\geq 0}$ for some compact sets $\mathcal{S}_{\varpi}$ and $\mathcal{S}_t$:
\begin{align}
\label{Eq:learning_error}
\|\Delta_L(\varpi,t)\| \leq \epsilon_{\ell0}+\epsilon_{\ell1}\|\xi_1-\xi_0\|,~\forall (\varpi,t) \in \mathcal{S}~~~~
\end{align}
where $\mathcal{S}=\mathcal{S}_{\varpi}\times\mathcal{S}_t$, $\xi_0=q(0,t)$ and $\xi_1=q(1,t)$ are particular solutions of \eqref{learning_x} and \eqref{learning_x_sto}, and $\epsilon_{\ell0},\epsilon_{\ell1}\in\mathbb{R}_{\geq 0}$ are given learning errors.

Examples~\ref{EX:ex:learning_prob1}--\ref{EX:ex:learning_prob3} illustrate how the problem \eqref{Eq:learning_error} with the nonlinear systems \eqref{learning_x} and \eqref{learning_x_sto} can be used to describe several learning-based and data-driven control problems to be discussed in the subsequent sections, regarding $\varpi$ as $x$, $x_d$, $\hat{x}$, $u_d$, etc.
\begin{example}{Learning-based Tracking Control}{ex:learning_prob1}
Let us consider the problem of learning a computationally-expensive (or unknown) feedback control law $u^*(x,x_d,u_d,t)$ that tracks a target trajectory $(x_d,u_d)$ given by $\dot{x}_d=f(x_d,t)+B(x_d,t)u_d$ as in \eqref{sdc_dynamicsd}, assuming that $f$ and $B$ are known. If the dynamics is perturbed by $d_c(x,t)$ as in \eqref{original_dynamics}, we have that
\begin{align}
\dot{x} &= f(x,t)+B(x,t)u_L+d_c(x,t) = f(x,t)+B(x,t)u^*+B(x,t)(u_L-u^*)+d_c(x,t)
\end{align}
where $u_L(x,x_d,u_d,t)$ denotes a learning-based control law that models $u^*(x,x_d,u_d,t)$. As long as $u^*$ renders the closed-loop dynamics $\dot{x}=f(x,t)+B(x,t)u^*(x,x_d,u_d,t)$ contracting with respect to $x$~\cite{mypaperTAC,ccm,ncm,7989693}, we can define the functions of \eqref{learning_x} as follows:
\begin{align}
\label{ex_control_l_1}
\textsl{g}(q,\varpi,t) &= f(q,t)+B(q,t)u^*(q,x_d,u_d,t) \\
\label{ex_control_l_2}
\Delta_L(\varpi,t) &= B(x,t)(u_L(x,x_d,u_d,t)-u^*(x,x_d,u_d,t))
\end{align}
with $d(\mu,\varpi,t)=\mu d_c(x,t)$, where $\varpi=[x^{\top},x_d^{\top},u_d^{\top}]^{\top}$ in this case. It can be easily verified that \eqref{learning_x} indeed has $q(\mu=0,t)=x_d$ and $q(\mu=1,t)=x$ as particular solutions if $u^*(x_d,x_d,u_d,t)=u_d$. Similarly, we can use \eqref{learning_x_sto} with $G=\mu G_c(x,t)$ if the dynamics is stochastically perturbed by $G_c(x,t)d\mathscr{W}$ as in \eqref{sdc_dynamics}.

The learning objective here is to make $\|u_L(x,x_d,u_d,t)-u^*(x,x_d,u_d,t)\|$ as small as possible, which aligns with the aforementioned objective in \eqref{Eq:learning_error}. Note that if $\|B\|$ is bounded in the compact set $\mathcal{S}$ of \eqref{Eq:learning_error}, we can bound $\|\Delta_L\|$ of \eqref{ex_control_l_2} with $\epsilon_{\ell0}\neq 0$ and $\epsilon_{\ell1}=0$ as to be explained in Remark~\ref{remark_learning_error}. See Theorems~\ref{THM:Thm:NCMstability1}, \ref{THM:Thm:NCMstability2}, and \ref{THM:Thm:lagros_stability} for details.
\end{example}
\begin{example}{Learning-based State Estimation}{ex:learning_prob2}
Next, let us consider the problem of learning a computationally-expensive (or unknown) state estimator, approximating its estimation gain $L(\hat{x},t)$ by $L_L(\hat{x},t)$. If there is no disturbance in \eqref{sdc_dynamics_est} and \eqref{sdc_dynamics_est_sto}, we have that
\begin{align}
\dot{\hat{x}} &= f(\hat{x},t)+L_L(h(x,t)-h(\hat{x},t)) \\
&= f(\hat{x},t)+L(h(x,t)-h(\hat{x},t))+(L_L-L)(h(x,t)-h(\hat{x},t))
\end{align}
where $\dot{x}=f(x,t)$ is the true system, $y=h(x,t)$ is the measurement, and we assume that $f$ and $h$ are known. If $L$ is designed to render $\dot{q}=f(q,t)+L(\hat{x},t)(h(x,t)-h(q,t))$ contracting with respect to $q$~\cite{ncm,mypaperTAC,Ref:Stochastic,6849943}, we could define the functions of \eqref{learning_x} and \eqref{learning_x_sto} as follows:
\begin{align}
\label{ex_estimation_l_1}
\textsl{g}(q,\varpi,t) &= f(q,t)+L(\hat{x},t)(h(x,t)-h(q,t)) \\
\label{ex_estimation_l_2}
\Delta_L(q,\varpi,t) &= (L_L(\hat{x},t)-L(\hat{x},t))(h(x,t)-h(\hat{x},t))
\end{align}
where $\varpi=[x^{\top},\hat{x}^{\top}]^{\top}$. It can be seen that \eqref{learning_x} and \eqref{learning_x_sto} with the relations \eqref{ex_estimation_l_1} and \eqref{ex_estimation_l_2} indeed has $q(\mu=0,t)=x$ and $q(\mu=1,t)=\hat{x}$ as its particular solutions when perturbed by deterministic and stochastic disturbances as in \eqref{sdc_dynamics_est} and \eqref{sdc_dynamics_est_sto}, respectively. We can bound $\|\Delta_L\|$ of \eqref{ex_estimation_l_2} in the compact set of \eqref{Eq:learning_error} with $\epsilon_{\ell0}=0$ and $\epsilon_{\ell1}\neq0$ if $h$ is Lipschitz, and with $\epsilon_{\ell0}\neq0$ and $\epsilon_{\ell1}=0$ if $h$ is bounded in $\mathcal{S}$, using the techniques to be discussed in Remark~\ref{remark_learning_error}. See Theorems~\ref{THM:Thm:NCMstability1_est} and~\ref{THM:Thm:NCMstability2} for details.
\end{example}
\begin{example}{Learning-based System Identification}{ex:learning_prob3}
We can use the problem \eqref{Eq:learning_error} with the systems \eqref{learning_x} and \eqref{learning_x_sto} also if $f_{\mathrm{true}}$ of the underlying dynamics $\dot{x}^* = f_{\mathrm{true}}(x^*,t)$ is unknown and learned by $\dot{x} = f_L(x,t)$. Since we have
\begin{align}
\dot{x}^* = f_{\mathrm{true}}(x^*,t) = f_L(x^*,t)+(f_{\mathrm{true}}(x^*,t)-f_L(x,t))
\end{align}
we could define $\textsl{g}$ and $\Delta_L$ of \eqref{learning_x} and \eqref{learning_x_sto} as follows, to have $q(\mu=0,t)=x$ and $q(\mu=1,t)=x^*$ as its particular solutions:
\begin{align}
\label{ex_systemID_l_2}
\begin{aligned}
    \textsl{g}(q,\varpi,t) &= f_L(q,t) \\
    \Delta_L(\varpi,t) &= f_{\mathrm{true}}(x^*,t)-f_L(x^*,t)
\end{aligned}
\end{align}
where $\varpi=x^*$, as long as $\dot{x} = f_L(x,t)$ is contracting with respect to $x$~\cite{47710,boffi2020learning}. Since $\Delta_L$ of \eqref{ex_systemID_l_2} is the learning error itself, $\|\Delta_L\|$ can be bounded in $\mathcal{S}$ using the techniques of Remark~\ref{remark_learning_error}. See Theorems~\ref{THM:Thm:NCMstability_modelfree} and \ref{THM:Thm:neurallander} for details.
\end{example}
\begin{remark}
\label{remark_learning_error}
As seen in Examples~\ref{EX:ex:learning_prob1}--\ref{EX:ex:learning_prob2}, $\Delta_L$ of \eqref{Eq:learning_error} is typically given by a learning error, $\phi(z)-\varphi(z)$, multiplied by a bounded or Lipschitz continuous function, where $z=(\varpi,t)\in\mathcal{S}$ for a compact set $\mathcal{S}$, and $\phi$ is the true computationally-expensive/unknown function to be learned by $\varphi$ as in \eqref{ex_control_l_2}, \eqref{ex_estimation_l_2}, \eqref{ex_systemID_l_2}, etc.

Let $\mathcal{T}$ be a set of training data $\{(z_i,\phi(z_i))\}_{i=1}^N$ sampled in $\mathcal{S}$. For systems with the true function $\phi$ and its approximation $\varphi$ being Lipschitz (\eg{}, by spectral normalization~\cite{miyato2018spectral} to be discussed in Definition~\ref{DEF:Def:SN} or by using contraction theory as in~\cite{revay2020lipschitz,revay2021recurrent,cdc_systemid}), we can analytically find a bound for $\|\phi(z)-\varphi(z)\|$ as follows if target data samples $z$ are in $B(r) = \{z\in\mathcal{S}|\sup_{z'\in\mathcal{T}}\|z-z'\|\leq r\}$ for some $r \in\mathbb{R}_{\geq0}$:
\begin{align}
\label{marginbound}
\sup_{z\in B(r)}\|\phi(z)-\varphi(z)\| \leq \sup_{z'\in \mathcal{T}}\|\phi(z')-\varphi(z')\|+(L_{\phi}+L_{\varphi})r
\end{align}
where $L_{\phi}$ and $L_{\varphi}$ are the Lipschitz constants of $\phi$ and $\varphi$, respectively. The term $\sup_{z'\in \mathcal{T}}\|\phi(z')-\varphi(z')\|$ can then be bounded by a constant, \eg{}, by using a deep robust regression model as proven in~\cite{pmlr-v51-chen16d,nnbound} with spectral normalization, under standard training data distribution assumptions.

Deep Neural Networks (DNNs) have been shown to generalize well to the set of unseen events that are from almost the same distribution as their training set~\cite{rethinknet,dnnlearn,neurallander,DBLP:conf/uai/DziugaiteR17,NIPS2017_10ce03a1}, and consequently, obtaining a tighter and more general upper bound for the learning error as in \eqref{marginbound} has been an active field of research~\cite{marginbounds1,marginbounds2,nnbound,pmlr-v51-chen16d}. Thus, the condition \eqref{Eq:learning_error} has become a common assumption in analyzing the performance of learning-based and data-driven control techniques~\cite{neurallander,8967820,pmlr-v97-cheng19a,8651519,lagros,ancm,cdc_ncm,swarm1,swarm2}.
\end{remark}
\subsection{Formal Stability Guarantees via Contraction Theory}
One drawback of naively using existing learning-based and data-driven control approaches for the perturbed nonlinear systems \eqref{learning_x} and \eqref{learning_x_sto} without analyzing contraction is that, as shall be seen in the following theorem, we can only guarantee the trajectory error to be bounded by a function that increases exponentially with time.
\begin{theorem}{Naive Learning-based Control}{Thm:naive_learning}
Suppose that $\Delta_L$ of \eqref{learning_x} is learned to satisfy \eqref{Eq:learning_error}, and that $\textsl{g}(q,\varpi,t)$ of \eqref{learning_x} is Lipschitz with respect to $q$ with its 2-norm Lipschitz constant $L_{\textsl{g}}\in\mathbb{R}_{\geq 0}$, \ie{},
\begin{align}
\|\textsl{g}(q,\varpi,t)-\textsl{g}(q',\varpi,t)\| \leq L_{\textsl{g}}\|q-q'\|,~\forall (\varpi,t) \in \mathcal{S}
\end{align}
where $q,q' \in \mathbb{R}^n$ and $\mathcal{S}$ is the compact set of \eqref{Eq:learning_error}. Then we have the following upper bound for all $(\varpi,t) \in \mathcal{S}$:
\begin{align}
\label{naive_learning_error}
\|\mathtt{e}(t)\| &\leq \|\mathtt{e}(0)\|e^{(L_{\textsl{g}}+\epsilon_{\ell1})t}+\frac{\epsilon_{\ell0}+\bar{d}}{L_{\textsl{g}}+\epsilon_{\ell1}}(e^{(L_{\textsl{g}}+\epsilon_{\ell1})t}-1)
\end{align}
where $\mathtt{e}(t) = \xi_1(t)-\xi_0(t)$, $\xi_0(t)=q(0,t)$ and $\xi_1(t)=q(1,t)$ for $q$ of \eqref{learning_x}, and $\bar{d}=\sup_{\mu,\varpi,t}\|\partial d/\partial \mu\|$ for $d$ of \eqref{Eq:learning_error}.
\end{theorem}
\begin{proof}
See the Gronwall-Bellman lemma~\cite[p. 651]{Khalil:1173048} and Theorem~3.4 of~\cite[pp. 96-97]{Khalil:1173048}. \qed
\end{proof}

The bound obtained in Theorem~\ref{THM:Thm:naive_learning} is useful in that it gives mathematical guarantees even for naive learning-based frameworks without a contracting property (\eg{}, it can be used to prove safety in the learning-based Model Predictive Control (MPC) framework~\cite{learningmpc}). However, the exponential term $e^{(L_{\textsl{g}}+\epsilon_{\ell1})t}$ in \eqref{naive_learning_error} causes the upper bound to diverge, which could result in more conservative automatic control designs than necessary.

In contrast, contraction theory gives an upper bound on the trajectory tracking error $\|\mathtt{e}(t)\|$ which is exponentially bounded linearly in the learning error, even under the presence of external disturbances~\cite{ncm,ancm,lagros,cdc_ncm}.
\begin{theorem}{Learning-based Control with Deterministic Contraction}{Thm:contraction_learning}
Let us consider the virtual system of a smooth path $q(\mu,t)$ in \eqref{learning_x} and suppose that $\Delta_L$ of \eqref{learning_x} is learned to satisfy \eqref{Eq:learning_error}. If the condition $\underline{m}\mathrm{I}\preceq M\preceq\overline{m}\mathrm{I}$ of \eqref{Mcon} holds and the system \eqref{learning_x} with $\Delta_L=0$ and $d=0$ is contracting, \ie{},
\begin{align}
\dot{M}+M\frac{\partial \textsl{g}}{\partial q}+\frac{\partial \textsl{g}}{\partial q}^{\top}M \preceq -2\alpha M
\end{align}
of Theorem~\ref{THM:Thm:contraction} or \ref{THM:Thm:partial_contraction} holds for $M\succ 0$ that defines a contraction metric with the contraction rate $\alpha$, and if the learning error $\epsilon_{\ell1}$ of \eqref{Eq:learning_error} is sufficiently small to satisfy
\begin{align}
\label{epsilon_ell2_condition}
\exists \alpha_{\ell} \in\mathbb{R}_{>0}\text{ \st{} }\alpha_{\ell}=\alpha-\epsilon_{\ell1}\sqrt{\frac{\overline{m}}{\underline{m}}} > 0
\end{align}
then we have the following bound for all $(\varpi,t) \in \mathcal{S}$:
\begin{equation}
\label{cont_learning_bound}
\|\mathtt{e}(t)\| \leq \frac{V_{\ell}(0)}{\sqrt{\underline{m}}}e^{-\alpha_{\ell} t}+\frac{\epsilon_{\ell0}+\bar{d}}{\alpha_{\ell}}\sqrt{\frac{\overline{m}}{\underline{m}}}(1-e^{-\alpha_{\ell} t})
\end{equation}
where $\mathtt{e}(t)=\xi_1(t)-\xi_0(t)$ with $\xi_0(t)=q(0,t)$ and $\xi_1(t)=q(1,t)$ for $q$ of \eqref{learning_x}, $\bar{d}=\sup_{\mu,\varpi,t}\|\partial d/\partial \mu\|$ as in \eqref{learning_x}, $\mathcal{S}$ is the compact set of Theorem~\ref{THM:Thm:naive_learning}, $\epsilon_{\ell0}$ and $\epsilon_{\ell1}$ are the learning errors of \eqref{Eq:learning_error}, and $V_{\ell}(t) = \int_{\xi_0}^{\xi_1} \|\Theta(q(t),t)\delta q(t)\|$ for $M=\Theta^{\top}\Theta$ as in Theorem~\ref{THM:Thm:path_integral}.
\end{theorem}
\begin{proof}
Since \eqref{learning_x} with $\Delta_L=0$ and $d=0$ is contracting and we have that 
\begin{align}\label{lip_learning_error_theta}
\begin{aligned}
\left\|\frac{\partial (\Delta_L+d)}{\partial \mu}\right\| &\leq \epsilon_{\ell0}+\epsilon_{\ell1}\|\xi_1-\xi_0\|+\bar{d} \\
&\leq \epsilon_{\ell0}+\frac{\epsilon_{\ell1}}{\sqrt{\underline{m}}}\int_{\xi_0}^{\xi_1}\|\Theta \delta q\|+\bar{d}=\epsilon_{\ell0}+\frac{\epsilon_{\ell1}}{\sqrt{\underline{m}}}V_{\ell}+\bar{d} 
\end{aligned}
\end{align}
for all $\mu\in[0,1]$ and $(\varpi,t)\in\mathcal{S}$ due to $\sup_{\mu,\varpi,t}\|\partial d/\partial \mu\|=\bar{d}$, the direct application of Theorem~\ref{THM:Thm:Robust_contraction_original} to the system \eqref{learning_x}, along with the condition \eqref{epsilon_ell2_condition}, yields \eqref{cont_learning_bound}. \qed
\end{proof}

Using Theorem~\ref{THM:Thm:robuststochastic}, we can easily derive a stochastic counterpart of Theorem~\ref{THM:Thm:contraction_learning} for the system \eqref{learning_x_sto}~\cite{nscm,cdc_ncm}.
\begin{theorem}{Learning-based Control with Stochastic Contraction}{Thm:contraction_learning_sto}
Consider the virtual system of a smooth path $q(\mu,t)$ in \eqref{learning_x_sto} and suppose $\Delta_L$ of \eqref{learning_x_sto} is learned to satisfy \eqref{Eq:learning_error}. If $\underline{m}\mathrm{I}\preceq M\preceq\overline{m}\mathrm{I}$ of \eqref{Mcon} holds and the system \eqref{learning_x_sto} is contracting, \ie{},
\begin{align}
\dot{M}+M\frac{\partial \textsl{g}}{\partial q}+\frac{\partial \textsl{g}}{\partial q}^{\top}M \preceq -2\alpha M-\alpha_s \mathrm{I}
\end{align}
of Theorem~\ref{THM:Thm:robuststochastic} holds, and if the learning error $\epsilon_{\ell1}$ of \eqref{Eq:learning_error} and an arbitrary constant $\alpha_d \in \mathbb{R}_{>0}$ (see \eqref{alpha_d_def}) are selected to satisfy
\begin{align}
\label{epsilon_ell2_condition_sto}
\exists \alpha_{\ell} \in\mathbb{R}_{>0}\text{ \st{} }\alpha_{\ell}=\alpha-\left(\frac{\alpha_d}{2}+\epsilon_{\ell1}\sqrt{\frac{\overline{m}}{\underline{m}}}\right) > 0
\end{align}
then we have the following bound for all $(\varpi,t) \in \mathcal{S}$:
\begin{align}
\label{cont_learning_bound_sto}
\mathop{\mathbb{E}}\left[\|\mathtt{e}(t)\|^2\right] \leq \frac{\mathop{\mathbb{E}}[V_{s\ell}(0)]}{\underline{m}}e^{-2\alpha_{\ell} t}+\frac{C}{2\alpha_{\ell}}\frac{\overline{m}}{\underline{m}}
\end{align}
where $\mathtt{e}(t)=\xi_1(t)-\xi_0(t)$ with $\xi_0(t)=q(0,t)$ and $\xi_1(t)=q(1,t)$ for $q$ of \eqref{learning_x_sto}, $\mathcal{S}$ is the compact set given in Theorem~\ref{THM:Thm:naive_learning}, $\epsilon_{\ell0}$ and $\epsilon_{\ell1}$ are the learning errors of \eqref{Eq:learning_error}, $V_{s\ell}(t) = \int_{\xi_0}^{\xi_1} \|\Theta(q(t),t)\delta q(t)\|^2$ for $M=\Theta^{\top}\Theta$ as given in Theorem~\ref{THM:Thm:path_integral}, and $C = \bar{g}^2({2}{\alpha_G}^{-1}+1)+\epsilon_{\ell0}^2\alpha_d^{-1}$ for $\bar{g}=\sup_{\mu,\varpi,t}\|\partial G/\partial \mu\|_F$ of \eqref{learning_x_sto} with an arbitrary constant $\alpha_G \in \mathbb{R}_{>0}$ as in Theorem~\ref{THM:Thm:robuststochastic}. Furthermore, the probability that $\|\mathtt{e}\|$ is greater than or equal to $\varepsilon\in\mathbb{R}_{> 0}$ is given as
\begin{align}
\label{cont_learning_bound_sto_prob}
\mathop{\mathbb{P}}\left[\|\mathtt{e}(t)\|\geq\varepsilon\right] \leq \frac{1}{\varepsilon^2}\left(\frac{\mathop{\mathbb{E}}[V_{s\ell}(0)]}{\underline{m}}e^{-2\alpha_{\ell} t}+\frac{C}{2\alpha_{\ell}}\frac{\overline{m}}{\underline{m}}\right).
\end{align}
\end{theorem}
\begin{proof}
Computing $\mathscr{L}V_{s\ell}$ with the virtual system \eqref{learning_x_sto} as in the proof of Theorem~\ref{THM:Thm:robuststochastic}, we get an additional term $2\int_{0}^{1}\partial_{\mu} q^{\top}M\Delta_L$. Using the learning error assumption \eqref{Eq:learning_error} to have $\|\Delta_L\| \leq \epsilon_{\ell0}+(\epsilon_{\ell1}/\sqrt{\underline{m}})\int_{0}^{1}\|\Theta \partial_{\mu} q\|d\mu$ as in \eqref{lip_learning_error_theta}, we have that
\begin{align}
\begin{aligned}
    2\int_{0}^{1}\partial_{\mu} q^{\top}M\Delta_L&\leq 2\sqrt{\overline{m}}V_{\ell}\left(\epsilon_{\ell0}+\left(\frac{\epsilon_{\ell1}}{\sqrt{\underline{m}}}\right)V_{\ell}\right) \\
    &\leq \alpha_d^{-1}\epsilon_{\ell0}^2\overline{m}+\left(\alpha_d+2\epsilon_{\ell1}\sqrt{\frac{\overline{m}}{\underline{m}}}\right) V_{\ell}^2 
\end{aligned}
\label{alpha_d_def}
\end{align}
where $V_{\ell}=\int_{0}^{1}\|\Theta \partial_{\mu} q\|d\mu$ as in Theorem~\ref{THM:Thm:path_integral}, and the relation $2ab \leq \alpha_d^{-1}a^2+\alpha_d b^2$, which holds for any $a,b\in\mathbb{R}$ and $\alpha_d \in \mathbb{R}_{>0}$ ($a = \epsilon_{\ell0}\sqrt{\overline{m}}$ and $b = V_{\ell}$ in this case), is used to obtain the second inequality. Since we have $V_{\ell}^2 \leq V_{s\ell}$ as proven in Theorem~\ref{THM:Thm:path_integral}, selecting $\alpha_d$ and $\epsilon_{\ell1}$ sufficiently small to satisfy \eqref{epsilon_ell2_condition_sto} gives the desired relations \eqref{cont_learning_bound_sto} and \eqref{cont_learning_bound_sto_prob} due to Theorem~\ref{THM:Thm:robuststochastic}. \qed
\end{proof}

As discussed in Theorems~\ref{THM:Thm:contraction_learning} and~\ref{THM:Thm:contraction_learning_sto}, using contraction theory for learning-based and data-driven control, we can formally guarantee the system trajectories to stay in a tube with an exponentially convergent bounded radius centered around the target trajectory, even with the external disturbances and learning errors $\epsilon_{\ell0}$ and $\epsilon_{\ell1}$. The exponential bounds \eqref{cont_learning_bound}, \eqref{cont_learning_bound_sto}, and \eqref{cont_learning_bound_sto_prob} become tighter as we achieve smaller $\epsilon_{\ell0}$ and $\epsilon_{\ell1}$ using more training data for verifying \eqref{Eq:learning_error} (see Remark~\ref{remark_learning_error}). It is also worth noting that the steady-state bounds of \eqref{cont_learning_bound} and \eqref{cont_learning_bound_sto} are again some functions of the condition number $\chi = \overline{m}/\underline{m}$ as in Theorems~\ref{THM:Thm:CVSTEMobjective} and~\ref{THM:Thm:CV-STEMobjective-est}, which renders the aforementioned CV-STEM approach valid and effective also in the learning-based frameworks. Theorems~\ref{THM:Thm:contraction_learning} and~\ref{THM:Thm:contraction_learning_sto} play a central role in providing incremental exponential stability guarantees for learning-based and data-driven automatic control to be introduced in Sec.~\ref{Sec:ncm}--\ref{Sec:adaptive}.
\subsection{Motivation for Neural Contraction Metrics}
The CV-STEM schemes in Theorems~\ref{THM:Thm:CV-STEM},~\ref{THM:Thm:CV-STEM-estimation}, and~\ref{THM:Thm:ccm_cvstem} permit the construction of optimal contraction metrics via convex optimization for synthesizing optimal, provably stable, and robust feedback control and state estimation. It is also shown in Theorems~\ref{THM:Thm:contraction_learning} and~\ref{THM:Thm:contraction_learning_sto} that these contraction metrics are useful for obtaining stability guarantees for machine learning-based and data-driven control techniques while retaining the CV-STEM-type optimality due to the tracking error bounds \eqref{cont_learning_bound} and \eqref{cont_learning_bound_sto}, which are comparable to those in Theorems~\ref{THM:Thm:CVSTEMobjective} and~\ref{THM:Thm:CV-STEMobjective-est}. However, the CV-STEM requires that a nonlinear system of equations or an optimization problem be solved at each time instant, which is not suitable for systems with limited online computational capacity. 

The Neural Contraction Metric (NCM) and its extensions~\cite{ncm,nscm,ancm,lagros} have been developed to address such a computational issue by modeling the CV-STEM optimization scheme using a DNN~\cite{neural1,neural2,neural3}, making it implementable in real-time. We will see in the subsequent sections that Theorems~\ref{THM:Thm:contraction_learning} and~\ref{THM:Thm:contraction_learning_sto} provide one formal approach to prove robustness and stability properties of the NCM methodologies~\cite{cdc_ncm}.

\newpage
\section{Learning-based Robust Control and Estimation}
\label{Sec:ncm}
\begin{figure}
    \centering
    \includegraphics[width=132mm]{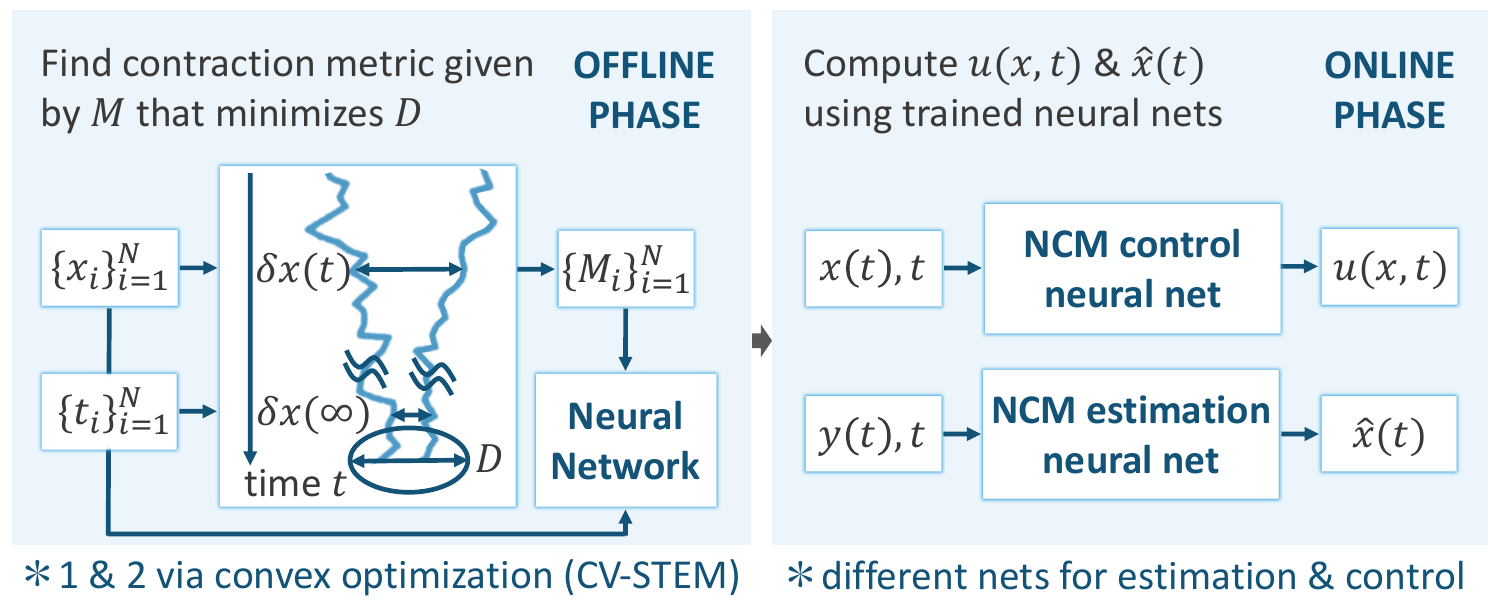}
    \caption{Illustration of NCM ($x$: system state; $M$: positive definite matrix that defines optimal contraction metric; $x_i$ and $M_i$: sampled $x$ and $M$; $\hat{x}$: estimated system state; $y$: measurement; and $u$: system control input). Note that the target trajectory $(x_d,u_d)$ is omitted in the figure for simplicity.}
    \label{ncmdrawing}
\end{figure}

Let us again consider the systems \eqref{original_dynamics}--\eqref{sdc_dynamicsd} for control, \ie{},
\begin{align}
\label{ncm_original_dynamics}
\dot{x} &= f(x,t) + B(x,t)u + d_c(x,t) \\
\label{ncm_sdc_dynamics}
dx &= (f(x,t)+B(x,t)u)dt+G_c(x,t)d\mathscr{W}(t) \\
\label{ncm_sdc_dynamicsd}
\dot{x}_d &= f(x_d,t)+B(x_d,t)u_d
\end{align}
with $\sup_{x,t}\|d_c\|=\bar{d}_c\in\mathbb{R}_{\geq0}$, $\sup_{x,t}\|G_c\|_F = \bar{g}_c\in\mathbb{R}_{\geq0}$, and \eqref{sdc_dynamics_est}--\eqref{sdc_dynamics_est_sto} for estimation, \ie{},
\begin{align}
\label{ncm_sdc_dynamics_est}
\dot{x} =& f(x,t)+d_{e0}(x,t),~y = h(x,t)+d_{e1}(x,t) \\
\label{ncm_sdc_dynamics_est_sto}
dx =& f(x,t)dt+G_{e0}d\mathscr{W}_0,~ydt = h(x,t)dt+G_{e1}d\mathscr{W}_1
\end{align}
with $\sup_{x,t}\|d_{e0}\|=\bar{d}_{e0}\in\mathbb{R}_{\geq0}$, $\sup_{x,t}\|d_{e1}\|=\bar{d}_{e1}\in\mathbb{R}_{\geq0}$, $\sup_{x,t}\|G_{e0}\|_F = \bar{g}_{e0}\in\mathbb{R}_{\geq0}$, and $\sup_{x,t}\|G_{e1}\|_F = \bar{g}_{e1}\in\mathbb{R}_{\geq0}$. The following definitions of a DNN~\cite{neural1,neural2,neural3} and NCM~\cite{ncm,nscm,cdc_ncm} will be utilized extensively hereinafter.
\begin{definition}{Neural Networks}{Def:NN_definition}
A neural network is a nonlinear mathematical model to approximately represent training samples $\{(x_i,y_i)\}_{i=1}^{N}$ of $y = \phi(x)$ by optimally tuning its hyperparameters $W_{\ell}$, and is given as follows:
\begin{align}
\label{neuralnet}
y_i = \varphi(x_i;W_{\ell}) = T_{L+1}\circ\sigma\circ T_{L}\circ\cdots\circ\sigma\circ T_{1}(x_i)
\end{align}
where $T_{\ell}(x) = W_{\ell}x$, $\circ$ denotes composition of functions, and $\sigma$ is an activation function (\eg{} $\sigma(x) = \tanh(x)$, note that $\varphi(x)$ is smooth in this case). We call a neural network that has more than two layers (\ie{}, $L\geq 2$) a Deep Neural Network (DNN).
\end{definition}
\begin{definition}{Neural Contraction Metrics}{Def:NCM}
A Neural Contraction Metric (NCM) is a DNN model of a contraction metric given in Theorem~\ref{THM:Thm:contraction}. Its training data could be sampled by solving the CV-STEM presented in Theorem~\ref{THM:Thm:CV-STEM} for control (or Theorem~\ref{THM:Thm:ccm_cvstem} for differential feedback control) and Theorem~\ref{THM:Thm:CV-STEM-estimation} for estimation. The NCM framework is summarized in Fig.~\ref{ncmdrawing}.
\end{definition}
\begin{remark}
\label{remark_NSCM_notation}
Although a stochastic version of the NCM is called a Neural Stochastic Contraction Metric (NSCM) in~\cite{nscm}, we also denote it as an NCM in this paper for simplicity of presentation.
\end{remark}

Since the NCM only requires one function evaluation at each time instant to get $M$ that defines a contraction metric, without solving any optimization problems unlike the CV-STEM approaches, it enables real-time optimal feedback control and state estimation in most engineering and scientific applications.
\subsection{Stability of NCM-based Control and Estimation}
The NCM exhibits superior incremental robustness and stability due to its internal contracting property.
\begin{theorem}{Learning-based Control with Neural Contraction Metrics}{Thm:NCMstability1}
Let $\mathcal{M}$ define the NCM in Definition~\ref{DEF:Def:NCM}, and let $u^*=u_d-R^{-1}B^{\top}M(x-x_d)$ for $M$ and $R$ given in Theorem~\ref{THM:Thm:CV-STEM}. Suppose that the systems \eqref{ncm_original_dynamics} and \eqref{ncm_sdc_dynamics} are controlled by $u_L$, which computes the CV-STEM controller $u^*$ of Theorem~\ref{THM:Thm:CV-STEM} replacing $M$ by $\mathcal{M}$, \ie{},
\begin{align}
\label{NCM_controller}
u_L=u_d-R(x,x_d,u_d,t)^{-1}B(x,t)^{\top}\mathcal{M}(x,x_d,u_d,t)\mathtt{e}
\end{align}
for $\mathtt{e}=x-x_d$. Note that we could use the differential feedback control $u=u_d+\int^1_0kd\mu$ of Theorem~\ref{THM:Thm:ccm_cvstem} for $u^*$ and $u_L$. Define $\Delta_L$ of Theorems~\ref{THM:Thm:contraction_learning} and \ref{THM:Thm:contraction_learning_sto} as follows:
\begin{align}
\label{ncm_DeltaL}
\Delta_L(\varpi,t) = 
B(x,t)(u_L(x,x_d,u_d,t)-u^*(x,x_d,u_d,t))
\end{align}
where we use $q(0,t)=\xi_0(t)=x_d(t)$, $q(1,t)=\xi_1(t)=x(t)$, and $\varpi=[x^{\top},x_d^{\top},u_d^{\top}]^{\top}$ in the learning-based control formulation \eqref{learning_x} and \eqref{learning_x_sto}. If the NCM is learned to satisfy the learning error assumptions of Theorems~\ref{THM:Thm:contraction_learning} and~\ref{THM:Thm:contraction_learning_sto} for $\Delta_L$ given by \eqref{ncm_DeltaL}, then \eqref{cont_learning_bound}, \eqref{cont_learning_bound_sto}, and \eqref{cont_learning_bound_sto_prob} of Theorems~\ref{THM:Thm:contraction_learning} and~\ref{THM:Thm:contraction_learning_sto} hold as follows:
\begin{align}
\label{Eq:NCM_control_error_bound}
\|\mathtt{e}(t)\| &\leq \frac{V_{\ell}(0)}{\sqrt{\underline{m}}}e^{-\alpha_{\ell} t}+\frac{\epsilon_{\ell0}+\bar{d}_c}{\alpha_{\ell}}\sqrt{\frac{\overline{m}}{\underline{m}}}(1-e^{-\alpha_{\ell} t}) \\
\label{Eq:NCM_control_error_bound_sto}
\mathop{\mathbb{E}}\left[\|\mathtt{e}(t)\|^2\right] &\leq \frac{\mathop{\mathbb{E}}[V_{s\ell}(0)]}{\underline{m}}e^{-2\alpha_{\ell} t}+\frac{C_C}{2\alpha_{\ell}}\frac{\overline{m}}{\underline{m}} \\
\label{Eq:NCM_control_error_bound_sto_prob}
\mathop{\mathbb{P}}\left[\|\mathtt{e}(t)\|\geq\varepsilon\right] &\leq \frac{1}{\varepsilon^2}\left(\frac{\mathop{\mathbb{E}}[V_{s\ell}(0)]}{\underline{m}}e^{-2\alpha_{\ell} t}+\frac{C_C}{2\alpha_{\ell}}\frac{\overline{m}}{\underline{m}}\right)
\end{align}
where $\mathtt{e}=x-x_d$, $C_C = \bar{g}_c^2({2}{\alpha_G}^{-1}+1)+\epsilon_{\ell0}^2\alpha_d^{-1}$, the disturbance upper bounds $\bar{d}_c$ and $\bar{g}_c$ are given in \eqref{ncm_original_dynamics} and \eqref{ncm_sdc_dynamics}, respectively, and the other variables are defined in Theorems~\ref{THM:Thm:CV-STEM},~\ref{THM:Thm:contraction_learning} and~\ref{THM:Thm:contraction_learning_sto}.
\end{theorem}
\begin{proof}
Let $\textsl{g}$ of the virtual systems given in Theorems~\ref{THM:Thm:contraction_learning} and~\ref{THM:Thm:contraction_learning_sto} be $\textsl{g}=\zeta$, where $\zeta$ is given by \eqref{Eq:virtual_sys_sto} (\ie{}, $\zeta = (A-BR^{-1}B^{\top}M)(q-x_d)+\dot{x}_d$). By definition of $\Delta_L$, this has $q=x$ and $q=x_d$ as its particular solutions (see Example~\ref{EX:ex:learning_prob1}). As defined in the proof of Theorems~\ref{THM:Thm:CVSTEM:LMI}, we have $d= \mu d_c\text{ and }G = \mu G_c$ for these virtual systems, resulting in the upper bounds of disturbances in the learning-based control formulation \eqref{cont_learning_bound} and \eqref{cont_learning_bound_sto} given as
\begin{align}
\label{Eq:NCM_control_bound}
\begin{aligned}
    \bar{d}&=\bar{d}_c \\
    \bar{g}&=\bar{g}_c.
\end{aligned}
\end{align}
Since $M$ and $u^*$ are constructed to render $\dot{q}=\zeta(q,\varpi,t)$ contracting as presented in Theorem~\ref{THM:Thm:CV-STEM}, and we have $\|\Delta_L\| \leq \epsilon_{\ell0}+\epsilon_{\ell1}\|\xi_1-\xi_0\|$ due to the learning assumption \eqref{Eq:learning_error}, Theorem~\ref{THM:Thm:contraction_learning} implies the deterministic bound \eqref{Eq:NCM_control_error_bound} and Theorem~\ref{THM:Thm:contraction_learning_sto} implies the stochastic bounds \eqref{Eq:NCM_control_error_bound_sto} and \eqref{Eq:NCM_control_error_bound_sto_prob}. Note that using the differential feedback control of Theorem~\ref{THM:Thm:ccm_cvstem} as $u^*$ also gives the bounds \eqref{Eq:NCM_control_error_bound}--\eqref{Eq:NCM_control_error_bound_sto_prob} following the same argument~\cite{cdc_ncm}. \qed
\end{proof}

Due to the estimation and control duality in differential dynamics observed in Sec.~\ref{sec:convex}, we have a similar result to Theorem~\ref{THM:Thm:NCMstability1} for the NCM-based state estimator. 
\begin{theorem}{Learning-based Estimation with Neural Contraction Metrics}{Thm:NCMstability1_est}
Let $\mathcal{M}$ define the NCM in Definition~\ref{DEF:Def:NCM}, and let $L(\hat{x},t)=M\bar{C}^{\top}R^{-1}$ for $M$, $\bar{C}$ and $R$ given in Theorem~\ref{THM:Thm:CV-STEM-estimation}. Suppose that the systems \eqref{ncm_sdc_dynamics_est} and \eqref{ncm_sdc_dynamics_est_sto} are estimated with an estimator gain $L_L$, which computes $L$ of the CV-STEM estimator $\dot{\hat{x}} = f(\hat{x},t)+L(\hat{x},t)(y-h(\hat{x},t))$ in Theorem~\ref{THM:Thm:CV-STEM-estimation} replacing $M$ by $\mathcal{M}$, \ie{},
\begin{align}
\label{NCM_estimator}
\dot{\hat{x}} &= f(\hat{x},t)+L_L(\hat{x},t)(y-h(\hat{x},t)) \\
&= f(\hat{x},t)+\mathcal{M}(\hat{x},t)\bar{C}(\varrho_c,\hat{x},t)^{\top}R(\hat{x},t)^{-1}(y-h(\hat{x},t)).
\end{align}
Define $\Delta_L$ of Theorems~\ref{THM:Thm:contraction_learning} and \ref{THM:Thm:contraction_learning_sto} as follows:
\begin{align}
\label{ncm_DeltaL_est}
\Delta_L(\varpi,t) = (L_L(\hat{x},t)-L(\hat{x},t))(h(x,t)-h(\hat{x},t))
\end{align}
where we use $q(0,t)=\xi_0(t)=x(t)$, $q(1,t)=\xi_1(t)=\hat{x}(t)$, and $\varpi=[x^{\top},\hat{x}^{\top}]^{\top}$ in the learning-based control formulation \eqref{learning_x} and \eqref{learning_x_sto}. If the NCM is learned to satisfy the learning error assumptions of Theorems~\ref{THM:Thm:contraction_learning} and~\ref{THM:Thm:contraction_learning_sto} for $\Delta_L$ given by \eqref{ncm_DeltaL_est}, then \eqref{cont_learning_bound}, \eqref{cont_learning_bound_sto}, and \eqref{cont_learning_bound_sto_prob} of Theorems~\ref{THM:Thm:contraction_learning} and~\ref{THM:Thm:contraction_learning_sto} hold as follows:
\begin{align}
\label{Eq:NCM_estimation_error_bound}
\|\mathtt{e}(t)\| &\leq \sqrt{\overline{m}}{V_{\ell}(0)}e^{-\alpha_{\ell} t}+\frac{\bar{d}_{a}\sqrt{\frac{\overline{m}}{\underline{m}}}+\bar{d}_{b}\overline{m}}{\alpha_{\ell}}(1-e^{-\alpha_{\ell} t})~~~~ \\
\label{Eq:NCM_estimation_error_bound_sto}
\mathop{\mathbb{E}}\left[\|\mathtt{e}(t)\|^2\right] &\leq \overline{m}{\mathop{\mathbb{E}}[V_{s\ell}(0)]}e^{-2\alpha_{\ell} t}+\frac{C_E}{2\alpha_{\ell}}\frac{\overline{m}}{\underline{m}} \\
\label{Eq:NCM_estimation_error_bound_sto_prob}
\mathop{\mathbb{P}}\left[\|\mathtt{e}(t)\|\geq\varepsilon\right] &\leq \frac{1}{\varepsilon^2}\left(\overline{m}{\mathop{\mathbb{E}}[V_{s\ell}(0)]}e^{-2\alpha_{\ell} t}+\frac{C_E}{2\alpha_{\ell}}\frac{\overline{m}}{\underline{m}}\right)
\end{align}
where $\mathtt{e}=\hat{x}-x$, $\bar{d}_{a} = \epsilon_{\ell0}+\bar{d}_{e0}$, $\bar{d}_{b}=\bar{\rho}\bar{c}\bar{d}_{e1}$, $C_E = (\bar{g}_{e0}^2+\bar{\rho}^2\bar{c}^2\bar{g}_{e1}^2\overline{m}^2)({2}{\alpha_G}^{-1}+1)+\epsilon_{\ell0}^2\alpha_d^{-1}$, the disturbance upper bounds $\bar{d}_{e0}$ $\bar{d}_{e1}$, $\bar{g}_{e0}$, and $\bar{g}_{e1}$ are given in \eqref{ncm_sdc_dynamics_est} and \eqref{ncm_sdc_dynamics_est_sto}, and the other variables are defined in Theorems~\ref{THM:Thm:CV-STEM-estimation},~\ref{THM:Thm:contraction_learning} and~\ref{THM:Thm:contraction_learning_sto}.
\end{theorem}
\begin{proof}
Let $\textsl{g}$ of the virtual systems given in Theorems~\ref{THM:Thm:contraction_learning} and~\ref{THM:Thm:contraction_learning_sto} be $\textsl{g}=\zeta$, where $\zeta$ is given by \eqref{Eq:virtual_sys_est_sto} (\ie{}, $\zeta = (A-LC)(q-x)+f(x,t)$). By definition of $\Delta_L$, this has $q=\hat{x}$ and $q=x$ as its particular solutions (see Example~\ref{EX:ex:learning_prob2}). As defined in the proof Theorem~~\ref{THM:Thm:CVSTEM-LMI-est}, we have
\begin{align}
d&=(1-\mu) d_{e0}+\mu L_Ld_{e1} \\
G&=\begin{bmatrix}
    (1-\mu) G_{e0}&\mu L_LG_{e1}
    \end{bmatrix}
\end{align}
for these virtual systems, resulting in the upper bounds of external disturbances in the learning-based control formulation \eqref{cont_learning_bound} and \eqref{cont_learning_bound_sto} given as
\begin{align}
\label{Eq:NCM_estimation_bound}
\begin{aligned}
\bar{d}&=\bar{d}_{e0}+\bar{\rho}\bar{c}\bar{d_{e1}}\sqrt{\overline{m}\underline{m}} \\
\bar{g}&=\sqrt{\bar{g}_{e0}^2+\bar{\rho}^2\bar{c}^2\bar{g}_{e1}^2\overline{m}^2}.
\end{aligned}
\end{align}
Since $W=M^{-1}$ is constructed to render $\dot{q}=\zeta(q,\varpi,t)$ contracting as presented in Theorem~\ref{THM:Thm:CV-STEM-estimation}, and we have $\|\Delta_L\| \leq \epsilon_{\ell0}+\epsilon_{\ell1}\|\xi_1-\xi_0\|$ due to the learning assumption \eqref{Eq:learning_error}, Theorem~\ref{THM:Thm:contraction_learning} implies the bound \eqref{Eq:NCM_estimation_error_bound}, and Theorem~\ref{THM:Thm:contraction_learning_sto} implies the bounds \eqref{Eq:NCM_estimation_error_bound_sto} and \eqref{Eq:NCM_estimation_error_bound_sto_prob}. \qed
\end{proof}
\begin{remark}
As discussed in Examples~\ref{EX:ex:learning_prob1} and~\ref{EX:ex:learning_prob2} and in Remark~\ref{remark_learning_error}, we have $\epsilon_{\ell1}=0$ for the learning error $\epsilon_{\ell1}$ of \eqref{Eq:learning_error} as long as $B$ and $h$ are bounded a compact set by the definition of $\Delta_L$ in \eqref{ncm_DeltaL} and \eqref{ncm_DeltaL_est}. If $h$ and $u^*$ are Lipschitz with respect to $x$ in the compact set, we have non-zero $\epsilon_{\ell1}$ with $\epsilon_{\ell0}=0$ in \eqref{Eq:learning_error} (see Theorem~\ref{THM:Thm:NCMstability2}).
\end{remark}

\begin{figure}
    \centering
    \includegraphics[width=132mm]{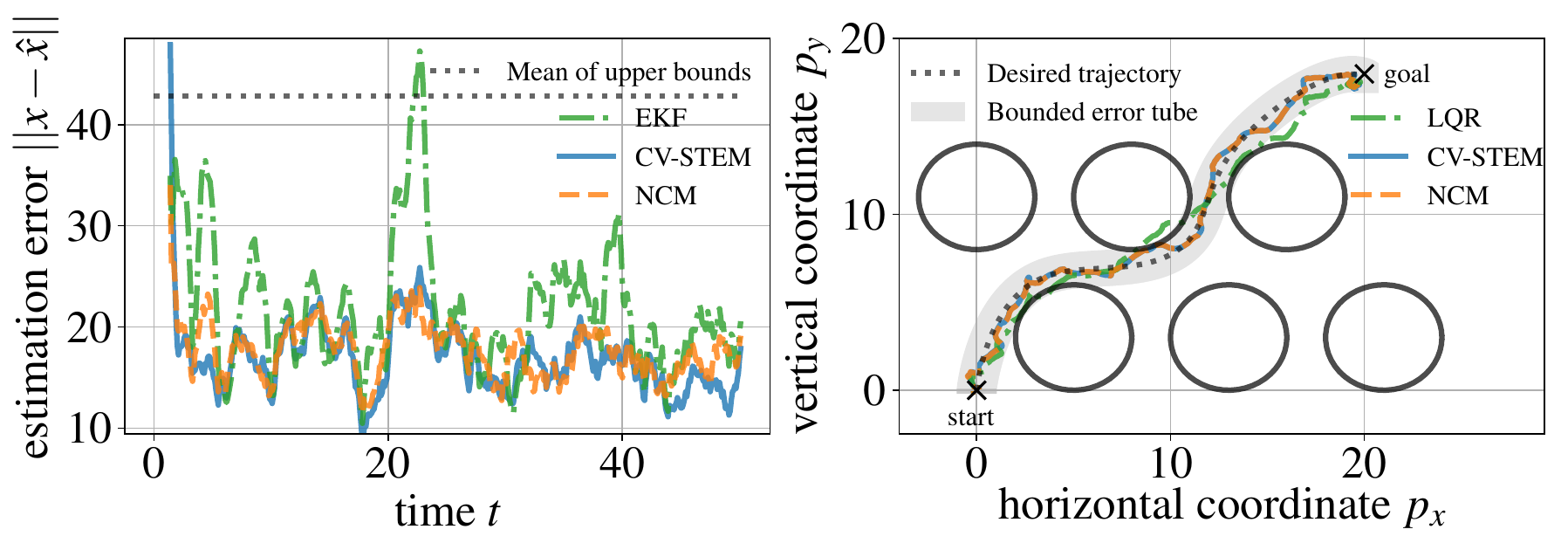}
    \caption{Lorenz oscillator state estimation error smoothed using a $15$-point moving average filter in Example~\ref{EX:ex:lorenz} (left), and spacecraft motion $(p_x,p_y)$ on a planar field in Example~\ref{EX:ex:sc_robust_control} (right).}
        \label{fig:NCMexample}
\end{figure}

Using Theorems~\ref{THM:Thm:NCMstability1} and~\ref{THM:Thm:NCMstability1_est}, we can relate the learning error of the matrix that defines the NCM, $\|\mathcal{M}-M\|$, to the error bounds \eqref{Eq:NCM_control_error_bound}--\eqref{Eq:NCM_control_error_bound_sto_prob} and \eqref{Eq:NCM_estimation_error_bound}--\eqref{Eq:NCM_estimation_error_bound_sto_prob}, if $M$ is given by the CV-STEM of Theorems~\ref{THM:Thm:CV-STEM} and~\ref{THM:Thm:CV-STEM-estimation}.
\begin{theorem}{Robustness against Metric Construction Errors}{Thm:NCMstability2}
Let $\mathcal{M}$ define the NCM in Definition~\ref{DEF:Def:NCM}, and let $\mathcal{S}_s\subseteq\mathbb{R}^n$, $\mathcal{S}_u\subseteq\mathbb{R}^m$, and $t\in\mathcal{S}_t\subseteq\mathbb{R}_{\geq 0}$ be some compact sets. Suppose that the systems \eqref{ncm_original_dynamics}--\eqref{ncm_sdc_dynamics_est_sto} are controlled by \eqref{NCM_controller} and estimated by \eqref{NCM_estimator}, respectively, as in Theorems~\ref{THM:Thm:NCMstability1} and~\ref{THM:Thm:NCMstability1_est}. Suppose also that $\exists \bar{b},\bar{c},\bar{\rho}\in\mathbb{R}_{\geq 0}$ \st{} $\|B(x,t)\|\leq\bar{b}$, $\|C(\varrho_c,x,\hat{x},t)\|\leq\bar{c}$, and $\|R^{-1}\|\leq\bar{\rho},~\forall x,\hat{x}\in\mathcal{S}_s$ and $t\in\mathcal{S}_t$, for $B$ in \eqref{ncm_original_dynamics}, $C$ in Theorem~\ref{THM:Thm:CV-STEM-estimation}, and $R$ in \eqref{NCM_controller} and \eqref{NCM_estimator}. If we have for all $x,x_d,\hat{x}\in\mathcal{S}_s$, $u_d\in\mathcal{S}_u$, and $t\in\mathcal{S}_t$ that 
\begin{align}
\label{Eq:Merror_control}
\|\mathcal{M}(x,x_d,u_d,t)-M(x,x_d,u_d,t)\| &\leq \epsilon_{\ell} \text{ for control} \\
\label{Eq:Merror_estimation}
\|\mathcal{M}(\hat{x},t)-M(\hat{x},t)\| &\leq \epsilon_{\ell} \text{ for estimation}
\end{align}
where $M(x,x_d,u_d,t)$ and $M(\hat{x},t)$ (or $M(x,t)$ and $M(\hat{x},t)$, see Theorem~\ref{THM:Thm:fixed_sdc}) are the CV-STEM solutions of Theorems~\ref{THM:Thm:CV-STEM} and \ref{THM:Thm:CV-STEM-estimation} with $\exists\epsilon_{\ell}\in\mathbb{R}_{\geq 0}$ being the learning error, then the bounds \eqref{Eq:NCM_control_error_bound}--\eqref{Eq:NCM_control_error_bound_sto_prob} of Theorem~\ref{THM:Thm:NCMstability1} and \eqref{Eq:NCM_estimation_error_bound}--\eqref{Eq:NCM_estimation_error_bound_sto_prob} of Theorem~\ref{THM:Thm:NCMstability1_est} hold with $\epsilon_{\ell0} = 0$ and $\epsilon_{\ell1} = \bar{\rho}\bar{b}^2\epsilon_{\ell}$ for control, and $\epsilon_{\ell0} = 0$ and $\epsilon_{\ell1} = \bar{\rho}\bar{c}^2\epsilon_{\ell}$ for estimation, as long as $\epsilon_{\ell}$ is sufficiently small to satisfy the conditions \eqref{epsilon_ell2_condition} and \eqref{epsilon_ell2_condition_sto} of Theorems~\ref{THM:Thm:contraction_learning} and~\ref{THM:Thm:contraction_learning_sto} for deterministic and stochastic systems, respectively.
\end{theorem}
\begin{proof}
For $\Delta_L$ defined in \eqref{ncm_DeltaL} and \eqref{ncm_DeltaL_est} with the controllers and estimators given as $u^*=u_d-R^{-1}B^{\top}M(x-x_d)$, $u_L=u_d-R^{-1}B^{\top}\mathcal{M}(x-x_d)$,  $L=MC^{\top}R^{-1}$, and $L_L=\mathcal{M}C^{\top}R^{-1}$, we have the following upper bounds:
\begin{align}
\label{ncm_DeltaL_bound}
\|\Delta_L\| \leq
\begin{cases}
\bar{\rho}\bar{b}^2\epsilon_{\ell}\|x-x_d\| & (\text{controller}) \\
\bar{\rho}\bar{c}^2\epsilon_{\ell}\|\hat{x}-x\| & (\text{estimator})
\end{cases}
\end{align}
where the relation $\|h(x,t)-h(\hat{x},t)\| \leq \bar{c}\|\hat{x}-x\|$, which follows from the equality $h(x,t)-h(\hat{x},t)=C(\varrho_c,x,\hat{x},t)(x-\hat{x})$ (see Lemma~\ref{sdclemma}), is used to obtain the second inequality. This implies that we have $\epsilon_{\ell0}=0$ and $\epsilon_{\ell1}=\bar{\rho}\bar{b}^2\epsilon_{\ell}$ for control, and $\epsilon_{\ell0}=0$ and $\epsilon_{\ell1}=\bar{\rho}\bar{c}^2\epsilon_{\ell}$ for estimation in the learning error of \eqref{Eq:learning_error}. The rest follows from Theorems~\ref{THM:Thm:NCMstability1} and~\ref{THM:Thm:NCMstability1_est}. \qed
\end{proof}
\begin{example}{Lorenz Oscillator}{ex:lorenz}
The left-hand side of Fig.~\ref{fig:NCMexample} shows the state estimation errors of the following Lorenz oscillator perturbed by process noise $d_0$ and measurement noise $d_1$~\cite{ncm}:
\begin{align}
\dot{x} &= [\sigma(x_2-x_1),x_1(\rho-x_3)-x_2,x_1 x_2-\beta x_3]^{\top}+d_0(x,t) \\
y &= [1,0,0]x+d_2(x,t)
\end{align}
where $x = [x_1,x_2,x_3]^{\top}$, $\sigma = 10$, $\beta = 8/3$, and $\rho = 28$. It can be seen that the steady-state estimation errors of the NCM and CV-STEM are below the optimal steady-state upper bound of Theorem~\ref{THM:Thm:CV-STEM-estimation} (dotted line in the left-hand side of Fig.~\ref{fig:NCMexample}) while the EKF has a larger error compared to the other two. In this example, training data is sampled along $100$ trajectories with uniformly distributed initial conditions ($-10 \leq x_i \leq 10,~i = 1,2,3$) using Theorem~\ref{THM:Thm:CV-STEM-estimation}, and then modeled by a DNN as in Definition~\ref{DEF:Def:NCM}. Additional implementation and network training details can be found in~\cite{ncm}.
\end{example}
\begin{example}{Planar Spacecraft Simulators}{ex:sc_robust_control}
Consider a robust feedback control problem of the planar spacecraft perturbed by deterministic disturbances, the unperturbed dynamics of which is given as follows~\cite{SCsimulator}:
\begin{align}
\label{eq:sc_dynamics}
\begin{bmatrix}m & 0 & 0 \\ 0 & m & 0 \\ 0 & 0 & I\end{bmatrix}\ddot{x}=\begin{bmatrix}\cos\phi & \sin\phi & 0 \\ -\sin\phi & \cos\phi & 0 \\ 0 & 0 & 1\end{bmatrix}\begin{bmatrix}-1 & -1 & 0 & 0 & 1 & 1 & 0 & 0 \\0 & 0 & -1 & -1 & 0 & 0 & 1 & 1 \\-\ell & \ell & -b & b & -\ell & \ell & -b & b\end{bmatrix}u~~~~~~~~~
\end{align}
where $x = [p_x,p_y,\phi,\dot{p}_x,\dot{p}_y,\dot{\phi}]^T$ with $p_x$, $p_y$, and $\phi$ being the horizontal coordinate, vertical coordinate, and yaw angle of the spacecraft, respectively, with the other variables defined as $m = \text{mass of spacecraft}$, $I = \text{yaw moment of inertia}$, $u = \text{thruster force vector}$, $\ell=\text{half-depth of spacecraft}$, and $b=\text{half-width of spacecraft}$ (see Fig.~8 of~\cite{SCsimulator} for details).

As shown in the right-hand side of Fig.~\ref{fig:NCMexample}, the NCM keeps their trajectories within the bounded error tube of Theorem~\ref{THM:Thm:CV-STEM} (shaded region) around the target trajectory (dotted line) avoiding collision with the circular obstacles, even under the presence of disturbances, whilst requiring much smaller computational cost than the CV-STEM. Data sampling and the NCM training are performed as in Example~\ref{EX:ex:lorenz}~\cite{ncm}.
\end{example}
\begin{remark}
We could also simultaneously synthesize a controller $u$ and the NCM, and Theorem~\ref{THM:Thm:NCMstability1} can provide formal robustness and stability guarantees even for such cases~\cite{chuchu,cdc_ncm}. In Sec.~\ref{Sec:lagros}, we generalize the NCM to learn contraction theory-based robust control using a DNN only taking $x$, $t$, and a vector containing local environment information as its inputs, to avoid the online computation of $x_d$ for the sake of automatic guidance and control implementable in real-time.
\end{remark}
\subsection{NCMs as Lyapunov Functions}
We could also utilize the NCM constructed by the CV-STEM contraction metric for designing a Control Lyapunov function (CLF)~\cite{doi:10.1137/0321028,SONTAG1989117,Khalil:1173048}, resulting in a stability result different from Theorems~\ref{THM:Thm:NCMstability1} and~\ref{THM:Thm:NCMstability2}. To this end, let us recall that designing optimal $k$ of $u = k(x,x_d,u_d,t)$ reduces to designing optimal $K(x,x_d,u_d,t)$ of $u=u_d-K(x,x_d,u_d,t)(x-x_d)$ due to Lemma~\ref{u_equivalence_lemma} of Sec.~\ref{sec:HinfKYP}. For such $u$, the virtual system of \eqref{ncm_original_dynamics}/\eqref{ncm_sdc_dynamics} and \eqref{ncm_sdc_dynamicsd} without any perturbation, which has $q=x$ and $q=x_d$ as its particular solutions, is given as follows:
\begin{align}
\label{virtual_dynamics_clf}
\dot{q} &= (A(\varrho,x,x_d,u_d,t)-B(x,t)K(x,x_d,u_d,t))(q-x_d) +f(x_d,t)+B(x_d,t)u_d
\end{align}
where $A$ is the SDC matrix of the system \eqref{ncm_original_dynamics} given by Lemma~\ref{sdclemma}, \ie{}, $A(\varrho,x,x_d,u_d,t)(x-x_d) = f(x,t)+B(x,t)u_d-f(x_d,t)-B(x_d,t)u_d$. Using the result of Theorem~\ref{THM:Thm:path_integral}, one of the Lyapunov functions for the virtual system \eqref{virtual_dynamics_clf} with unknown $K$ can be given as the following transformed squared length integrated over $x$ of \eqref{ncm_original_dynamics}/\eqref{ncm_sdc_dynamics} and $x_d$ of \eqref{ncm_sdc_dynamicsd}:
\begin{align}
\label{eq_ncmclf}
V_{\mathrm{NCM}} = \int_{x_d}^{x}\delta q^{\top}\mathcal{M}(x,x_d,u_d,t)\delta q
\end{align}
where $\mathcal{M}$ defines the NCM of Definition~\ref{DEF:Def:NCM} modeling $M$ of the CV-STEM contraction metric in Theorem~\ref{THM:Thm:CV-STEM}.
\begin{theorem}{Control Lyapunov Functions with Neural Contraction Metrics}{Thm:ncm_clf}
Suppose \eqref{ncm_original_dynamics} and \eqref{ncm_sdc_dynamics} are controlled by $u = u_d-K^*(x,x_d,u_d,t)\mathtt{e}$, where $\mathtt{e}=x-x_d$, and that $K^*$ is designed by the following convex optimization for given $(x,x_d,u_d,t)$:
\begin{align}
\label{ncm_robust_control}
&(K^*,p^*) = \mathrm{arg}\min_{K\in \mathcal{C}_K,p\in\mathbb{R}}\|K\|^2_F+p^2 \\
&\text{\st{} } \int_{x_d}^x\delta q^{\top}(\dot{\mathcal{M}}+2\sym{}(\mathcal{M}A-\mathcal{M}BK))\delta q 
\leq -2\alpha\int_{x_d}^x\delta q^{\top}( \mathcal{M}+\beta \mathrm{I})\delta q+p
\label{stability_clf}
\end{align}
where $\mathcal{M}$ defines the NCM of Definition~\ref{DEF:Def:NCM} that models $M$ of the CV-STEM contraction metric in Theorem~\ref{THM:Thm:CV-STEM} as in \eqref{Eq:Merror_control}, $\mathcal{C}_K$ is a convex set containing admissible $K$, $\alpha$ is the contraction rate, and $\beta$ is as defined in theorem~\ref{THM:Thm:CVSTEM:LMI} (\ie{}, $\beta=0$ for deterministic systems and $\beta = \alpha_s$ for stochastic systems). Then Theorems~\ref{THM:Thm:Robust_contraction_original} and~\ref{THM:Thm:robuststochastic} still hold for the Lyapunov function defined in \eqref{eq_ncmclf}, yielding the following bounds:
\begin{align}
\label{Eq_NCM_CLF_bound}
\|\mathtt{e}(t)\|^2 &\leq \frac{V_{\mathrm{NCM}}(0)}{\underline{m}}e^{-2\alpha_p t}+\frac{\frac{\bar{p}}{\underline{m}}+\frac{\bar{d}_c^2\overline{m}}{\alpha_d\underline{m}}}{2\alpha_p}(1-e^{-2\alpha_p t}) \\
\label{Eq_NCM_CLF_bound_sto}
\mathop{\mathbb{E}}\left[\|\mathtt{e}(t)\|^2\right] &\leq \frac{\mathop{\mathbb{E}}[V_{\mathrm{NCM}}(0)]}{\underline{m}}e^{-2\alpha t}+\frac{C_C\overline{m}+\bar{p}}{2\alpha\underline{m}} \\
\label{Eq_NCM_CLF_bound_sto_prob}
\mathop{\mathbb{P}}\left[\|\mathtt{e}(t)\|\geq\varepsilon\right] &\leq \frac{1}{\varepsilon^2}\left(\frac{\mathop{\mathbb{E}}[V_{\mathrm{NCM}}(0)]}{\underline{m}}e^{-2\alpha t}+\frac{C_C\overline{m}+\bar{p}}{2\alpha\underline{m}}\right)~~~~~~~~~~
\end{align}
where $\bar{p}=\sup_{x,x_d,u_d,t}p$, $\underline{m}\mathrm{I}\preceq\mathcal{M}\preceq\overline{m}\mathrm{I}$ as in \eqref{Mcon}, $\alpha_d\in\mathbb{R}_{>0}$ is an arbitrary positive constant (see \eqref{Vncm_tochu1}) selected to satisfy $\alpha_p=\alpha-\alpha_d/2>0$, $C_C = \bar{g}_c^2({2}{\alpha_G}^{-1}+1)$, $\alpha_G\in\mathbb{R}_{>0}$ is an arbitrary constant as in Theorem~\ref{THM:Thm:robuststochastic}, $\varepsilon\in\mathbb{R}_{\geq0}$, and the disturbance terms $\bar{d}_c$ and $\bar{g}_c$ are given in \eqref{ncm_original_dynamics} and \eqref{ncm_sdc_dynamics}, respectively. Note that the problem \eqref{ncm_robust_control} is feasible due to the constraint relaxation variable $p$.

Furthermore, if $\epsilon_{\ell} = 0$ in \eqref{Eq:Merror_control}, \eqref{ncm_robust_control} with $p=0$ is always feasible, and the optimal feedback gain $K^*$ minimizes its Frobenius norm under the contraction constraint \eqref{stability_clf}. 
\end{theorem}
\begin{proof}
Computing $\dot{V}_{\mathrm{NCM}}$ for the deterministic system \eqref{ncm_original_dynamics} along with the virtual dynamics \eqref{virtual_dynamics_clf} and the stability condition \eqref{stability_clf} yields
\begin{align}
\dot{V}_{\mathrm{NCM}} &\leq -2\alpha V_{\mathrm{NCM}}+\bar{d}_c\sqrt{\overline{m}}\sqrt{V_{\mathrm{NCM}}}+\bar{p} 
\leq -2(\alpha-\alpha_d/2) V_{\mathrm{NCM}}+\bar{d}_c^2\overline{m}/\alpha_d+\bar{p}
\label{Vncm_tochu1}
\end{align}
where the relation $2ab \leq \alpha_d^{-1}a^2+\alpha_d b^2$, which holds for any $a,b\in\mathbb{R}$ and $\alpha_d \in \mathbb{R}_{>0}$ ($a = \bar{d}_c\sqrt{\overline{m}}$ and $b = \sqrt{V_{\mathrm{NCM}}}$ in this case), is used to obtain the second inequality. Since we can arbitrarily select $\alpha_d$ to have $\alpha_p=\alpha-\alpha_d/2>0$, \eqref{Vncm_tochu1} gives the exponential bound \eqref{Eq_NCM_CLF_bound} due to the comparison lemma of Lemma~\ref{Lemma:comparison}. Similarly, computing $\mathscr{L}V_{\mathrm{NCM}}$ for the stochastic system \eqref{ncm_sdc_dynamics} yields
\begin{align}
\mathscr{L}V_{\mathrm{NCM}} &\leq -2\alpha V_{\mathrm{NCM}}+C_c\overline{m}+\bar{p}
\end{align}
which gives \eqref{Eq_NCM_CLF_bound_sto} and \eqref{Eq_NCM_CLF_bound_sto_prob} due to Lemma~\ref{Lemma:comparison_sto}. If $\epsilon_{\ell} = 0$ in \eqref{Eq:Merror_control}, $K = R^{-1}B^{\top}M$ satisfies \eqref{stability_clf} with $p=0$ as we have $\mathcal{M}=M$ and $M$ defines the contraction metric of Theorem~\ref{THM:Thm:CV-STEM}, which implies that \eqref{ncm_robust_control} with $p=0$ is always feasible.
\qed
\end{proof}

We could also use the CCM-based differential feedback formulation of Theorem~\ref{THM:Thm:ccm_cvstem} in Theorem~\ref{THM:Thm:ncm_clf} (see~\cite{ccm,7989693}). To this end, define $E_{\mathrm{NCM}}$ (Riemannian energy) as follows:
\begin{align}
\label{riemannian_energy}
E = \int^1_0\gamma_{\mu}(\mu,t)^{\top}\mathcal{M}(\gamma(\mu,t),t)\gamma_{\mu}(\mu,t)d\mu
\end{align}
where $\mathcal{M}$ defines the NCM of Definition~\ref{DEF:Def:NCM} that models $M$ of the CCM in Theorem~\ref{THM:Thm:ccm_cvstem}, $\gamma$ is the minimizing geodesic connecting $x(t)=\gamma(1,t)$ of \eqref{ncm_original_dynamics} and $x_d(t)=\gamma(0,t)$ of \eqref{ncm_sdc_dynamicsd} (see Theorem~\ref{THM:Thm:path_integral}), and $\gamma_{\mu}={\partial \gamma}/{\partial \mu}$.
\begin{theorem}{Control Lyapunov Functions with Control Contraction Metrics}{Thm:ncm_ccm_clf}
Suppose $u=u^*$ of \eqref{ncm_original_dynamics} is designed by the following convex optimization for given $(x,x_d,u_d,t)$:
\begin{align}
\label{eq_ccm_clf}
(u^*,p^*) &= \mathrm{arg}\min_{u\in \mathcal{C}_u,p\in\mathbb{R}}\|u-u_d\|^2+p^2 \\
\label{eq_ccm_clf_cond}
&\begin{aligned}
&\text{\st{} }\frac{\partial E}{\partial t}+2\gamma_{\mu}(1,t)^{\top}\mathcal{M}(x,t)(f(x,t)+B(x,t)u) \\
&\text{\textcolor{white}{\st{}} }-2\gamma_{\mu}(0,t)^{\top}\mathcal{M}(x_d,t)(f(x_d,t)+B(x_d,t)u_d) \leq -2\alpha E+p
\end{aligned}
\end{align}
where $\mathcal{M}$ defines the NCM of Definition~\ref{DEF:Def:NCM} that models the CCM of Theorem~\ref{THM:Thm:ccm_cvstem}, and $\mathcal{C}_{u}$ is a convex set containing admissible $u$. Then the bound \eqref{Eq_NCM_CLF_bound} of Theorem~\ref{THM:Thm:ncm_clf} holds.

Furthermore, if $\epsilon_{\ell}=0$ for the NCM learning error $\epsilon_{\ell}$ in \eqref{Eq:Merror_control}, then the problem \eqref{eq_ccm_clf} with $p=0$ is always feasible, and $u^*$ minimizes the deviation of $u$ from $u_d$ under the contraction constraint \eqref{eq_ccm_clf_cond}.
\end{theorem}
\begin{proof}
Since the right-hand side of \eqref{eq_ccm_clf_cond} represents $\dot{E}$ of \eqref{riemannian_energy} if $d_c=0$ in \eqref{ncm_original_dynamics}, the comparison lemma of Lemma~\ref{Lemma:comparison} gives \eqref{Eq_NCM_CLF_bound} as in Theorem~\ref{THM:Thm:ncm_clf}. The rest follows from the fact that $u$ of Theorem~\ref{THM:Thm:ccm_cvstem} is a feasible solution of \eqref{eq_ccm_clf} if $p=0$~\cite{ccm,7989693}.
\end{proof}
\begin{remark}
As discussed in Remark~\ref{remark_ccm_kgamma}, Theorem~\ref{THM:Thm:ncm_ccm_clf} can also be formulated in a stochastic setting as in Theorem~\ref{THM:Thm:ncm_clf} using the technique discussed in~\cite{cdc_ncm}.
\end{remark}

Theorems~\ref{THM:Thm:ncm_clf} and~\ref{THM:Thm:ncm_ccm_clf} provide a perspective on the NCM stability different from Theorems~\ref{THM:Thm:NCMstability1} and~\ref{THM:Thm:NCMstability2} written in terms of the constraint relaxation variable $p$, by directly using the NCM as a contraction metric instead of the CV-STEM contraction metric or the CCM. Since the CLF problems are feasible with $p=0$ when we have zero NCM modeling error, it is implied that the upper bound of $p$ (\ie{}, $\bar{p}$ in \eqref{Eq_NCM_CLF_bound} and \eqref{Eq_NCM_CLF_bound_sto}) decreases as we achieve a smaller learning error $\epsilon_{\ell}$ using more training data for verifying \eqref{Eq:Merror_control} (see Remark~\ref{remark_learning_error}). This intuition will be formalized later in Theorem~\ref{THM:Thm:data_driven_cont_thm} of Sec.~\ref{Sec:datadriven}~\cite{boffi2020learning}. See~\cite{ccm,7989693} for how we compute the minimizing geodesic $\gamma$ and its associated Riemannian energy $E$ in Theorem~\ref{THM:Thm:ncm_ccm_clf}.
\begin{example}{Analytical Solutions of CLF Convex Optimization}{}
If $\mathcal{C}_K=\mathbb{R}^{m\times n}$ and $\mathcal{C}_u=\mathbb{R}^{m}$ in \eqref{ncm_robust_control} and \eqref{eq_ccm_clf}, respectively, they can both be expressed as the following quadratic optimization problem:
\begin{align}
&v^* = \mathrm{arg}\min_{v\in\mathbb{R}^\mathtt{v}} \frac{1}{2}v^{\top}v+c(x,x_d,u_d,t)^{\top}v \text{ \st{} } \varphi_0(x,x_d,u_d,t)+\varphi_1(x,x_d,u_d,t)^{\top}v \leq 0
\end{align}
by defining ${c}\in\mathbb{R}^{\mathtt{v}}$, $\varphi_0\in\mathbb{R}$, $\varphi_1\in\mathbb{R}^{\mathtt{v}}$, and $v\in\mathbb{R}^\mathtt{v}$ appropriately. Applying the KKT condition~\cite[pp. 243-244]{citeulike:163662} to this problem, we can show that
\begin{align}
v^* =
\begin{cases}
-c-\frac{\varphi_0-\varphi_1^{\top}c}{\varphi_1^{\top}\varphi_1}\varphi_1 & \text{if }\varphi_0-\varphi_1^{\top}c > 0 \\
-c & \text{otherwise}.
\end{cases}
\end{align}
This implies that the controller utilizes $u_d$ with the feedback term $(({\varphi_0-\varphi_1^{\top}c})/{\varphi_1^{\top}\varphi_1})\varphi_1$ only if $\dot{x}=f(x,t)+B(x,t)u$ with $u=u_d$ is not contracting, \ie{}, $\varphi_0-\varphi_1^{\top}c > 0$.
\end{example}
\begin{example}{Control Input Constraints}{}
When $q$ is parameterized linearly by $\mu$ in Theorem~\ref{THM:Thm:ncm_clf}, \ie{}, $q = \mu x+(1-\mu)x_d$, we have $V_{\mathrm{NCM}}=(x-x_d)^{\top}\mathcal{M}(x-x_d)$ as in the standard Lyapunov function formulation~\cite{doi:10.1137/0321028,SONTAG1989117,Khalil:1173048}. Also, a linear input constraint, $u_{\min} \leq u \leq u_{\max}$ can be implemented by using
\begin{align}
\mathcal{C}_K &= \{K\in\mathbb{R}^{m\times n}|u_d-u_{\max} \leq K\mathtt{e} \leq u_d-u_{\min}\} \\
\mathcal{C}_u &= \{u\in\mathbb{R}^{m}|u_{\min} \leq u \leq u_{\max}\}
\end{align}
in \eqref{ncm_robust_control} and \eqref{eq_ccm_clf}, respectively.
\end{example}
\begin{remark}
For the CV-STEM and NCM state estimation in Theorems~\ref{THM:Thm:CV-STEM-estimation}, \ref{THM:Thm:NCMstability1_est}, and \ref{THM:Thm:NCMstability2}, we could design a similar Lyapunov function-based estimation scheme if its contraction constraints of Theorem~\ref{THM:Thm:CV-STEM-estimation} do not explicitly depend on the actual state $x$. This can be achieved by using $A(\varrho_A,\hat{x},t)$ and $C(\varrho_C,\hat{x},t)$ instead of $A(\varrho_A,x,\hat{x},t)$ and $C(\varrho_C,x,\hat{x},t)$ in Theorem~\ref{THM:Thm:CV-STEM-estimation} as in Theorem~\ref{THM:Thm:fixed_sdc}~\cite{Ref:Stochastic}, leading to exponential boundedness of system trajectories as derived in~\cite{mypaperTAC,Ref:Stochastic,mypaper}.
\end{remark}
\subsection{Remarks in NCM Implementation}
\label{Sec:SN}
This section briefly summarizes several practical techniques in constructing NCMs using the CV-STEM of Theorems~\ref{THM:Thm:CV-STEM},~\ref{THM:Thm:CV-STEM-estimation}, and~\ref{THM:Thm:ccm_cvstem}. 
\subsubsection{Neural Network Training}
\label{data_preprocessing}
We model $M$ of the CV-STEM contraction metric sampled offline by the DNN of Definition~\ref{DEF:Def:NN_definition}, optimizing its hyperparameters by stochastic gradient descent~\cite{sgd,beyondconvexity} to satisfy the learning error condition \eqref{Eq:learning_error} (see Remark~\ref{remark_learning_error}). The positive definiteness of $M$ could be exploited to reduce the dimension of the DNN target output due to the following lemma~\cite{ncm}.
\begin{lemma}
\label{cholesky}
A matrix $A \succ 0$ has a unique Cholesky decomposition, \ie{}, there exists a unique upper triangular matrix $U\in\mathbb{R}^{n\times n}$ with strictly positive diagonal entries \st{} $A = U^TU$.
\end{lemma}
\begin{proof}
See~\cite[pp. 441]{10.5555/2422911}.
\end{proof}
Selecting $\Theta$ of $M=\Theta^{\top}\Theta$ as the unique Cholesky decomposition of $M$ and training the DNN using only the non-zero entries of $\Theta$, we can reduce the dimension of the target output by $n(n-1)/2$ without losing any information on $M$. The pseudocode to obtain NCMs, using this approach depicted in Fig.~\ref{ncmdrawing}, can be found in~\cite{ncm}.

Instead of solving the convex optimization in Theorems~\ref{THM:Thm:CV-STEM},~\ref{THM:Thm:CV-STEM-estimation}, and~\ref{THM:Thm:ccm_cvstem} to sample training data, we could also use them directly for training and optimizing the DNN hyperparameters, treating the constraints and the objective functions as the DNN loss functions as demonstrated in~\cite{chuchu}. Although this approach no longer preserves the convexity and can lead only to a sub-optimal solution, this still gives the exponential tracking error bounds as long as we can get sufficiently small $\epsilon_{\ell0}$ and $\epsilon_{\ell1}$ in the learning error assumption \eqref{Eq:learning_error}, as discussed in Theorems~\ref{THM:Thm:contraction_learning},~\ref{THM:Thm:contraction_learning_sto}, and~\ref{THM:Thm:NCMstability1}--\ref{THM:Thm:NCMstability2}. See~\cite{cdc_ncm} for more details on the robustness and stability properties of this approach.
\subsubsection{Lipschitz Condition and Spectral Normalization}
\label{sec_sn_nscm}
Let us first define a spectrally-normalized DNN, a useful mathematical tool designed to overcome the instability of network training by constraining \eqref{neuralnet} of Definition~\ref{DEF:Def:NN_definition} to be Lipschitz, \ie{}, $\exists \ L_{nn} \in\mathbb{R}_{\geq0}$ \st{} $\|\varphi(x)-\varphi(x')\| \leq L_{nn}\|x-x'\|,~\forall x,x'$~\cite{miyato2018spectral,neurallander}. 
\begin{definition}{Spectrally-Normalized DNNs}{Def:SN}
A spectrally-normalized DNN is a DNN of Definition~\ref{DEF:Def:NN_definition} with its weights $W_{\ell}$ normalized as $W_{\ell} = (C_{nn}\Omega_{\ell})/\|\Omega_{\ell}\|$, where $C_{nn} \in\mathbb{R}_{\geq0}$ is a given constant.
\end{definition}
\begin{lemma}
\label{lemma:SNneuralnet}
A spectrally-normalized DNN given in Definition~\ref{DEF:Def:SN} is Lipschitz continuous with its 2-norm Lipschitz constant $C_{nn}^{L+1}L_{\sigma}^L$, where $L_{\sigma}$ is the Lipschitz constant of the activation function $\sigma$ in \eqref{neuralnet}. Also, it is robust to perturbation in its input.
\end{lemma}
\begin{proof}
Let $\|f\|_{\mathrm{Lip}}$ represent the Lipschitz constant of a function $f$ and let $W_{\ell} = (C_{nn}\Omega_{\ell})/\|\Omega_{\ell}\|$. Using the property $\|f_1 \circ f_2\|_{\mathrm{Lip}} \leq \|f_1\|_{\mathrm{Lip}}\|f_2\|_{\mathrm{Lip}}$, we have for a spectrally-normalized DNN in Definition~\ref{DEF:Def:SN} that
\begin{align}
\|\varphi(x;W_{\ell})\| \leq \|\sigma\|_{\mathrm{Lip}}^L\prod_{\ell = 1}^{L+1}\|T_{\ell}\|_{\mathrm{Lip}} = L_{\sigma}^L\prod_{\ell = 1}^{L+1}\|W_{\ell}\|
\end{align}
where we used the fact that $\|f\|_{\mathrm{Lip}}$ of a differentiable function $f$ is equal to the maximum spectral norm (induced $2$-norm) of its gradient over its domain to obtain the last equality. Since $\|W_{\ell}\|=C_{nn}$ holds, the Lipschitz constant of $\varphi(x_i;W_{\ell})$ is $C_{nn}^{L+1}L_{\sigma}^L$ as desired. Since $\varphi$ is Lipschitz, we have for small perturbation $x_{\epsilon}$ that
\begin{align}
\|\varphi(x+x_{\epsilon};W_{\ell})-\varphi(x;W_{\ell})\| &\leq C_{nn}^{L+1}L_{\sigma}^L\|(x+x_{\epsilon})-x\| = C_{nn}^{L+1}L_{\sigma}^L\|x_{\epsilon}\|
\end{align}
implying robustness to input perturbation. \qed
\end{proof}

In general, a spectrally-normalized DNN is useful for improving the robustness and generalization properties of DNNs, and for obtaining the learning error bound as in \eqref{Eq:learning_error}, as delineated in Remark~\ref{remark_learning_error}~\cite{miyato2018spectral,marginbounds1,marginbounds2}. Using it for system identification yields a DNN-based nonlinear feedback controller with a formal stability guarantee as shall be seen in Sec.~\ref{Sec:neurallander}.

For the NCM framework with stochastic perturbation, we can utilize the spectrally-normalized DNN of Definition~\ref{DEF:Def:SN} to guarantee the Lipschitz condition on $\partial M/\partial x_i$, appeared in the stochastic contraction condition of Theorem~\ref{THM:Thm:robuststochastic} (see Proposition~1 of~\cite{nscm}). We could also utilize the technique proposed in~\cite{revay2020lipschitz,revay2021recurrent,cdc_systemid}, which designs Lipschitz-bounded equilibrium neural networks using contraction theory, for obtaining a result analogous to Lemma~\ref{lemma:SNneuralnet}. The pseudocode for the NCM construction in a stochastic setting can be found in~\cite{nscm}.
\begin{figure}
    \centering
    \includegraphics[width=52.5mm]{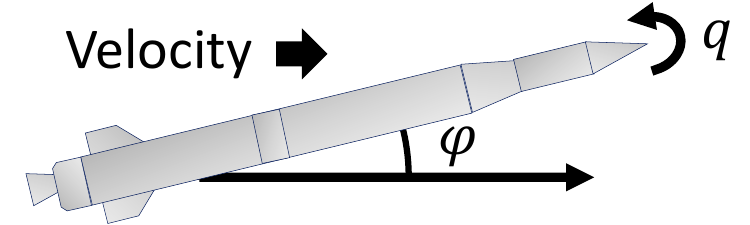}
    \caption{Rocket model (angle of attack $\varphi$, pitch rate $q$).}
    \label{rocket_model}
\end{figure}
\begin{table}[tb]
\caption{Notations in Example~\ref{EX:ex:rocket}. \label{tab:notations_rocket}}
\footnotesize
\begin{center}
\renewcommand{\arraystretch}{1.2}
\rowcolors{1}{uiucbluedark!5}{uiucbluedark!10}
\begin{tabular}{ l l } 
\hline\hline
$\varphi$ & angle of attack (state variable) \\ \arrayrulecolor{mygray}\hline
$q$ & pitch rate (state variable) \\ \hline
$u$ & tail fin deflection (control input) \\ \hline
$y$ & measurement \\ \hline
$M$ & Mach number \\ \hline
$p_0$ & static pressure at 20,000 ft \\ \hline
$S$ & reference area \\ \hline
$d$ & reference diameter \\ \hline
$I_y$ & pitch moment of inertia \\ \hline
$g$ & acceleration of gravity \\ \hline
$m$ & rocket mass \\ \hline
$V$ & rocket speed \\ \arrayrulecolor{black}\hline\hline
\end{tabular}
\end{center}
\end{table}
\begin{example}{Rocket Autopilot}{ex:rocket}
Let us demonstrate how having the Lipschitz condition on $\partial M/\partial x_i$ in Theorem~\ref{THM:Thm:robuststochastic} would affect the NCM-based control and estimation performance. We consider a rocket autopilot problem perturbed by stochastic disturbances as in~\cite{nscm}, the unperturbed dynamics of which is depicted in Fig.~\ref{rocket_model} and given as follows, assuming that the pitch rate $q$ and specific normal force are available as a measurement via rate gyros and accelerometers~\cite{doi:10.2514/3.20997,doi:10.2514/6.1997-3641}:
\begin{align}
\label{rocket_eq}
\dfrac{d}{dt}\begin{bmatrix}\varphi \\ q\end{bmatrix} &= \begin{bmatrix}\frac{gC(M)\cos(\varphi)\phi_z(\varphi,M)}{mV}+q \\ \frac{C(M)d\phi_m(\varphi)}{I_{y}}\end{bmatrix}+\begin{bmatrix}\frac{gC(M)\bar{d}_n\cos(\varphi)}{mV} \\ \frac{C(M)\bar{d}_m}{I_{y}}\end{bmatrix}u \\
\label{rocket_ms}
y &= \begin{bmatrix}q\\\frac{C(M)\phi_z(\varphi,M)}{m}\end{bmatrix}+\begin{bmatrix}0\\\frac{C(M)\bar{d}_n}{m}\end{bmatrix}u
\end{align}
where $C(M) = p_0M^2S$, $\phi_z(\varphi,M)=0.7(a_n\varphi^3+b_n\varphi|\varphi|+c_n(2+M/3)\varphi)$, $\phi_m(\varphi,M)=0.7(a_m\varphi^3+b_m\varphi|\varphi|-c_m(7-8M/3)\varphi)$, $\bar{d}_n=0.7d_n$, $\bar{d}_m=0.7d_m$, $(a_n,b_n,c_n,d_n)$ and $(a_m,b_m,c_m,d_m)$ are given in~\cite{doi:10.2514/6.1997-3641}, and the notations are defined in Table~\ref{tab:notations_rocket}. Since this example explicitly takes into account stochastic perturbation, the spectrally-normalized DNN of Definition~\ref{DEF:Def:SN} is used to guarantee the Lipschitz condition of Theorem~\ref{THM:Thm:robuststochastic} by Lemma~\ref{lemma:SNneuralnet}.

Figure~\ref{est_con_sim} shows the state estimation and tracking error performance of each estimator and controller, where the NSCM is the stochastic counterpart of the NCM aforementioned in Remark~\ref{remark_NSCM_notation}. It is demonstrated that the steady-state errors of the NSCM and CV-STEM are indeed smaller than its steady-state upper bounds of Theorems~\ref{THM:Thm:CV-STEM} and~\ref{THM:Thm:CV-STEM-estimation} (solid black line), while the other controllers violate this condition. In particular, the optimal contraction rate of the deterministic NCM for state estimation is turned out to be much larger than the NSCM as it does not account for stochastic perturbation, which makes the NCM trajectory diverge around $t=5.8$ in Fig.~\ref{est_con_sim}. The NSCM circumvents this difficulty by imposing the Lipschitz condition on $\partial M/\partial x_i$ as in Theorem~\ref{THM:Thm:robuststochastic} using spectral normalization of Lemma~\ref{lemma:SNneuralnet}. See~\cite{nscm} for additional details.
\end{example}
\begin{figure}
    \centering
    \includegraphics[width=132mm]{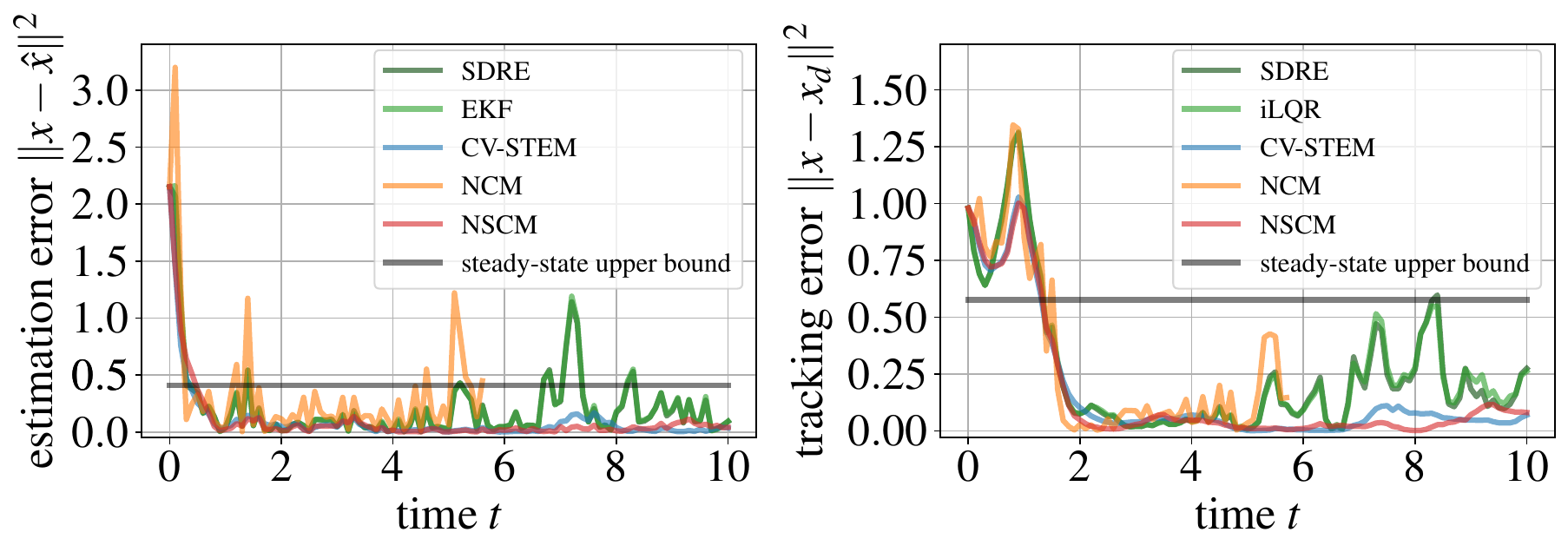}
    \caption{Rocket state estimation and tracking errors in Example~\ref{EX:ex:rocket} ($x = [\varphi,q]^T$).}
    \label{est_con_sim}
\end{figure}
\newpage
\section{Learning-based Robust Motion Planning}
\label{Sec:lagros}
The CV-STEM and NCM are also useful for designing a real-time robust motion planning algorithm for systems perturbed by deterministic and stochastic disturbances as in \eqref{ncm_original_dynamics} and \eqref{ncm_sdc_dynamics}. The purpose of this section is not for proposing a new motion planner to compute $x_d$ and $u_d$ of \eqref{ncm_sdc_dynamicsd}, but for augmenting existing motion planners with real-time learning-based robustness and stability guarantees via contraction theory.
\subsection{Overview of Existing Motion Planners}
Let us briefly review the following standard motion planning techniques and their limitations:
\setlength{\leftmargini}{17pt}     
\begin{enumerate}[label={\color{uiucbluedark}{(\alph*)}}]
	\setlength{\itemsep}{1pt}      
	\setlength{\parskip}{0pt}   
    \item Learning-based motion planner (see, \eg{},~\cite{glas,8593871,NIPS2017_766ebcd5,9001182,NIPS2016_cc7e2b87,8578338,7995721}): $(o_{\ell}(x,o_g),t) \mapsto u_d(o_g,t)$, modeled by a DNN, where $o_g\in\mathbb{R}^{g}$ is a vector containing global environment information, and $o_{\ell}:\mathbb{R}^n\times\mathbb{R}^g\mapsto\mathbb{R}^{\ell}$ with $\ell \leq g$ is local environment information extracted from $o_g \in \mathbb{R}^g$~\cite{glas}. \label{itemFF}
    \item Robust tube-based motion planner (see, \eg{},~\cite{7989693,10.1007/BFb0109870,tube_mpc,tube_nmpc,doi:10.1177/0278364914528132,9290355,ccm,mypaperTAC,ncm,nscm,chuchu,8814758,9303957,L1contraction2,sun2021uncertaintyaware,mypaper,zhao2021tubecertified}): $(x,x_d,u_d,t) \mapsto u^*$, where $u^*$ is a robust feedback tracking controller given by Theorems~\ref{THM:Thm:CV-STEM},~\ref{THM:Thm:ccm_cvstem},~\ref{THM:Thm:ncm_clf},~or~\ref{THM:Thm:ncm_ccm_clf}.\label{itemMP}
\end{enumerate}         

\begin{figure}[b]
    \centering
    \includegraphics[width=127.5mm]{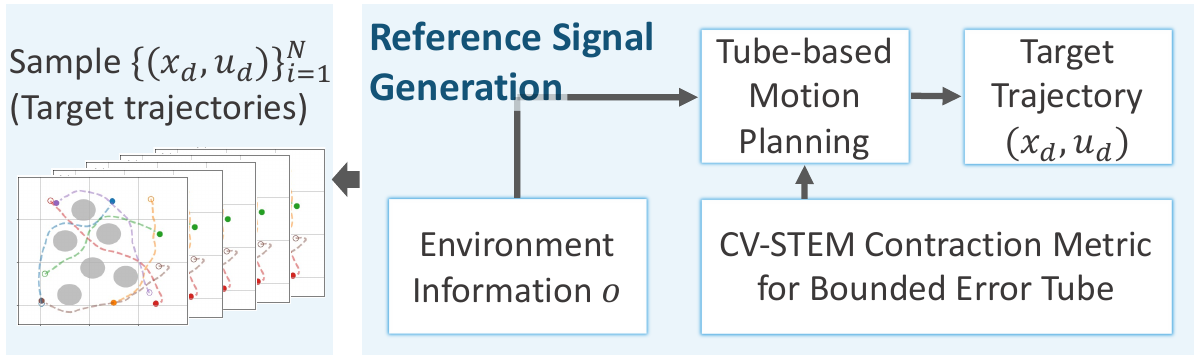}
    \caption{Block diagram of tube-based motion planning using contraction theory.}
        \label{slide3_mp}
\end{figure}

The robust tube-based motion planner~\ref{itemMP} ensures that the perturbed trajectories $x$ of \eqref{ncm_original_dynamics} and \eqref{ncm_sdc_dynamics} stay in an exponentially bounded error tube around the target trajectory $x_d$ of \eqref{ncm_sdc_dynamicsd}~\cite{ccm,7989693,mypaperTAC,ncm,nscm,chuchu,8814758,9303957,L1contraction2,sun2021uncertaintyaware,mypaper,zhao2021tubecertified} given as in Fig.~\ref{slide3_mp}, using Theorems~\ref{THM:Thm:Robust_contraction_original} and~\ref{THM:Thm:robuststochastic}. However, it requires the online computation of $(x_d,u_d)$ as an input to their control policy given in Theorems~\ref{THM:Thm:CV-STEM},~\ref{THM:Thm:ccm_cvstem},~\ref{THM:Thm:ncm_clf}, and~\ref{THM:Thm:ncm_ccm_clf}, which is not realistic for systems with limited computational resources. 
The learning-based motion planner~\ref{itemFF} circumvents this issue by modeling the target policy $(o_{\ell},t) \mapsto u_d$ by a DNN. In essence, our approach, to be proposed in Theorem~\ref{THM:Thm:lagros_stability}, is for providing~\ref{itemFF} with the contraction theory-based stability guarantees of~\ref{itemMP}, thereby significantly enhancing the performance of~\ref{itemFF} that only assures the tracking error $\|x-x_d\|$ to be bounded by a function which increases exponentially with time, as derived previously in Theorem~\ref{THM:Thm:naive_learning} of Sec.~\ref{Sec:learning_stability}.
\subsection{Stability Guarantees of LAG-ROS}
The method of LAG-ROS $-$ for Learning-based Autonomous Guidance with RObustness and Stability guarantees~\cite{lagros}, bridges the gap between~\ref{itemFF} and~\ref{itemMP} by ensuring that the distance between the target and controlled trajectories to be exponentially bounded.
\setlength{\leftmargini}{17pt}     
\begin{enumerate}[label={\color{uiucbluedark}{(\alph*)}},start=3]
	\setlength{\itemsep}{2pt}      
	\setlength{\parskip}{0pt}   
    \item LAG-ROS (see Fig.~\ref{lagrosdrawing} and~\cite{lagros}): $(x,o_{\ell}(x,o_g),t) \mapsto u^*(x,x_d(o_g,t),u_d(o_g,t),t)$, modeled by $u_L$ to be given in Theorem~\ref{THM:Thm:lagros_stability}, where $u^*(x,x_d,u_d,t)$ of \ref{itemMP} is viewed as a function of $(x,o_{\ell},t)$.\label{itemLAGROS}
\end{enumerate} 
Table~\ref{learning_summary} summarizes the differences of the existing motion planners~\ref{itemFF} and~\ref{itemMP} from~\ref{itemLAGROS}, which is defined as follows.
\begin{definition}{LAG-ROS}{Def:lagros}
~Learning-based Autonomous Guidance with RObustness and Stability guarantees (LAG-ROS) is a DNN model for approximating a function which maps state $x$, local environment information $o_{\ell}$, and time $t$ to an optimal robust feedback control input $u^*$ given by \eqref{controller} of Theorem~\ref{THM:Thm:CV-STEM}, \ie{}, $u^*=u_d-K(x-x_d)$, or \eqref{ccm_controller} of Theorem~\ref{THM:Thm:ccm_cvstem}, \ie{}, $u=u_d+\int_{0}^1kd\mu$, where its training data is sampled as explained in Sec.~\ref{sec_robust_sampling} (see Figures~\ref{samplingfig},~\ref{slide4_data}, and~\ref{lagrosdrawing}).
\end{definition}

The LAG-ROS in Definition~\ref{DEF:Def:lagros} achieves online computation of $u$ without solving a motion planning problem, and it still possesses superior robustness and stability properties due to its internal contracting architecture~\cite{lagros}.
\begin{theorem}{Learning-based Motion Planning with Contraction Theory}{Thm:lagros_stability}
Let $u_L=u_L(x,o_{\ell}(x,o_g),t)$ be the LAG-ROS in Definition~\ref{DEF:Def:lagros}, and let $u^*$ be given by \eqref{controller} of Theorem~\ref{THM:Thm:CV-STEM}, \ie{}, $u^*=u_d-K(x-x_d)$, or \eqref{ccm_controller} of Theorem~\ref{THM:Thm:ccm_cvstem}, \ie{}, $u^*=u_d+\int^1_{0}kd\mu$. Define $\Delta_L$ of Theorems~\ref{THM:Thm:contraction_learning} and \ref{THM:Thm:contraction_learning_sto} as follows:
\begin{align}
\label{Eq:DeltaL_lagros}
\Delta_L = B(u_L(x,o_{\ell}(x,o_g),t)-u^*(x,x_d(o_g,t),u_d(o_g,t),t))
\end{align}
where $B=B(x,t)$ is the actuation matrix given in \eqref{ncm_original_dynamics} and \eqref{ncm_sdc_dynamics}. Note that we use $q(0,t)=\xi_0(t)=x_d(o_g,t)$ and $q(1,t)=\xi_1(t)=x(t)$ in the learning-based control formulation of Theorems~\ref{THM:Thm:contraction_learning} and~\ref{THM:Thm:contraction_learning_sto}.

If the LAG-ROS $u_L$ is learned to satisfy \eqref{Eq:learning_error} with $\epsilon_{\ell1}=0$, \ie{}, $\|\Delta_L\| \leq \epsilon_{\ell0}$ for all $x\in\mathcal{S}_s$, $o_g\in\mathcal{S}_o$, and $t\in\mathcal{S}_t$, where $\mathcal{S}_s\subseteq\mathbb{R}^n$, $\mathcal{S}_o\subseteq\mathbb{R}^{g}$, and $\mathcal{S}_t\subseteq\mathbb{R}_{\geq0}$ are some compact sets, and if the control-affine nonlinear systems \eqref{ncm_original_dynamics} and \eqref{ncm_sdc_dynamics} are controlled by $u=u_L$, then \eqref{cont_learning_bound} of Theorem~\ref{THM:Thm:contraction_learning} and \eqref{cont_learning_bound_sto} of Theorem~\ref{THM:Thm:contraction_learning_sto} hold with $\bar{d}$ and $\bar{g}$ replaced as in \eqref{Eq:NCM_control_bound}, \ie{}, we have the following for $\mathtt{e}(t)=x(t)-x_d(o_g,t)$:
\begin{align}
\label{learning_bound_lagros}
\|\mathtt{e}(t)\| &\leq \frac{V_{\ell}(0)}{\sqrt{\underline{m}}}e^{-\alpha t}+\frac{\bar{d}_{\epsilon_{\ell}}}{\alpha}\sqrt{\frac{\overline{m}}{\underline{m}}}(1-e^{-\alpha t})=r_{\ell}(t)~~~~~~~~ \\
\label{learning_bound_lagros_sto}
\mathop{\mathbb{E}}\left[\|\mathtt{e}(t)\|^2\right] &\leq \frac{\mathop{\mathbb{E}}[V_{s\ell}(0)]}{\underline{m}}e^{-2\alpha t}+\frac{C_{\epsilon_{\ell}}}{2\alpha}\frac{\overline{m}}{\underline{m}}=r_{s\ell}(t) \\
\label{learning_bound_lagros_sto_prob}
\mathop{\mathbb{P}}\left[\|\mathtt{e}(t)\|\geq\varepsilon\right] &\leq \frac{1}{\varepsilon^2}\left(\frac{\mathop{\mathbb{E}}[V_{s\ell}(0)]}{\underline{m}}e^{-2\alpha t}+\frac{C_{\epsilon_{\ell}}}{2\alpha}\frac{\overline{m}}{\underline{m}}\right)
\end{align}
where $x_d=x_d(o_g,t)$ and $u_d=u_d(o_g,t)$ are as given in \eqref{ncm_sdc_dynamicsd}, $\bar{d}_{\epsilon_{\ell}}=\epsilon_{\ell0}+\bar{d}_c$, $C_{\epsilon_{\ell}} = \bar{g}_c^2({2}{\alpha_G}^{-1}+1)+\epsilon_{\ell0}^2\alpha_d^{-1}$, and the other variables are as defined in Theorems~\ref{THM:Thm:contraction_learning} and \ref{THM:Thm:contraction_learning_sto} with the disturbance terms in \eqref{Eq:NCM_control_bound} of Theorem~\ref{THM:Thm:NCMstability1}.
\end{theorem}
\begin{proof}
We have $f(x,t)+B(x,t)u_L=f(x,t)+B(x,t)u^*+B(x,t)(u_L-u^*)$ and $\|\Delta_L\|=\|B(x,t)(u_L-u^*)\| \leq \epsilon_{\ell0}$ by \eqref{Eq:learning_error} with $\epsilon_{\ell1}=0$ for $x_d=x_d(o_g,t)$ and $u_d=u_d(o_g,t)$. Since the virtual system which has $x$ of $\dot{x} = f(x,t)+B(x,t)u^*$ and $x_d$ of \eqref{ncm_sdc_dynamicsd} as its particular solutions is contracting due to Theorems~\ref{THM:Thm:CV-STEM} or~\ref{THM:Thm:ccm_cvstem}, the application of Theorems~\ref{THM:Thm:contraction_learning} and \ref{THM:Thm:contraction_learning_sto} yields the desired relations \eqref{learning_bound_lagros}--\eqref{learning_bound_lagros_sto_prob}. \qed
\end{proof}
\begin{table}[tb]
\caption{Comparison of LAG-ROS~\cite{lagros} with the Learning-based and Robust Tube-based Planners. \label{learning_summary}}
\footnotesize
\begin{center}
\renewcommand{\arraystretch}{1.2}
\rowcolors{1}{uiucbluedark!5}{uiucbluedark!10}
\begin{tabular}{ l l m{4.5cm} m{3.7cm}}
\hline\hline
Motion planning scheme & Control policy & State tracking error $\|x-x_d\|$ & Computational load \\
\hline
\ref{itemFF} Learning-based & $(o_{\ell},t) \mapsto u_d$ & Increases exponentially (Theorem~\ref{THM:Thm:naive_learning}) & One DNN evaluation \\ \arrayrulecolor{mygray}\hline
\ref{itemMP} Robust tube-based & $(x,x_d,u_d,t) \mapsto u^*$ & Exponentially bounded (Theorems~\ref{THM:Thm:CV-STEM},~\ref{THM:Thm:ccm_cvstem},~\ref{THM:Thm:ncm_clf},~\ref{THM:Thm:ncm_ccm_clf}) & Computation of $(x_d,u_d)$ \\ \hline
\ref{itemLAGROS} LAG-ROS & $(x,o_{\ell},t) \mapsto u^*$ & Exponentially bounded (Theorem~\ref{THM:Thm:lagros_stability}) & One DNN evaluation \\
\arrayrulecolor{black}\hline\hline
\end{tabular}
\end{center}
\end{table}
\renewcommand{\arraystretch}{1.0}

Due to its internal feedback structure, the LAG-ROS achieves exponential boundedness of the trajectory tracking error as in Theorem~\ref{THM:Thm:lagros_stability}, thereby improving the performance of existing learning-based feedforward control laws to model $(o_{\ell},t)\mapsto u_d$ as in~\ref{itemFF}, whose tracking error bound increases exponentially with time as in Theorem~\ref{THM:Thm:naive_learning} (see~\cite{lagros}). This property enables its use in safety-critical guidance and control tasks which we often encounter in modern machine learning applications.
\begin{figure}[tb]
\centerline{
\subfloat{\includegraphics[clip, width=37.5mm]{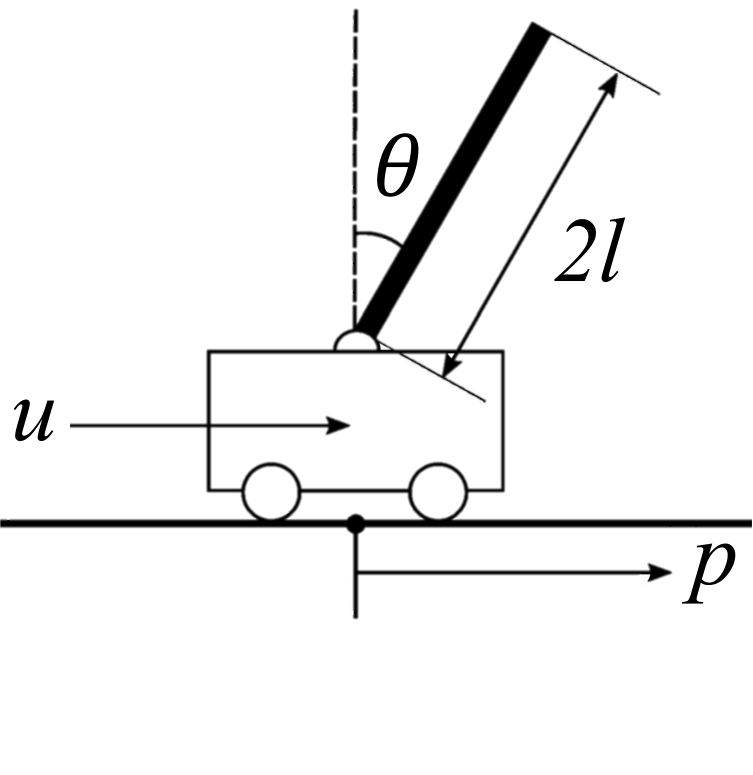}}
\hfil
\subfloat{\includegraphics[clip, width=60mm]{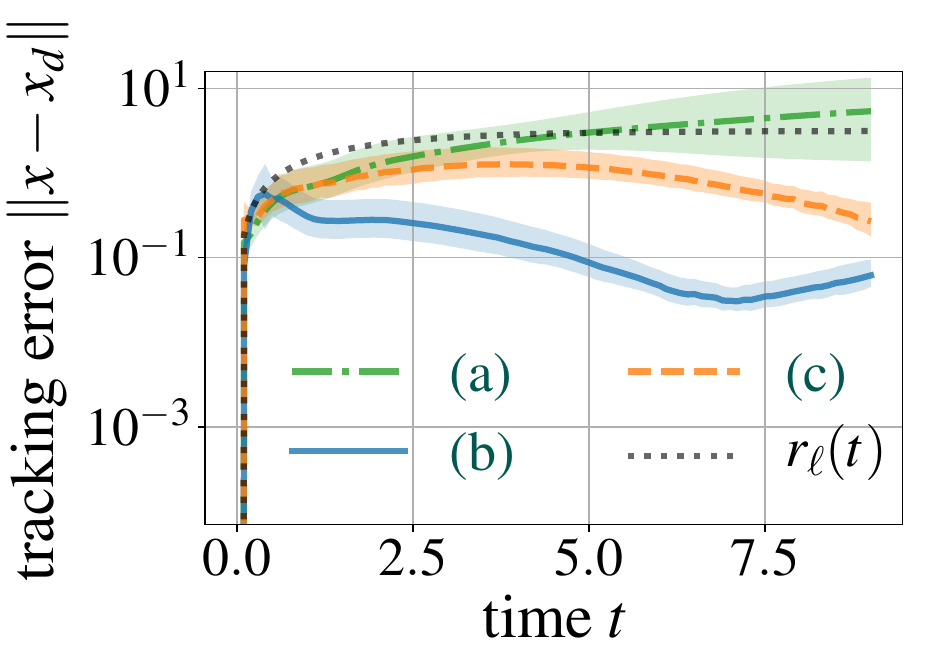}}
}
\caption{Cart-pole balancing task: $x=[p,\theta,\dot{p},\dot{\theta}]^{\top}$, $x$ and $x_d$ are given in \eqref{ncm_original_dynamics} and \eqref{ncm_sdc_dynamicsd}, \ref{itemFF}--\ref{itemLAGROS} are given in Table~\ref{learning_summary}, and $r_{\ell}$ is given in \eqref{exp_bound_cp}. The shaded area denotes the standard deviation ($+1\sigma$ and $-0.5\sigma$).}
\label{lagros_cp_dist_fig}
\end{figure}
\begin{example}{Cart-Pole Balancing}{ex:cartpole}
Let us consider the cart-pole balancing task in Fig.~\ref{lagros_cp_dist_fig}, perturbed externally by deterministic perturbation, to demonstrate the differences of \ref{itemFF}--\ref{itemLAGROS} summarized in Table~\ref{learning_summary}. Its dynamical system is given in~\cite{ancm,6313077} with its values defined in~\cite{6313077}.

The target trajectories $(x_d,u_d)$ are sampled with the objective function $\int_0^{T}\|u\|^2dt$, and the LAG-ROS of Theorem~\ref{THM:Thm:lagros_stability} is modeled by a DNN independently of the target trajectory as detailed in~\cite{lagros}. The environment information $o_g$ is selected as random initial and target terminal states, and we let $V_{\ell}(0)=0$ and $\epsilon_{\ell0}+\bar{d}_c = 0.75$ in \eqref{learning_bound_lagros} of Theorem~\ref{THM:Thm:lagros_stability}, yielding the following exponential bound:
\begin{align}
\label{exp_bound_cp}
r_{\ell}(t) = \frac{V_{\ell}(0)}{\sqrt{\underline{m}}}+\frac{\bar{d}_{\epsilon_{\ell}}}{\alpha}\sqrt{\frac{\overline{m}}{\underline{m}}}(1-e^{-\alpha t}) = 3.15(1-e^{-0.60 t})~~~~~~
\end{align}
where $\bar{d}_{\epsilon_{\ell}}=\epsilon_{\ell0}+\bar{d}_c$.

Figure~\ref{lagros_cp_dist_fig} shows the tracking errors of each motion planner averaged over $50$ simulations at each time instant $t$. The LAG-ROS \ref{itemLAGROS} and robust tube-based planner~\ref{itemMP} indeed satisfy the bound \eqref{exp_bound_cp} (dotted black curve) for all $t$ with a small standard deviation $\sigma$, unlike learning-based motion planner~\ref{itemFF} with a diverging bound in Theorem~\ref{THM:Thm:naive_learning} and increasing deviation $\sigma$. This example demonstrates one of the major advantages of contraction theory, which enables such quantitative analysis on robustness and stability of learning-based planners.
\end{example}

As implied in Example~\ref{EX:ex:cartpole}, the LAG-ROS indeed possesses the robustness and stability guarantees of $u^*$ as proven in Theorem~\ref{THM:Thm:lagros_stability}, unlike~\ref{itemFF}, while retaining significantly lower computational cost than that of~\ref{itemMP}. See~\cite{lagros} for a more detailed discussion of this simulation result.
\subsection{Tube-based State Constraint Satisfaction}
\label{sec_robust_sampling}
We exploit the result of Theorem~\ref{THM:Thm:lagros_stability} in the tube-based motion planning~\cite{7989693} for generating LAG-ROS training data, which satisfies given state constraints even under the existence of the learning error and disturbance of Theorem~\ref{THM:Thm:contraction_learning}. In particular, we sample target trajectories $(x_d,u_d)$ of \eqref{ncm_sdc_dynamicsd} in $\mathcal{S}_s\times\mathcal{S}_u$ of Theorem~\ref{THM:Thm:lagros_stability} by solving the following, assuming the learning error $\epsilon_{\ell0}$ and disturbance upper bound $\bar{d}_c$ of Theorem~\ref{THM:Thm:lagros_stability} are selected \textit{a priori}~\cite{neurallander}:
\begin{align}
\label{tube_motion_plan}
&\min_{\substack{\bar{x}=\bar{x}(o_g,t)\\\bar{u}=\bar{u}(o_g,t)}}~\int_{0}^{T}c_0\|\bar{u}\|^2+c_1P(\bar{x},\bar{u},t)dt \\
&\text{\st{} $\dot{\bar{x}} = f(\bar{x},t)+B(\bar{x},t)\bar{u}$, $\bar{x}\in\bar{\mathcal{X}}(o_g,t)$, and $\bar{u} \in\bar{\mathcal{U}}(o_g,t)$}
\end{align}
where $c_0>0$, $c_1\geq0$, $P(\bar{x},\bar{u},t)$ is some performance-based cost function (\eg{}, information-based cost~\cite{doi:10.2514/6.2021-1103}), $\bar{\mathcal{X}}$ is robust admissible state space defined as $\bar{\mathcal{X}}(o_g,t)=\{v(t)\in\mathbb{R}^n|\forall \xi(t) \in \{\xi(t)|\|v(t)-\xi(t)\|\leq r_{\ell}(t)\},~\xi(t)\in\mathcal{X}(o_g,t)\}$, $\mathcal{X}(o_g,t)$ is given admissible state space, $r_{\ell}(t)$ is given by \eqref{learning_bound_lagros} of Theorem~\ref{THM:Thm:lagros_stability}, and $\bar{u} \in\bar{\mathcal{U}}(o_g,t)$ is an input constraint. The following theorem shows that the LAG-ROS ensures the perturbed state $x$ of \eqref{ncm_original_dynamics} to satisfy $x \in \mathcal{X}$, due to the contracting property of $u^*$.
\begin{lemma}
\label{tube_lemma}
If the solution of \eqref{tube_motion_plan} gives $(x_d,u_d)$, the LAG-ROS $u_L$ of Theorem~\ref{THM:Thm:lagros_stability} ensures the perturbed solution $x(t)$ of $\dot{x}=f(x,t)+B(x,t)u_L+d_c(x,t)$ in \eqref{ncm_original_dynamics} to stay in the admissible state space $\mathcal{X}(o_g,t)$, \ie{}, $x(t) \in \mathcal{X}(o_g,t)$, even with the learning error $\epsilon_{\ell0}$ of $u_L$ in Theorem~\ref{THM:Thm:lagros_stability}.
\end{lemma}
\begin{proof}
See~\cite{lagros}. \qed
\end{proof}

Lemma~\ref{tube_lemma} implies that the perturbed trajectory \eqref{ncm_original_dynamics} controlled by LAG-ROS will not violate the given state constraints as long as $x_d$ is sampled by \eqref{tube_motion_plan}, which helps greatly reduce the need for safety control schemes such as~\cite{8405547}. The localization method in~\cite{glas} allows extracting $o_{\ell}$ of~\ref{itemFF} by $o_g$ of \eqref{tube_motion_plan}, to render LAG-ROS applicable to different environments in a distributed way with a single policy.

\begin{figure}[tb]
    \centering
    \includegraphics[width=105mm]{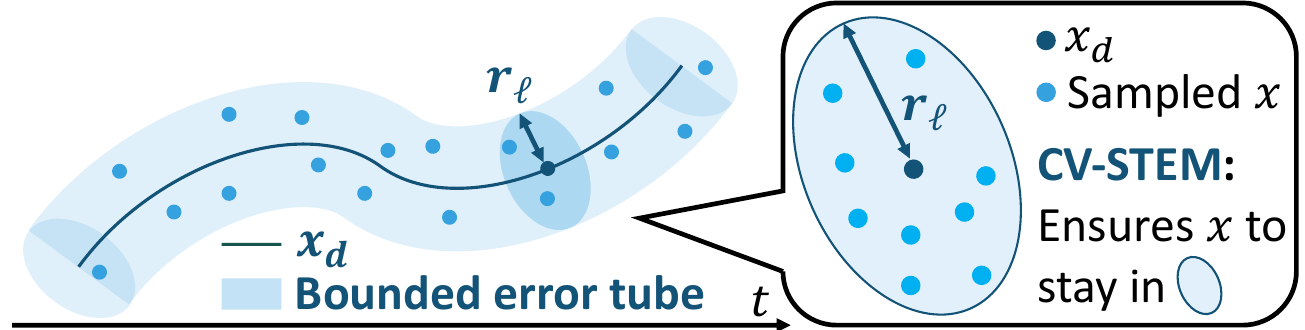}
    \caption{Illustration of state sampling in robust bounded error tube.}
    \label{samplingfig}
\end{figure}

\begin{figure}[tb]
    \centering
    \includegraphics[width=127.5mm]{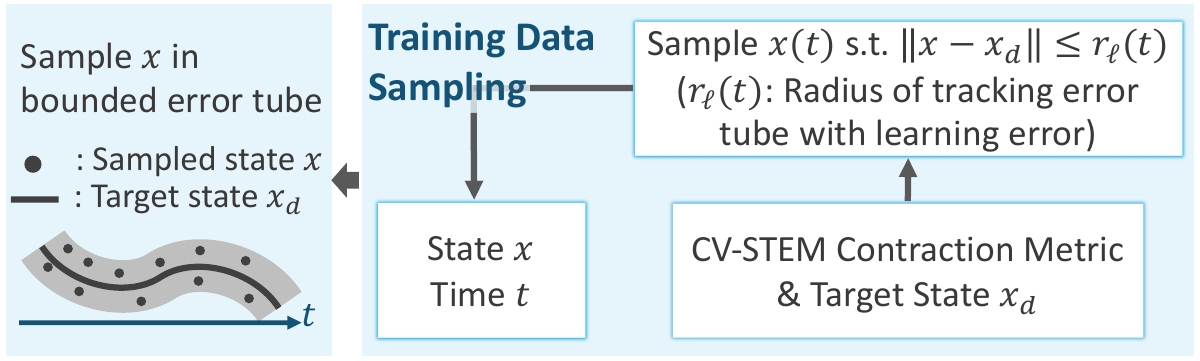}
    \caption{Block diagram of state sampling in robust bounded error tube.}
        \label{slide4_data}
\end{figure}

\begin{figure}[tb]
    \centering
    \includegraphics[width=160mm]{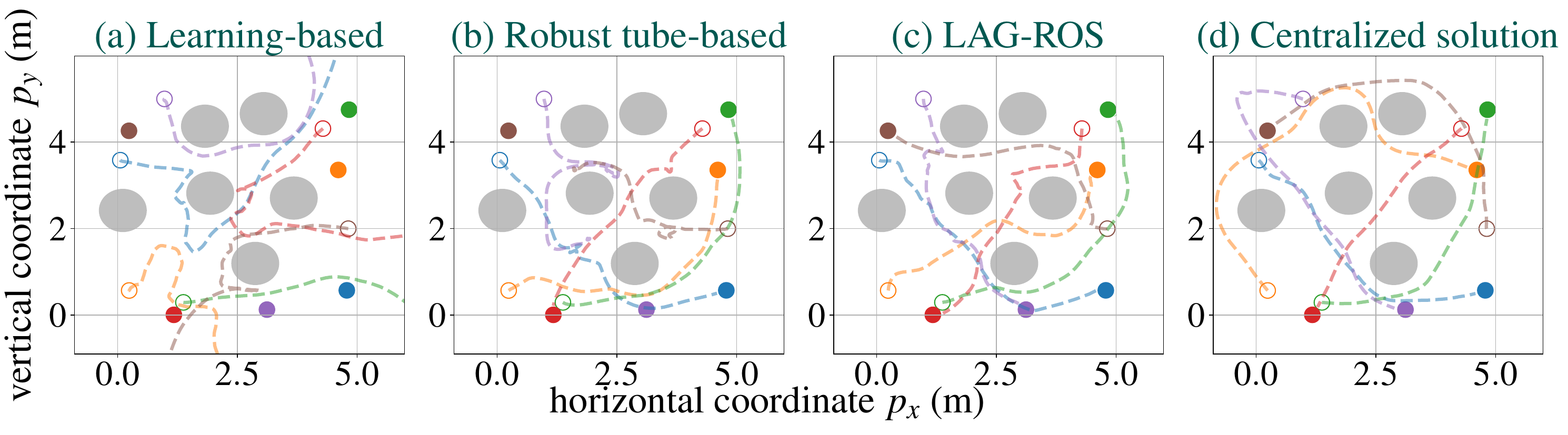}
    \caption{Trajectories for the learning-based planner~\ref{itemFF}, robust tube-based planner~\ref{itemMP}, LAG-ROS~\ref{itemLAGROS}, and offline centralized solution~\ref{itemC} ($\circ$: start, $\bullet$: goal).}
    \label{lagros_sc_fig}
\end{figure}

\begin{figure}[b]
    \centering
    \includegraphics[width=160mm]{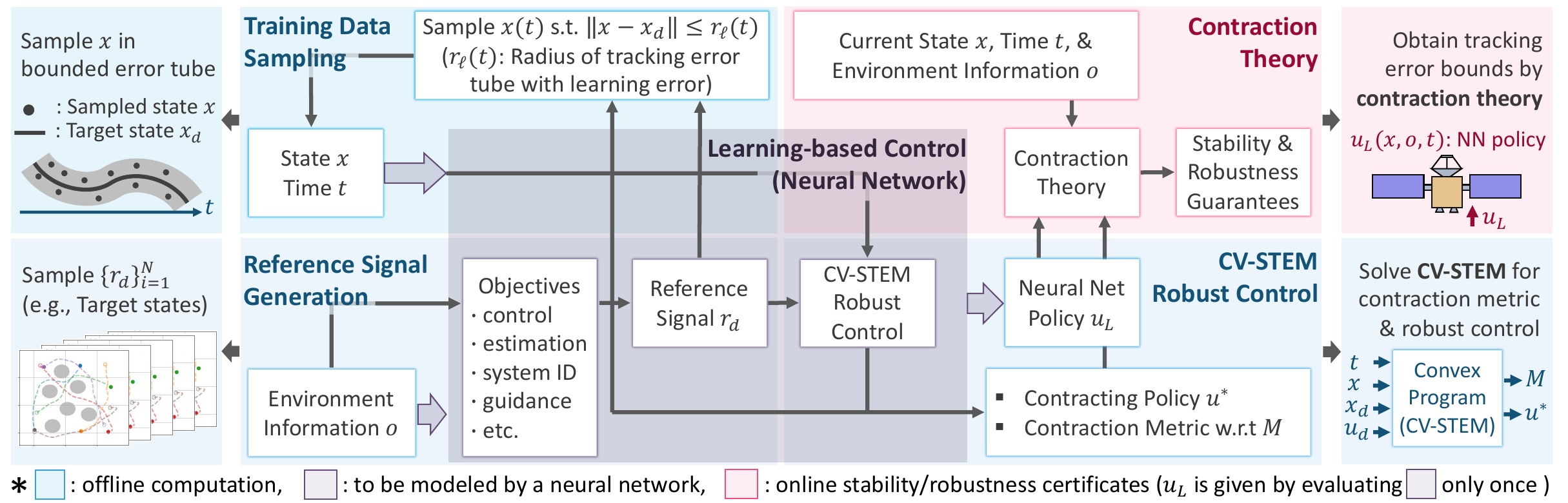}
    \caption{Detailed block diagram of machine learning-based control using contraction theory, including LAG-ROS, where Fig.~\ref{slide2_cvstem}, Fig.~\ref{slide3_mp}, and Fig.~\ref{slide4_data} are utilized as building blocks.}
    \label{lagrosdrawing}
\end{figure}

We therefore sample artificially perturbed states $x$ in the tube $S(x_d(o_g,t)) = \{\xi(t)\in\mathbb{R}^n|\|\xi(t)-x_d(o_g,t)\|\leq r_{\ell}(t)\}$ as depicted in Figures~\ref{samplingfig} and~\ref{slide4_data}~\cite{lagros}. The robust control inputs $u^*$ for training LAG-ROS of Theorem~\ref{THM:Thm:lagros_stability} is then sampled by computing the CV-STEM of Theorem~\ref{THM:Thm:CV-STEM} or Theorem~\ref{THM:Thm:ccm_cvstem}. The LAG-ROS framework is summarized in Fig.~\ref{lagrosdrawing}, and a pseudocode for its offline construction can be found in~\cite{lagros}.
\begin{remark}
For stochastic systems, we could sample $x$ around $x_d$ using a given probability distribution of the disturbance using \eqref{learning_bound_lagros_sto} and \eqref{learning_bound_lagros_sto_prob} of Theorem~\ref{THM:Thm:lagros_stability}.
\end{remark}
\begin{example}{Planar Spacecraft Simulators}{ex:sc_lagros}
Let us again consider the spacecraft dynamics of Example~\ref{EX:ex:sc_robust_control}, but now for designing a real-time robust motion planner using the LAG-ROS in a multi-agent setting. In this example, each agent is assumed to have access only to the states of the other agents/obstacles located in a circle with a limited radius, $2.0$m, centered around it. The performance of LAG-ROS is compared with~\ref{itemFF}, \ref{itemMP}, and the following centralized motion planner~\ref{itemC} which is not computable in real-time:
\setlength{\leftmargini}{17pt}     
\begin{enumerate}[label={\color{uiucbluedark}{(\alph*)}},start=4]
	\setlength{\itemsep}{1pt}      
	\setlength{\parskip}{0pt}   
    \item Centralized robust motion planner:\\
    $(x,x_d,u_d,t) \mapsto u^*$, offline centralized solution of \ref{itemMP}. \label{itemC}
\end{enumerate}
The objective function for sampling target trajectories of \eqref{ncm_sdc_dynamicsd} is selected as $\int_0^{T}\|u\|^2dt$ with their target terminal positions $x_f=[p_{xf},p_{yf},0,0]^{\top}$, where $(p_{xf},p_{yf})$ is a random position in $(0,0) \leq (p_{xf},p_{yf}) \leq (5,5)$. For the state constraint satisfaction in Lemma~\ref{tube_lemma}, the following error tube \eqref{learning_bound_lagros}:
\begin{align}
r_{\ell}(t) = \frac{\bar{d}_{\epsilon_{\ell}}}{{\alpha}}\sqrt{\chi}(1-e^{-\alpha t}) = 0.125(1-e^{-0.30 t}) \nonumber
\end{align}
is used with an input constraint $u_i\geq 0,~\forall i$, to avoid collisions with a random number of multiple circular obstacles and of other agents, even under the learning error and disturbances. See~\cite{lagros} for the LAG-ROS training details~\cite{lagros}.

Figure~\ref{lagros_sc_fig} shows one example of the trajectories of the motion planners \ref{itemFF}--\ref{itemC} under deterministic external disturbances. It implies the following:
\setlength{\leftmargini}{17pt}     
\begin{itemize}
	\setlength{\itemsep}{1pt}      
	\setlength{\parskip}{0pt}      
    \item For~\ref{itemFF}, the tracking error accumulates exponentially with time due to the lack of robustness as proven in Theorem~\ref{THM:Thm:naive_learning}.
    \item \ref{itemMP}~only yields locally optimal $(x_d,u_d)$ of \eqref{ncm_sdc_dynamicsd}, as its time horizon has to be small enough to make the problem solvable online with limited computational capacity, only with local environment information. This renders some agents stuck in local minima as depicted in Fig.~\ref{lagros_sc_fig}.
    \item LAG-ROS \ref{itemLAGROS} tackles these two problems by providing formal robustness and stability guarantees of Theorem~\ref{THM:Thm:lagros_stability}, whilst implicitly knowing the global solution (only from the local information $o_{\ell}$ as in~\cite{glas}) without computing it online. It indeed satisfies the state constraints due to Lemma~\ref{tube_lemma} as can be seen from Fig.~\ref{lagros_sc_fig}.
\end{itemize}
See~\cite{lagros} for the in-depth discussion on this simulation result.
\end{example}
\subsection{Safe Exploration and Learning of Disturbances}
All the theorems so far assume that the sizes of unknown external disturbances are fixed (\ie{}, $\bar{d}$ in Theorem~\ref{THM:Thm:Robust_contraction_original} and $(\bar{g}_{0},\bar{g}_1)$ in Theorem~\ref{THM:Thm:robuststochastic} are known and fixed). Since such an assumption could yield conservative state constraints, \eg{}, in utilizing motion planning of Lemma~\ref{tube_lemma} with the tube of Theorem~\ref{THM:Thm:lagros_stability}, we could consider better estimating the unknown parts, $d(x,t)$ in Theorem~\ref{THM:Thm:Robust_contraction_original} or $G_{0}(x,t)$ and $G_{1}(x,t)$ in Theorem~\ref{THM:Thm:robuststochastic}, also using contraction theory and machine learning. For example, it is demonstrated in~\cite{9290355} that the stochastic bounds of Theorem~\ref{THM:Thm:robuststochastic} can be used to ensure safe exploration and reduction of uncertainty over epochs in learning unknown, control non-affine residual dynamics. Here, its optimal motion plans are designed by solving chance-constrained trajectory optimization and executed using a feedback controller such as Theorem~\ref{THM:Thm:CV-STEM} with a control barrier function-based safety filter~\cite{7040372,8405547}. It is also shown in~\cite{9303957,L1contraction2,sun2021uncertaintyaware} that the state-dependent disturbance $d(x)$ in Theorem~\ref{THM:Thm:robuststochastic} can be learned adaptively with the Gaussian process while ensuring safety. 

These techniques are based on the robustness and stability properties of contraction theory, which guarantee their state trajectories to stay in a tube around the target trajectory as in Theorems~\ref{THM:Thm:Robust_contraction_original},~\ref{THM:Thm:robuststochastic},~\ref{THM:Thm:contraction_learning}, and~\ref{THM:Thm:contraction_learning_sto}, and at the same time, enable safely sampling training data for learning the unknown part of the dynamics. Sections~\ref{Sec:adaptive} and~\ref{Sec:datadriven} are for giving an overview of these concepts in the context of contraction theory-based adaptive control~\cite{ancm,lopez2021universal,9109296} and robust control synthesis for learned models~\cite{neurallander,47710,boffi2020learning,cdc_systemid,7992901}.
\newpage
\section{Learning-based Adaptive Control}
\label{Sec:adaptive}
In this section, we consider the following smooth nonlinear system with an uncertain parameter $\theta\in\mathbb{R}^{c}$:
\begin{align}
\label{adaptive_general}
\begin{aligned}
    \dot{x} &= f(x,\theta)+B(x,\theta)u \\
    \dot{x}_d &= f(x_d,\theta)+B(x_d,\theta)u_d
\end{aligned}
\end{align}
where $x$, $u$, $x_d$, and $u_d$ are as defined in \eqref{ncm_original_dynamics} and \eqref{ncm_sdc_dynamicsd}, and ${f}:
\mathbb{R}^n\times\mathbb{R}^c\mapsto\mathbb{R}^n$ and $B:\mathbb{R}^n\times\mathbb{R}^c\mapsto\mathbb{R}^{n\times m}$ are known smooth functions with the uncertain parameter $\theta\in\mathbb{R}^{c}$. Due to Theorems~\ref{THM:Thm:Robust_contraction_original} and \ref{THM:Thm:contraction_learning}, we can see that the robust control techniques presented earlier in Theorems~\ref{THM:Thm:CV-STEM},~\ref{THM:Thm:ccm_cvstem},~\ref{THM:Thm:ncm_clf}, and~\ref{THM:Thm:ncm_ccm_clf} are still useful in this case if the modeling errors $\|f(x,\theta)-f(x,\theta_n)\|$ and $\|B(x,\theta)-B(x,\theta_n)\|$ are bounded, where $\theta_n$ is a nominal guess of the true parameter $\theta$. However, there are situations where such assumptions are not necessarily true.

We present a method of deep learning-based adaptive control for nonlinear systems with parametric uncertainty, thereby further improving the real-time performance of robust control in Theorems~\ref{THM:Thm:contraction_learning} and~\ref{THM:Thm:contraction_learning_sto} for model-based systems, and Theorems~\ref{THM:Thm:NCMstability_modelfree} and~\ref{THM:Thm:neurallander} for model-free systems. Although we consider continuous-time dynamical systems in this section, discrete-time changes could be incorporated in this framework using~\cite{regretboffi,1469901}. Also, note that the techniques in this section~\cite{ancm} can be used with differential state feedback frameworks~\cite{ccm,7989693,47710,WANG201944,vccm,rccm} as described in Theorem~\ref{THM:Thm:ccm_cvstem}, trading off added computational cost for generality (see Table~\ref{tab:sdcccm_summary} and~\cite{9109296,lopez2021universal}).
\subsection{Adaptive Control with CV-STEM and NCM}
\label{sec:matched_uncertainty}
Let us start with the following simple case.
\begin{assumption}
\label{Asmp:affine}
The matrix $B$ in \eqref{adaptive_general} does not depend on $\theta$, and $\exists \Pi(x) \in \mathbb{R}^{{c}\times n}$ \st{} $\Pi(x)^{\top}\vartheta = f(x,\theta_n)-f(x,\theta)$, where $\vartheta=\theta_n-\theta$ and $\theta_n$ is a nominal guess of the uncertain parameter $\theta$.
\end{assumption}
Under Assumption~\ref{Asmp:affine}, we can write \eqref{adaptive_general} as
\begin{align}
\label{adaptive_nominal_sys}
\dot{x}=f(x,\theta_n)+B(x)u-\Pi(x)^{\top}\vartheta
\end{align}
leading to the following theorem~\cite{ancm} for the NCM-based adaptive control. Note that we could also use the SDC formulation with respect to a fixed point as delineated in Theorem~\ref{THM:Thm:fixed_sdc}~\cite{mypaperTAC,Ref:Stochastic,mypaper}.
\begin{theorem}{Learning-based Adaptive Control with Contraction Theory}{Thm:adaptive_robust_affine}
Suppose that Assumption~\ref{Asmp:affine} holds and let $\mathcal{M}$ defines the NCM of Definition~\ref{DEF:Def:NCM}, which models $M$ of the CV-STEM contraction metric in Theorem~\ref{THM:Thm:CV-STEM} for the nominal system \eqref{adaptive_general} with $\theta = \theta_n$, constructed with an additional convex constraint given as $\partial_{b_i(x)}\bar{W}+\partial_{b_i(x_d)}\bar{W}=0$, where $\partial_{b_i(q)}\bar{W}=\sum_{i}(\partial \bar{W}/\partial q_i)b_i(q)$ for $B=[b_1,\cdots,b_m]$ (see~\cite{ccm,9109296}). Suppose also that the matched uncertainty condition~\cite{9109296} holds, \ie{}, $(\Pi(x)-\Pi(x_d))^{\top}\vartheta \in \textrm{span}(B(x))$ for $\Pi(x)$, and that \eqref{adaptive_general} is controlled by the following adaptive control law:
\begin{align}
\label{affine_adaptive_u_ncm}
u &= u_L+\varphi(x,x_d)^{\top} \hat{\vartheta} \\
\label{affine_adaptation_ncm}
\dot{\hat{\vartheta}} &= -\Gamma (\varphi(x,x_d)B(x)^{\top}\mathcal{M}(x,x_d,u_d)(x-x_d)+\sigma \hat{\vartheta})
\end{align}
where $u_L$ is given by \eqref{NCM_controller} of Theorem~\ref{THM:Thm:NCMstability2}, \ie{}, $u_L=u_d-R^{-1}B^{\top}\mathcal{M}(x-x_d)$, $\Gamma \in \mathbb{R}^{{c}\times {c}}$ is a diagonal matrix with positive elements that governs the rate of adaptation, $\sigma \in \mathbb{R}_{\geq 0}$, and $(\Pi(x)-\Pi(x_d))^{\top}\vartheta=B(x)\varphi(x,x_d)^{\top}\vartheta$.

If $\exists \underline{\gamma},\overline{\gamma},\bar{b},\bar{\rho},\bar{\phi},\bar{\vartheta} \in \mathbb{R}_{>0}$ \st{} $\underline{\gamma} \mathrm{I} \preceq \Gamma \preceq \overline{\gamma}\mathrm{I}$, $\|B(x)\| \leq \bar{b}$, $\|R^{-1}(x,x_d,u_d)\| \leq \bar{\rho}$, $\|\varphi(x,x_d)\| \leq \bar{\phi}$, and $\|\vartheta\| \leq \bar{\vartheta}$, and if $\Gamma$ and $\sigma$ of \eqref{affine_adaptation_ncm} are selected to satisfy the following relation for the learning error $\|\mathcal{M}-M\| \leq \epsilon_{\ell}$ in some compact set $\mathcal{S}$ as in \eqref{Eq:Merror_control} of Theorem~\ref{THM:Thm:NCMstability2}:
\begin{align}
\label{condition_mNCM}
\begin{bmatrix}
-2\alpha_{\ell}\underline{m} & \bar{\phi}\bar{b}\epsilon_{\ell} \\
\bar{\phi}\bar{b}\epsilon_{\ell} & -2\sigma
\end{bmatrix} \preceq -2 \alpha_{a}\begin{bmatrix}
\overline{m} & 0 \\
0 & 1/\underline{\gamma}
\end{bmatrix}
\end{align}
for $\exists \alpha_{a} \in \mathbb{R}_{>0}$, $\alpha_{\ell}$ given in Theorem~\ref{THM:Thm:NCMstability2}, and $\underline{m}$ and $\overline{m}$ given in $\underline{m}\mathrm{I}\preceq M \preceq \overline{m}\mathrm{I}$ of \eqref{Mcon}, then the system \eqref{adaptive_general} is robust against bounded deterministic and stochastic disturbances with $\sigma\neq0$, and we have the following bound in the compact set $\mathcal{S}$:
\begin{align}
\label{adaptive_bound_1}
\|\mathtt{e}(t)\| \leq \frac{V_{\ell}(0)e^{-\alpha_a t}+\alpha_a^{-1}\sigma\sqrt{\overline{\gamma}}\bar{\vartheta}(1-e^{-\alpha_a t})}{\sqrt{\underline{m}}}
\end{align}
where $\mathtt{e}=x-x_d$, and $V_{\ell}=\int^{\xi_1}_{\xi_0}\|\Theta\delta q\|$ is defined in Theorem~\ref{THM:Thm:path_integral} with $M=\Theta^{\top}\Theta$ replaced by $\mathrm{diag}(M,\Gamma^{-1})$ for $\xi_0=[x_d^{\top},\vartheta^{\top}]^{\top}$ and $\xi_1=[x^{\top},\hat{\vartheta}^{\top}]^{\top}$. Furthermore, if the learning error $\epsilon_{\ell}=0$ (CV-STEM control), \eqref{affine_adaptation_ncm} with $\sigma = 0$ guarantees asymptotic stability of $x$ to $x_d$ in \eqref{adaptive_general}.
\end{theorem}
\begin{proof}
The proof can be found in~\cite{ancm}, but here we emphasize the use of contraction theory. For $u$ given by \eqref{affine_adaptive_u_ncm}, the virtual system of a smooth path $q(\mu,t) = [q_x^{\top},q_{\vartheta}^{\top}]^{\top}$ parameterized by $\mu\in[0,1]$, which has $q(\mu=0,t)=[x_d^{\top},\vartheta^{\top}]^{\top}$ and $q(\mu=1,t) =[x^{\top},\hat{\vartheta}^{\top}]^{\top}$ as its particular solutions, is given as follows:
\begin{align}
\label{virtual_adaptive}
\dot{q} = \begin{bmatrix}\zeta(q_x,x,x_d,u_d)+B\varphi^{\top} q_{\vartheta}-\Pi(x)^{\top}\vartheta \\
-\Gamma(\varphi B^{\top}\mathcal{M}(q_x-x_d)+\sigma(q_{\vartheta}-\vartheta))+d_{q\vartheta}(\mu,\vartheta)
\end{bmatrix}
\end{align}
where $d_{q\vartheta}(\mu,\vartheta) = -\mu\sigma\vartheta$. Note that $\zeta$ is as given in \eqref{Eq:virtual_sys_sto}, \ie{}, $\zeta = (A-BR^{-1}B^{\top}\mathcal{M})(q_x-x_d)+\dot{x}_d$, where the SDC matrix $A$ is defined as (see Lemma~\ref{sdclemma}) 
\begin{align}
A(x-x_d)=f(x,\theta_n)+B(x)u_d-f(x_d,\theta_n)-B(x_d)u_d.
\end{align}
The arguments of $A(\varrho,x,x_d,u_d)$, $B(x)$, $\mathcal{M}(x,x_d,u_d)$, and $\varphi(x,x_d)$ are omitted for notational simplicity. Since $\dot{q}_x=\zeta(q_x,x,x_d,u_d,t)$ is contracting due to Theorems~\ref{THM:Thm:CV-STEM} and~\ref{THM:Thm:NCMstability2} with a contraction rate given by $\alpha_{\ell}$ in Theorem~\ref{THM:Thm:NCMstability2}, we have for a Lyapunov function $V = \delta q_x^{\top}M\delta q_x+\delta q_{\vartheta}^{\top}\Gamma^{-1}\delta q_{\vartheta}$ that
\begin{align}
\frac{\dot{V}}{2} &\leq -\alpha_{\ell}\delta q_x^{\top}M\delta q_x+\delta q_x^{\top}(M-\mathcal{M})B\varphi^{\top}\delta q_{\vartheta} -\sigma\|\delta q_{\vartheta}\|^2 +\delta q_{\vartheta}^{\top}\delta d_{q\vartheta}. \label{ncm_adaptive_V_computation}
\end{align}
Applying \eqref{condition_mNCM} with $\|\mathcal{M}-M\|\leq\epsilon_{\ell}$ of \eqref{Eq:Merror_control}, we get
\begin{align}
\dot{V}/2-\delta q_{\vartheta}^{\top}\delta d_{q\vartheta} &\leq -(\alpha_{\ell}\underline{m})\|\delta q_x\|^2+\bar{\phi}\bar{b}\epsilon_{\ell}\|\delta q_x\|\|\delta q_{\vartheta}\| -\sigma\|\delta q_{\vartheta}\|^2 \\
&\leq -\alpha_{a}\left(\overline{m}\|\delta q_x\|^2+\frac{\|\delta q_{\vartheta}\|^2}{\underline{\gamma}}\right) \leq -\alpha_{a}V.
\end{align}
Since we have $\|\partial d_{q\vartheta}/\partial \mu\| = \sigma\bar{\vartheta}$, this implies $\dot{V}_{\ell} \leq -\alpha_a V_{\ell}+\sigma\sqrt{\overline{\gamma}}\bar{\vartheta}$, yielding the bound \eqref{adaptive_bound_1} due to Lemma~\ref{Lemma:comparison}~\cite{ancm}. Robustness against deterministic and stochastic disturbances follows from Theorem~\ref{THM:Thm:CVSTEM:LMI} if $\sigma\neq0$. If $\epsilon_{\ell}=0$ and $\sigma = 0$, the relation \eqref{ncm_adaptive_V_computation} reduces to $\dot{V}/2 \leq -\alpha\delta q_x^{\top}M\delta q_x$, which results in asymptotic stability of $x$ to $x_d$ in \eqref{adaptive_general} due to Barbalat's lemma~\cite[p. 323]{Khalil:1173048} as in the proof of Theorem~2 in~\cite{9109296}. \qed
\end{proof}
\begin{remark}
\label{remark_sliding}
Although Theorem~\ref{THM:Thm:adaptive_robust_affine} is for the case where $f(x)$ is affine in its parameter, it is also useful for the following types of systems with an uncertain parameter $\theta\in\mathbb{R}^{c}$ and a control input $u$ (see~\cite{ancm}):
\begin{align}
\label{lagrangian_dynamics}
H(x){p}^{(n)}+h(x)+\Pi (x)\theta=u
\end{align}
where ${p}\in \mathbb{R}^n$, $u\in \mathbb{R}^n$, $h:\mathbb{R}^{n}\rightarrow\mathbb{R}^{n}$, $H:\mathbb{R}^{n}\rightarrow\mathbb{R}^{n\times n}$, $\Pi:\mathbb{R}^{n}\rightarrow\mathbb{R}^{n\times {c}}$, $x=[({p}^{(n-2)})^{\top},\cdots,({p})^{\top}]^{\top}$, and ${p}^{(k)}$ denotes the $k$th time derivative of ${p}$. In particular, adaptive sliding control~\cite{adaptive_sliding} designs $u$ to render the system of the composite variable $s$ given as $s = {p}^{(n-1)}-{p}_r^{(n-1)}$ to be contracting, where ${p}_r^{(n-1)} = p_d^{(n-1)}-\sum_{i=0}^{n-2}\lambda_{i}\mathtt{e}^{(i)}$, $\mathtt{e}=p-p_d$, $p_d$ is a target trajectory, and $\kappa^{n-1}+\lambda_{n-2}\kappa^{n-2}+\cdots+\lambda_0$ is a stable (Hurwitz) polynomial in the Laplace variable $\kappa$ (see Example~\ref{EX:ex:lag_metric}). Since we have $\mathtt{e}^{(n-1)}=s-\sum_{i=0}^{n-2}\lambda_{i}\mathtt{e}^{(i)}$ and the system for $[\mathtt{e}^{(0)},\cdots,\mathtt{e}^{(n-2)}]$ is also contracting if $s=0$ due to the Hurwitz property, the hierarchical combination property~\cite{Ref:contraction2,Ref:contraction5} of contraction in Theorem~\ref{THM:Thm:Robust_contraction_hierc} guarantees $\lim_{t\to\infty}\|p-p_d\| = 0$~\cite[p. 352]{Ref_Slotine} (see Example~\ref{EX:ex:hiera}).
\end{remark}
\begin{example}{Adaptive Control of Lagrangian Systems}{ex:lag_adaptive0}
Using the technique of Remark~\ref{remark_sliding}, we can construct adaptive control for the Lagrangian system in Example~\ref{EX:ex:lag_metric} as follows:
\begin{align}
\label{lagrange_partial_adaptive}
&\mathcal{H}(\mathtt{q})\ddot{\mathtt{q}}+\mathcal{C}(\mathtt{q},\dot{\mathtt{q}})\dot{\mathtt{q}}+\mathcal{G}(\mathtt{q})=u(\mathtt{q},\dot{\mathtt{q}},t) \\
\label{lagrange_u_adaptive}
&u(\mathtt{q},\dot{\mathtt{q}},t) = -\mathcal{K}(t)(\dot{\mathtt{q}}-\dot{\mathtt{q}}_r)+\hat{\mathcal{H}}(\mathtt{q})\ddot{\mathtt{q}}_r+\hat{\mathcal{C}}(\mathtt{q},\dot{\mathtt{q}})\dot{\mathtt{q}}_r+\hat{\mathcal{G}}(\mathtt{q})~~~~~
\end{align}
where $\hat{\mathcal{H}}$, $\hat{\mathcal{C}}$, and $\hat{\mathcal{G}}$ are the estimates of $\mathcal{H}$, $\mathcal{C}$, and $\mathcal{G}$, respectively, and the other variables are as defined in Example~\ref{EX:ex:lag_metric}. Suppose that the terms $\mathcal{H}$, $\mathcal{C}$, and $\mathcal{G}$ depend linearly on the unknown parameter vector $\theta$ as follows~\cite[p. 405]{Ref_Slotine}:
\begin{align}
\label{lagrange_linear_assump}
\mathcal{H}(\mathtt{q})\ddot{\mathtt{q}}_r+\mathcal{C}(\mathtt{q},\dot{\mathtt{q}})\dot{\mathtt{q}}_r+\mathcal{G}(\mathtt{q})=Y(\mathtt{q},\dot{\mathtt{q}},\dot{\mathtt{q}}_r,\ddot{\mathtt{q}}_r)\theta.
\end{align}
Updating the parameter estimate $\hat{\theta}$ using the adaptation law, $\dot{\hat{\theta}} = -\Gamma Y(\mathtt{q},\dot{\mathtt{q}},\dot{\mathtt{q}}_r,\ddot{\mathtt{q}}_r)^{\top}(\dot{\mathtt{q}}-\dot{\mathtt{q}}_r)$, as in \eqref{affine_adaptation_ncm} where $\Gamma \succ 0$, we can define the following virtual system which has $q = \xi_0 = [\dot{\mathtt{q}}_r^{\top},{\theta}^{\top}]^{\top}$ and $q = \xi_1 = [\dot{\mathtt{q}}^{\top},\hat{\theta}^{\top}]^{\top}$ as its particular solutions:
\begin{align}
\label{lagrange_cl_adaptive}
\begin{bmatrix}\mathcal{H} & 0 \\ 0 & \Gamma^{-1}\end{bmatrix}(\dot{q}-\dot{\xi}_0)+\begin{bmatrix}\mathcal{C}+\mathcal{K} & -Y \\ Y^{\top} & 0\end{bmatrix}(q-\xi_0) =0
\end{align}
where the relation $u=-\mathcal{K}(t)(\dot{\mathtt{q}}-\dot{\mathtt{q}}_r)+Y(\mathtt{q},\dot{\mathtt{q}},\dot{\mathtt{q}}_r,\ddot{\mathtt{q}}_r)\hat{\theta}$ is used. Thus, for a Lyapunov function $V=\delta q^{\top}\left[\begin{smallmatrix}\mathcal{H} & 0\\ 0 & \Gamma^{-1}\end{smallmatrix}\right]\delta q$, we have that
\begin{align}
\dot{V} &= \delta q^{\top}\begin{bmatrix}\mathcal{K} & Y \\ -Y^{\top} & 0\end{bmatrix}\delta q= \delta q^{\top}\begin{bmatrix}\mathcal{K} & 0 \\ 0 & 0\end{bmatrix}\delta q
\end{align}
which results in asymptotic stability of the differential state $\delta q$ (\ie{}, semi-contraction~\cite{Ref:contraction3}, see also Barbalat's lemma~\cite[p. 405-406]{Ref_Slotine}). 
\end{example}
\subsection{Parameter-Dependent Contraction Metric (aNCM)}
\label{Sec:adaptiveNCM}
Although Theorem~\ref{THM:Thm:adaptive_robust_affine} utilizes the NCM designed for the nominal system \eqref{adaptive_general} with $\theta = \theta_n$, we could improve its representational power by explicitly taking the parameter estimate $\hat{\theta}$ as one of the NCM arguments~\cite{ancm}, leading to the concept of an adaptive NCM (aNCM).

In this section, we consider multiplicatively-separable nonlinear systems given in Assumption~\ref{multiplicative_asmp}, which holds for many types of systems including robotics systems~\cite{Ref_Slotine}, spacecraft high-fidelity dynamics~\cite{battin,doi:10.2514/1.55705}, and systems modeled by basis function approximation and DNNs~\cite{Nelles2001,SannerSlotine1992}.
\begin{assumption}
\label{multiplicative_asmp}
The dynamical system \eqref{adaptive_general} is multiplicatively separable in terms of $x$ and $\theta$, \ie{}, $\exists$ $Y_f:\mathbb{R}^n\mapsto\mathbb{R}^{n\times {c}_z}$, $Y_{b_i}:\mathbb{R}^n\mapsto\mathbb{R}^{n\times {c}_z}$, and $Z:\mathbb{R}^{c}\mapsto\mathbb{R}^{{c}_z}$ \st{} 
\begin{align}
\label{linear_adaptive}
\begin{aligned}
    Y_f(x)Z(\theta) &= f(x,\theta) \\
    Y_{b_i}(x)Z(\theta) &= b_i(x,\theta)
\end{aligned}
\end{align}
where $B(x,\theta)=[b_1(x,\theta),\cdots,b_m(x,\theta)]$. We could define $\theta$ as $[\theta^{\top},Z(\theta)^{\top}]^{\top}$ so we have $Y_f(q)\theta = f(q,\theta)$ and $Y_{b_i}(q)\theta = b_i(q;\theta)$. Such an over-parameterized system could be regularized using the Bregman divergence as in~\cite{boffi2020implicit} (see Example~\ref{EX:ex:bregman}), and thus we denote $[\theta^{\top},Z(\theta)^{\top}]^{\top}$ as $\theta$ in the subsequent discussion.
\end{assumption}

\begin{figure}[htbp]
    \centering
    \includegraphics[width=132mm]{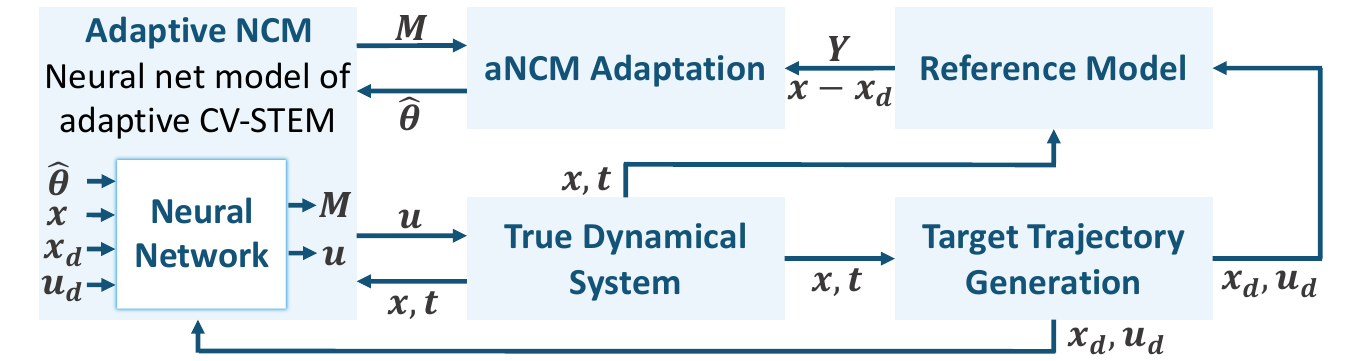}
    \caption{Illustration of aNCM ($M$: positive definite matrix that defines aNCM; $\hat{\theta}$: estimated parameter; $Y$: error signal, see \eqref{definition_Y}; $x$: system state; $u$: system control input; and $(x_d,u_d)$: target state and control input trajectory).}
    \label{ancmdrawing}
\end{figure}

Under Assumption~\ref{multiplicative_asmp}, the dynamics for $\mathtt{e}=x-x_d$ of \eqref{adaptive_general} can be expressed as follows: 
\begin{equation}
\label{error_dynamics}
\dot{\mathtt{e}} = A(\varrho,x,x_d,u_d,\hat{\theta})\mathtt{e}+B(x,\hat{\theta})(u-u_d)-\tilde{Y}(\hat{\theta}-\theta)
\end{equation}
where $A$ is the SDC matrix of Lemma~\ref{sdclemma}, $\hat{\theta}$ is the current estimate of $\theta$, and $\tilde{Y}$ is defined as
\begin{equation}
\label{definition_Y}
\tilde{Y} = Y-Y_d= (Y_f(x)+Y_b(x,u))-(Y_f(x_d)+Y_b(x_d,u_d))
\end{equation}
where $Y_b(x,u)=\sum_{i=1}^mY_{b_i}(q)u_i$.
\begin{definition}{Adaptive Neural Contraction Metrics}{Def:aNCM}
The adaptive Neural Contraction Metric (aNCM) in Fig.~\ref{ancmdrawing} is a DNN model for the optimal parameter-dependent contraction metric, given by solving the adaptive CV-STEM, \ie{}, \eqref{cvstem_eq} of Theorem~\ref{THM:Thm:CV-STEM} (or Theorem~\ref{THM:Thm:ccm_cvstem} for differential feedback) with its contraction constraint replaced by the following convex constraint:
\begin{align}
\label{deterministic_contraction_adaptive_tilde}
&-\Xi+2\sym{\left(A\bar{W}\right)}-2\nu BR^{-1}B^{\top} \preceq -2\alpha \bar{W}
\end{align}
for deterministic systems, and
\begin{equation}
\label{stochastic_contraction_adaptive_tilde}
\begin{bmatrix}
-\Xi+2\sym{}(A\bar{W})-2\nu BR^{-1}B^{\top}+2\alpha \bar{W}& \bar{W} \\
\bar{W} & -\frac{\nu}{\alpha_s}\mathrm{I} \end{bmatrix} \preceq 0
\end{equation}
for stochastic systems, where $W=M(x,x_d,u_d,\hat{\theta})^{-1}\succ0$ (or $W=M(x,\hat{\theta})^{-1}\succ0$, see Theorem~\ref{THM:Thm:fixed_sdc}), $\bar{W} = \nu W$, $\nu = \overline{m}$, $R=R(x,x_d,u_d)\succ0$ is a weight matrix on $u$, $\Xi=({d}/{dt})|_{\hat{\theta}}{\bar{W}}$ is the time derivative of $\bar{W}$ computed along \eqref{adaptive_general} with $\theta = \hat{\theta}$, $A$ and $B$ are given in \eqref{error_dynamics}, $\alpha$, $\underline{m}$, $\overline{m}$, and $\alpha_s$ are as given in Theorem~\ref{THM:Thm:CV-STEM}, and the arguments $(x,x_d,u_d,\hat{\theta})$ are omitted for notational simplicity.
\end{definition}

The aNCM given in Definition~\ref{DEF:Def:aNCM} has the following stability property along with its optimality due to the CV-STEM of Theorem~\ref{THM:Thm:CV-STEM}~\cite{ancm}.
\begin{theorem}{Learning-based Adaptive Control with aNCMs}{Thm:adaptiveNCMthm}
Suppose that Assumption~\ref{multiplicative_asmp} holds and let $\mathcal{M}$ define the aNCM of Definition~\ref{DEF:Def:aNCM}. Suppose also that the true dynamics \eqref{adaptive_general} is controlled by the following adaptive control law:
\begin{align}
\label{adaptive_general_u}
u &= u_d-R(x,x_d,u_d)^{-1}B(x,\hat{\theta})^{\top}\mathcal{M}(x,x_d,u_d,\hat{\theta})\mathtt{e}\\
\label{adaptation_general}
\dot{\hat{\theta}} &= \Gamma((Y^{\top}d\mathcal{M}_x+Y_d^{\top}d\mathcal{M}_{x_d}+\tilde{Y}^{\top}\mathcal{M}) \mathtt{e}-\sigma\hat{\theta})
\end{align}
where $d\mathcal{M}_q = [({\partial \mathcal{M}}/{\partial q_1})\mathtt{e},\cdots,({\partial \mathcal{M}}/{\partial q_n})\mathtt{e}]^{\top}/2$, $\mathtt{e}=x-x_d$, $\Gamma \succ 0$, $\sigma \in \mathbb{R}_{\geq 0}$, and $Y$, $Y_d$, $\tilde{Y}$ are given in \eqref{definition_Y}. Suppose further that the learning error in $\|\mathcal{M}-M\|\leq\epsilon_{\ell}$ of Theorem~\ref{THM:Thm:NCMstability2} additionally satisfies $\|d\mathcal{M}_{x_d} -dM_{x_d} \| \leq \epsilon_{\ell}$ and $\|d\mathcal{M}_x -dM_x \| \leq \epsilon_{\ell}$ in some compact set $\mathcal{S}$.

If $\exists \bar{b},\bar{\rho},\bar{y} \in \mathbb{R}_{>0}$ \st{} $\|B(x,\hat{\theta})\| \leq \bar{b}$, $\|R^{-1}(x,x_d,u_d)\| \leq \bar{\rho}$, $\|Y\|\leq \bar{y}$, $\|Y_d\|\leq \bar{y}$, and $\|\tilde{Y}\|\leq \bar{y}$ in \eqref{adaptive_general_u} and \eqref{adaptation_general}, and if $\Gamma$ and $\sigma$ of \eqref{adaptation_general} are selected to satisfy the following as in Theorem~\ref{THM:Thm:adaptive_robust_affine}:
\begin{align}
\begin{bmatrix}
\label{condition_aNCM}
-2\alpha_{\ell}\underline{m} & \bar{y}\epsilon_{\ell} \\
\bar{y}\epsilon_{\ell} & -2\sigma
\end{bmatrix} \preceq -2 \alpha_{a}\begin{bmatrix}
\overline{m} & 0 \\
0 & 1/\underline{\gamma}
\end{bmatrix}
\end{align}
for $\exists \alpha_a\in\mathbb{R}_{>0}$, $\alpha_{\ell}$ given in Theorem~\ref{THM:Thm:NCMstability2}, and $\underline{m}$ and $\overline{m}$ given in $\underline{m}\mathrm{I}\preceq M \preceq \overline{m}\mathrm{I}$ of \eqref{Mcon}, then the system \eqref{adaptive_general} is robust against bounded deterministic and stochastic disturbances, and we have the exponential bound \eqref{adaptive_bound_1} in the compact set $\mathcal{S}$. Furthermore, if $\epsilon_{\ell}=0$ (adaptive CV-STEM control), \eqref{adaptation_general} with $\sigma = 0$ guarantees asymptotic stability of $x$ to $x_d$ in \eqref{adaptive_general}.
\end{theorem}
\begin{proof}
Replacing the contraction constraints of the CV-STEM in Theorem~\ref{THM:Thm:CV-STEM} by \eqref{deterministic_contraction_adaptive_tilde} and \eqref{stochastic_contraction_adaptive_tilde}, the bound \eqref{adaptive_bound_1} and the asymptotic stability result can be derived as in the proof of Theorem~\ref{THM:Thm:adaptive_robust_affine} (see Theorem~4 and Corollary~2 of~\cite{ancm} for details). Theorems~\ref{THM:Thm:Robust_contraction_original} and \ref{THM:Thm:robuststochastic} guarantee robustness of \eqref{adaptive_general} against bounded deterministic and stochastic disturbances for $\sigma\neq0$. \qed
\end{proof}

Let us again emphasize that, by using Theorem~\ref{THM:Thm:ccm_cvstem}, the results of Theorems~\ref{THM:Thm:adaptive_robust_affine} and~\ref{THM:Thm:adaptiveNCMthm} can be extended to adaptive control with CCM-based differential feedback~\cite{9109296,lopez2021universal} (see~Table~\ref{tab:sdcccm_summary} for the trade-offs).

Since the adaptation laws \eqref{affine_adaptation_ncm} and \eqref{adaptation_general} in Theorems~\ref{THM:Thm:adaptive_robust_affine} and~\ref{THM:Thm:adaptiveNCMthm} yield an explicit bound on the steady-state error as in \eqref{adaptive_bound_1}, it could be used as the objective function of the CV-STEM in Theorem~\ref{THM:Thm:CV-STEM}, regarding $\Gamma$ and $\sigma$ as extra decision variables to get $M$ optimal in a sense different from Theorem~\ref{THM:Thm:CV-STEM}. Smaller $\epsilon_{\ell}$ would lead to a weaker condition on them in \eqref{condition_mNCM} and \eqref{condition_aNCM}. Also, the size of $\|\vartheta\| \leq \bar{\vartheta}$ in \eqref{adaptive_bound_1} can be adjusted simply by rescaling it (\eg{}, $\vartheta\to\theta/\bar{\vartheta}$). However, such a robustness guarantee comes with a drawback of having $\lim_{t\to\infty}\|\hat{\theta}(t)\| = 0$ for $\sigma \neq 0$ in \eqref{affine_adaptation_ncm}, leading to the trade-offs in different types of adaptation laws, some of which are given in Examples~\ref{EX:ex:projection_operator}--\ref{EX:ex:bregman} (see also Remark~\ref{remark_PE_condition}).
\begin{example}{Projection Operators}{ex:projection_operator}
Let us briefly describe the following projection operator-based adaptation law, again for the Lagrangian system~\eqref{lagrange_partial_adaptive} of Example~\ref{EX:ex:lag_adaptive0} with the unknown parameter vector $\theta$:
\begin{align}
\label{projection_adaptation}
\dot{\hat{\theta}} = \proj{}(\hat{\theta},-\Gamma Y(\mathtt{q},\dot{\mathtt{q}},\dot{\mathtt{q}}_r,\ddot{\mathtt{q}}_r)^{\top}(\dot{\mathtt{q}}-\dot{\mathtt{q}}_r),p)
\end{align}
where $\proj{}$ is the projection operator and $p$ is a convex boundary function (\eg{}, $p(\hat{\theta})=(\hat{\theta}^{\top}\hat{\theta}-\theta_{\max}^2)/(\epsilon_{\theta}\theta_{\max}^2)$ for given positive constants $\theta_{\max}$ and $\epsilon_{\theta}$). If $p(\hat{\theta})>0$ and $\nabla p(\hat{\theta})^{\top}\xi > 0$,
\begin{align}
\label{projection_operator}
\proj{}(\hat{\theta},\xi,p)=
\begin{cases}
    \xi-\frac{\nabla p(\hat{\theta})\nabla p(\hat{\theta})^{\top}}{\|\nabla p(\hat{\theta})\|^2}\xi p(\hat{\theta}) & \text{if $p(\hat{\theta})>0$ and $\nabla p(\hat{\theta})^{\top}\xi > 0$} \\
    \xi & \text{otherwise}
\end{cases}.
\end{align}
The projection operator has the following useful property which allows bounding the parameter estimate $\hat{\theta}$.
\begin{lemma}
\label{lemma_projection}
If $\hat{\theta}$ is given by \eqref{projection_adaptation} with $\hat{\theta}(0) \in \Omega_{\theta} = \{\theta\in\mathbb{R}^c|p(\theta) \leq 1\}$ for a convex function $p(\theta)$, then $\hat{\theta}(t) \in \Omega_{\theta},~\forall t\geq 0$ (\eg{}, $\|\hat{\theta}\| \leq \theta_{\max}\sqrt{1+\epsilon_{\theta}}$ if $p(\hat{\theta})=(\hat{\theta}^{\top}\hat{\theta}-\theta_{\max}^2)/(\epsilon_{\theta}\theta_{\max}^2)$).  
\end{lemma}
\begin{proof}
See~\cite{lavretsky2012projection}. \qed
\end{proof}
Since Lemma~\ref{lemma_projection} guarantees the boundedness of $\|\hat{\theta}\|$, the adaptive controller \eqref{lagrange_u_adaptive}, $u=-\mathcal{K}(\dot{\mathtt{q}}-\dot{\mathtt{q}}_r)+Y\theta+Y\tilde{\theta}$, can be viewed as the exponentially stabilizing controller $-\mathcal{K}(\dot{\mathtt{q}}-\dot{\mathtt{q}}_r)+Y\theta$ (see Example~\ref{EX:ex:lag_metric}) plus a bounded external disturbance $Y\tilde{\theta}$, implying robustness due to Theorem~\ref{THM:Thm:Robust_contraction_original}~\cite{1576865,lavretsky2012projection,Ref:phasesync,lopez2021universal}.

Let us also remark that, as in Example~\ref{EX:ex:lag_adaptive0}, the projection operator-based adaptation law \eqref{projection_adaptation} still achieves asymptotic stabilization. Applying the control law \eqref{lagrange_u_adaptive} with the adaptation \eqref{projection_adaptation} yields the following virtual system which has $q = \xi_0 = [\dot{\mathtt{q}}_r^{\top},{\theta}^{\top}]^{\top}$ and $q = \xi_1 = [\dot{\mathtt{q}}^{\top},\hat{\theta}^{\top}]^{\top}$ as its particular solutions:
\begin{align}
\label{lagrange_cl_adaptive_proj}
\begin{bmatrix}\mathcal{H} & 0 \\ 0 & \mathrm{I}\end{bmatrix}(\dot{q}-\dot{\xi}_0)+\begin{bmatrix}(\mathcal{C}+\mathcal{K})(q_s-\dot{\mathtt{q}}_r)-Y(q_{\theta}-\theta) \\ \proj{}(\hat{\theta},\Gamma Y^{\top}(q_s-\dot{\mathtt{q}}_r),p)\end{bmatrix} =0
\end{align}
where $q = [q_s^{\top},q_{\theta}^{\top}]^{\top}$. Thus, for a Lyapunov function $V_{s\ell}=\frac{1}{2}\int_{\xi_0}^{\xi_1}\delta q^{\top}\left[\begin{smallmatrix}\mathcal{H}(\mathtt{q}) & 0\\ 0 & \Gamma^{-1}\end{smallmatrix}\right]\delta q$, we have that
\begin{align}
\dot{V}_{s\ell} &= \int_{\xi_0}^{\xi_1}-\delta q_s^{\top}\mathcal{K}\delta q_s+\delta q_{\theta}^{\top}(\proj{}(\hat{\theta},Y^{\top}\delta q_s,p)-Y^{\top}\delta q_s). \nonumber
\end{align}
Using the convex property of the projection operator, \ie{}, $\tilde{\theta}^{\top}(\proj{}(\hat{\theta},\xi,p)-\xi) \leq 0$~\cite{1576865,lavretsky2012projection}, this gives $\dot{V}_{s\ell} \leq -\int_{\xi_0}^{\xi_1}\delta q_s^{\top}\mathcal{K}\delta q_s$, which results in asymptotic stability of $\delta q$ due to Barbalat's lemma~\cite[p. 405-406]{Ref_Slotine} as in Example~\ref{EX:ex:lag_adaptive0}.
\end{example}
\begin{example}{Adaptation Rate Scaling}{ex:adaptation_rate_scale}
The dependence on $u$ and $\dot{\hat{\theta}}$ in $(d/dt)|_{\hat{\theta}}M$ can be removed by using $\partial_{b_i(x)}M+\partial_{b_i(x_d)}M=0$ as in Theorem~\ref{THM:Thm:adaptive_robust_affine}, and by using adaptation rate scaling introduced in~\cite{lopez2021universal}. In essence, the latter multiplies the adaptation law \eqref{adaptation_general} by any strictly-increasing and strictly-positive scalar function $v(2\rho)$, and update $\rho$ as
\begin{align}
\label{rho_adaptation}
\dot{\rho} = \frac{1}{2}\frac{v(2\rho)}{v_{\rho}(2\rho)}\sum_{i=1}^{c}\frac{1}{V_{\mathtt{e}}+\eta}\frac{\partial V_{\mathtt{e}}}{\partial \hat{\theta}_i}\dot{\theta_i}
\end{align}
where $v_{\rho}=\partial v/\partial \rho$, $\eta \in\mathbb{R}_{>0}$, and $V_{\mathtt{e}} = \mathtt{e}^{\top}M(x,x_d,u_d,\hat{\theta})\mathtt{e}$ for $\mathtt{e}=x-x_d$ and $M$ given in Definition~\ref{DEF:Def:aNCM}, so the additional term due to \eqref{rho_adaptation} cancels out the term involving $\dot{\hat{\theta}}$ in $(d/dt)|_{\hat{\theta}}\bar{W}$ of \eqref{deterministic_contraction_adaptive_tilde} (see~\cite{lopez2021universal} for details). Its robustness property follows from Theorem~\ref{THM:Thm:adaptiveNCMthm} also in this case.
\end{example}
\begin{example}{Bregman Divergence}{ex:bregman}
Using the Bregman divergence-based adaptation in~\cite{boffi2020implicit}, we could implicitly regularize the parameter estimate $\hat{\theta}$ as follows:
\begin{align}
\label{implicit_reg}
\lim_{t\to\infty}\hat{\theta}=\mathrm{arg}\min_{\vartheta\in \mathcal{A}}d_{\psi}(\vartheta\|\theta^*)=\mathrm{arg}\min_{\vartheta\in \mathcal{A}}d_{\psi}\left(\vartheta\left\|\hat{\theta}(0)\right.\right)
\end{align}
where $d_{\psi}$ is the Bregman divergence defined as $d_{\psi}(x\|y)=\psi(x)-\psi(y)-(x-y)^{\top}\nabla\psi(y)$ for a convex function $\psi$, and $\mathcal{A}$ is a set containing only parameters that interpolate the dynamics along the entire trajectory. If $\hat{\theta}(0)=\mathrm{arg}\min_{b\in \mathbb{R}^p}\psi(b)$, we have $\lim_{t\to\infty}\hat{\theta}=\mathrm{arg}\min_{\vartheta\in \mathcal{A}}\psi(\vartheta)$, which regularizes $\hat{\theta}$ to converge to a parameter that minimizes $\psi$. For example, using $1$-norm for $\psi$ would impose sparsity on the steady-state parameter estimate $\hat{\theta}$~\cite{boffi2020implicit}.
\end{example}

These extensions of adaptive control techniques described in Examples~\ref{EX:ex:projection_operator}--\ref{EX:ex:bregman} could be utilized with contraction theory and learning-based control as in Theorems~\ref{THM:Thm:adaptive_robust_affine} and~\ref{THM:Thm:adaptiveNCMthm}.
\begin{remark}
\label{remark_PE_condition}
Note that the results presented earlier in this section do not necessarily mean parameter convergence, $\lim_{t\to\infty}\|\tilde{\theta}\|=0$, as the adaptation objective is to drive the tracking error $\|x-x_d\|$ to zero~\cite[p. 331]{Ref_Slotine}, not to find out the true parameter $\theta$ out of the many that achieve
perfect tracking.

Asymptotic parameter convergence, $\lim_{t\to\infty}\|\tilde{\theta}\|=0$ for $\theta$ of \eqref{adaptive_general} under Assumption~\ref{multiplicative_asmp}, could be guaranteed if there is no disturbance and learning error with $\sigma=0$, and we have $\exists T,\alpha_{PE}\in\mathbb{R}_{>0}\text{ \st{} }\int_{t}^{t+T}\tilde{Y}^{\top}\tilde{Y}d\tau \succeq \alpha_{PE}\mathrm{I},~\forall{t}$ for $\tilde{Y}$ given in \eqref{definition_Y} (the persistent excitation condition~\cite[p. 366]{Ref_Slotine}). We could also utilize the Bregman divergence-based adaptation to regularize the behavior of $\lim_{t\to\infty}\|\hat{\theta}\|$ as in Example~\ref{EX:ex:bregman}.
\end{remark}
\newpage
\section{Contraction Theory for Learned Models}
\label{Sec:datadriven}
In recent applications of learning-based and data-driven automatic control frameworks, we often encounter situations where we only have access to a large amount of system trajectory data. This section, therefore, considers the cases where the true underlying dynamical system is poorly modeled or completely unknown, and assumptions in Sec.~\ref{Sec:adaptive} for using learning-based adaptive control techniques are no longer valid. 

A typical approach is to perform system identification~\cite{neurallander,47710,cdc_systemid,7992901} using trajectory data generated by \eqref{sysIDxd}:
\begin{align}
\label{sysIDx}
\dot{x} &= f_L(x,u(x,t),t)  \\
\label{sysIDxd}
\dot{x}^* &= f_\mathrm{true}(x^*,u(x^*,t),t)
\end{align}
where ${x}:\mathbb{R}_{\geq 0}\rightarrow\mathbb{R}^n$ is the system state, $u:\mathbb{R}^n\times\mathbb{R}_{\geq 0}\rightarrow\mathbb{R}^m$ is the system control input, $f_\mathrm{true}: \mathbb{R}^n\times\mathbb{R}^m\times\mathbb{R}_{\geq 0}\rightarrow\mathbb{R}^n$ is a smooth function of the true dynamical system \eqref{sysIDxd}, which is unknown and thus modeled by a learned smooth function $f_L: \mathbb{R}^n\times\mathbb{R}^m\times\mathbb{R}_{\geq 0}\rightarrow\mathbb{R}^n$ of \eqref{sysIDx}. If we can learn $f_L$ to render the system \eqref{sysIDx} contracting, contraction theory still allows us to ensure robustness and stability of these systems.
\begin{theorem}{Contraction Theory for Learned Models}{Thm:NCMstability_modelfree}
Let $q(0,t)=\xi_0(t)=x(t)$, $q(1,t)=\xi_1(t)=x^*(t)$, and $\textsl{g}=f_L$ in Theorems~\ref{THM:Thm:contraction_learning} and~\ref{THM:Thm:contraction_learning_sto} as in Example~\ref{EX:ex:learning_prob3}, and define $\Delta_L$ as
\begin{align}
\label{Eq:DeltaL_sysID}
\Delta_L = f_\mathrm{true}(x^*,u(x^*,t),t)-f_L(x^*,u(x^*,t),t)
\end{align}
for the learning error condition $\|\Delta_L\|\leq\epsilon_{\ell0}+\epsilon_{\ell1}\|\xi_1-\xi_0\|$ in \eqref{Eq:learning_error}. If the function $f_L$ is learned to satisfy \eqref{Eq:learning_error} with $\epsilon_{\ell1}=0$, \ie{}, $\|\Delta_L\| \leq \epsilon_{\ell0}$ in some compact set $\mathcal{S}$, and if there exists a contraction metric defined by $M$ bounded as $\underline{m}\mathrm{I}\preceq M \preceq \overline{m}\mathrm{I}$ as in \eqref{Mcon}, which renders \eqref{sysIDx} contracting as in \eqref{eq_MdotContracting} of Theorem~\ref{THM:Thm:contraction} for deterministic systems, \ie{},
\begin{align}
\label{contracting_system_id}
\dot{M}+M\frac{\partial f_L}{\partial x}+\frac{\partial f_L}{\partial x}^{\top}M\preceq-2\alpha M,    
\end{align}
and \eqref{eq_MdotContracting_sto} of Theorem~\ref{THM:Thm:robuststochastic} for stochastic systems, \ie{},
\begin{align}
\label{contracting_system_id_sto}
\dot{M}+M\frac{\partial f_L}{\partial x}+\frac{\partial f_L}{\partial x}^{\top}M\preceq-2\alpha M-\alpha_s\mathrm{I},
\end{align}
then we have the following in the compact set $\mathcal{S}$:
\begin{align}
\label{system_id_bound}
\|x(t)-x^*(t)\| \leq \frac{V_{\ell}(0)}{\sqrt{\underline{m}}}e^{-\alpha t}+\frac{\epsilon_{\ell0}}{\alpha}\sqrt{\frac{\overline{m}}{\underline{m}}}(1-e^{-\alpha t})
\end{align}
for $x$ and $x^*$ in \eqref{sysIDx} and \eqref{sysIDxd}, where $V_{\ell}=\int^{x}_{x^*}\|\Theta\delta x\|$ as in Theorem~\ref{THM:Thm:path_integral} with $M=\Theta^{\top}\Theta$.
Furthermore, the systems \eqref{sysIDx} and \eqref{sysIDxd} are robust against bounded deterministic and stochastic disturbances.
\end{theorem}
\begin{proof}
Let $p_t^* = (x^*,u(x^*(t),t),t)$ for notational simplicity. Since \eqref{sysIDxd} can be written as $\dot{x}^* = f_\mathrm{true}(p_t^*) = f_L(p_t^*)+(f_\mathrm{true}(p_t^*)-f_L(p_t^*))$ (see Example~\ref{EX:ex:learning_prob3}) and $\|\Delta_L\|=\|f_\mathrm{true}(p_t^*)-f_L(p_t^*)\| \leq \epsilon_{\ell0}$, Theorem~\ref{THM:Thm:Robust_contraction_original} holds with $\bar{d}=\epsilon_{\ell0}$ as \eqref{sysIDx} is contracting, resulting in \eqref{system_id_bound}. Also, defining $\Delta_L$ as \eqref{Eq:DeltaL_sysID} in Theorems~\ref{THM:Thm:contraction_learning} and \ref{THM:Thm:contraction_learning_sto} results in robustness of \eqref{sysIDx} and \eqref{sysIDxd} against bounded deterministic and stochastic disturbances due to \eqref{contracting_system_id} and \eqref{contracting_system_id_sto}, respectively. \qed
\end{proof}

Theorem~\ref{THM:Thm:NCMstability_modelfree} is the theoretical foundation for stability analysis of model-free nonlinear dynamical systems. The bound \eqref{system_id_bound} becomes tighter as we achieve smaller $\epsilon_{\ell0}$ using more training data for verifying  $\|\Delta_L\| \leq \epsilon_{\ell0}$ (see Remark~\ref{remark_learning_error}). From here onwards, we utilize contraction theory to provide stability guarantees to such model-free systems, partially enabling the use of the aforementioned model-based techniques.
\subsection{Robust Control of Systems Modeled by DNNs}
\label{Sec:neurallander}
One challenge in applying Theorem~\ref{THM:Thm:NCMstability_modelfree} in practice is to find contraction metrics for the control non-affine nonlinear systems \eqref{sysIDx}. This section delineates one way to construct a contraction metric in Theorem~\ref{THM:Thm:NCMstability_modelfree} for provably-stable feedback control, using the CV-STEM and NCM of Theorems~\ref{THM:Thm:CV-STEM},~\ref{THM:Thm:ccm_cvstem},~\ref{THM:Thm:NCMstability1}, \ref{THM:Thm:NCMstability2}--\ref{THM:Thm:ncm_ccm_clf}, and~\ref{THM:Thm:lagros_stability}, along with the spectrally-normalized DNN of Definition~\ref{DEF:Def:SN}.

To this end, let us assume that $f_\mathrm{true}$ of the dynamics \eqref{sysIDxd} can be decomposed into a known control-affine part $f(x^*,t)+B(x^*,t)u$ and an unknown control non-affine residual part $r(x^*,u,t)$ as follows:
\begin{equation}
\label{nonaffine_residual}
\dot{x}^* = f_\mathrm{true} = f(x^*,t)+B(x^*,t)(u+r(x^*,u,t)).
\end{equation}
Ideally, we would like to design $u$ as 
\begin{align}
\label{ideal_nonaffine_u}
u = u^*(x,t)-r_L(x,u,t) 
\end{align}
to cancel out the unknown term $r(x,u,t)$ of the dynamical system \eqref{nonaffine_residual} by the model $r_L(x,u,t)$ learned using trajectory data, where $u^*$ is a nominal stabilizing control input for $\dot{x}=f(x,t)+B(x,t)u$ given by, \eg{}, Theorems~\ref{THM:Thm:CV-STEM} and \ref{THM:Thm:ccm_cvstem}. However, the equation \eqref{ideal_nonaffine_u} depends implicitly on $u$, which brings extra computational burden especially if the learned model $r_L(x,u,t)$ is highly nonlinear as in DNNs. In~\cite{neurallander}, a discrete-time nonlinear controller is proposed to iteratively solve \eqref{ideal_nonaffine_u}.
\begin{lemma}
\label{lemma:discrete_neurallander}
Define a mapping $\mathcal{F}$ as $\mathcal{F}(u) = u^*(x,t)-r_L(x,u,t)$, where $u^*$ and $r_L$ are given in \eqref{ideal_nonaffine_u}. If $r_L$ is Lipschitz in $u$ with a 2-norm Lipschitz constant $L_u < 1$, \ie{}, $\|r_L(x,u,t)-r_L(x,u',t)\| \leq L_u\|u-u'\|,~\forall u,u'$, then $\mathcal{F}$ is a contraction mapping for fixed $x,t$. Therefore, if $x,t$ are fixed, discrete-time nonlinear control $u_k$ defined as
\begin{align}
\label{discrete_nonaffine_u}
u_k = \mathcal{F}(u_{k-1}) = u^*(x,t)-r_L(x,u_{k-1},t) 
\end{align}
converges to a unique solution $u$ given by $u=\mathcal{F}(u)$.
\end{lemma}
\begin{proof}
Since $r_L$ is Lipschitz, we have that
\begin{equation}
\|\mathcal{F}(u)-\mathcal{F}(u')\| \leq \|r_L(x,u,t)-r_L(x,u',t)\| \leq L_u\|\Delta u\|
\end{equation}
where $\Delta u = u-u'$. Thus, the assumption $L_u < 1$ ensures that $\mathcal{F}$ is a contraction mapping for fixed $x,t$~\cite{neurallander}. \qed
\end{proof}

By applying contraction theory to the discrete-time controller \eqref{discrete_nonaffine_u} of Lemma~\ref{lemma:discrete_neurallander}, we can guarantee the stability of \eqref{nonaffine_residual} if $r_L$ is modeled by a spectrally-normalized DNN of Definition~\ref{DEF:Def:SN}.
\begin{theorem}{Robust Nonlinear Control of Learned Models}{Thm:neurallander}
Let $x$ be the trajectory of the following ideal system without the unknown part $r$ of the dynamics \eqref{nonaffine_residual}:
\begin{align}
\label{ideal_trajectory}
\dot{x} = f(x,t)+B(x,t)u^*(x,t)
\end{align}
where $u^*$ is a stabilizing controller that renders \eqref{ideal_trajectory} contracting as in Theorem~\ref{THM:Thm:NCMstability_modelfree} for $M$ which satisfies $\underline{m}\mathrm{I}\preceq M \preceq \overline{m}\mathrm{I}$ of \eqref{Mcon}. Note that such $u^*$ can be designed by using, \eg{}, Theorems~\ref{THM:Thm:CV-STEM} and \ref{THM:Thm:ccm_cvstem}. Suppose that the true dynamics \eqref{nonaffine_residual} is controlled by \eqref{discrete_nonaffine_u} and
\begin{align}
\label{disc_assump}
\exists \rho \in \mathbb{R}_{\geq 0}\text{ \st{} } \|u_k-u_{k-1}\| \leq \rho \|x-x^*\|
\end{align}
for $x^*$ in \eqref{nonaffine_residual}~\cite{neurallander}. If $\exists\bar{b}\in\mathbb{R}_{\geq 0}$ \st{} $\|B(x,t)\|\leq\bar{b}$, and if $r_L$ is modeled by a spectrally-normalized DNN of Definition~\ref{DEF:Def:SN} to have
\begin{align}
\label{learning_residual}
\|r_L(x,u,t)-r(x,u,t)\| \leq \epsilon_{\ell} 
\end{align}
for all $x\in\mathcal{S}_s$, $u\in\mathcal{S}_u$, and $t\in\mathcal{S}_t$, where $\mathcal{S}_s\subseteq\mathbb{R}^n$, $\mathcal{S}_u\subseteq\mathbb{R}^m$, and $\mathcal{S}_t\subseteq\mathbb{R}_{\geq 0}$ are some compact sets, then $r_L$ is Lipschitz continuous, and the controller \eqref{discrete_nonaffine_u} applied to \eqref{nonaffine_residual} gives the following bound in the compact set:
\begin{align}
\label{neural_lander_bound}
\|x(t)-x^*(t)\| \leq \frac{V_{\ell}(0)}{\sqrt{\underline{m}}}e^{-\alpha_{\ell}t}+\frac{\bar{b}\epsilon_{\ell}}{\alpha_{\ell}}\sqrt{\frac{\overline{m}}{\underline{m}}}(1-e^{-\alpha_{\ell}t})
\end{align}
as long as the Lipschitz constant of $r_L$ is selected to have 
\begin{align}
\label{condition_neuralland}
\exists \alpha_{\ell} \in\mathbb{R}_{>0}\text{ \st{} }\alpha_{\ell} = \alpha-\bar{b}L_u\rho\sqrt{\frac{\overline{m}}{\underline{m}}} > 0
\end{align}
where $V_{\ell}=\int^{x}_{x^*}\|\Theta\delta x\|$ as in Theorem~\ref{THM:Thm:path_integral} with $M=\Theta^{\top}\Theta$. Furthermore, the system \eqref{nonaffine_residual} with the controller \eqref{discrete_nonaffine_u} is robust against deterministic and stochastic disturbances.
\end{theorem}
\begin{proof}
If $r_L$ is modeled by a spectrally-normalized DNN, we can arbitrarily choose its Lipschitz constant $L_u$ by Lemma~\ref{lemma:SNneuralnet}. Applying \eqref{discrete_nonaffine_u} to \eqref{nonaffine_residual} yields
\begin{align}
\dot{x}^* = f_{cl}(x^*,t)+B(x^*,t)(r(x^*,u_k,t)-r_L(x^*,u_{k-1},t))
\end{align}
where $f_{cl}(x^*,t) = f(x^*,t)+B(x^*,t)u^*$. Using the Lipschitz condition on $r_L$ and the learning error assumption \eqref{learning_residual}, we have that
\begin{align}
\|r(x^*,u_k,t)-r_L(x^*,u_{k-1},t)\| &\leq \epsilon_{\ell}+L_u\|u_k-u_{k-1}\| \leq \epsilon_{\ell}+L_u\rho\|x-x^*\|
\end{align}
where \eqref{disc_assump} is used to obtain the second inequality. Since we have $\|B(x,t)\|\leq\bar{b}$ and the closed-loop system $\dot{x} = f_{cl}(x,t)$ is contracting, Theorem~\ref{THM:Thm:contraction_learning} holds with $\epsilon_{\ell0}=\bar{b}\epsilon_{\ell}$ and $\epsilon_{\ell1}=\bar{b}L_u\rho$ as long as we select $L_u$ to satisfy \eqref{condition_neuralland}, resulting in the bound \eqref{neural_lander_bound}. The final robustness statement follows from Theorems~\ref{THM:Thm:contraction_learning} and \ref{THM:Thm:contraction_learning_sto}.
\qed
\end{proof}

Theorem~\ref{THM:Thm:neurallander} implies that the control synthesis algorithms via contraction theory, including robust control of Theorems~\ref{THM:Thm:CV-STEM} and \ref{THM:Thm:ccm_cvstem} (CV-STEM), learning-based robust control of Theorems~\ref{THM:Thm:NCMstability1}, \ref{THM:Thm:NCMstability2}--\ref{THM:Thm:ncm_ccm_clf}, and~\ref{THM:Thm:lagros_stability} (NCM, LAG-ROS), can be enhanced to provide explicit robustness and stability guarantees even for systems partially modeled by DNNs that depend nonlinearly on $u$.
\begin{example}{Lagrangian Systems with Learned Residuals}{ex:neural_lander}
Let us consider the following Lagrangian system of Example~\ref{EX:ex:lag_metric} perturbed externally by unknown control non-affine residual disturbance:
\begin{align}
\label{lagrange_non_affine}
&\mathcal{H}(\mathtt{q})\ddot{\mathtt{q}}+\mathcal{C}(\mathtt{q},\dot{\mathtt{q}})\dot{\mathtt{q}}+\mathcal{G}(\mathtt{q})=u+r(x,u)
\end{align}
where $x=[\mathtt{q}^{\top},\dot{\mathtt{q}}^{\top}]^{\top}$ and the other variables are as given in Example~\ref{EX:ex:lag_metric}. Using the result of Theorem~\ref{THM:Thm:neurallander}, we can design a discrete-time nonlinear controller by augmenting the exponentially stabilizing controller of Example~\ref{EX:ex:lag_metric} with a learned residual part $r_L(x,u)$ as follows:
\begin{align}
\label{lagrange_non_affine_u}
u_k &= -\mathcal{K}(t)(\dot{\mathtt{q}}-\dot{\mathtt{q}}_r)+\mathcal{H}(\mathtt{q})\ddot{\mathtt{q}}_r+\mathcal{C}(\mathtt{q},\dot{\mathtt{q}})\dot{\mathtt{q}}_r+\mathcal{G}(\mathtt{q})-r_L(x,u_{k-1})
\end{align}
where $\dot{\mathtt{q}}_r=\dot{\mathtt{q}}_d(t)-\Lambda(t)(\mathtt{q}-\mathtt{q}_d(t))$, $\mathcal{K}: \mathbb{R}_{\geq 0}\rightarrow\mathbb{R}^{n\times n}$, $\Lambda: \mathbb{R}_{\geq 0}\rightarrow\mathbb{R}^{n\times n}$, and $(\mathtt{q}_d,\dot{\mathtt{q}}_d)$ is the target trajectory of the state $(\mathtt{q},\dot{\mathtt{q}})$, and $\mathcal{K},\Lambda \succ 0$ are control gain matrices (design parameters). Again, note that $\dot{\mathcal{H}}-2\mathcal{C}$ is skew-symmetric with $\mathcal{H} \succ 0$ by construction. This gives us the following closed-loop virtual system of a smooth path $q(\mu,t)$ parameterized by $\mu \in [0,1]$, which has $q(\mu=0,t)=\dot{\mathtt{q}}_r$ and $q(\mu=1,t)=\dot{\mathtt{q}}$ as its particular solutions as in Example~\ref{EX:ex:lag_metric}, but now with non-zero perturbation due to $r(x,u)$:
\begin{align}
\label{lagrange_cl_non_affine}
&\mathcal{H}(\dot{q}-\ddot{\mathtt{q}}_r)+(\mathcal{C}+\mathcal{K})(q-\dot{\mathtt{q}}_r)=\mu(r(x,u_k)-r_L(x,u_{k-1})).\nonumber
\end{align}
After some algebra as in the proof of Theorem~\ref{THM:Thm:neurallander}, we can show that
\begin{align}
\dfrac{d}{dt}\int_0^1\|\Theta\partial_{\mu} q\| &\leq -\left(\frac{k_{\ell}}{h_u}-\frac{L_u\rho}{h_{\ell}}\right)\int_0^1\|\Theta\partial_{\mu} q\|+\frac{\epsilon_{\ell}}{\sqrt{h_{\ell}}}
\end{align}
where $\mathcal{H}=\Theta^{\top}\Theta$, $h_{\ell}\mathrm{I}\preceq\mathcal{H}\preceq h_u\mathrm{I}$, $k_{\ell}\mathrm{I}\preceq\mathcal{K}$, $\|u_k-u_{k-1}\| \leq \rho\|\dot{\mathtt{q}}-\dot{\mathtt{q}}_r\|$, $\epsilon_{\ell}$ is the learning error of $r_L$, and $L_u$ is the Lipschitz constant of $r_L$ (assuming $r_L$ is modeled by a spectrally-normalized DNN of Definition~\ref{DEF:Def:SN}). This indeed indicates that the tracking error of the Lagrangian system \eqref{lagrange_non_affine} is exponentially bounded as proven in Theorem~\ref{THM:Thm:neurallander}.
\end{example}

In~\cite{neurallander}, the technique in Theorem~\ref{THM:Thm:neurallander} and in Example~\ref{EX:ex:neural_lander} is used to perform precise near-ground trajectory control of multi-rotor drones, by learning complex aerodynamic effects caused by high-order interactions between multi-rotor airflow and the environment. It is demonstrated that it significantly outperforms a baseline nonlinear tracking controller in both landing and cross-table trajectory tracking tasks. Theorem~\ref{THM:Thm:neurallander} enables applying it to general nonlinear systems with state and control dependent uncertainty, as long as we have a nominal exponentially stabilizing controller (which can be designed by Theorems~\ref{THM:Thm:CV-STEM} and~\ref{THM:Thm:ccm_cvstem}, or approximately by Theorems~\ref{THM:Thm:NCMstability1}, \ref{THM:Thm:NCMstability2}--\ref{THM:Thm:ncm_ccm_clf}, and~\ref{THM:Thm:lagros_stability}).
\subsection{Learning Contraction Metrics from Trajectory Data}
\label{Sec:learning_contractionmetric}
This section presents a data-driven method to design contraction metrics for stability guarantees directly from trajectory data, assuming that its underlying dynamical system is completely unknown, unlike \eqref{nonaffine_residual}. We specifically consider the situation where the underlying system is given by a continuous-time autonomous system, $\dot{x}=f(x)$ with $f$ being unknown, and the state $x \in \mathbb{R}^n$ being fully observed.

Let $\mathcal{S}_s \subseteq \mathbb{R}^n$ be a compact set and $\mathcal{S}_t\subseteq \mathbb{R}_{\geq 0}$ be the maximal interval starting at zero for which a unique solution $\varphi_t(\xi)$ exists for all initial conditions $\xi \in \mathcal{S}_s$ and $t \in \mathcal{S}_t$. Let us also assume access to sampled trajectories generated from random initial conditions. Notations to be used in this section are given in Table~\ref{tab:notations_lmetric}.
\begin{table}[htbp]
\caption{Notations in Sec.~\ref{Sec:learning_contractionmetric}~\cite{boffi2020learning}. \label{tab:notations_lmetric}}
\footnotesize
\begin{center}
\renewcommand{\arraystretch}{1.2}
\rowcolors{1}{uiucbluedark!5}{uiucbluedark!10}
\begin{tabular}{ l l } 
\hline \hline
$\mathbb{B}^n_2(r)$ & Closed $\ell_2$-ball in $\mathbb{R}^n$ of radius $r$ centered at $0$ \\ \arrayrulecolor{mygray}\hline
$\mathbb{B}^n_2(\xi,r)$ & Closed $\ell_2$-ball in $\mathbb{R}^n$ of radius $r$ centered at $\xi$ \\ \hline
$\mathbb{S}^{n-1}$ & Sphere in $\mathbb{R}^n$ \\ \hline
$\mu_{\mathrm{Leb}}(\cdot)$ & Lebesgue measure on $\mathbb{R}^n$ \\ \hline
$\psi_t(\cdot)$ & Induced flow on prolongated system given as $p(x,\delta x) = (f(x),(\partial f/\partial x)\delta x)^{\top}$ \\ \hline
$\theta_t(\delta \xi,\xi)$  &Second element
of $\psi_t(\xi,\delta \xi)$ \\ \hline
$\zeta_n(r)$ & Haar measure of a spherical cap in $\mathbb{S}^{n-1}$ with arc length $r$ \\ \hline
 $\nu$ & Uniform probability measure on $\mathcal{S}_s\times \mathbb{S}^{n-1}$ \\ \hline
$\mathcal{T}S$ & Tangent bundle of $S = \cup_{t\in \mathcal{S}_t}\varphi_t(\mathcal{S}_s)$ \\ \arrayrulecolor{black}\hline\hline
\end{tabular}
\end{center}
\end{table}

The following theorem states that a contraction metric defined by $\mathcal{M}$, learned from trajectory data, indeed guarantees contraction of the unknown underlying dynamical system~\cite{boffi2020learning}.
\begin{theorem}{Contraction in Learned Contraction Metrics}{Thm:data_driven_cont_thm}
Suppose that $\mathcal{S}_s \subseteq \mathbb{R}^n$ is full-dimensional, $n \geq 2$, and $\dot{x} = f(x)$ is contracting in a metric defined by a uniformly bounded $M(x)\succ 0$, \ie{},
\begin{align}
\dot{M}+M\frac{\partial f}{\partial x}+\frac{\partial f}{\partial x}^{\top}M\preceq-2\alpha M
\end{align}
of \eqref{eq_MdotContracting} and $\underline{m}\mathrm{I}\preceq M \preceq \overline{m}\mathrm{I}$ of \eqref{Mcon} hold for $M$. Suppose also that a matrix-valued function $\mathcal{M}(x)$ with $\mathcal{M}(x)\succeq \underline{m}_L \mathrm{I}$ is learned to satisfy $\nu(Z_b)\leq\epsilon_{\ell}$, where the contraction violation set $Z_b$ for $p(x,\delta x) = (f(x),(\partial f/\partial x)\delta x)^{\top}$ is defined as
\begin{align}
&Z_b = \left\{(\xi,\delta\xi)\in \mathcal{S}_s\times\mathbb{S}^{n-1}:\max_{t\in \mathcal{S}_t}(\nabla \mathcal{V}(\psi_{t}(\xi,\delta \xi))^{\top}p(\psi_{t}(\xi,\delta \xi))-\alpha \mathcal{V}(\psi_t(\xi,\delta \xi)) > 0)\right\}
\end{align}
with $\mathcal{V}: \mathcal{T}S \rightarrow\mathbb{R}$ being a learned differential Lyapunov function $\mathcal{V}(x,\delta x) = \delta x^{\top}\mathcal{M}(x)\delta x$ for $S = \cup_{t\in \mathcal{S}_t}\varphi_t(\mathcal{S}_s)$. Define $\bar{B}$, $B_H$, $B_{\nabla q}$, and $B_{\nabla \mathcal{V}}$ as
\begin{align}
\bar{B}&=B_H(B_{\nabla q}+\alpha B_{\nabla \mathcal{V}})\left(\frac{\overline{m}}{\underline{m}}\right)^{3/2} \\
B_H &= \sup_{x\in S}\left\|\frac{\partial^2f}{\partial x^2}\right\| \nonumber \\
B_{\nabla q} &= \sup_{x\in S}\|\nabla q(x)\| \\
B_{\nabla \mathcal{V}} &= \sup_{x\in S}\|\nabla \mathcal{V}(x)\|
\end{align}
where $q = \nabla \mathcal{V}(x)^{\top} f(x)$. Also, we define $r_{\epsilon_{\ell}}$ and $r_b$ as
\begin{align}
r_{\epsilon_{\ell}} &= \sup\left\{r\in\mathbb{R}_{>0}:r^n\zeta_n(r)\leq \frac{\epsilon_{\ell}\mu_{\mathrm{Leb}}(\mathcal{S}_s)}{\mu_{\mathrm{Leb}}(\mathbb{B}_2^{n}(1))}\right\} \nonumber \\
r_b &= \sqrt{\frac{r_{\epsilon_{\ell}}\bar{B}}{\eta \alpha \underline{m}_L}}
\end{align}
where $\eta \in (0,1)$. Finally, define $\tilde{X}_t(r_b)$ as 
\begin{align}
\tilde{X}_t(r_b) = \left\{\xi \in \tilde{X}:\inf_{\delta \xi \in \mathbb{S}^{n-1}} \|\theta_t(\delta \xi,\xi)\| \geq r_b\right\}
\end{align}
for $t \in \mathcal{S}_t$, where $\tilde{X} = \{\xi \in \mathcal{S}_s: \mathbb{B}^n_2(\xi,r_{\epsilon_{\ell}}) \subset \mathcal{S}_s\}$. Then the system is contracting in the learned metric defined by $\mathcal{M}(x)$ at the rate $(1-\eta)\alpha$ with $\eta \in (0,1)$, for every $x \in \tilde{S}(r_b) = \cup_{t\in \mathcal{S}_t}\varphi_t(\tilde{X}_t(r_b))$.
\end{theorem}
\begin{proof}
See Theorem~5.2 of~\cite{boffi2020learning}. \qed
\end{proof}

Agreeing with our intuition, Theorem~\ref{THM:Thm:data_driven_cont_thm} ensures that the region $\tilde{S}(r_b)$ where the contraction condition holds becomes larger, as the upper bound $\epsilon_{\ell}$ on $\nu(Z_b)$ for the contraction violation set $Z_b$ becomes smaller, yielding the learned contraction metric of higher quality. In~\cite{boffi2020learning}, it is proven that if the samples are drawn uniformly from $\mathcal{S}_s\times \mathbb{S}^{n-1}$, we have $\epsilon_{\ell} \leq O(k\cdot \text{polylog}(N)/N)$ decay rates for various parametric and nonparametric function classes, where $N$ is the number of sampled trajectories and $k$ is the effective number of parameters of the function class of interest. Such a learned contraction metric can be used in Theorems~\ref{THM:Thm:NCMstability_modelfree} and~\ref{THM:Thm:neurallander} for certifying robustness and stability of model-free systems.

\newpage
\section{Concluding Remarks}
\label{Sec:conclusion}
The main contribution of this paper is twofold. First, a tutorial overview of contraction theory is presented to generalize and simplify Lyapunov-based stability methods for incremental exponential stability analysis of nonlinear non-autonomous systems. The use of differential dynamics and its similarity to an LTV system allow for LMI and convex optimization formulations that are useful for systematic nonlinear control and estimation synthesis. Second, various methods of machine learning-based control using contraction theory are presented to augment the existing learning frameworks with formal robustness and stability guarantees, extensively using the results of the first part of the paper. Such formal guarantees are essential for their real-world applications but could be difficult to obtain without accounting for a contracting property. It is also emphasized that, especially in situations where ISS and uniform asymptotic stability-based arguments render nonlinear stability analysis unnecessarily complicated, the use of exponential stability and the comparison lemma in contraction theory helps to achieve significant conceptual and methodological simplifications. A connection to the KYP and bounded real lemmas is also shown in the context of contraction-based incremental stability analysis.

Considering the promising outcomes on its utilization for model-based learning in Sec.~\ref{Sec:learning_stability}--\ref{Sec:adaptive} and for model-free data-driven learning in Sec.~\ref{Sec:datadriven}, the methods of contraction theory that are surveyed in this paper provide important mathematical tools for formally providing safety and stability guarantees of learning-based and data-driven control, estimation, and motion planning techniques for high-performance robotic and autonomous systems. Examples are elucidated to provide clear guidelines for its use in deep learning-based stability analysis and its associated control design for various nonlinear systems.
\section*{Acknowledgments}
This work was in part funded by NASA's Jet Propulsion Laboratory, California Institute of Technology, and benefited from discussions with Nicholas Boffi (CMU), Winfried Lohmiller (MIT), Brett Lopez (UCLA), Ian Manchester (University of Sydney), Quang Cuong Pham (NTU), Sumeet Singh (Google Brain Robotics), Stephen Tu (USC), Patrick Wensing (University of Notre Dame), Chuchu Fan (MIT), and Guanya Shi (CMU).

\bibliographystyle{elsarticle-num-names}

\newpage
\bibliography{root}

\end{document}